\documentclass[journal]{IEEEtran}
\usepackage{amssymb}
\ifCLASSINFOpdf
\else
\fi

\usepackage[utf8]{inputenc} 
\usepackage[T1]{fontenc}    
\usepackage{cite}
\usepackage{url}            
\usepackage{booktabs}       
\usepackage{amsfonts}       
\usepackage{nicefrac}       
\usepackage{microtype}      
\usepackage{amsmath}

\usepackage{pdfpages}
\usepackage{makecell}
\usepackage{graphicx}
\usepackage{epstopdf}
\usepackage{multirow}
\usepackage{subfigure}
\begin{document}

\title{Rain Streak Removal for Single Image via Kernel Guided CNN}

\author{Ye-Tao Wang,
Xi-Le Zhao\IEEEauthorrefmark{1},
Tai-Xiang Jiang\IEEEauthorrefmark{1},
Liang-Jian Deng,
Yi Chang,
and Ting-Zhu Huang
\thanks{\IEEEauthorrefmark{1} Corresponding authors.}
\thanks{Y.-T. Wang, X.-L. Zhao, T.-X. Jiang, L.-J. Deng and T.-Z. Huang are with the School of Mathematical Sciences, University of Electronic Science and Technology of China, Chengdu, 611731, P. R. China. Y. Chang is with the Institute of Control and Information Technology, Huazhong University of Science and Technology, Wuhan, 430074, P. R. China. E-mails: xlzhao122003@163.com, taixiangjiang@gmail.com, liangjian1987112@126.com, yichang@hust.edu.cn, tingzhuhuang@126.com.}}%
\maketitle

\begin{abstract}
  Rain streak removal is an important issue and has recently been investigated extensively.
  Existing methods, especially the newly emerged deep learning methods, could remove the rain streaks well in many cases.
  However the essential factor in the generative procedure of the rain streaks, i.e., the motion blur, which leads to the line pattern appearances, were neglected by the deep learning rain streaks removal approaches and this resulted in over-derain or under-derain results.
  In this paper, we propose a novel rain streak removal approach using a kernel guided convolutional neural network (KGCNN), achieving the state-of-the-art performance with simple network architectures.
  We first model the rain streak interference with its motion blur mechanism.
  Then, our framework starts with learning the motion blur kernel, which is determined by two factors including angle and length, by a plain neural network, denoted as parameter net, from a patch of the texture component.
  Then, after a dimensionality stretching operation, the learned motion blur kernel is stretched into a degradation map with the same spatial size as the rainy patch.
  The stretched degradation map together with the texture patch is subsequently input into a derain convolutional network, which is a typical ResNet architecture and trained to output the rain streaks with the guidance of the learned motion blur kernel.
  Experiments conducted on extensive synthetic and real data demonstrate the effectiveness of the proposed method, which preserves the texture and the contrast while removing the rain streaks.
\end{abstract}


%
\IEEEpeerreviewmaketitle

\section{Introduction}\label{intro}
\begin{figure}[!htp]
\centering\scriptsize\renewcommand\arraystretch{1}
\begin{tabular}{cc}
\includegraphics[width=0.48\linewidth]{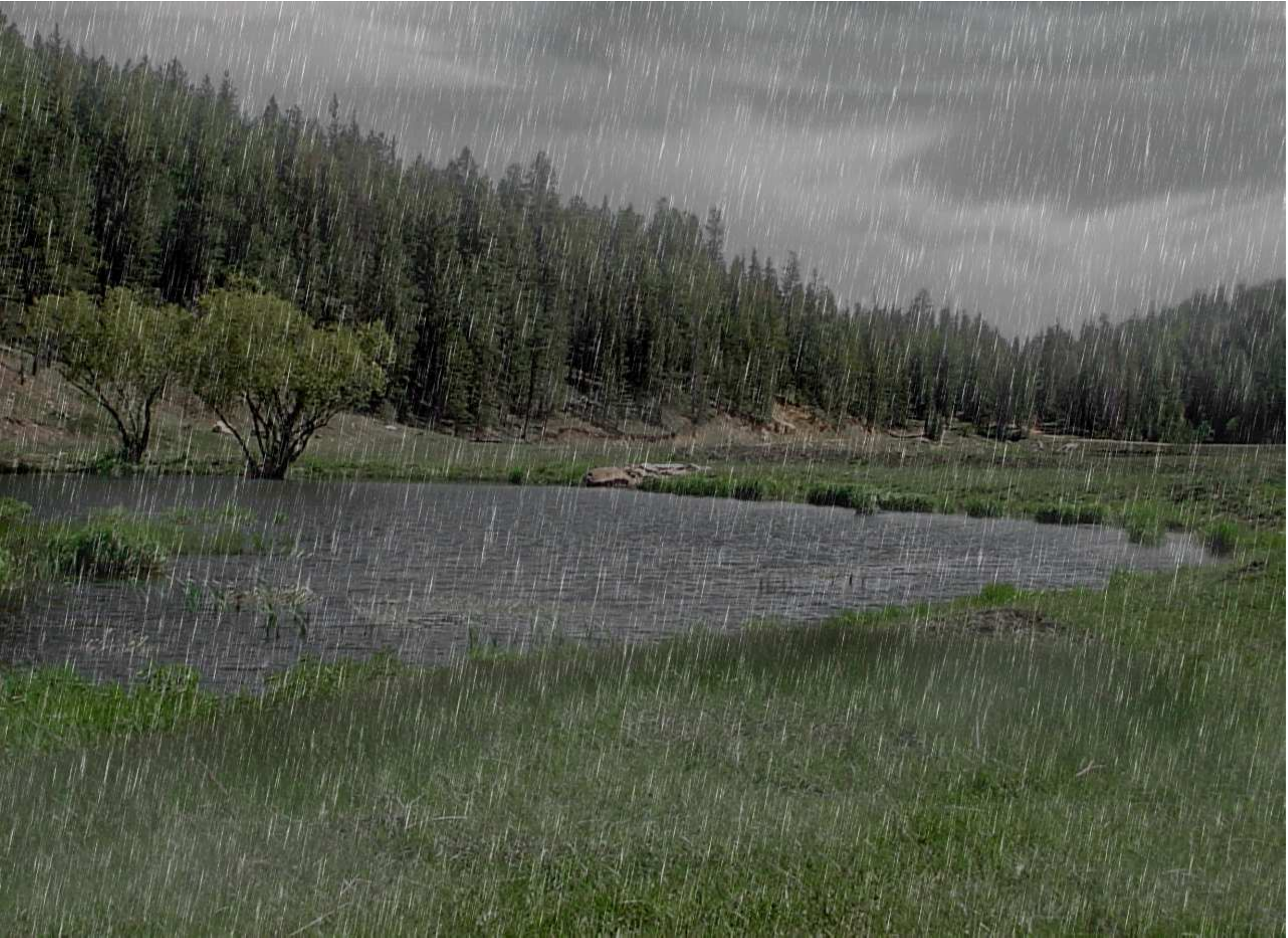}&\includegraphics[width=0.48\linewidth]{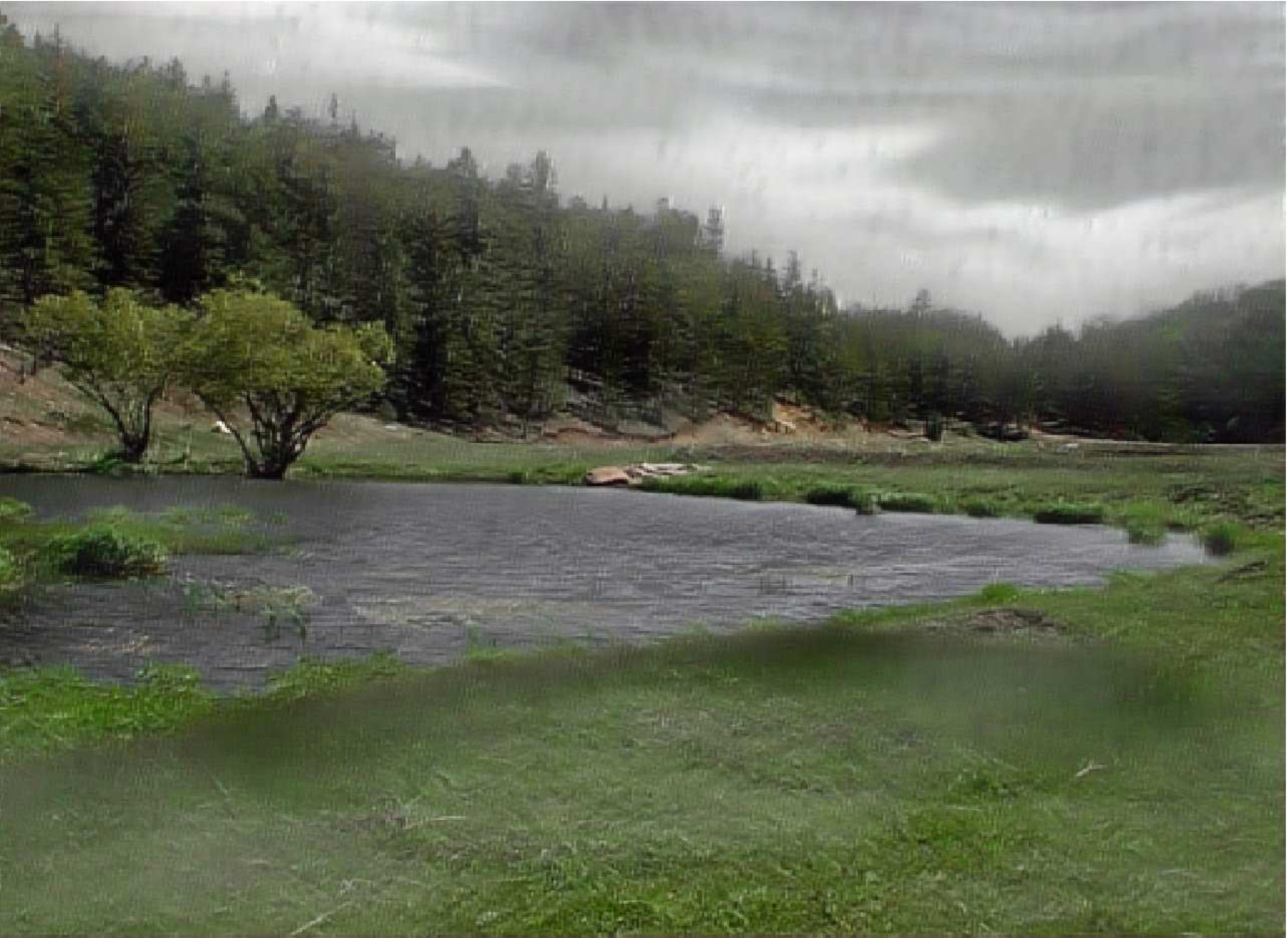}\\
(a)Rainy & (b)DID \cite{zhang2018density} \\
\includegraphics[width=0.48\linewidth]{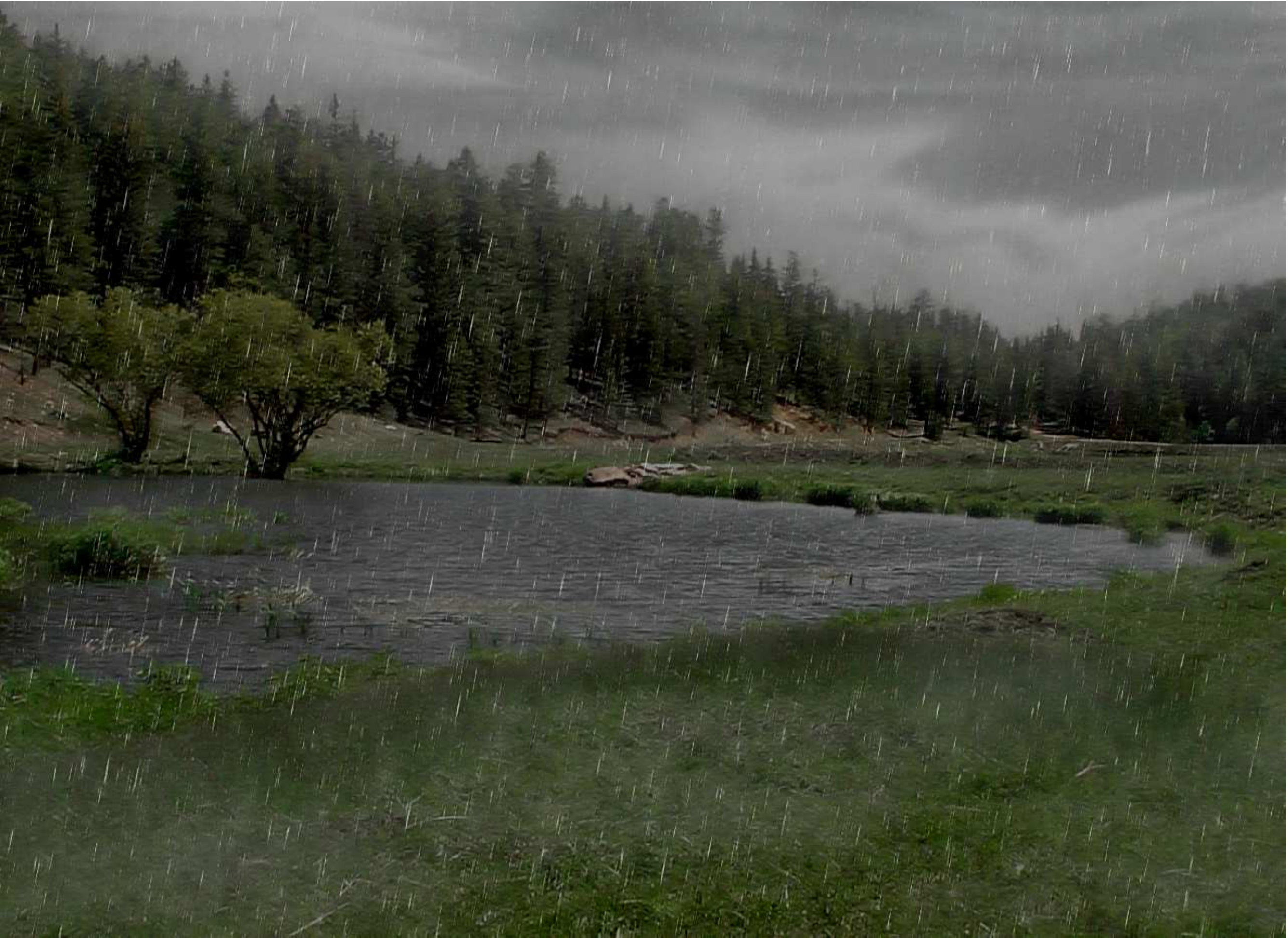}&\includegraphics[width=0.48\linewidth]{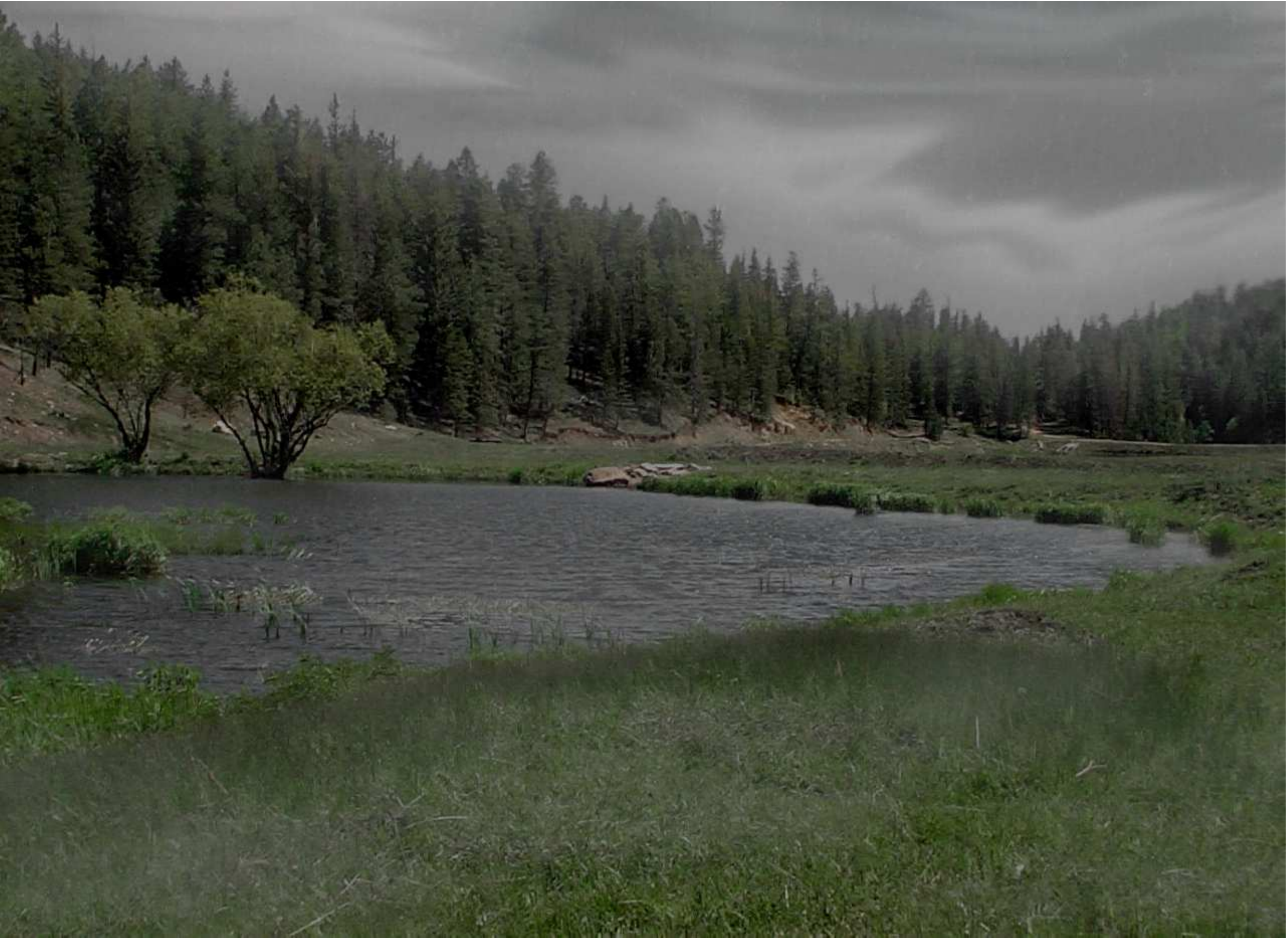}\\
(c)DDN \cite{fu2017removing}& (d)Proposed KGCNN
\end{tabular}
\caption{An example of rain streak removal for a real-world rainy image. \textbf{Top-left}: the rainy image. \textbf{Top-right}: the derain result by DID \cite{zhang2018density}. \textbf{Button-left}: the derain result by DDN \cite{fu2017removing}. \textbf{Button-right}: the derain result by the proposed KGCNN.}
\label{1st}
\end{figure}

Outdoor vision systems are frequently affected by bad weather conditions, one of which is the rain.
Because of the high motion velocities and the light scattering, raindrops usually introduce bright streaks into the images or videos acquired by cameras.
This undesirable interference will degrade the performance of various computer vision algorithms \cite{garg2007vision}, such as object detection \cite{zhang2017bayesian}, event detection \cite{shehata2008video}, action recognition \cite{song2018spatio}, and scene analysis \cite{itti1998model}.
Therefore, alleviating the effects from the rain is an essential task and has been investigated extensively.
Fig. \ref{1st} exhibits one example of the single image rain streak removal.

\begin{figure*}[htb]
 \begin{center}
    \includegraphics[width=1.0 \textwidth]{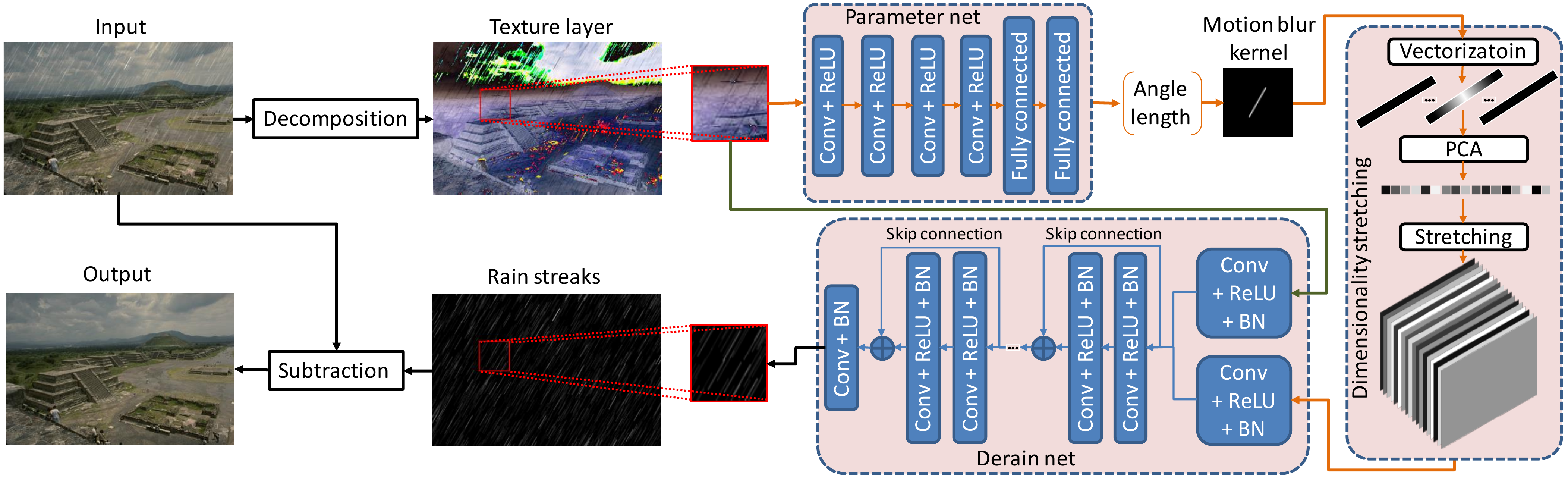}
 \end{center}
\caption{The flowchart of the proposed KGCNN  single image rain streak removal framework. 1) The rainy image is decomposed into the texture component and the structure component. 2) A rainy patch is feed to the parameter net to obtain the angle and the length of the motion blur kernel. 3) The motion blur kernel is stretched to the degradation map. 3) The rainy patch and the degradation map is then transmitted into the derain net, whose output is the  rain streak patch. 4) Finally, derain image is obtained by the subtraction of the rainy image and the rain streak image.}
 \label{architecture}
\end{figure*}

Most of the traditional methods always focus on the discriminative characteristics of the rain streaks and the clean background, for instance, the high-frequency property \cite{kang2012automatic,sun2014exploiting}, directionality \cite{chang2017transformed,deng2018directional,du2018single,Jiang_2017_CVPR,jiang2018fastderain} and repeatability \cite{chen2013generalized} of the rain streaks and the piecewise smoothness \cite{li2016rain,li2017single,chen2013generalized,Jiang_2017_CVPR,jiang2018fastderain} of the background (without loss of generality, we denote the rain-free content as ``background'' throughout this paper).
It is common for the model based methods to elaborately tailor an optimization model with the hand-crafted regularizer expressing the prior knowledge.
Although these model based methods go into much depth on the distinct characteristics of the rains streaks, they are often insufficient to cover the majority of the important factors in a real rainy scene since the degradation of rain streaks can be very complex.
The traditional learning based methods attempt to overcome this shortage by inferring the discriminative dictionaries\cite{luo2015removing}, the GMMs \cite{li2016rain,li2017single}, the stochastic distributions\cite{Wei_2017_ICCV} or the convolutional filters \cite{li2018video,gu2017joint,zhang2017convolutional} from the data.
Benefit from the complex representation ability of the convolutional neural network, the deep learning techniques \cite{qian2018attentive,li2018fast,zhang2018density,chen2018robust,Liu_2018_CVPR,Yang_2017_CVPR,li2018video,wei2018semi} further leverage the data to the most extent, and obtain promising results.

As it can be seen in Fig. \ref{1st}, state-of-the-art de-raining algorithms \cite{fu2017removing,zhang2018density} tend to obtain the under-derain result (bottom-left) or the over-derain result (top-right).
We attribute these phenomena to the fact that it is challenging for deraining methods, even the deep learning based methods, to distinguish the rain streaks and the line pattern textures (e.g. the grass in Fig. \ref{1st}).
The rain streaks removal performance of the neural network can be heighten by adopting deeper architecture as \cite{fu2017removing} or elaborately designing the architecture, in which the contexture information are taken into account, as \cite{Yang_2017_CVPR,zhang2018density}.
However, it is still difficult to purposefully enhance the capacity of the neural network to face the fore-mentioned challenge.
Meanwhile, anther question is that can we address this challenge and achieve promising performance with the common and simple neural network structures?

In this paper, referring to \cite{garg2007vision}, in which it was pointed out that the appearance of the rain streaks is mainly related to the motion blur mechanism, we propose a novel degradation model of the rain streak   interference taking the motion blur procedure into account.
By modeling the degradation of rain streaks as the motion blur, we are able to utilize two important distinct characteristics of the rain streaks, i.e., the repeatability and the directionality.
The line pattern textures do not possess the same generation mechanism as the rain streaks'.
Therefore, this modeling strategy would contribute to distinguish the rain streaks and the line pattern textures.
After modeling the rain streaks with motion blur kernel, the questions come to 1) how to infer the motion kernel from the data, and 2) how to utilize the information provided by the motion blur kernel when deraining.

In our approach, we assume that the rain streaks in a small patch approximately share the same motion blur kernel.
At the beginning, a rainy patch is feed to the parameter net, a plain 6-layer network, to infer the angle and the length of the motion blur kernel.
To enable the learned motion blur kernels to participate in the subsequent deraining process, we adopt the dimensionality stretching strategy \cite{zhang2018learning}, which stretched the motion blur kernels to degradation maps with the same spatial resolution as the detail patches.
Then, the detail patch together with the degradation map is input into a common 26 layer ResNet, whose output is a patch of rain streaks.
Finally, we obtain the derain results by subtracting the rain streaks from rainy image.
The core idea of our framework is exploiting the generation mechanism of the rain streaks to guide rain streak removal, and the flowchart of our framework is illustrated in Fig. \ref{architecture}.



\textbf{Contributions}:
The contributions of this paper mainly include four aspects.
\begin{itemize}
\item We build a novel rain streak generation model which takes the motion blur kernel into account. This modeling strategy enables us to utilize the repeatability and the directionality of the rain streaks. 

\item A sub-net, i.e., the parameter net, is built to learn the parameters (length and angle) of the motion blur kernel. Unlike existing methods using sub-net to embed the contextual information, our sub-net is designed to exploit  the generation information of the rain streaks. 

\item We propose an effective kernel guided CNN (KGCNN) framework, in which the network structures are common and simple, for rain streak removal. Within this framework, the automatically learned motion blur kernel thoroughly guides the process of rain streak removal.

\item Extensive experiments are conducted on publicly available real and synthetic data. Qualitative and quantitative comparisons with existing state-of-the-art methods are presented. The results show that the KGCNN removes rain streaks well while keeping the texture and the contrast of background.
\end{itemize}

The organization of this paper is as follows.
We provide an overview of the existing deraining methods in Section \ref{related_works}.
Section \ref{architecture_of_KGCNN} gives the detailed architecture of the proposed KGCNN.
In Section \ref{experimental_results}, experimental results on the synthetic data and the real-world data are reported.
Finally, we draw conclusions in Section \ref{conclusion}.

\section{Related Works}\label{related_works}
In the past decades, numerous methods have been proposed to improve the visibility of images/videos captured with rain streak interference. \cite{kang2012automatic,sun2014exploiting,chen2013generalized,luo2015removing,li2017single,li2016rain,Zhu_2017_ICCV,chen2017error,gu2017joint,
chang2017transformed,deng2018directional,du2018single,wang2017hierarchical,ren2018simultaneous,Yang_2017_CVPR,fu2017clearing,zhang2017image,zhang2018density,zhang2017convolutional,
qian2018attentive,li2018fast,garg2004detection,tripathi2012video,tripathi2014removal,kim2015video,santhaseelan2015utilizing,
you2016adherent,Jiang_2017_CVPR, jiang2018fastderain,  Wei_2017_ICCV,li2018video,ren2017video,chen2018robust,Liu_2018_CVPR,shen2018deep,wei2018semi}.
Traditionally, these methods can be divided into two categories: single image based methods and multiple-images/videos based methods. Nevertheless, the explosive development of the deep learning brings in a novel branch, i.e., the deep learning methods.
\subsection{Single Image Based Methods}
For the single image derain task, Kang {\em et al}. \cite{kang2012automatic} decomposed a rainy image into low-frequency (LF) and high-frequency (HF) components using a bilateral filter and then performed morphological component analysis (MCA)-based dictionary learning and sparse coding to separate the rain streaks in the HF component.
To alleviate the loss of the details when learning HF image bases, Sun {\em et al}. \cite{sun2014exploiting} tactfully exploited the structural similarity of the derived HF image bases.
Kim {\em et al.} \cite{kim2013single} took advantage of the nonlocal similarity.
Chen {\em et al}. \cite{chen2013generalized} considered the similar and repeated patterns of the rain streaks and the smoothness of the rain-free content.
Sparse coding and dictionary learning were adopted by Luo {\em et al.} \cite{luo2015removing} and Son {\em et al.} \cite{son2016rain}.
In their results, the details of backgrounds were well preserved.
The recent work by Li {\em et al.} \cite{li2017single} utilized the Gaussian mixture model (GMM) patch priors for rain streak removal, with the ability to account for rain streaks of different orientations and scales.
Meanwhile, the directional property of rain streaks received a lot of attention in \cite{chang2017transformed,Zhu_2017_ICCV,deng2018directional,du2018single} and these methods achieved promising performances.
Ren {\em et al.} \cite{ren2018simultaneous} removed the rain streaks from the image recovery perspective.
Wang {\em et al.} \cite{wang2017hierarchical} took advantage of the image decomposition and dictionary learning.

\subsection{Video Based Methods}
Garg {\em et al}. \cite{garg2004detection} first raised a video rain streak removal method with a comprehensive analysis of the visual effects of rain on an imaging system.
Since then, many approaches have been proposed for the video rain streak removal task and obtained good performances with different rain circumstances.
Tripathi {\em et al}. \cite{tripathi2012video} took the spatiotemporal properties into consideration.
In \cite{chen2013generalized}, the similarity and repeatability of rain streaks were considered, and a generalized low-rank appearance model was proposed.
Chen {\em et al.} \cite{chen2014rain} considered the highly dynamic scenes.
Whereafter, Kim {\em et al}. \cite{kim2015video} considered the temporal correlation of rain streaks and the low-rank nature of clean videos. 
Santhaseelan {\em et al.} \cite{santhaseelan2015utilizing} detected and removed the rain streaks based on phase congruency features.
Additionally, comprehensive early existing video based methods were reviewed in \cite{tripathi2014removal}.
You {\em et al.} \cite{you2016adherent} took the raindrops adhered to a windscreen or window glass into account.
In \cite{Jiang_2017_CVPR} and \cite{jiang2018fastderain}, a novel tensor-based video rain streak removal approach was proposed by considering the directional property.
The rain streaks and the clean background were stochastically modeled as a mixture of Gaussians by Wei {\em et al.} \cite{Wei_2017_ICCV}.
The convolutional sparse coding (CSC), which has shown its ability in image cartoon-texture decomposition \cite{zhang2017convolutional}, was also adopted by Li {\em et al.} \cite{li2018video} for the video rain streaks removal.
Ren {\em et al.} \cite{ren2017video} addressed the video desnow and derain task based on matrix decomposition.

\subsection{Deep Learning Based Methods}
The deep learning based method was first applied to derain in \cite{eigen2013restoring}, in which a 3-layer convolutional neural network (CNN) was designed to remove static raindrops and dirt spots from pictures taken through window glass.
Fu {\em et al.} \cite{fu2017clearing} was the first to successfully tailor a deep CNN for the rain streak removal task.
Moreover, in \cite{fu2017removing}, Fu {\em et al.} designed the deep detail network (DDN) to further improved the performance by adopting the well-known deep residual network (ResNet) \cite{he2016deep} structure.
Pan {\em et al.} \cite{pan2018learning} simultaneously operated on the texture component and the structure component.
Yang {\em et al.} \cite{Yang_2017_CVPR} added a binary map, which reflects the contextual information, in the rain streak observation model and constructed a deep network that jointly detected and removed rain streaks.
Meanwhile, the increasingly popular Generative Adversarial Networks (GAN) was first used in \cite{zhang2017image} for rain streak removal and recently applied to the task of dealing with adherent raindrop \cite{qian2018attentive}.
In \cite{chen2018robust}, Chen {\em et al.} proposed a CNN Framework for the video rain streak removal task, while the recurrent neural network was adopted by Liu {\em et al} \cite{Liu_2018_CVPR}.
For jointly rain-density estimation and derain, Zhang {\em et al}. \cite{zhang2018density} raised a density aware multi-stream densely connected convolutional neural network (DID).
In \cite{li2018fast}, both the rain component and the background component are considered to remove rain streaks.
Fan {\em et al}. \cite{fan2018residual} developed a residual-guide feature fusion network, which was detachable to meet different rainy conditions.
A lightweight pyramid of networks was proposed in \cite{fu2018lightweight}, using the domain-specific knowledge to simplify the learning process.

\section{The Framework of The Kernel Guided Convolutional Neural network}\label{architecture_of_KGCNN}
In this section, we will give our rain streak observation model and subsequently clarify the detail architecture of the proposed derain framework.
As exhibited in Fig. \ref{architecture}, there are mainly three parts, i.e., the parameter net, the dimensionality stretching, and the derain net.
The main stream is 1) decomposing the rainy image into texture component and structure component; 2) processing the patches in the texture component using the parameter net, the PCA operation, and the derain net; 3) subtracting the obtained rain streaks and obtaining the derain result.

\subsection{Observation Model}
As mentioned previously, the rain streaks can be approximately viewed as sharing the same motion blur kernel.
Hence the basic unit of our observation model is the patch.
Similar to many existing methods, the rainy image is modeled as a linear superposition:
\begin{equation}\label{ob1}
\mathbf O  = \mathbf B  + \mathbf R,
\end{equation}
where $\mathbf O$, $\mathbf B$, and $\mathbf R\in\mathbb{R}^{m\times n} $ are patches of the observed rainy image, the underlying background (i.e. the background) and the rain streaks, respectively.
After taking the motion blur into consideration, the observation model in Eq. \eqref{ob1} turns to be:
\begin{equation}\label{ob2}
\mathbf O  = \mathbf B  + \mathbf K(\theta,l) \otimes \mathbf {M},
\end{equation}
where $\theta$ and $l$ are respectively the {\em angle} and {\em length} of the motion blur kernel $\mathbf K\in\mathbb{R}^{p\times p}$, $\mathbf {M}$ is the raindrops mask, and $\otimes$ denotes the convolution operation.
Because of the high velocity of the raindrops, the appearance of the rain streaks are mostly linear.
Hence using the angle $\theta$ and the length $l$ to characterize the motion blur kernel of the rain streaks is reasonable and its advantage illustrated in the next subsection.

Meanwhile, many existing methods conduct the rain streak removal procedures on the detail component \cite{fu2017clearing,fu2017removing,pan2018learning} or the high-frequency (HF) component \cite{kang2012automatic,sun2014exploiting}.
Following this research line, we adopt the guided filter method in \cite{he2013guided} as the low-pass filter because it is simple and fast to implement\footnote{As discussed in \cite{fu2017clearing}, the choice of low-pass filter is not limited to guided filtering.}.
The rainy patch is decomposed into two parts the texture component $ \mathbf O_{\text{T}}$ (denoted as ``detail component'' in  \cite{fu2017clearing,fu2017removing,pan2018learning}) and the structure component $\mathbf O_{\text{S}}$, and they satisfy $\mathbf O = \mathbf O_{\text{S}} + \mathbf O_{\text{T}}$.


The advantages of processing on the texture component have been fully discussed in \cite{fu2017clearing,fu2017removing}.
In order to facilitate the readers, we briefly bring them herein.
It can be found in Fig. \ref{architecture} that the texture component consists of all rain streaks, i.e., $\mathbf O_{\text{T}} = \mathbf B_{\text{T}}+\mathbf R$, so that training and testing on the texture component $\mathbf O_{\text{T}}$ is sufficient and compact.
Meanwhile, the texture component is sparser and the range of the values is significantly decreased compared to the pixels in the original image domain.
This also decreases the mapping range of the neural network, making the network focus on the important information.

After the decomposition in Eq. \eqref{decomp}, the observation model becomes
\begin{equation}
\begin{aligned}
 \mathbf O_{\text{T}} =  \mathbf B_{\text{T}}  + \mathbf K(\theta,l) \otimes \mathbf {M},
\end{aligned}
\label{decomp}
\end{equation}
where $\mathbf B_{\text{T}}\in\mathbb{R}^{m \times n\times c}$ is the rain free content of the texture component and the goal turns to estimate the clean texture part and separate the rain streaks from the rainy texture component.
In this work, considering the benefits of processing on the texture component, we attempt to design and train a CNN derainer $\mathcal{F}_\text{D}(\cdot; \Theta_\text{D})$, which maps the texture $\mathbf O_{\text{T}}$ patch into the rain streaks patch  $\mathbf R = \mathbf K(\theta,l) \otimes \mathbf {M}$.

Modeling the rain streaks with the motion blur kernel $\mathbf K$ maintains two advantages.
One is that two important factors, i.e., the length $l$ and the angle $\theta$, of the rain streak appearance are uniformly encoded by the motion blur kernel.
Another one is that the repeatability of the rain streaks allows us to easily infer the two parameters from an input texture patch.
In the next subsection, we would present the detail of how to estimate the parameters and embed the learned motion blur kernel to the deraining procedure.

\subsection{The Parameter Sub-Network}

Since the CNN has shown its overwhelming superiority on feature extraction, we plan to use a CNN to learn the motion blur kernel.
Initially, given a CNN $\mathcal{F}_\text{K}(\cdot;\Theta_\text{K}): \mathbb{R}^{m\times n} \rightarrow\mathbb{R}^{p\times p}$, which maps the input texture patch to the motion blur kernel, with network parameter $\Theta_\text{K}$, the loss function for training this CNN architecture is
\begin{equation}
L_\text{K}(\Theta_\text{K})= \frac{1}{n} \sum\limits_{i = 1}^{n} \|\mathcal{F}_\text{K}(\mathbf O_\text{T}^i;\Theta_\text{K}) - \mathbf K^i \|_F^2,
\end{equation}
where $\|\cdot\|_F$ denotes the Frobenius norm and $i$ index the patches and motion blur kernels.

However, the performance of $\mathcal{F}_\text{K}(\cdot)$ is not satisfactory.
Without the fully connect layer, it dose not converge.
As we pointed out above, the motion blur kernel within the generation of rain streaks is conclusively decided by two parameters, i.e., the angle $\theta$ and length $l$.
This indicates that the intrinsic information lies in a parameter space with much lower dimension than the convolution filter space.
Working directly on the low dimension information can not only facilitate the task of motion blur kernel estimation, but also prevent possibly overfitting, which would be verified in the experimental part.
Therefore, we adopt the CNN $\mathcal{F}_\text{P}(\cdot;\Theta_\text{P}): \mathbb{R}^{m\times n} \rightarrow\mathbb{R}^{2}$, which maps the input texture patch to
the parameter vector, with network parameter $\Theta_\text{P}$ and the loss function thereof for training turns to:
\begin{equation}
L_\text{P}(\Theta_\text{P}) = \frac{1}{n} \sum\limits_{i = 1}^{n} \|\mathcal{F}_\text{P}(\mathbf O_\text{T}^i;\Theta_\text{P})) - \mathbf p^i \|_F^2,
\end{equation}
where $\mathbf p = [\theta,l]^\top$ is the parameter vector.
The architecture of $\mathcal{F}_\text{P}(\cdot)$ (denoted as ``parameter net'') is exhibited in Fig. \ref{architecture}.
Once the parameters $\theta$ and $l$ are determined, the motion blur kernel $\mathbf K$ is unique.

\subsection{Dimensionality Stretching}
After maintaining the motion blur kernel $\mathbf K$ , the question comes to how to utilize the motion blur kernel when deraining.
In general, the input of the derain net $\mathcal{F}_\text{D}(\cdot; \Theta_\text{D})$, which would be detailed in the next subsection, is supposed to be the texture patch together with the motion blur kernel learned from this texture patch, since the motion blur kernel consists of  the prior knowledge of the rain streaks.
If we simply splice the texture patch $\mathbf O_\text{T}\in\mathbb{R}^{m\times n}$ and motion blur kernel $\mathbf K \in\mathbb{R}^{p\times p}$, weight sharing in CNN makes that the texture patch could not get the entire information of the motion blur kernel.
Hence, a dimensionality stretching operation in \cite{zhang2018learning} is necessary.

The dimensionality stretching strategy is schematically illustrated in Fig. \ref{architecture}.
At the beginning, the motion blur kernel $\mathbf K$ is vectorized into a vector $\mathbf k\in\mathbb{R}^{p^2}$.
After the vectorization, $\mathbf{k}$ is projected onto a $t$-dimensional linear space by the principal component analysis (PCA) technique.
Then the projected vector $\mathbf{k}_t\in\mathbb{R}^{t}$ is stretched into degradation maps $\mathcal{M}\in\mathbb{R}^{m \times n \times t}$.
All values in the $j$-th horizontal slice with size $m \times n$ of the 3-dimensional $\mathcal{M}$ are same as  the $j$-th element of $\mathbf{k}_t$.
By doing so, the degradation maps then can be concatenated with the texture patch, making CNN possible to handle the two inputs.

However, different from \cite{zhang2018learning}, in the case of plain motion blur kernel with only two parameters, we found that $t$ is still too large compared with number of channels of the texture image.
Meanwhile, since that working on texture component leads to the pixel values being close to zero, the information of the texture image may be drowned in the information of motion blur kernel with a relatively large $t$.
To tackle this issue, degradation maps will be concatenated with the texture image after the first convolutional layer in the derain net, as shown in Fig. \ref{architecture}.

\subsection{Derain Net}
As previously mentioned in Sec \ref{intro}, instead of elaborately designing the architecture, we resort to the typical ResNet structure. 
A cascade of $3\times 3$ convolutional layers are applied to perform the deraining.
Each layer is composed of three types of operations, including convolution (denoted as ``Conv''), rectified linear units \cite{krizhevsky2012imagenet} (denoted as ``ReLU''), and batch normalization \cite{ioffe2015batch} (denoted as ``BN'').
We still use Frobenius norm in the loss function, which is
\begin{equation}
L_\text{D}(\Theta_\text{D}) = \frac{1}{n} \sum\limits_{i = 1}^{n} \|f_d(\mathbf O_{texture}^i, \mathbf K^i - \mathbf R^i \|_F^2,
\end{equation}
where $\mathbf R$ is rain streaks.
After subtracting the rain streaks $\mathbf R$ from the rainy image $\mathbf O$, we could get the background.

\textbf{Disscusion:}
As we mentioned above,  distinguishing the rain streaks and the line pattern textures is important but challenging.
In this work, we face this challenge by exploiting the generation mechanism of the rain streaks to guide the rain streak removal.
Within our framework, the generation mechanism of the rain streaks is taken into consideration, and the prior knowledge of the rain streaks, i.e., the angle and the length of the motion blur kernel, are automatically learned.
The embedding of the motion blur kernel into the derain net, which maintains a plain ResNet structure, greatly enhances the performance (see the comparisons in Sec. \ref{Sec:Exp:Kernel}).
To some extent, the utilization of the motion blur kernel in our method can be viewed as the traditional optimization model utilizing the regularizer to express the prior knowledge.

\section{Experimental results}\label{experimental_results}
To evaluate the performance of the proposed KGCNN framework, we test it on both synthetic and real-world rainy images.
The networks are trained on synthesized rainy images.
We compare our KGCNN with six state-of-the-art methods, including three traditional methods: the unidirectional global sparse model (UGSM\footnote{\url{http://www.escience.cn/people/dengliangjian/index.html}}) \cite{deng2018directional}, the discriminative sparse coding method (DSC\footnote{\url{http://www.math.nus.edu.sg/~matjh/research/ research.htm}}) \cite{luo2015removing}, and the method using layer prior (LP\footnote{\url{http://yu-li.github.io/}}) \cite{li2016rain}, as well as three deep learning based methods: the density-aware multi-stream deraining dense network  (DID\footnote{\url{https://github.com/hezhangsprinter/DID-MDN}}) \cite{zhang2018density}, a plain convolutional neural network deraining method (CNN\footnote{\url{https://xueyangfu.github.io/projects/tip2017.html}}) \cite{fu2017clearing}, and the deep detail network (DDN\footnote{\url{https://xueyangfu.github.io/projects/cvpr2017.html}}) \cite{fu2017removing}.

\subsection{Rainy Images Simulation}

With the observation model in Eq. \eqref{ob2}, the synthetic rainy images are generated by the following steps.
(1) Transform the background from RGB color space to YUV color space\footnote{\url{https://en.wikipedia.org/wiki/YUV}}.
(2) Generate the raindrops mask $\mathbf M$ by adding salt and pepper noise with signal-noise ratio from 0.9 to 1.0  to a zero matrix with the same size as the Y channel of the background, and adding a Gaussian blur with standard variance from 0.2 to 0.5.
(3) Generate the motion blur kernel $\mathbf K$ with angle $\theta$ sampled from $[45^\circ,135^\circ]$ and length $l$ varing from 15 to 30.
(5) Directly added the generated rain streaks $\mathbf R = \mathbf K\otimes\mathbf M$ to the background on Y channel, and the intensity values greater than 1 are set as 1.
(6) Finally, transform the image back to RGB color space.

\subsection{Experiments Setting}
For fair comparisons, we use the default parameters in the codes for traditional methods and the default trained models for the deep learning methods.
Since existing rainy datasets do not consist of the information of the motion blur kernel, we train our networks only on our synthetic data.
The patch size is set as $64\times 64 \times 3$.
Guided filter with radius 15 and regularization 1 is selected to decompose the rain images.
By preserving $99\%$ of the energy, the kernel is projected onto a space of dimension 162.
Because of the full connection, the input image should be split into several patches for experiments.
We use Adam \cite{kingma2014adam} optimizers with learning rate 0.01.
Our model is trained and tested on Python 3.5.2 with TensorFlow 1.0.1 framework on a desktop of GPU NVIDIA GeForce GTX 1060 with 6GB.
For other compared methods based in Matlab, they are running on Matlab 2017A.

\begin{figure*}[htb]
\centering
\small\renewcommand\arraystretch{1}
\setlength{\tabcolsep}{2pt}
\begin{tabular}{ccccc}
     The background &Rainy image & DID  \cite{zhang2018density} & DSC \cite{luo2015removing}& LP \cite{li2016rain} \\

                \includegraphics[width=0.12\linewidth]{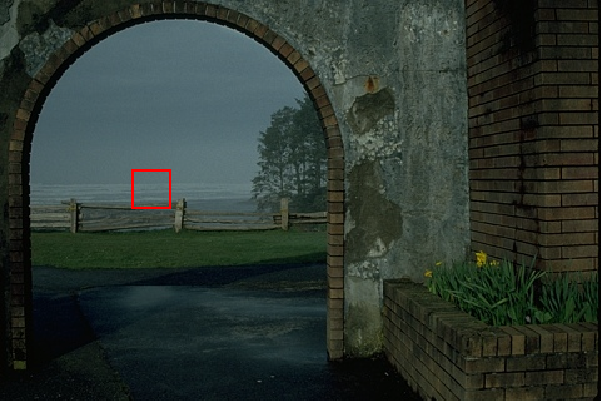}\hspace{0.4mm}\includegraphics[width=0.06\linewidth]{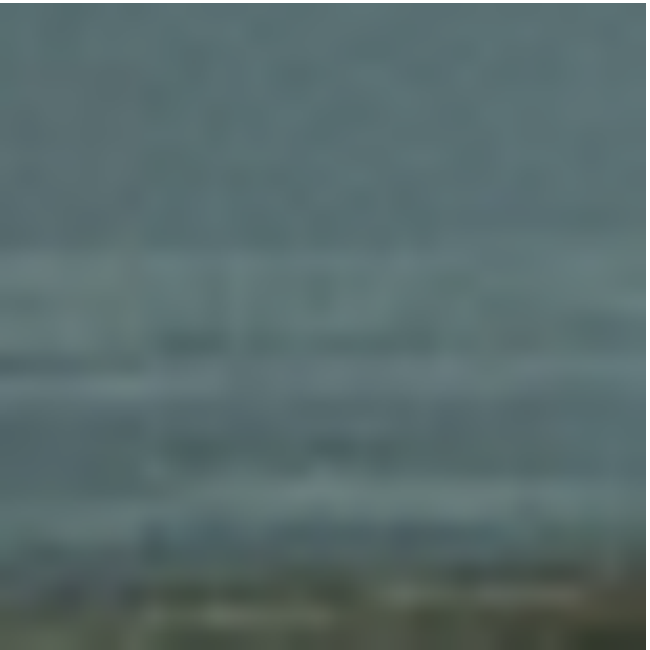}&
                \includegraphics[width=0.12\linewidth]{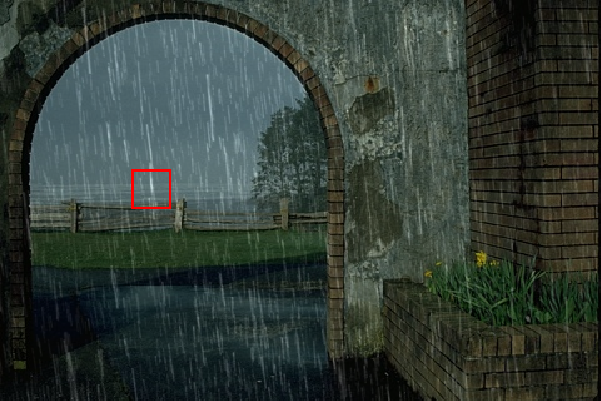}\hspace{0.4mm}\includegraphics[width=0.06\linewidth]{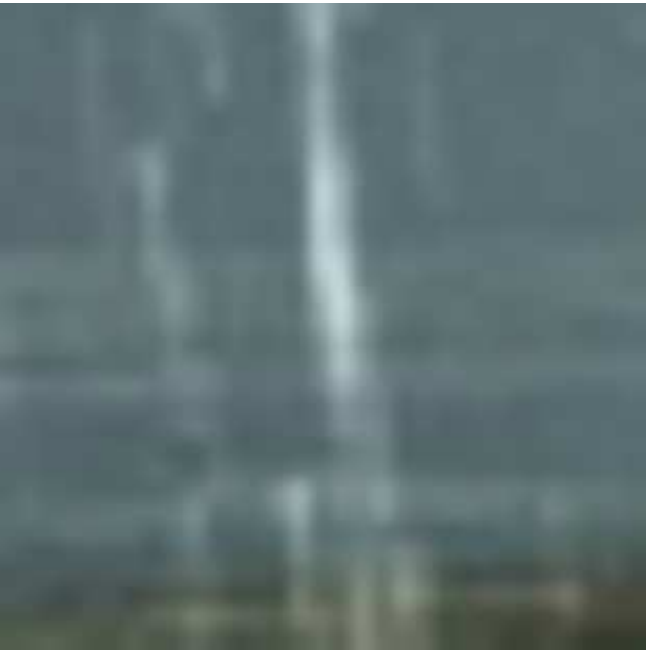}&
                \includegraphics[width=0.12\linewidth]{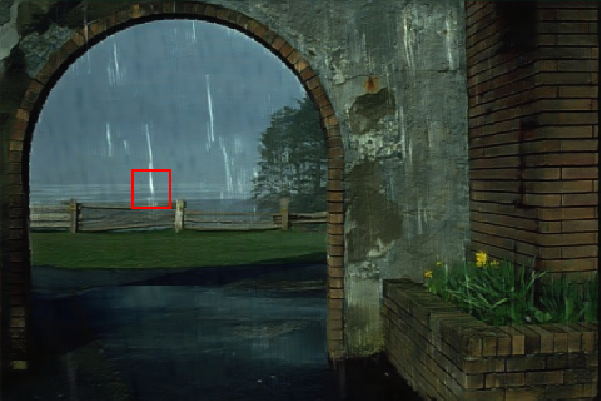}\hspace{0.4mm}\includegraphics[width=0.06\linewidth]{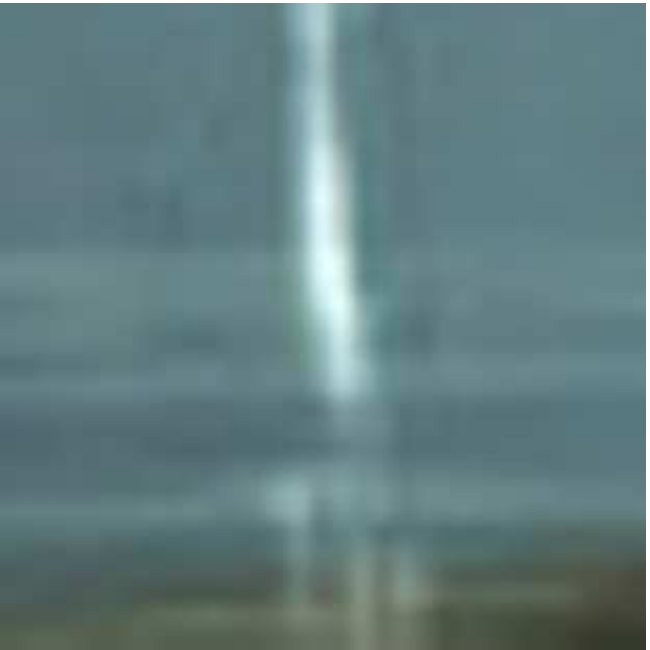}&
                \includegraphics[width=0.12\linewidth]{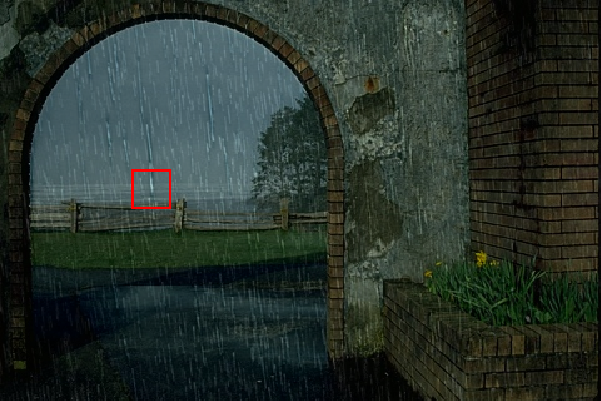}\hspace{0.4mm}\includegraphics[width=0.06\linewidth]{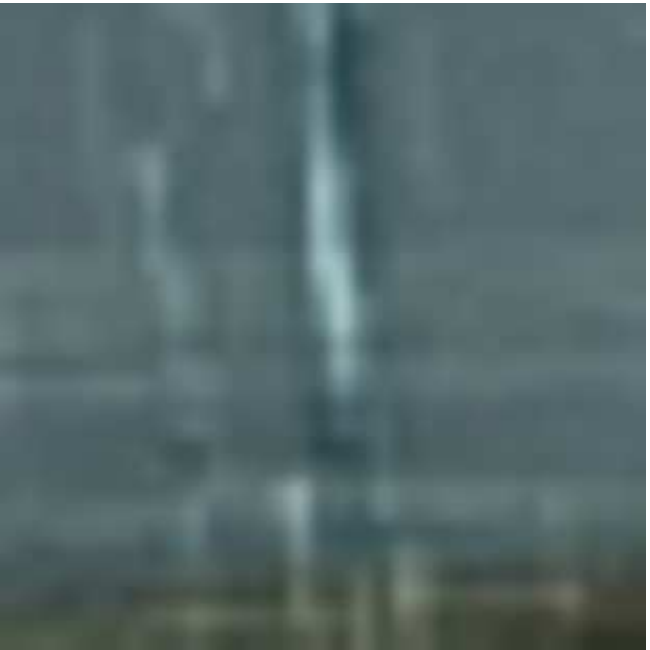}&
                 \includegraphics[width=0.12\linewidth]{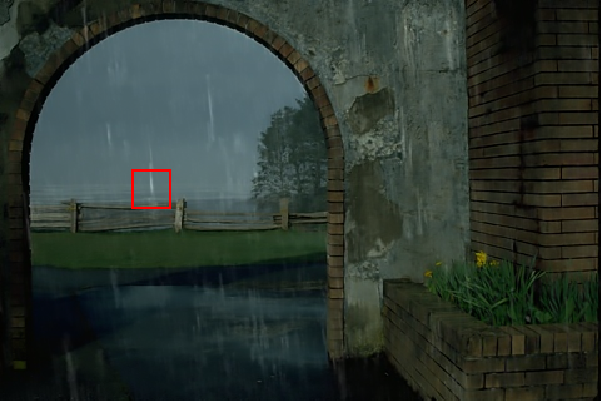}\hspace{0.4mm}\includegraphics[width=0.06\linewidth]{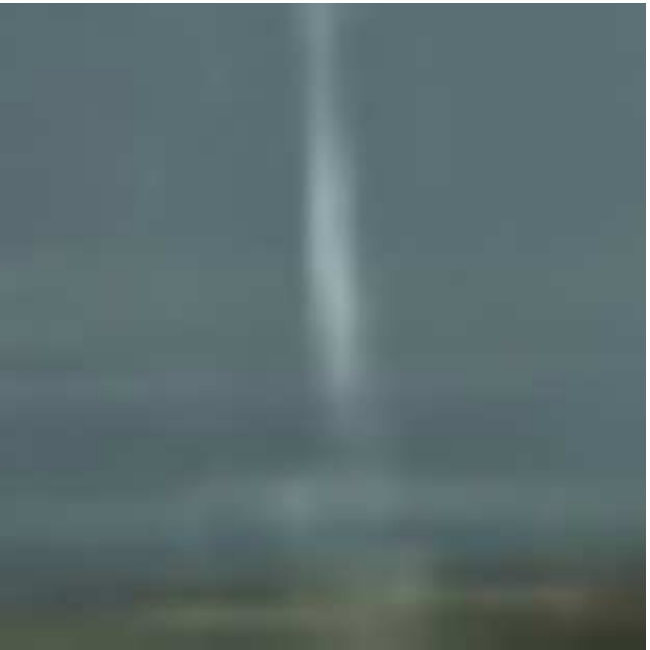}\\

                \includegraphics[width=0.12\linewidth]{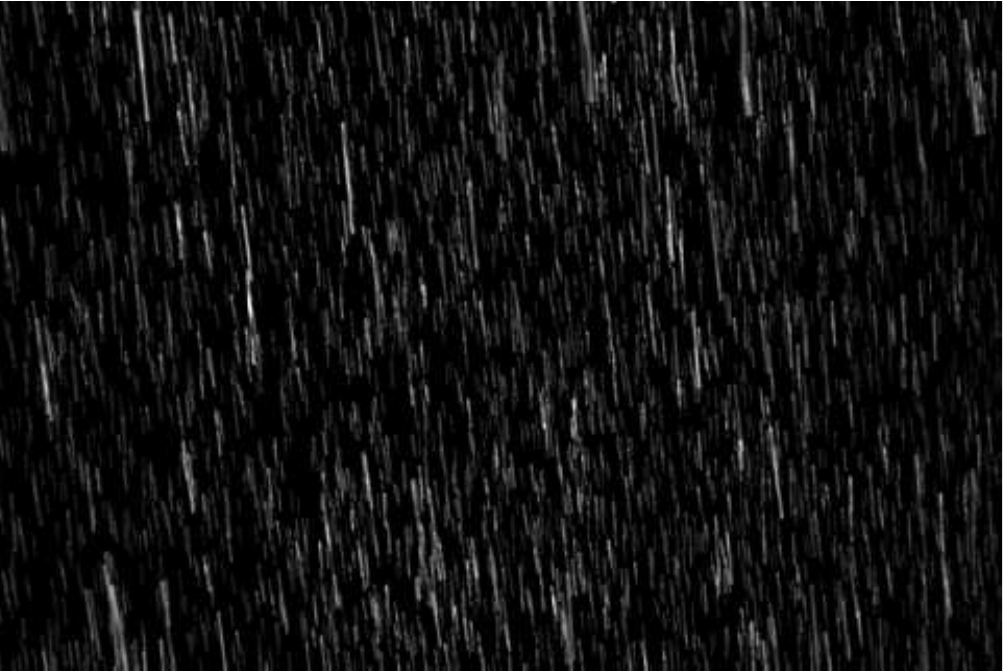}&
                \includegraphics[width=0.12\linewidth]{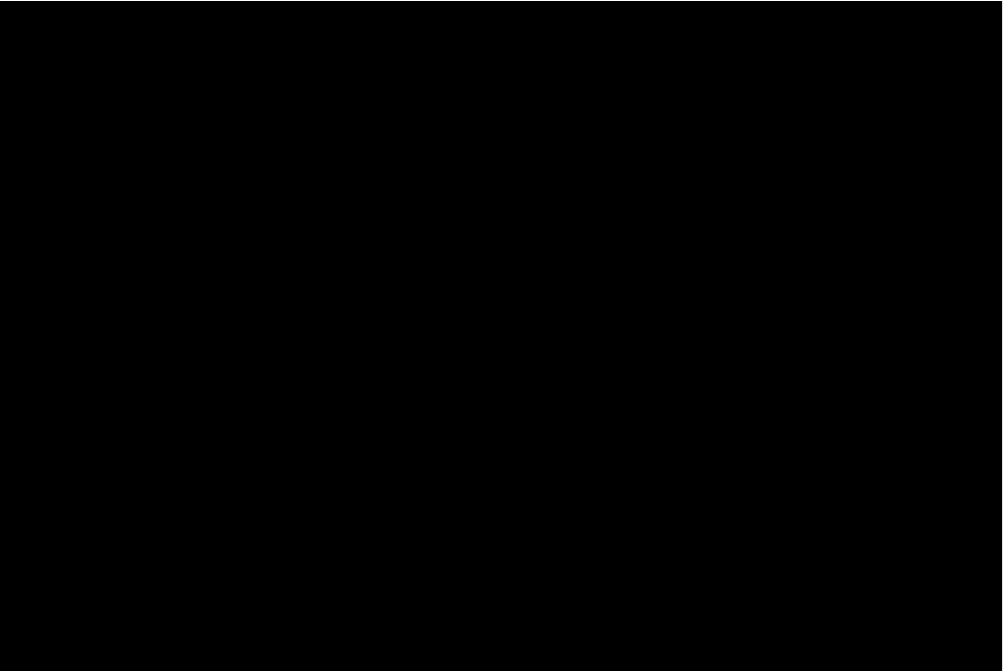}&
                \includegraphics[width=0.12\linewidth]{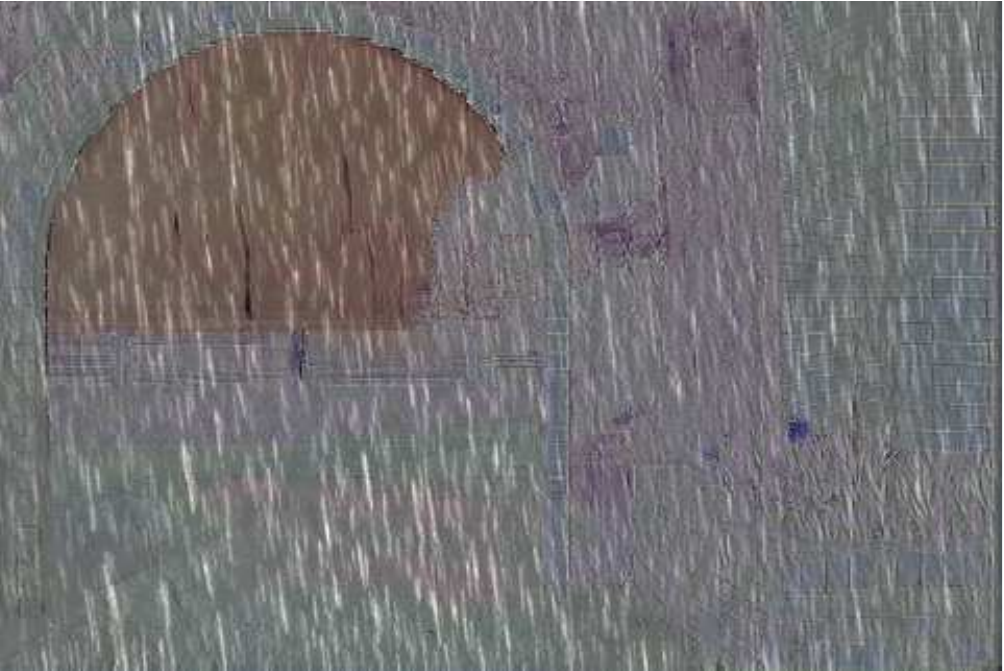}&
                \includegraphics[width=0.12\linewidth]{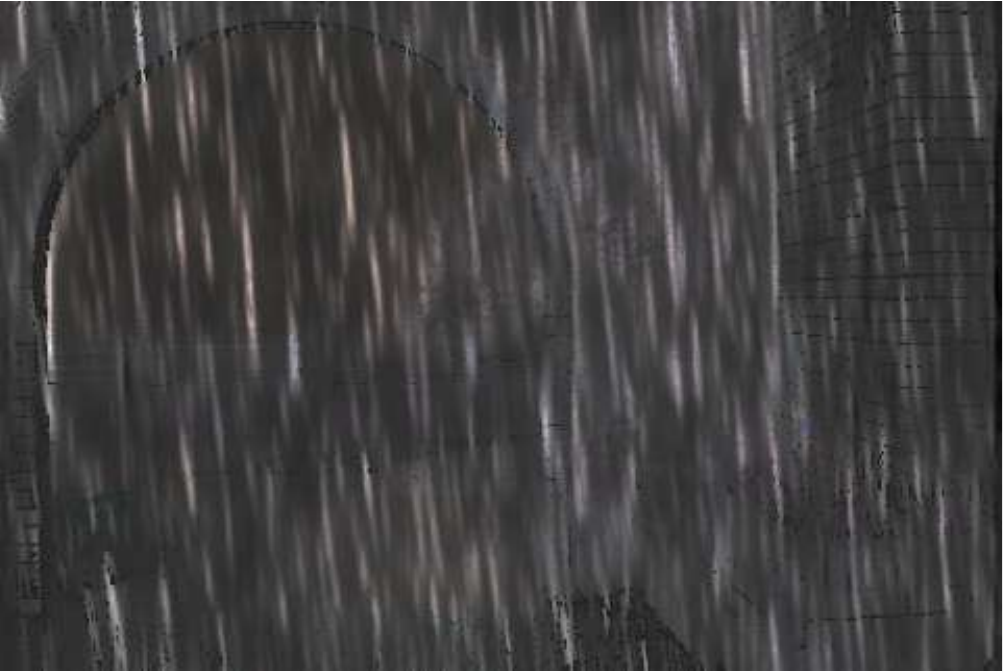}&
                \includegraphics[width=0.12\linewidth]{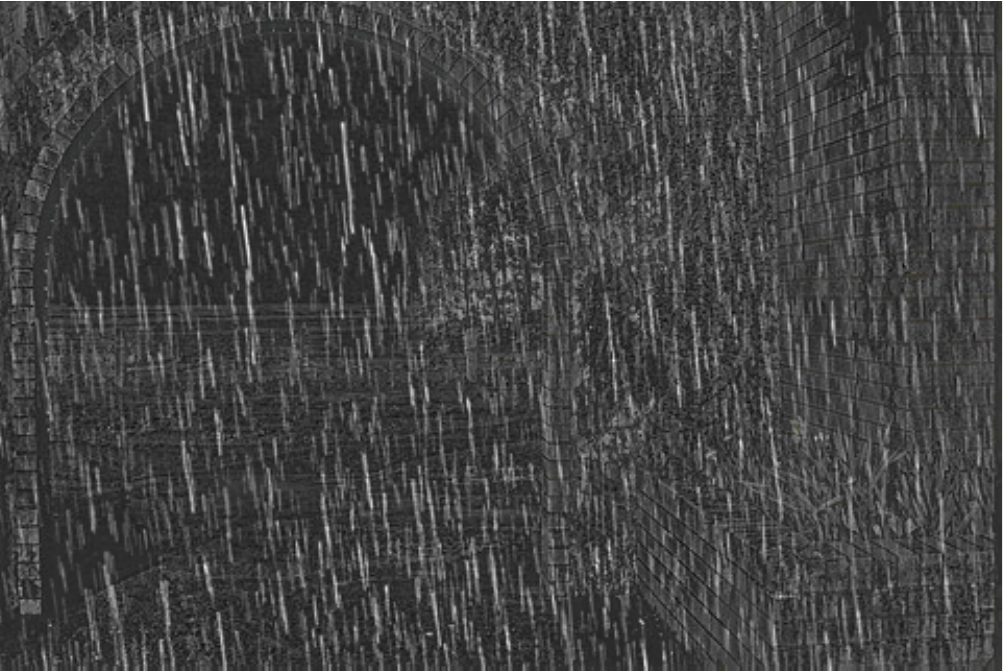}\\
                \end{tabular}
                \begin{tabular}{cccc}
                 UGSM \cite{deng2018directional} &CNN \cite{fu2017clearing} & DNN \cite{fu2017removing}  & Proposed KGCNN\\

                \includegraphics[width=0.12\linewidth]{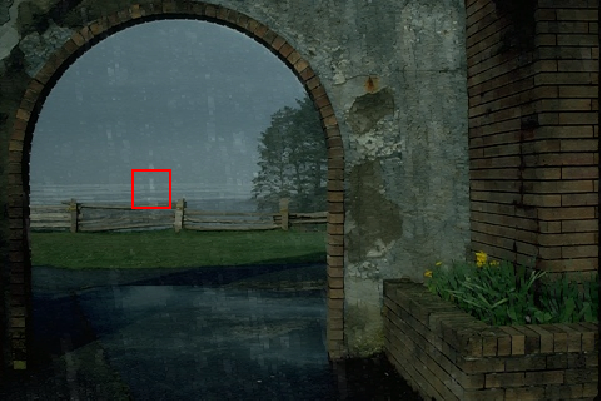}\hspace{0.4mm}\includegraphics[width=0.06\linewidth]{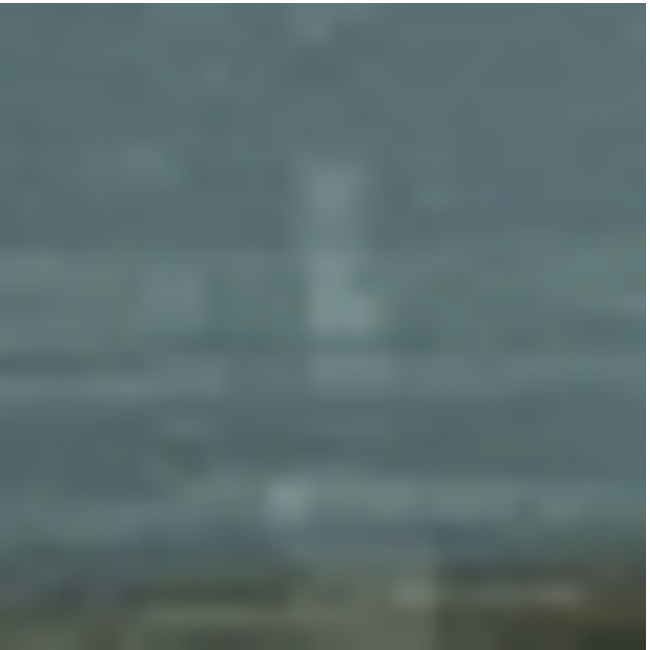}&
                \includegraphics[width=0.12\linewidth]{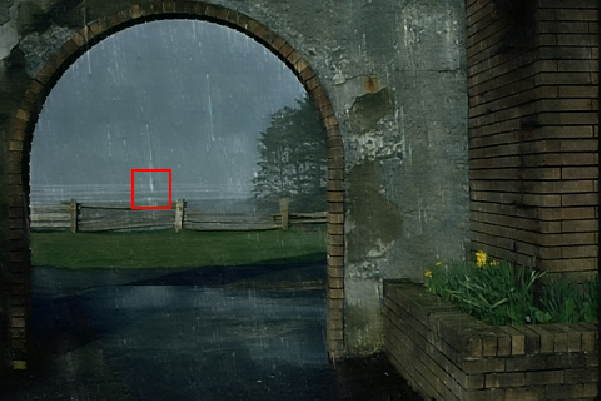}\hspace{0.4mm}\includegraphics[width=0.06\linewidth]{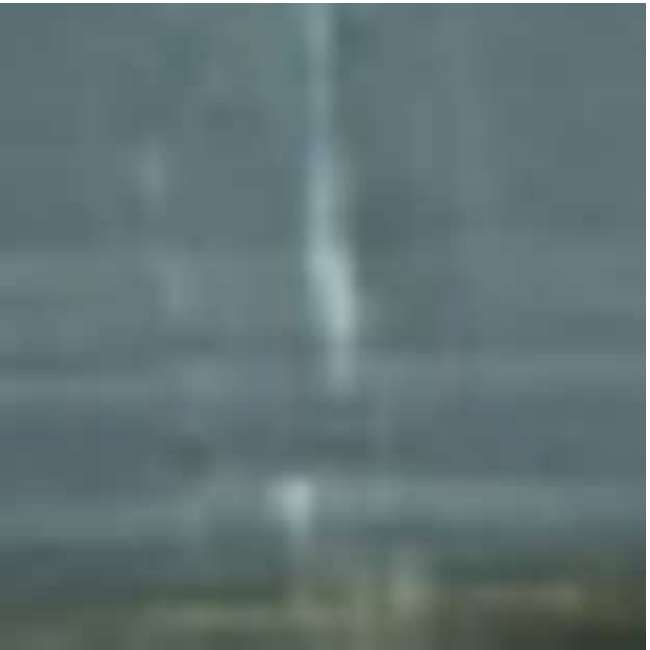}&
                \includegraphics[width=0.12\linewidth]{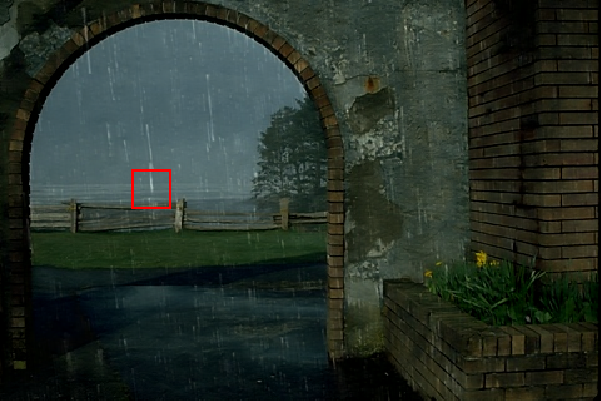}\hspace{0.4mm}\includegraphics[width=0.06\linewidth]{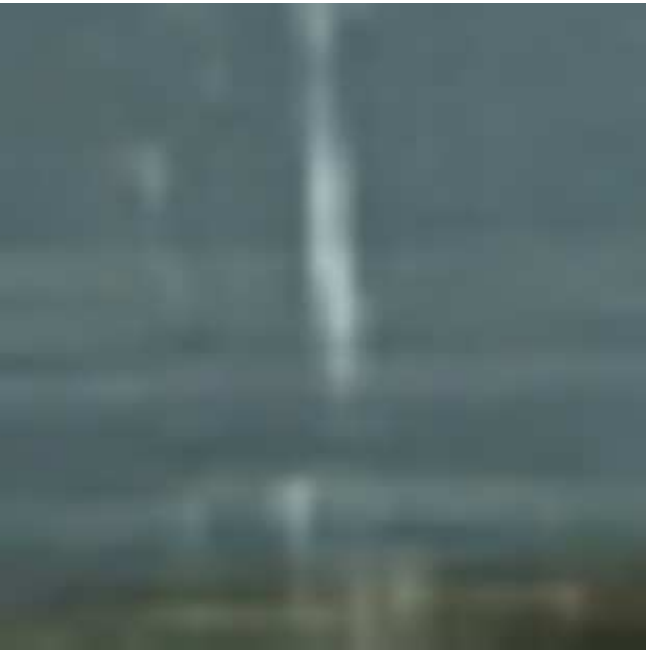}&
                \includegraphics[width=0.12\linewidth]{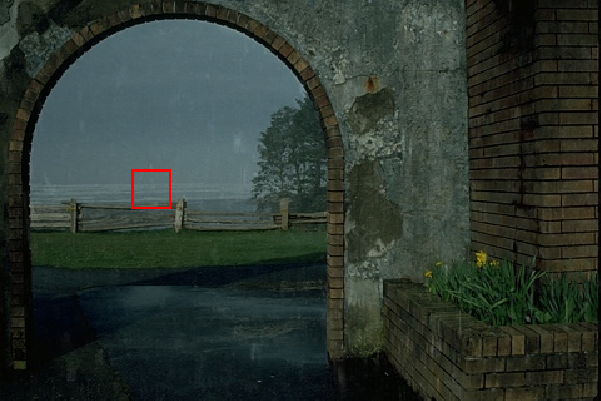}\hspace{0.4mm}\includegraphics[width=0.06\linewidth]{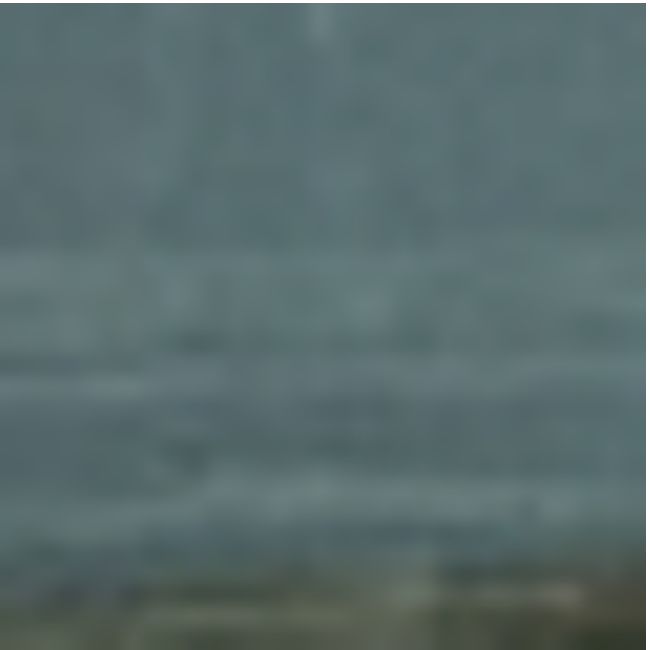}\\

                \includegraphics[width=0.12\linewidth]{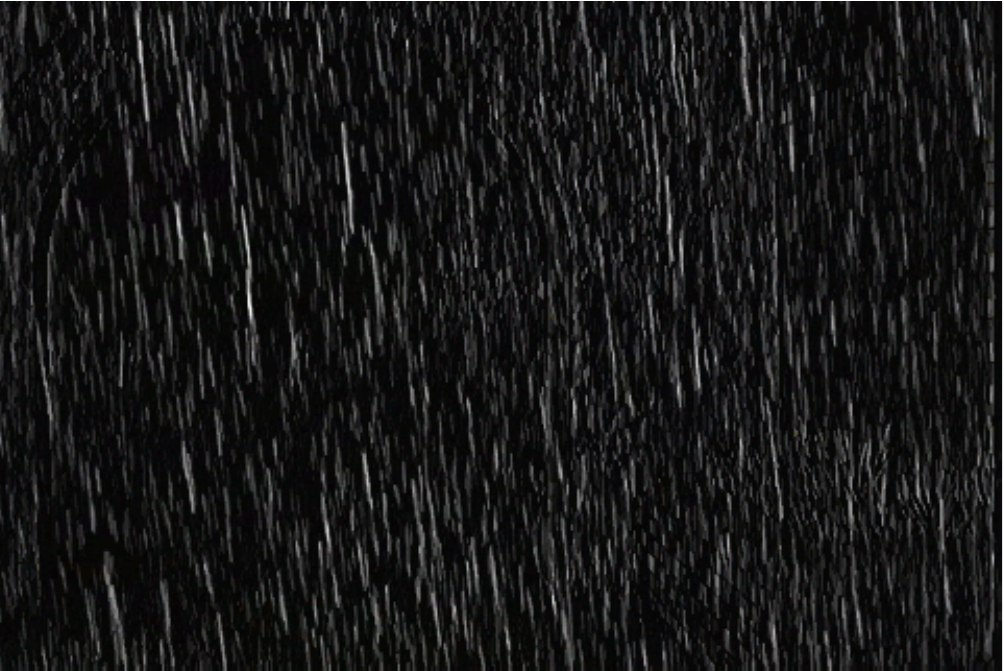}&
                \includegraphics[width=0.12\linewidth]{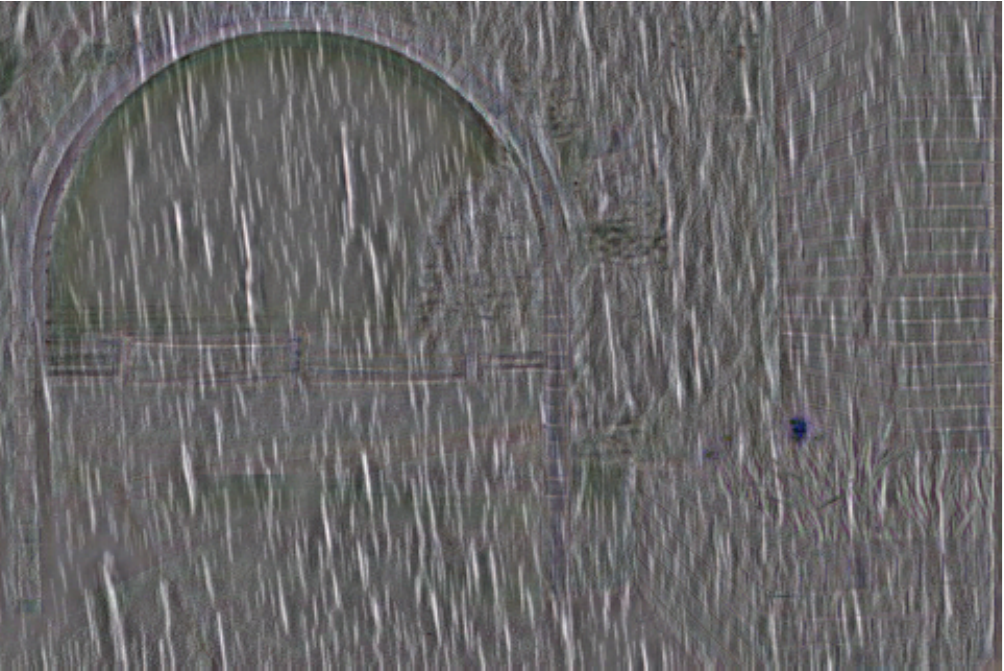}&
                \includegraphics[width=0.12\linewidth]{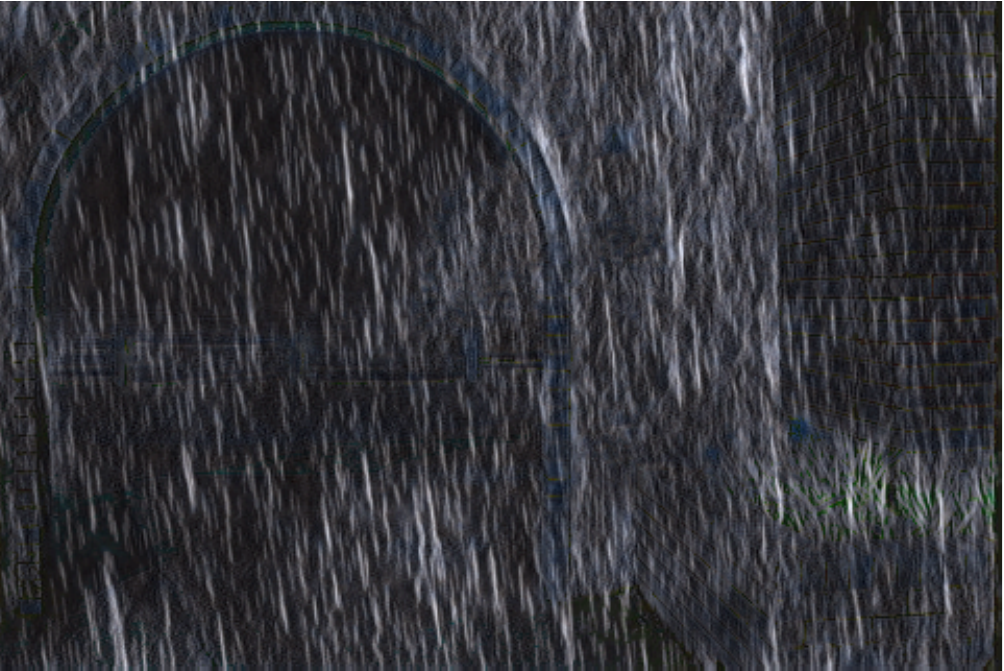}&
                \includegraphics[width=0.12\linewidth]{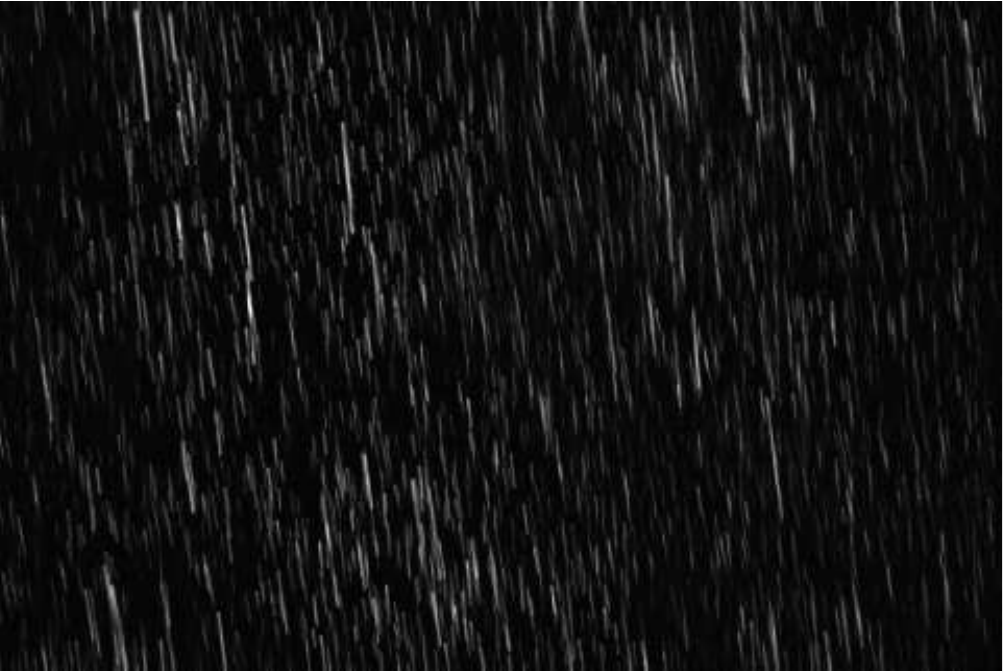}\\

\end{tabular}
\caption{Rain streak removal results by different methods on one rainy image of the Rain12 dataset.}
\label{synthetic-visual-R12-kuang}
\end{figure*}

\subsection{Synthesized Data}
In this subsection, we evaluate performance of different state-of-the-art methods on the synthetic rainy images.
Three datasets are selected:
1) the benchmark dataset provided by Dr. Yu Li using the rain streaks rendering technique in \cite{garg2006photorealistic} (denoted as Rain12),
2) 3 synthetic rainy images by our simulating method, and
3) several synthetic rainy images in \cite{deng2018directional}.
Due to the limit of space, we only show partial results in this section, and please see more results in the supplementary materials.

For quantitative comparisons, we adopt the peak signal to noise ratio (PSNR), structure similarity index (SSIM) \cite{wang2004image}, feature similarity index (FSIM) \cite{zhang2011fsim}, universal image quality index (UIQI) \cite{wang2002universal}, and gradient magnitude similarity deviation (GMSD, smaller is better) \cite{xue2014gradient} as the quality metrics of the deraining results.
Particularly, since the compared methods are implemented with different programming languages (or platforms), e.g., UGSM with Matlab and CNN with Python, we save all output images of different methods as png format, then reload them in Matlab and compute the corresponding quantitative results on RGB color space.

To show that KGCNN could remove rain streaks while keeping the texture and the contrast of background, we show the rain streak images (residual images between rainy images and resulted images).

Normalization is performed to the rain streak images so that we could distinguish whether the proposed method changes texture and contrast significantly or removes rain streaks completely.
For instance, if the rain streak images are too bright, it indicates the method significantly changes intensity contrast.
For the first dataset, Fig. \ref{synthetic-visual-R12-kuang} shows the visual results, local close-up images and rain streak images on one synthetic rainy image of the Rain12 dataset.
We can see that the proposed KGCNN method could remove the rain streaks completely while other approaches fail to do so (see local close-up images for better comparisons in Fig. \ref{synthetic-visual-R12-kuang}).
Especially, it is easy to see that the obtained rain streaks by the proposed approach do not contain the structures of background, which indicates KGCNN has a very good ability for rain streak removal.
From the perspective of quantitative results, KGCNN method performs best for the 12 synthesized images, compared with other six state-of-the-art methods (see Table \ref{synthetic-quant-R12} for more details).

\begin{table}[htb]
\renewcommand\arraystretch{0.9}\setlength{\tabcolsep}{1.8pt}
\caption{Quantitative comparisons of rain streak removal results by DID \cite{zhang2018density}, DSC \cite{luo2015removing}, LP \cite{Li2014Single}, UGSM \cite{Deng2018A}, CNN \cite{fu2017clearing}, DDN \cite{fu2017removing}, and KGCNN on the Rain12 (average value).}
 \label{synthetic-quant-R12}
\begin{center}
\begin{tabular}{c|cccccc}
 \Xhline{1.2pt}
           Method     &PSNR    &SSIM     &FSIM      &UIQI     &GMSD  &Time (s)  \\
\Xhline{0.8pt}
                      rainy    &28.822    &0.910    &0.910    &0.968    &0.134    &-    \\
                      DID    &27.485    &0.919    &0.918    &0.941    &0.086    &\bf{0.624}     \\
                      DSC    &28.584    &0.915    &0.917    &0.965    &0.097    &153.792      \\
                      LP    &30.825    &0.947    &0.935    &0.966    &0.070    &321.328      \\
                      UGSM    &32.185    &0.958    &0.947    &\bf{0.983}    &0.065    &2.767      \\
                      CNN    &28.155    &0.942    &0.935    &0.966    &0.071    &7.432      \\
                      DNN    &29.112    &0.935    &0.942    &0.931    &0.073    &0.660      \\
                      KGCNN    &\bf{34.731}    &\bf{0.971}    &\bf{0.965}    &\bf{0.983}    &\bf{0.055}    &8.850      \\
\Xhline{1.2pt}
\end{tabular}
\end{center}
\end{table}

\begin{figure*}[!htb]
\renewcommand\arraystretch{0.8}\setlength{\tabcolsep}{1.8pt}
\begin{center}
\begin{tabular}{ccccccccc}
                ~\includegraphics[width=0.7in]{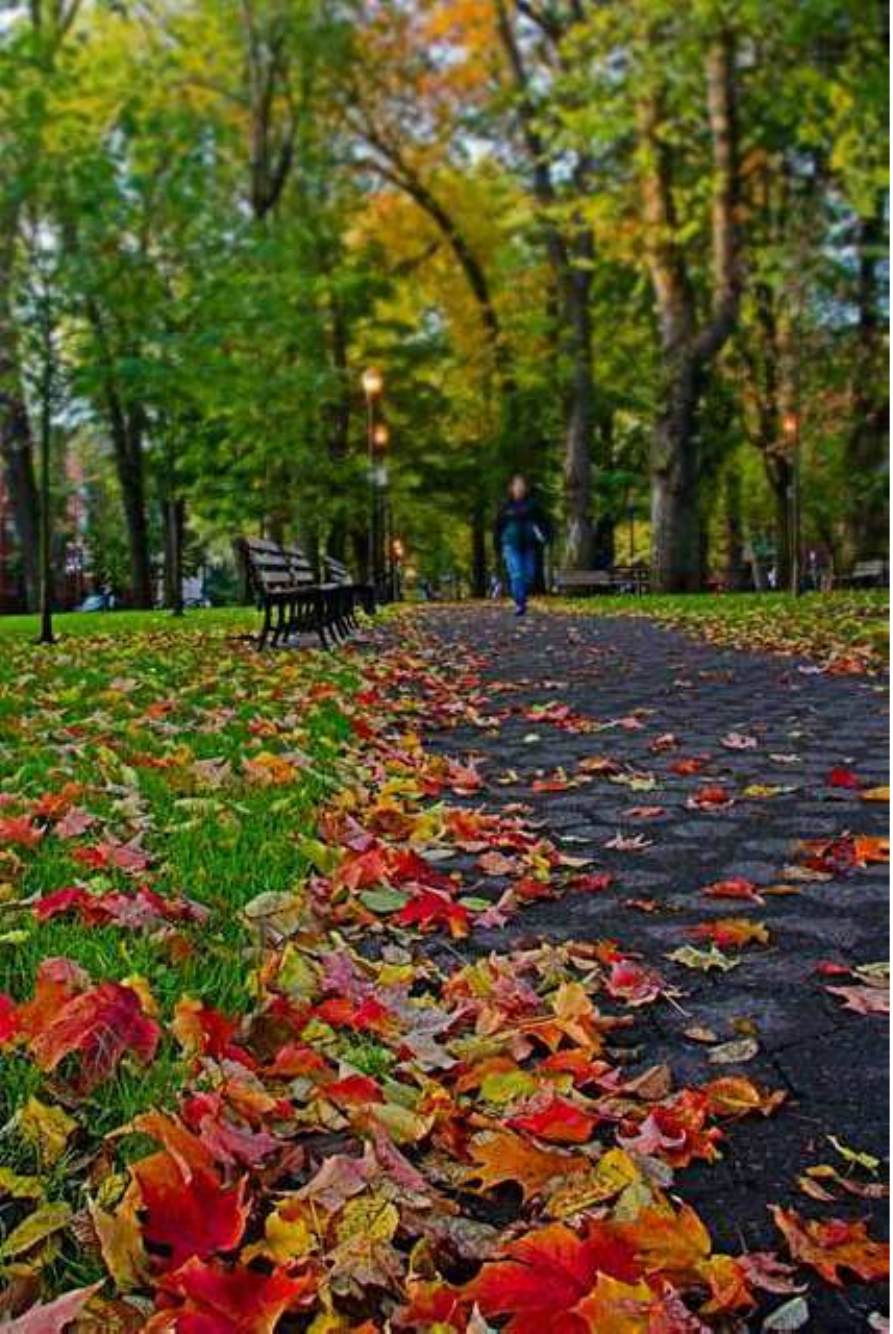}&
                \includegraphics[width=0.7in]{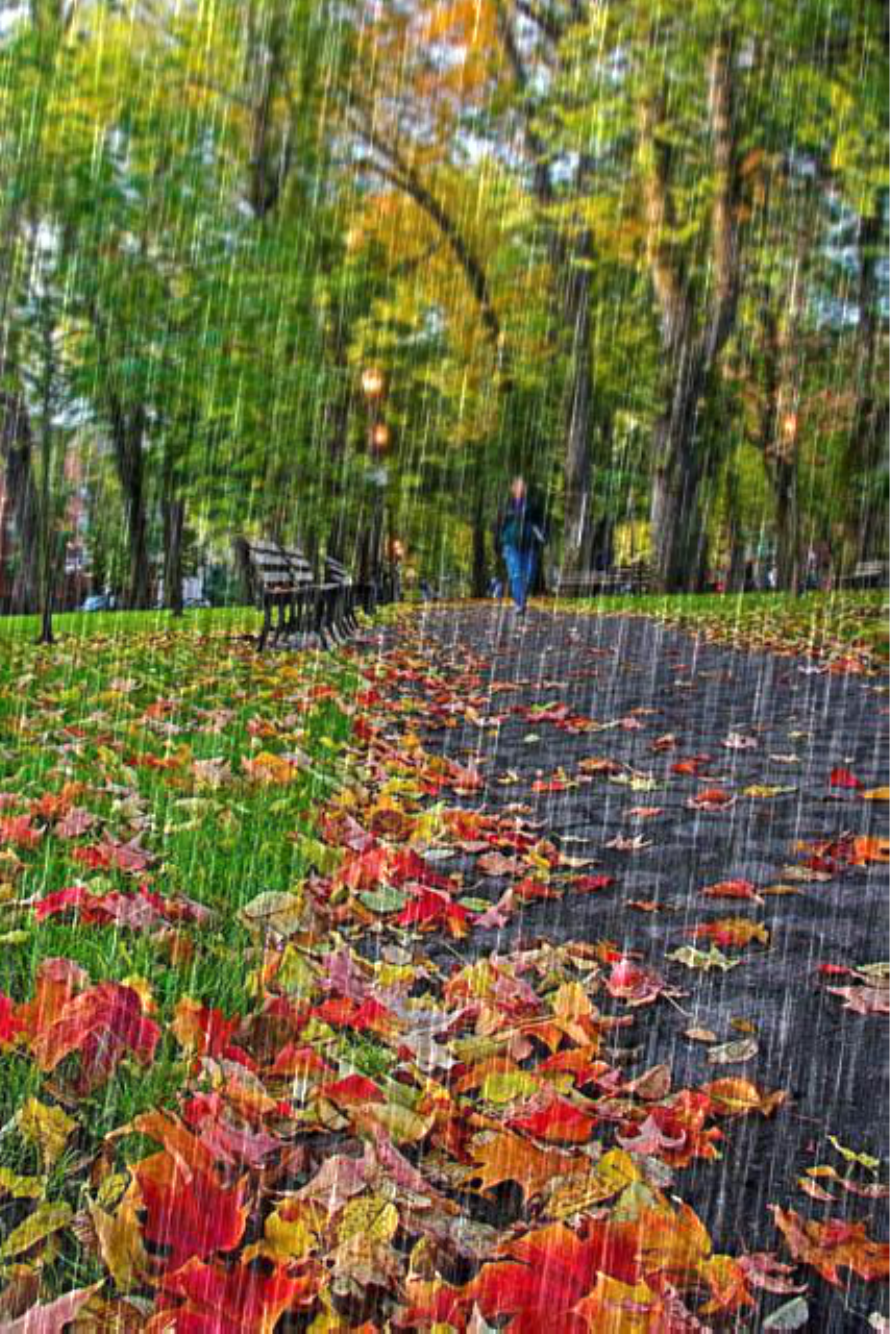}&
                \includegraphics[width=0.7in]{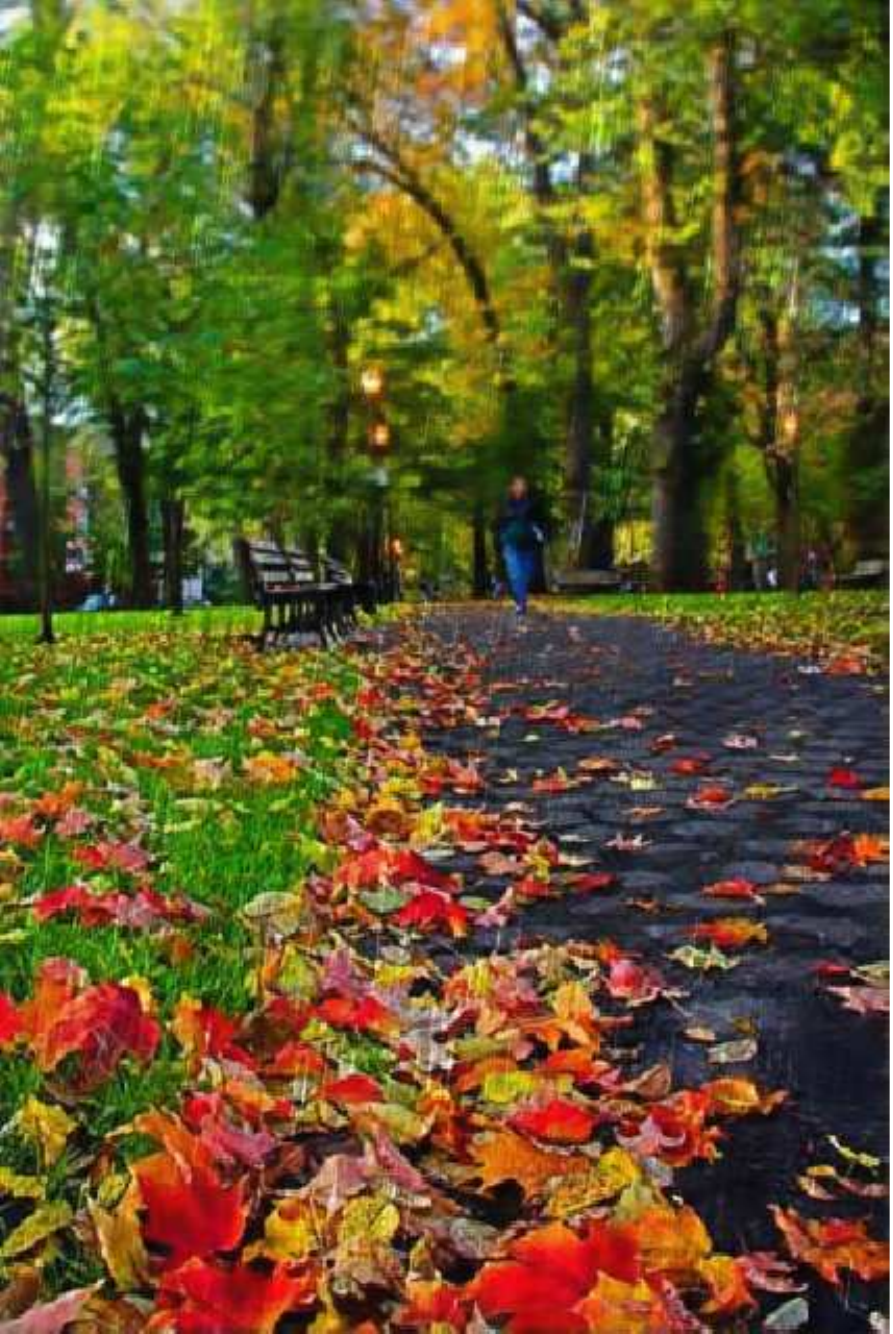}&
                \includegraphics[width=0.7in]{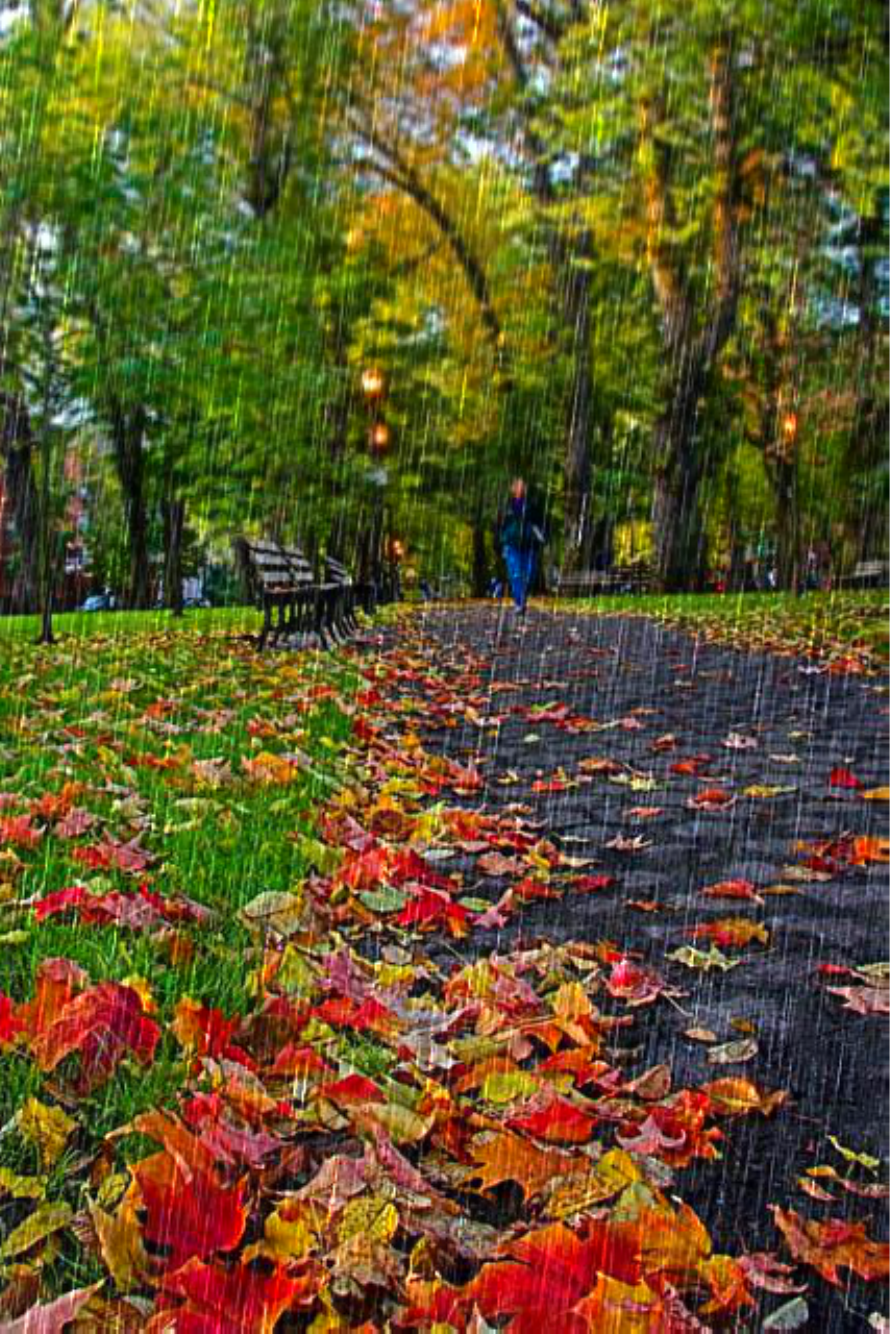}&
                \includegraphics[width=0.7in]{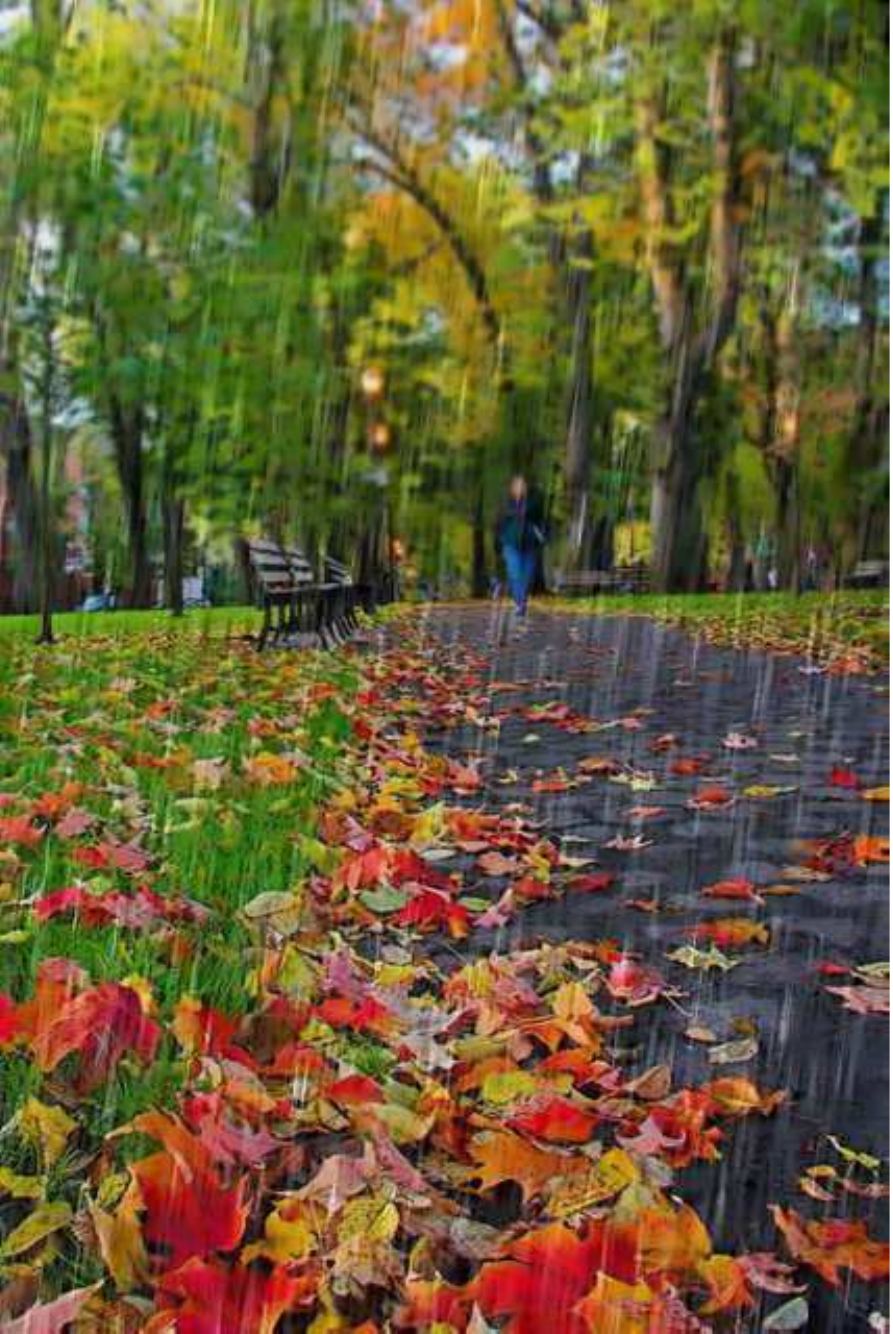}&
                \includegraphics[width=0.7in]{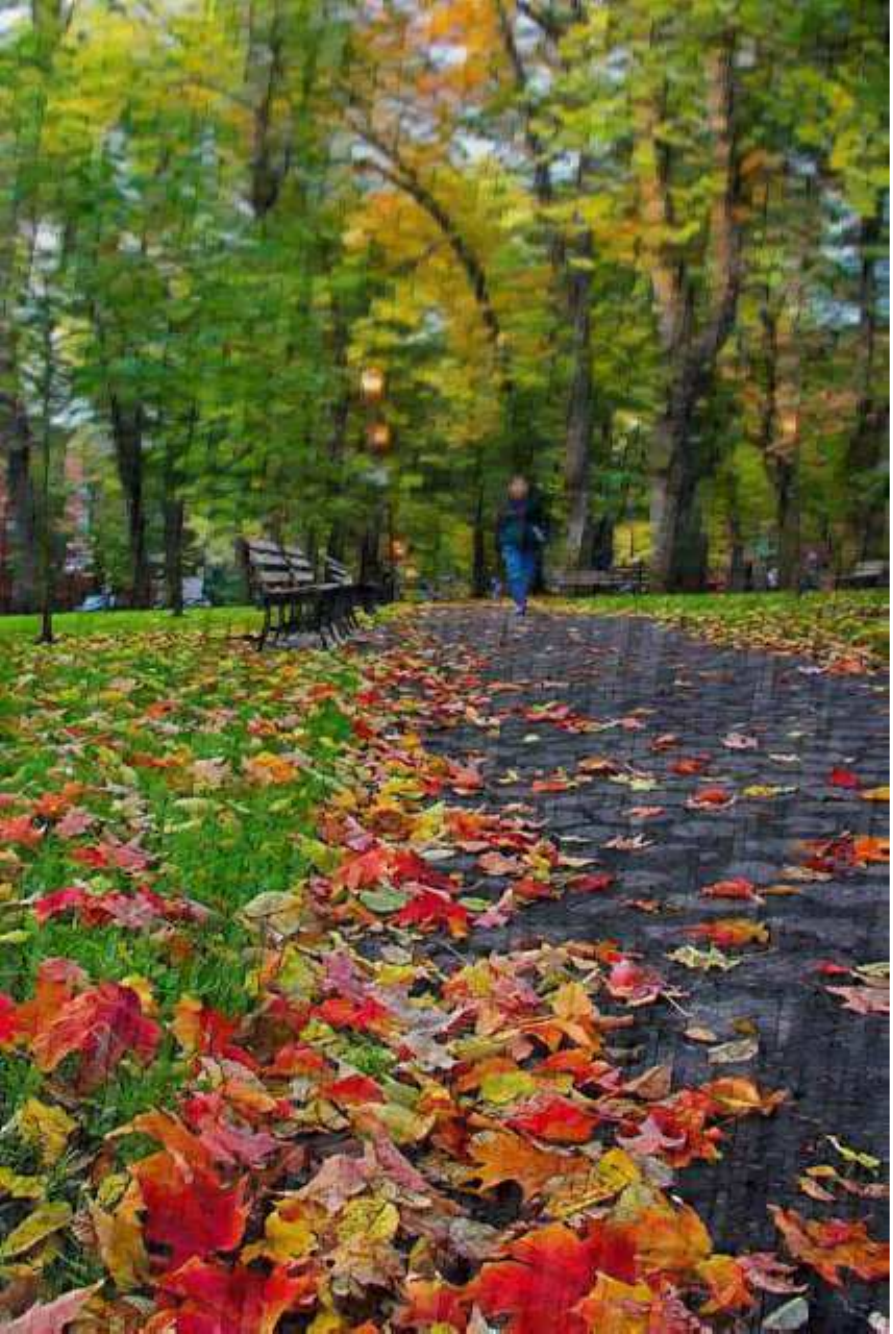}&
                \includegraphics[width=0.7in]{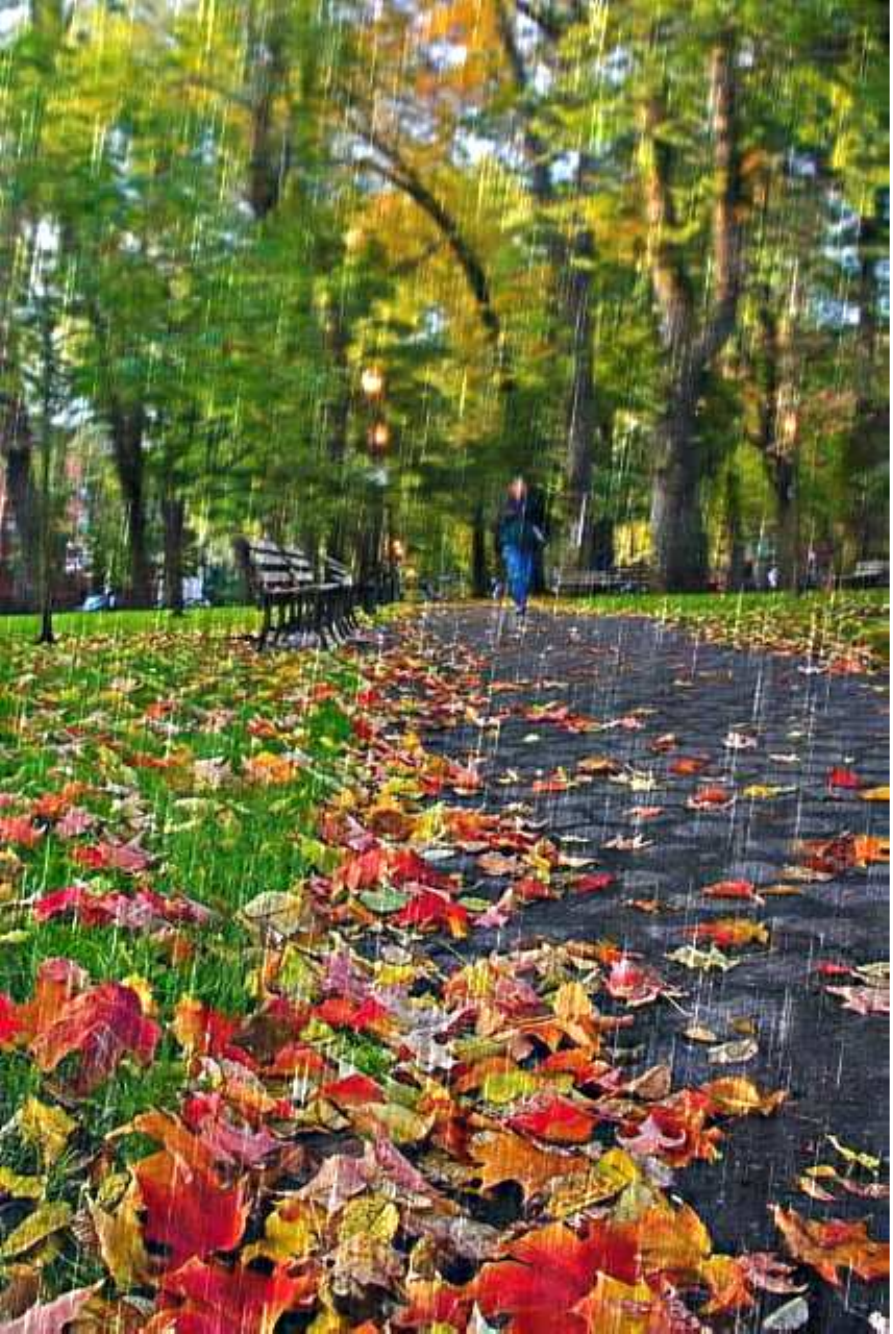}&
                \includegraphics[width=0.7in]{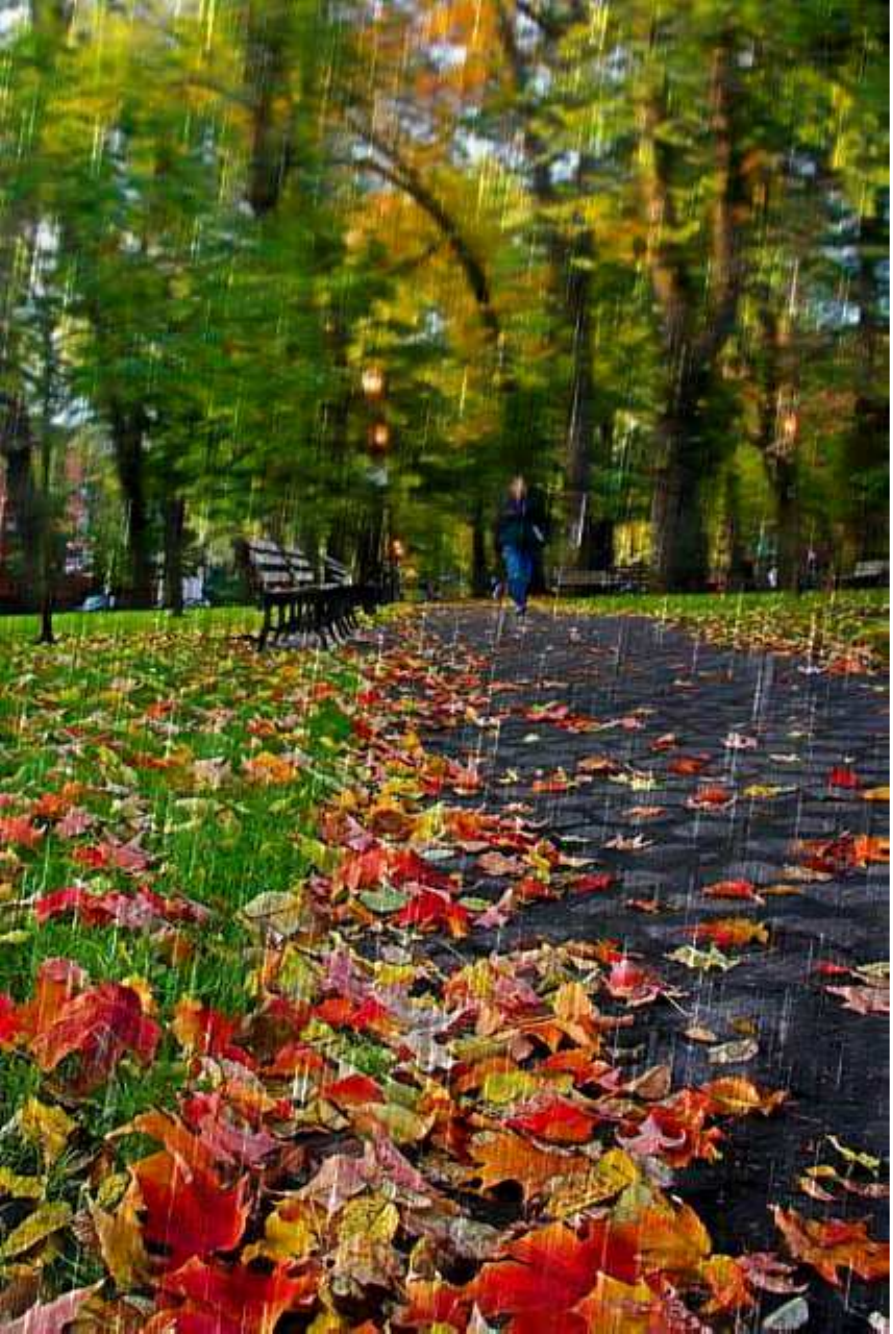}&
                \includegraphics[width=0.7in]{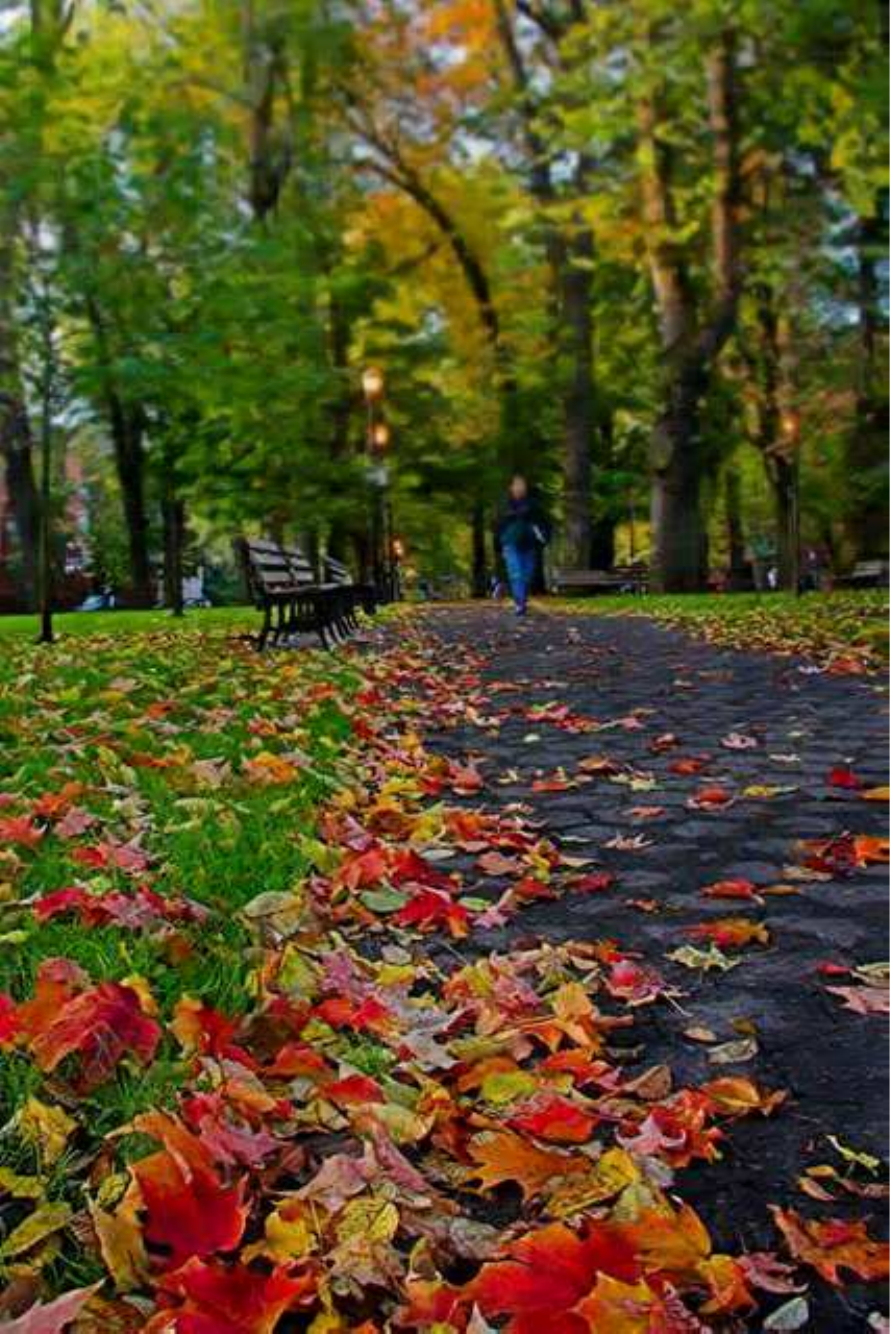}\\
                \vspace{0.5mm}

                \includegraphics[width=0.7in]{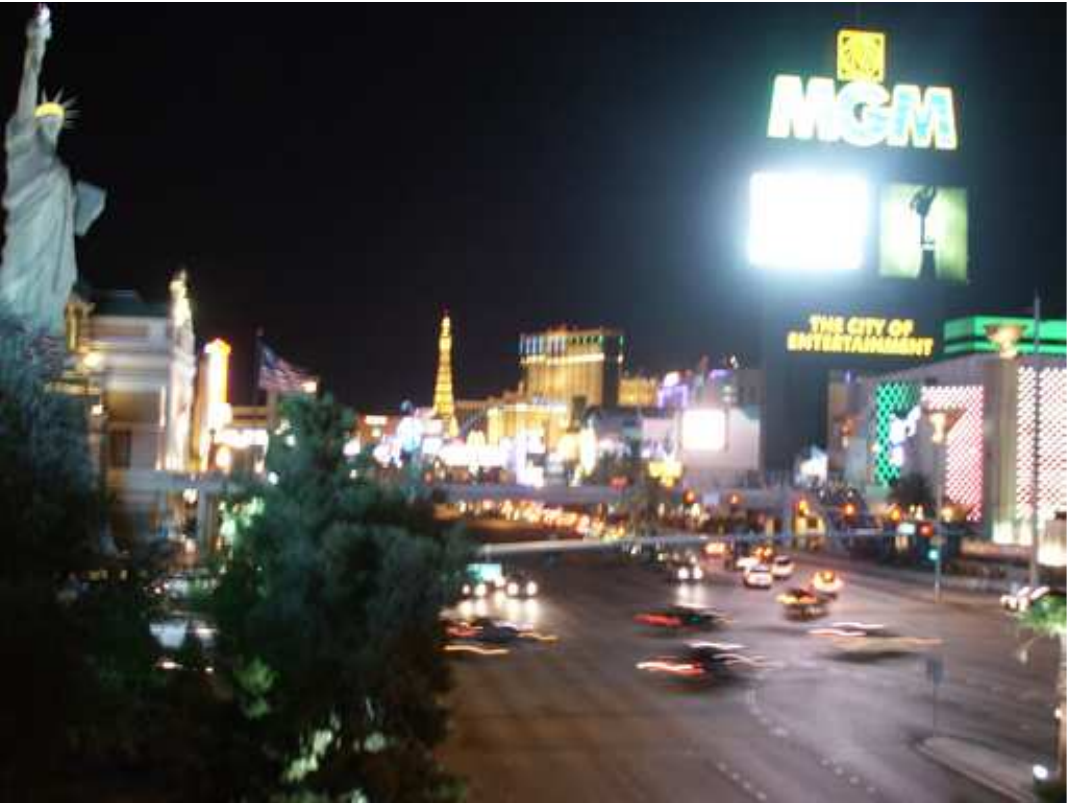}&
                \includegraphics[width=0.7in]{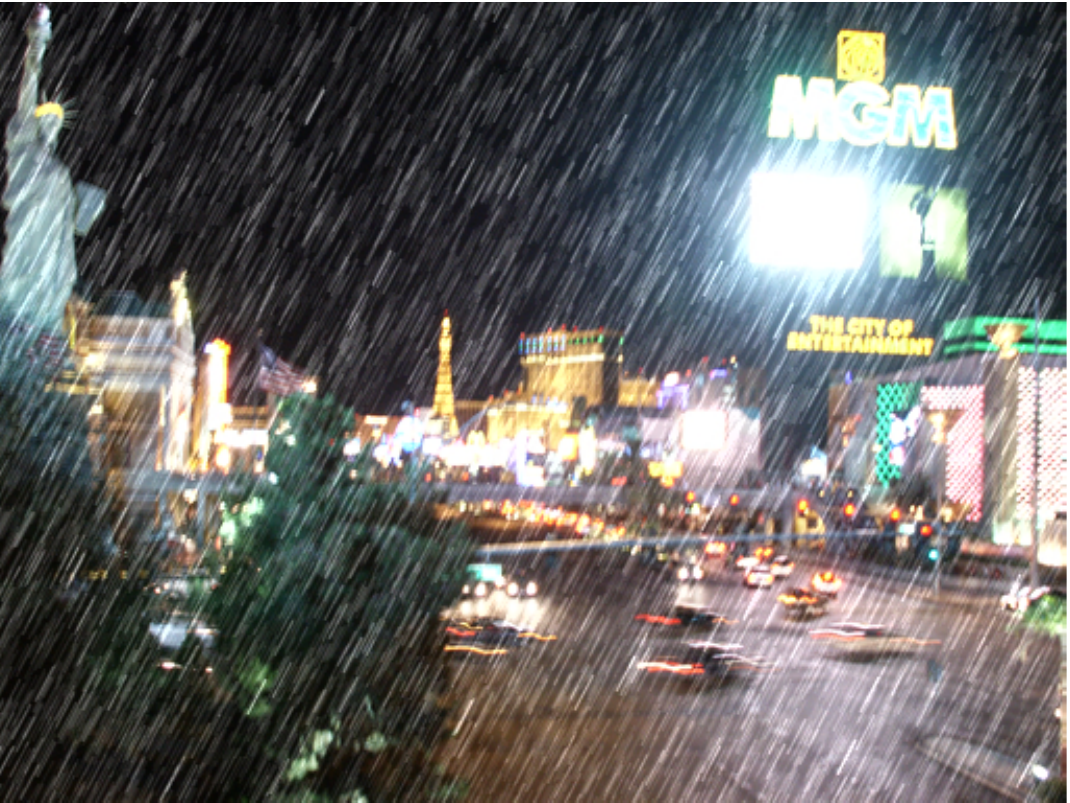}&
                \includegraphics[width=0.7in]{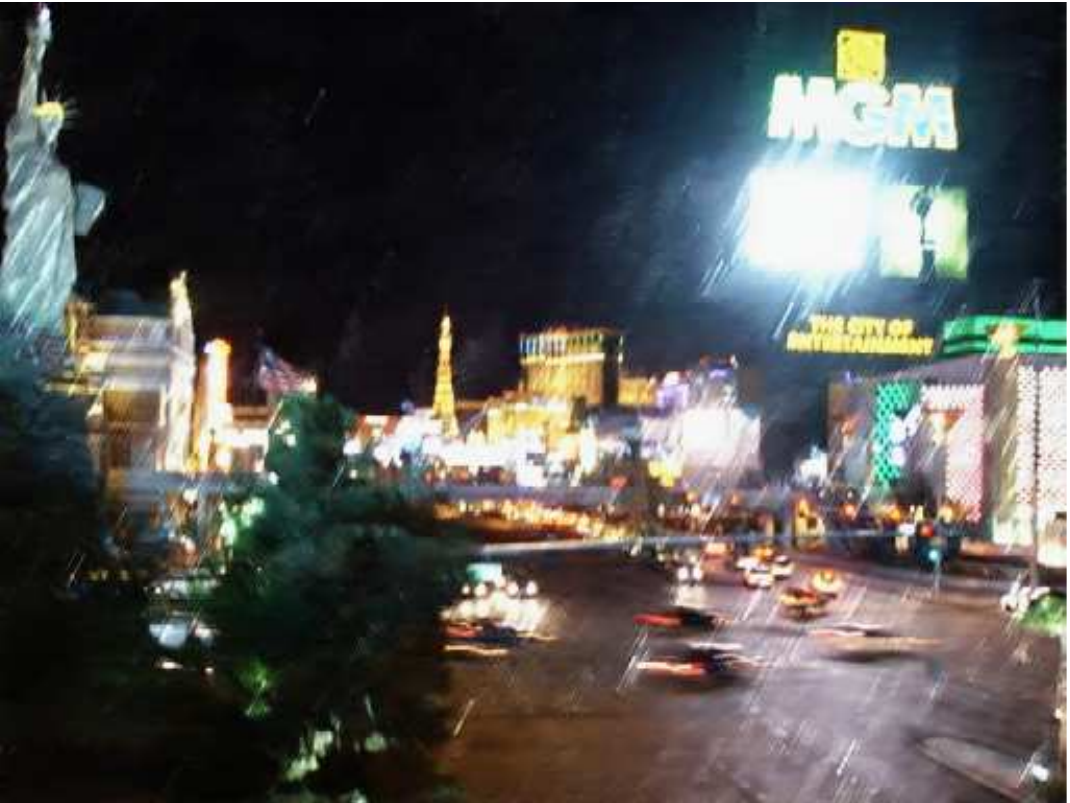}&
                \includegraphics[width=0.7in]{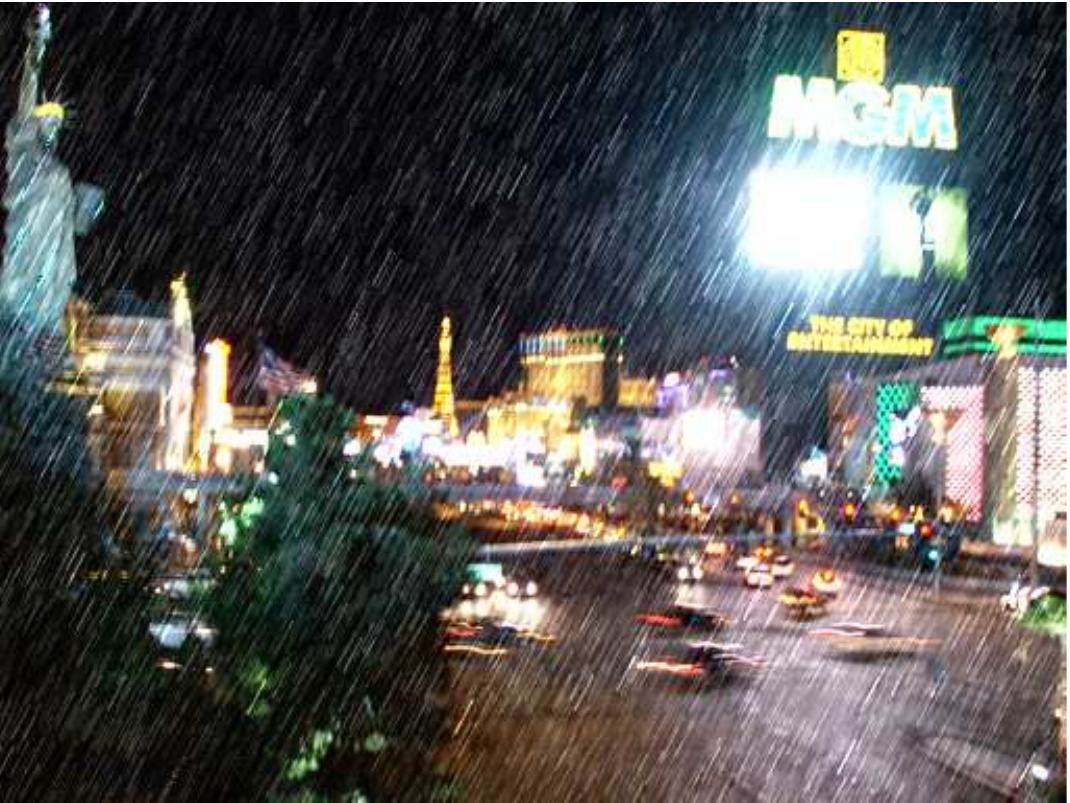}&
                \includegraphics[width=0.7in]{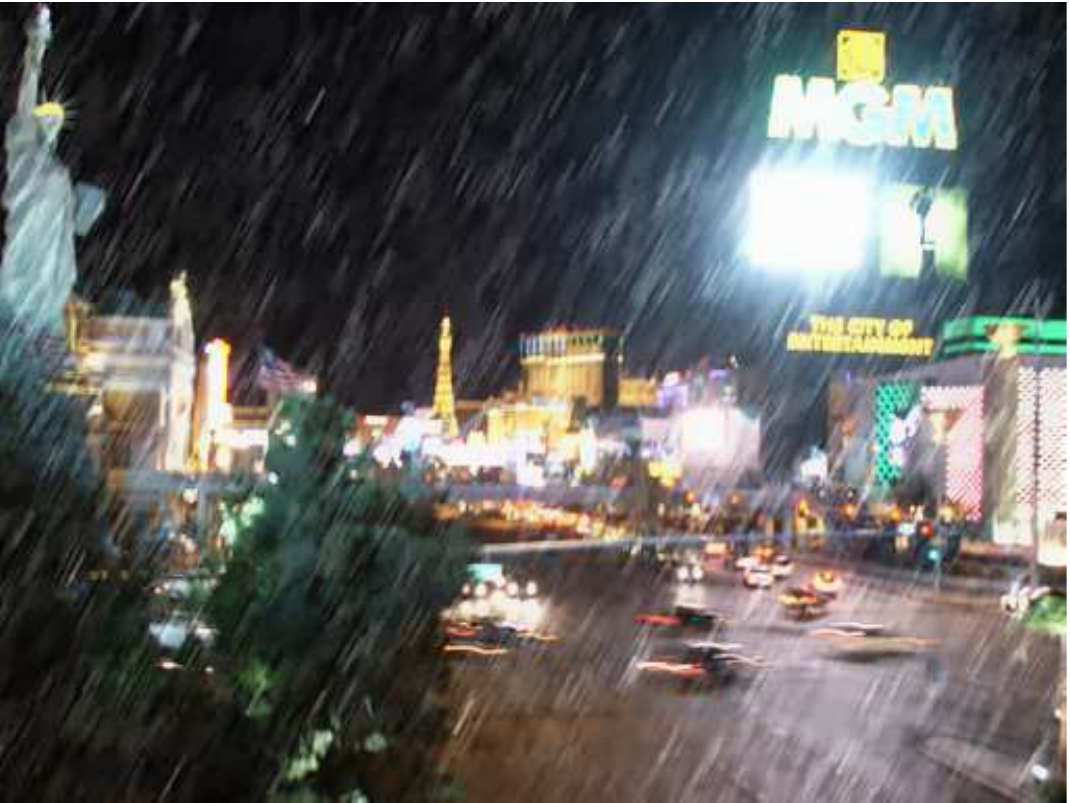}&
                \includegraphics[width=0.7in]{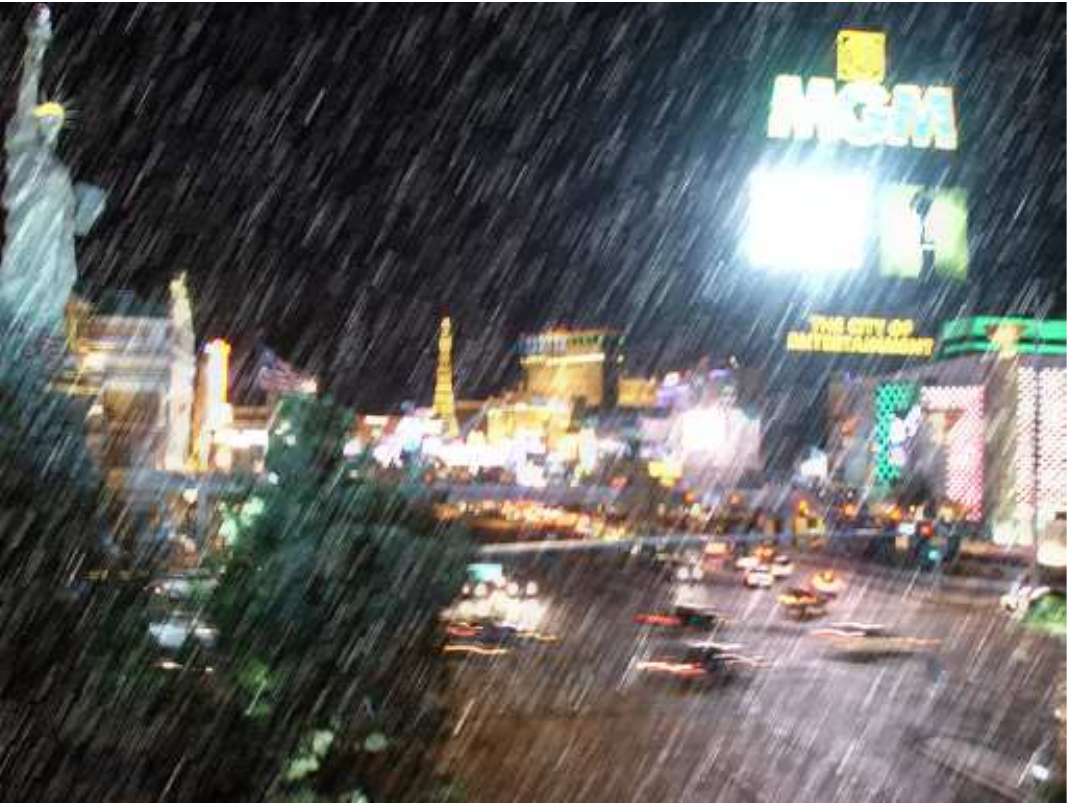}&
                \includegraphics[width=0.7in]{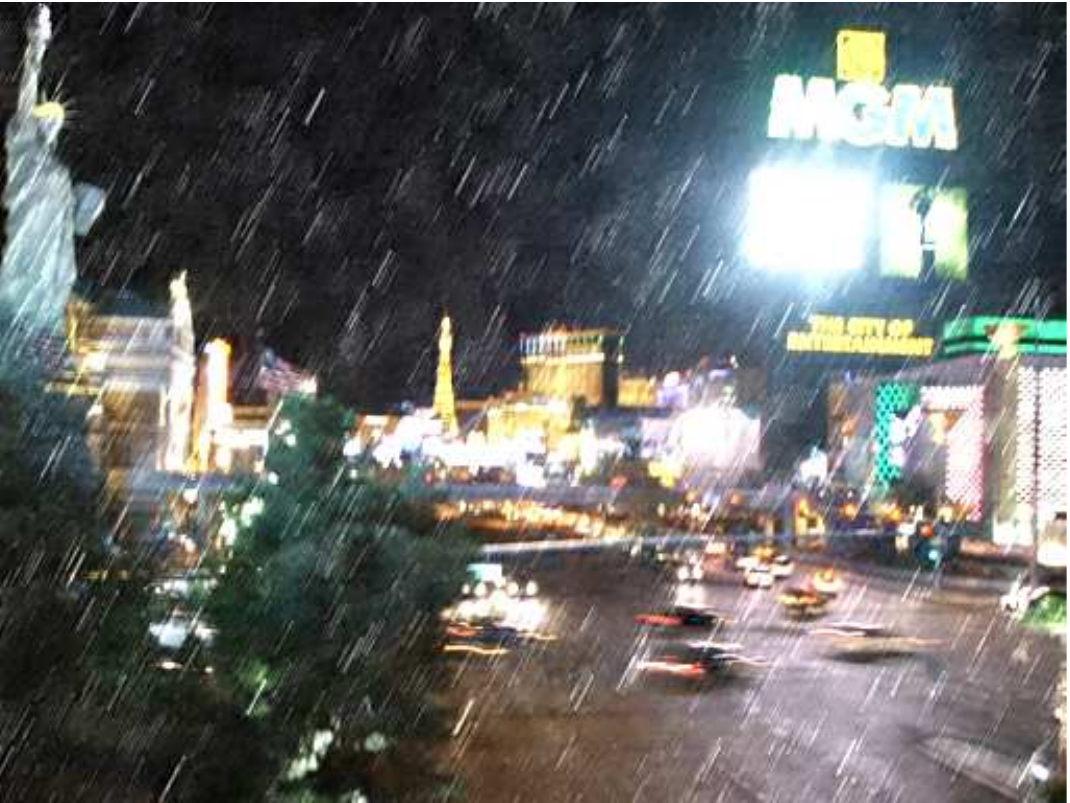}&
                \includegraphics[width=0.7in]{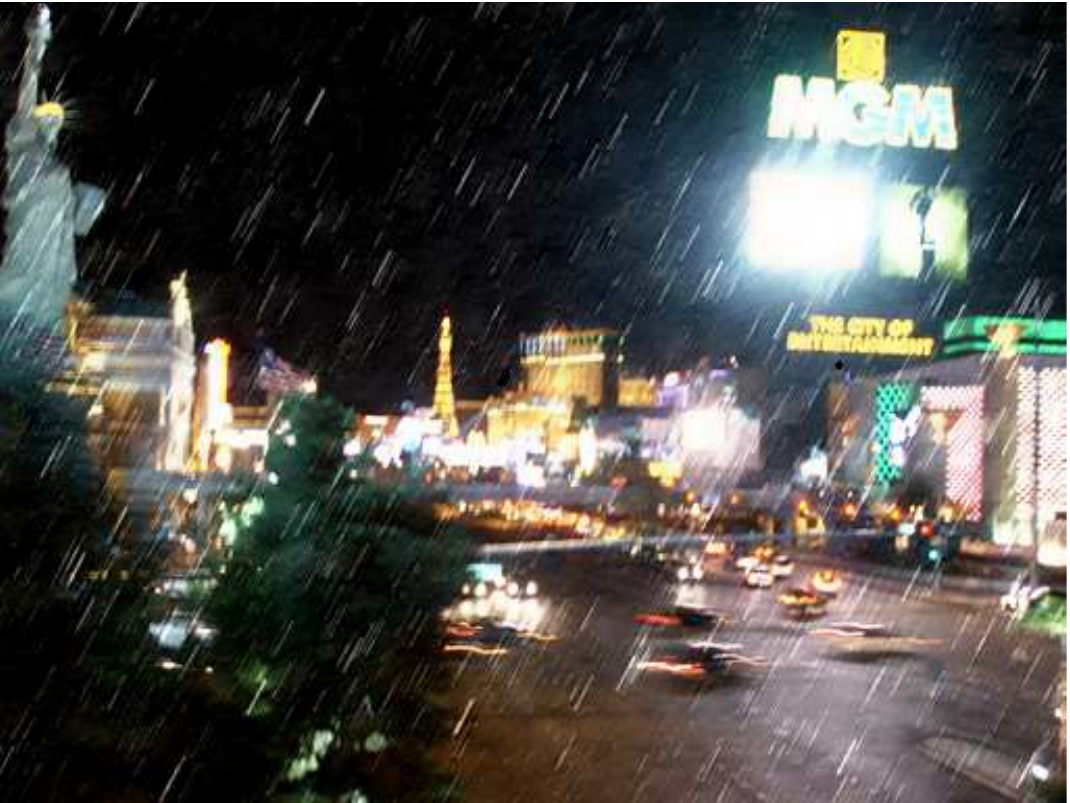}&
                \includegraphics[width=0.7in]{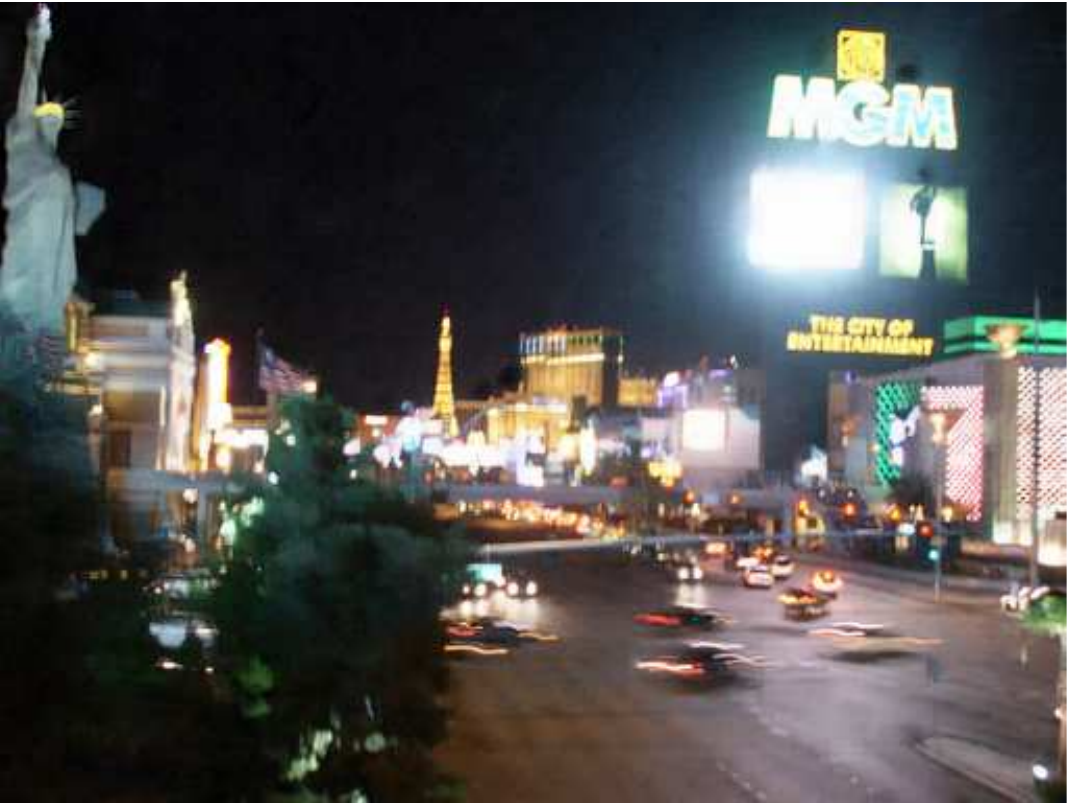}\\
                \vspace{0.5mm}

                \includegraphics[width=0.7in]{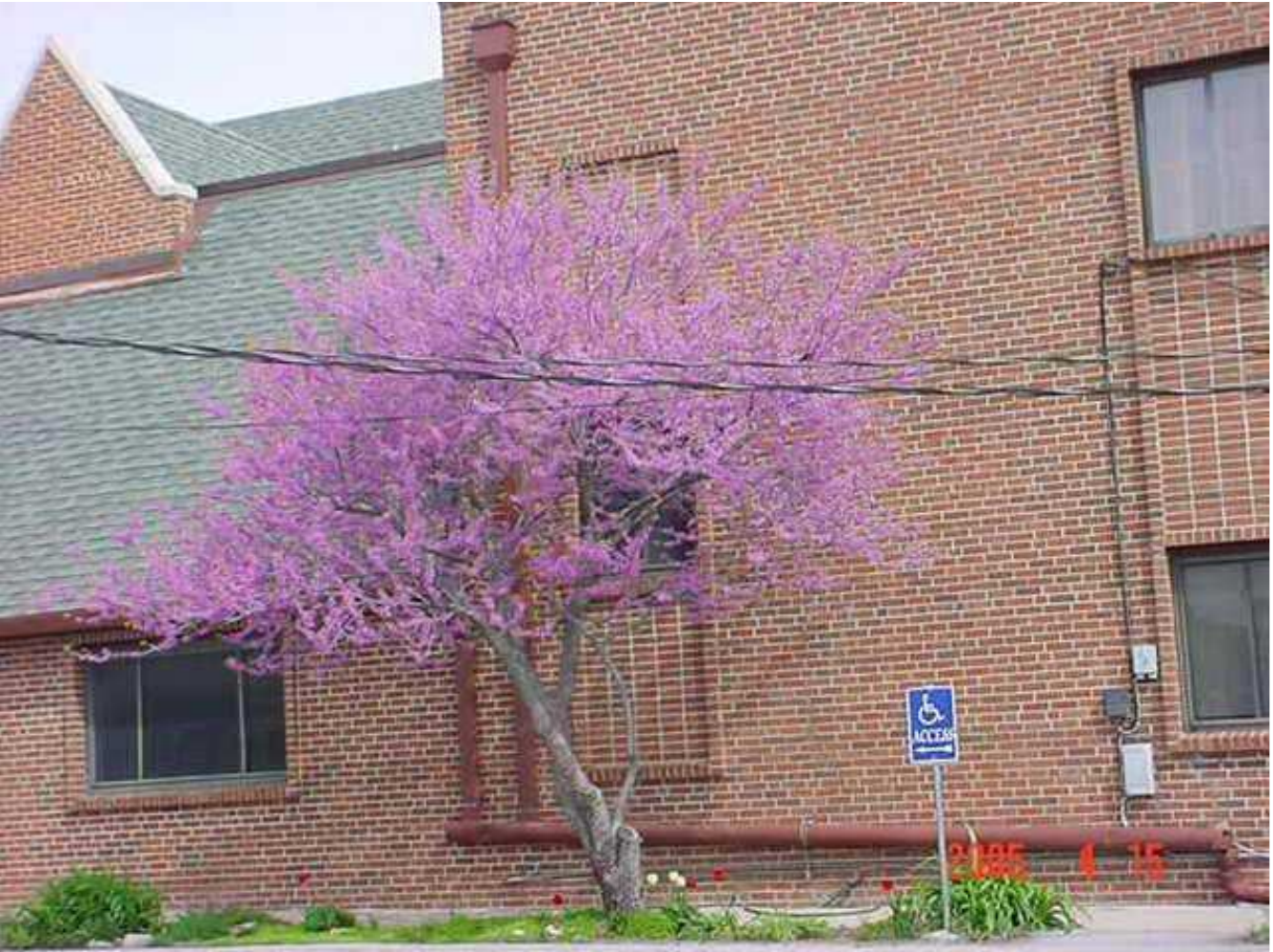}&
                \includegraphics[width=0.7in]{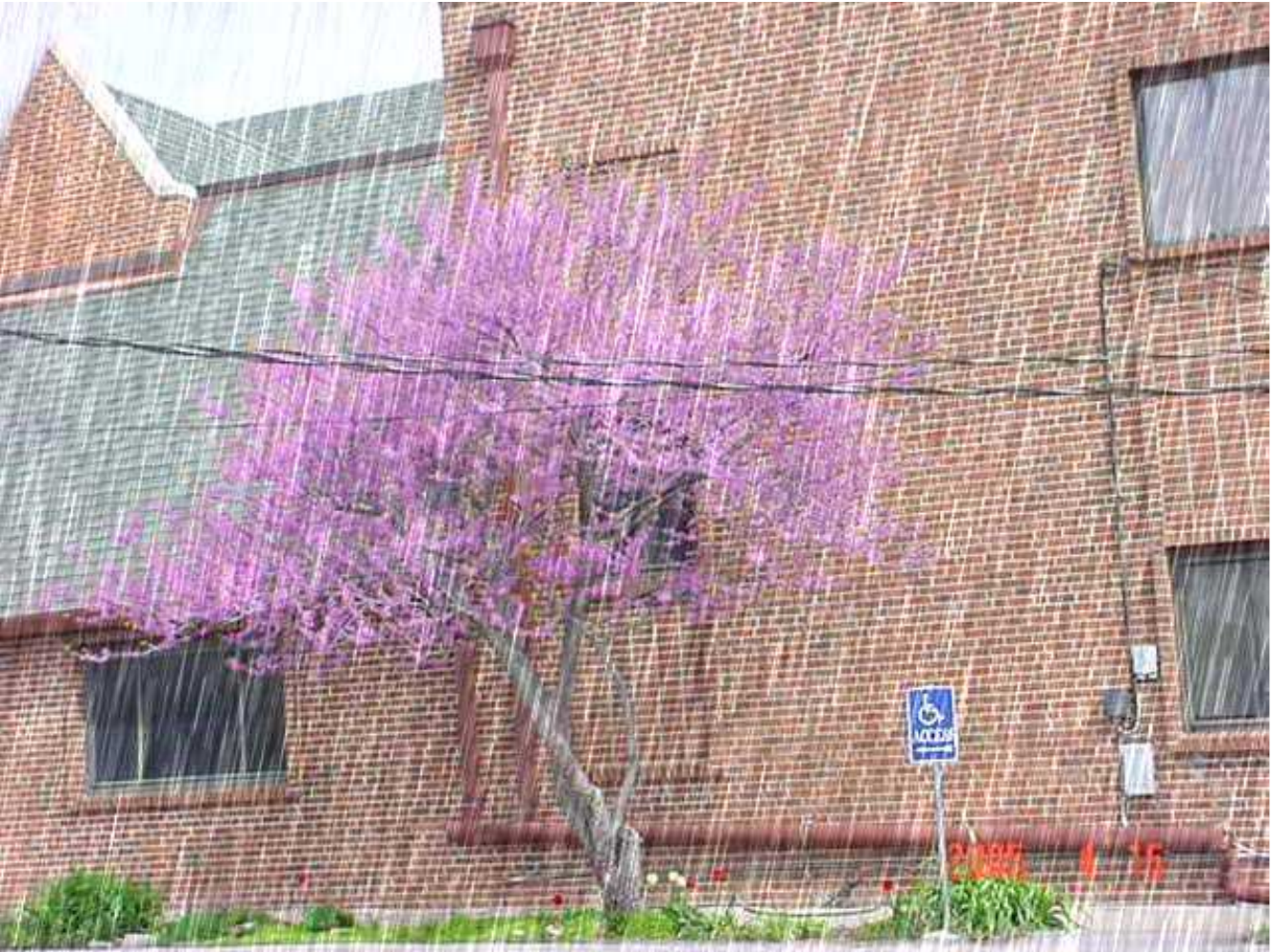}&
                \includegraphics[width=0.7in]{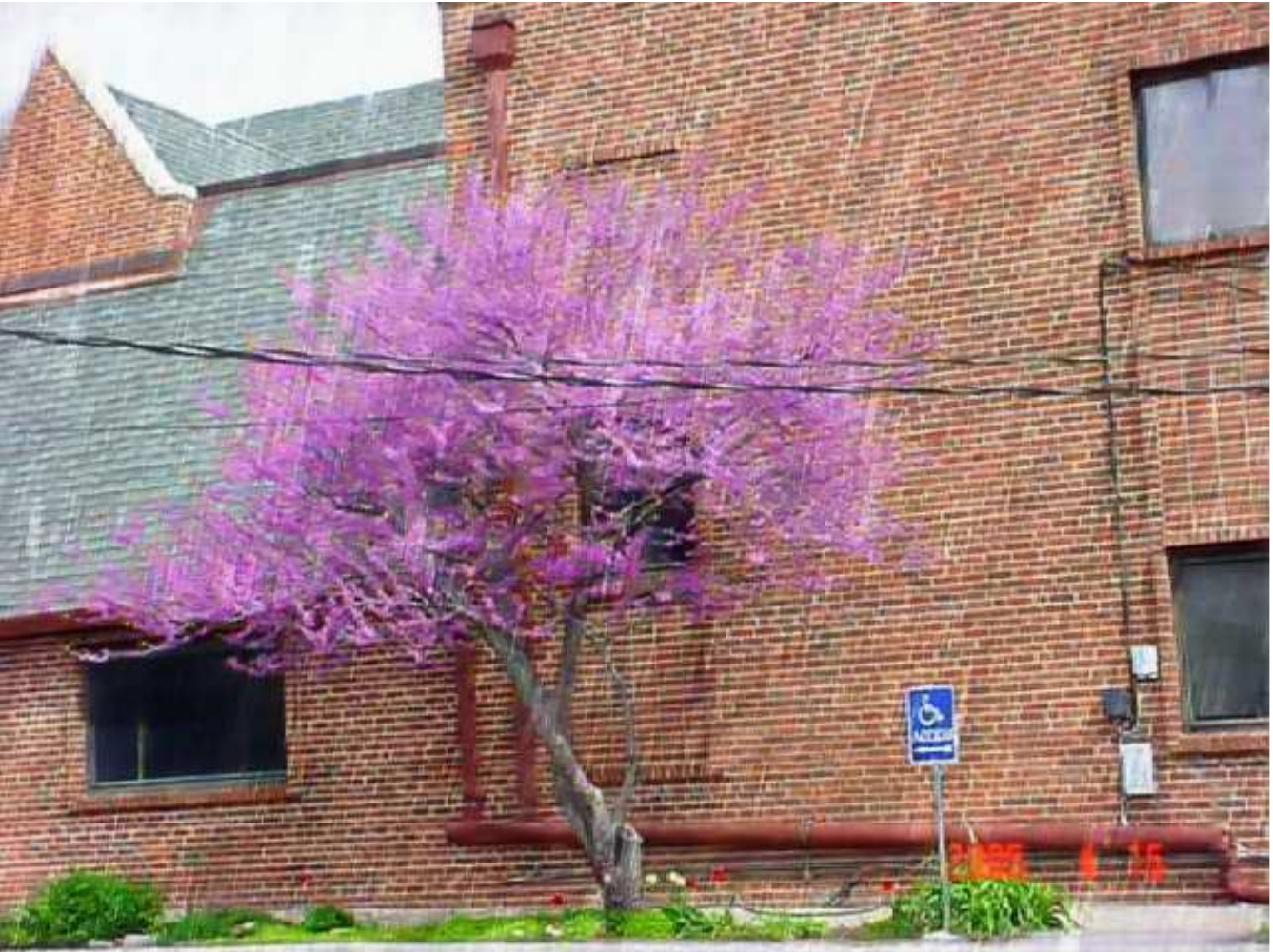}&
                \includegraphics[width=0.7in]{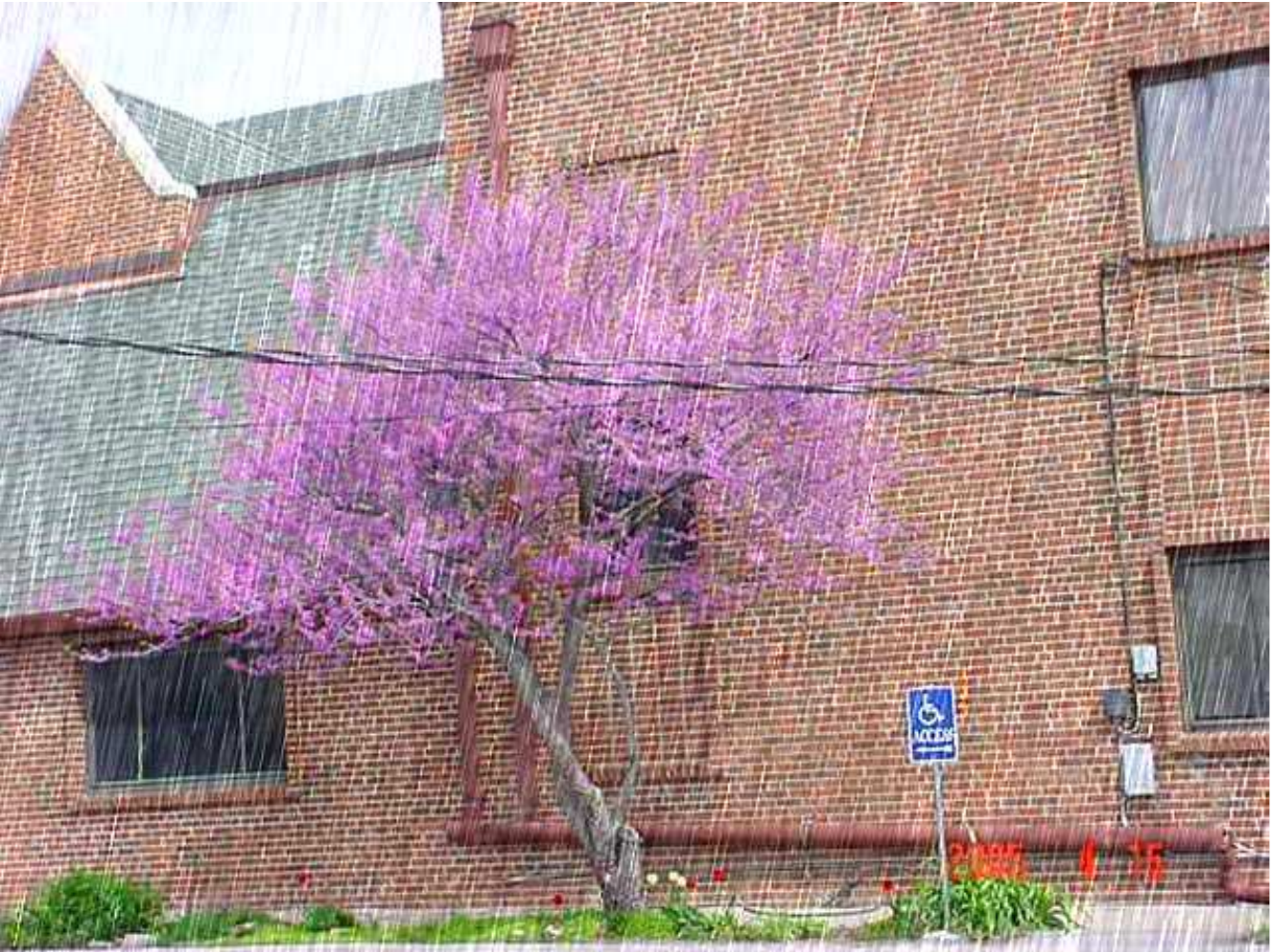}&
                \includegraphics[width=0.7in]{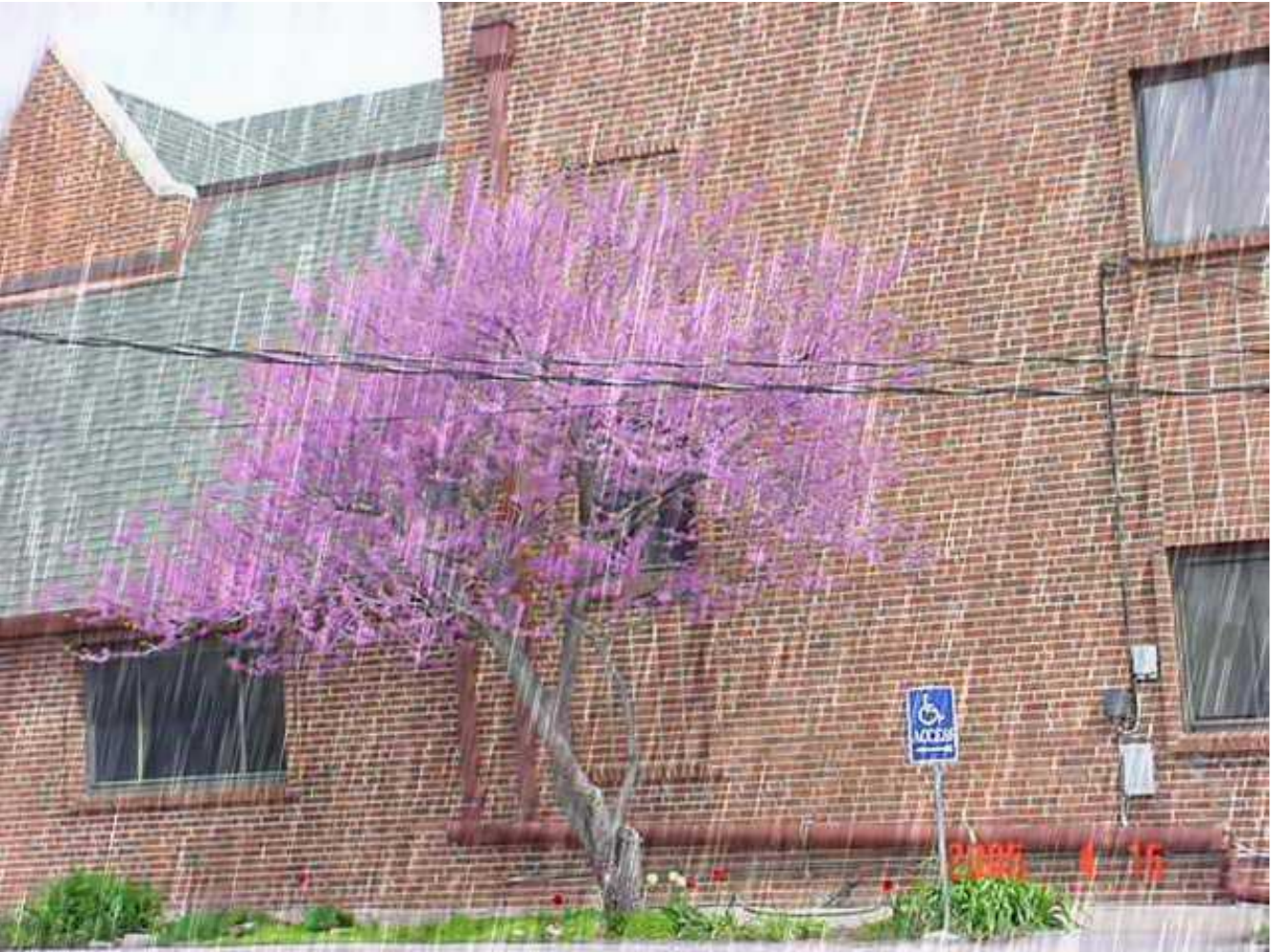}&
                \includegraphics[width=0.7in]{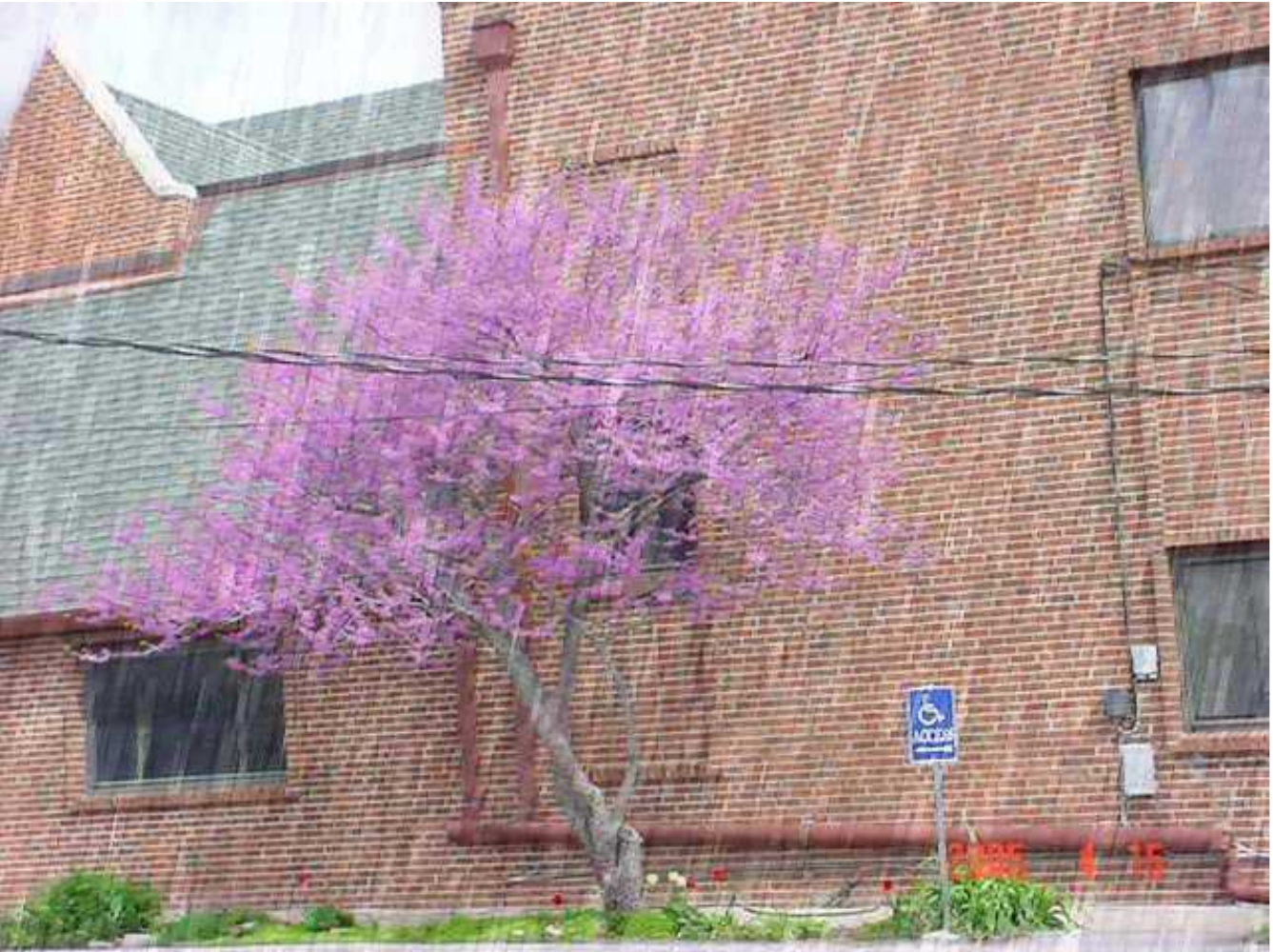}&
                \includegraphics[width=0.7in]{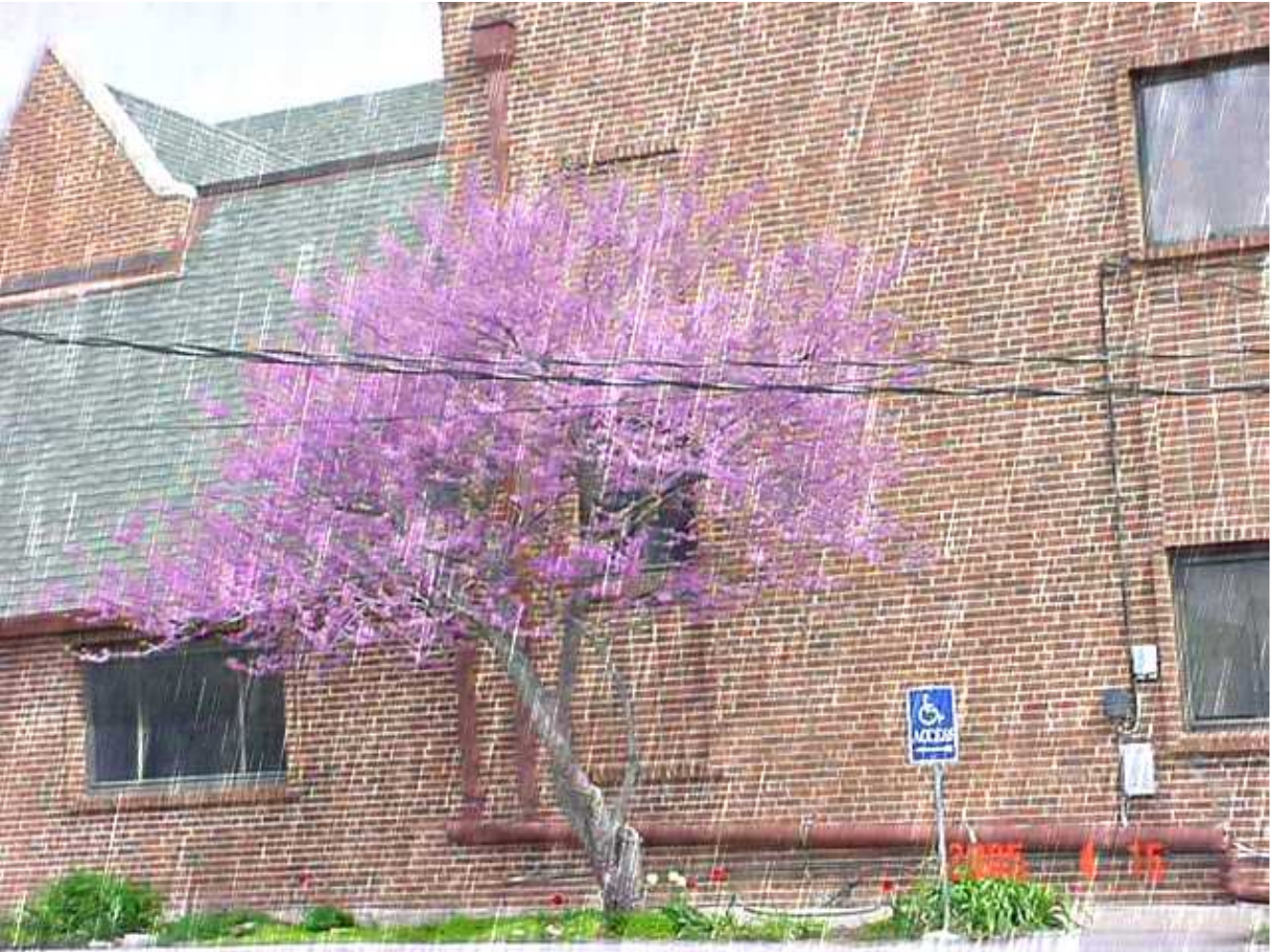}&
                \includegraphics[width=0.7in]{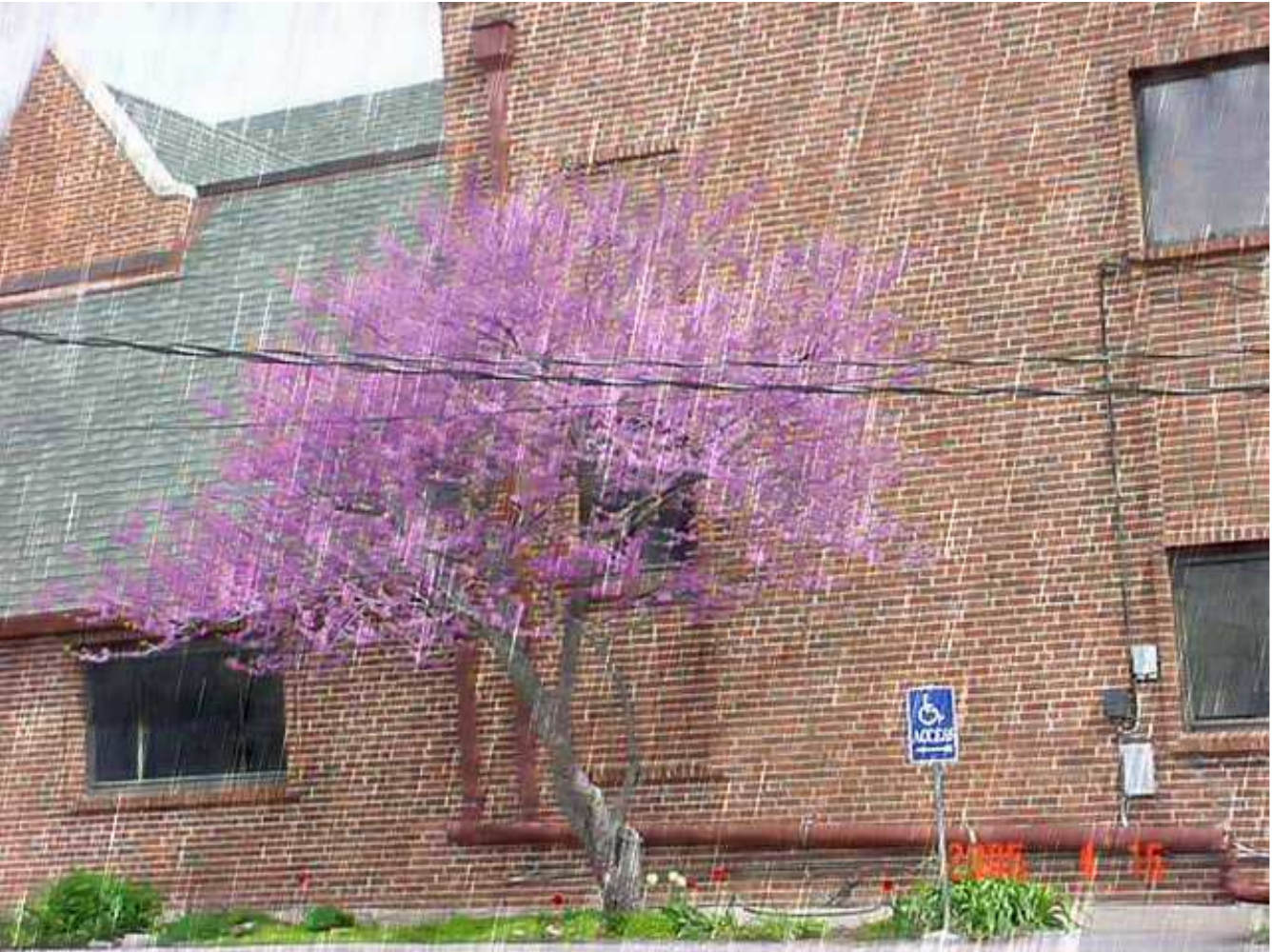}&
                \includegraphics[width=0.7in]{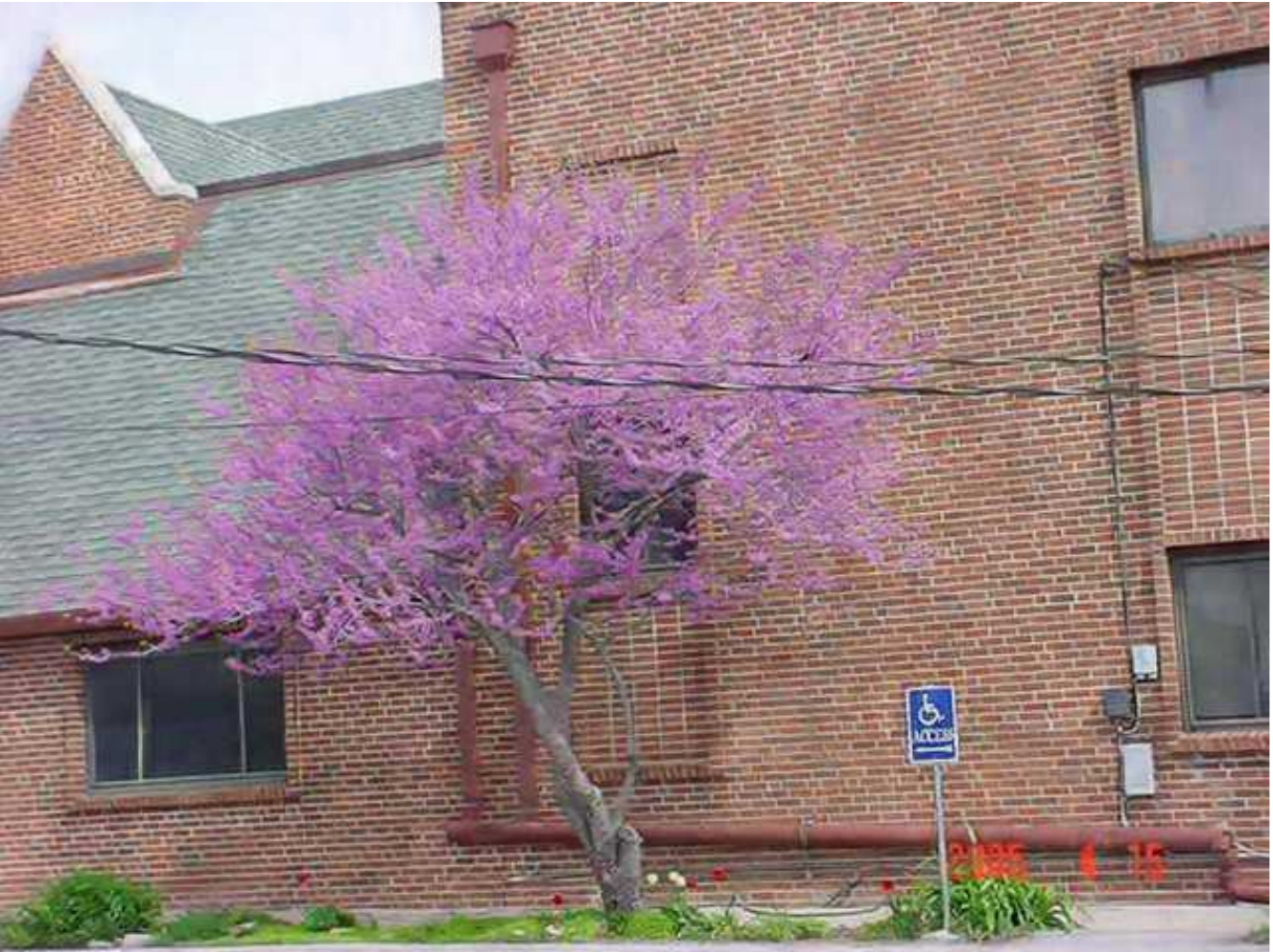}\\
                (a)&
                (b)&
                (c)&
                (d)&
                (e)&
                (f)&
                (g)&
                (h)&
                (i)\\

\end{tabular}
\caption{Rain streak removal results by different methods on 3 synthetic rain images (road, night, and street) by our simulating method. From left to right: (a) the background, (b) the rainy images, the derain results by (c) DID \cite{zhang2018density}, (d) DSC \cite{luo2015removing}, (e) LP \cite{Li2014Single}, (f) UGSM \cite{Deng2018A}, (g) CNN \cite{fu2017clearing}, (h) DDN \cite{fu2017removing}, and (i) KGCNN.}
\label{synthetic-visual-our}
\end{center}
\end{figure*}

\begin{figure*}[!htb]
\renewcommand\arraystretch{0.8}\setlength{\tabcolsep}{1.8pt}
\begin{center}
\begin{tabular}{ccccccccc}
                ~\includegraphics[width=0.7in]{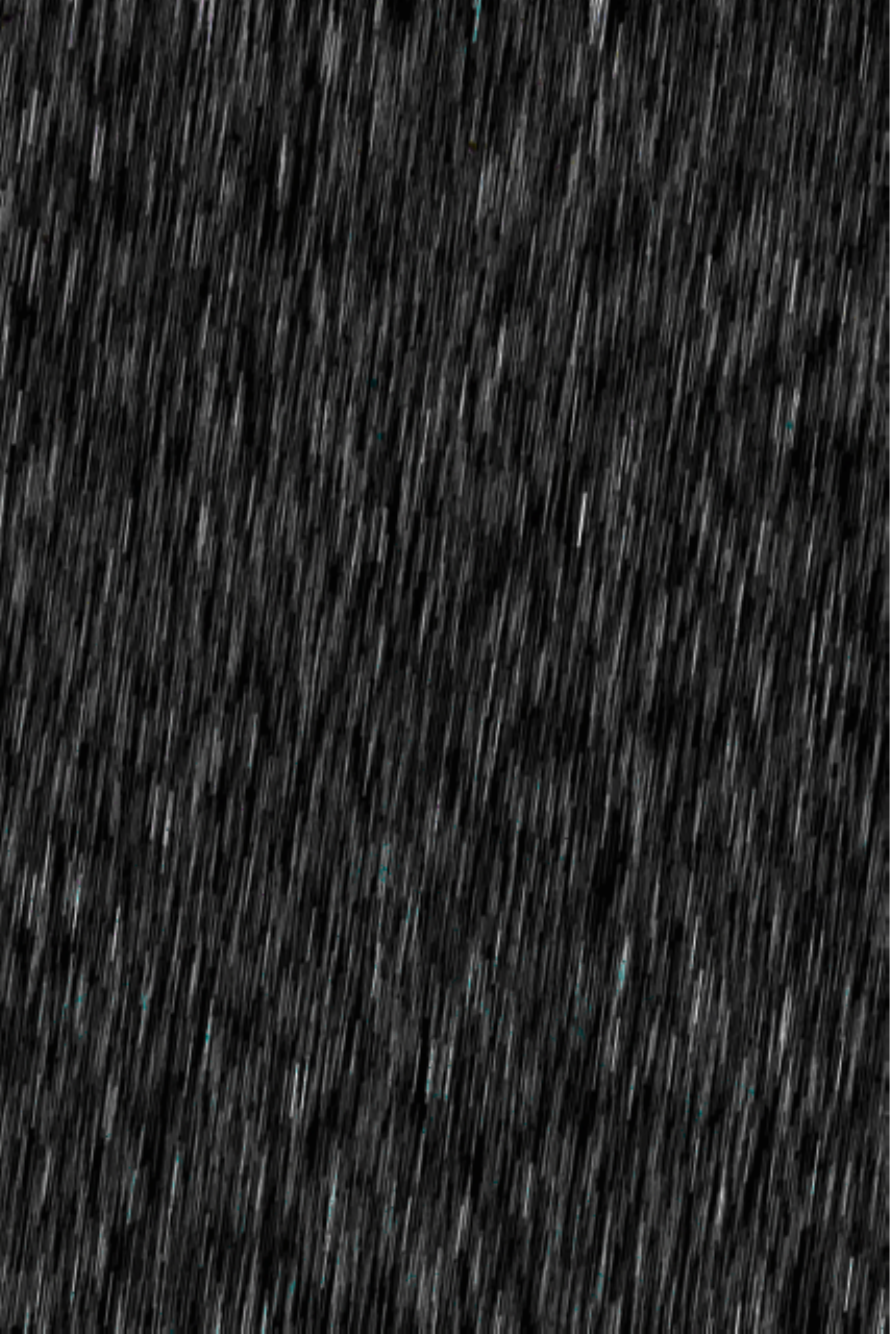}&
                \includegraphics[width=0.7in]{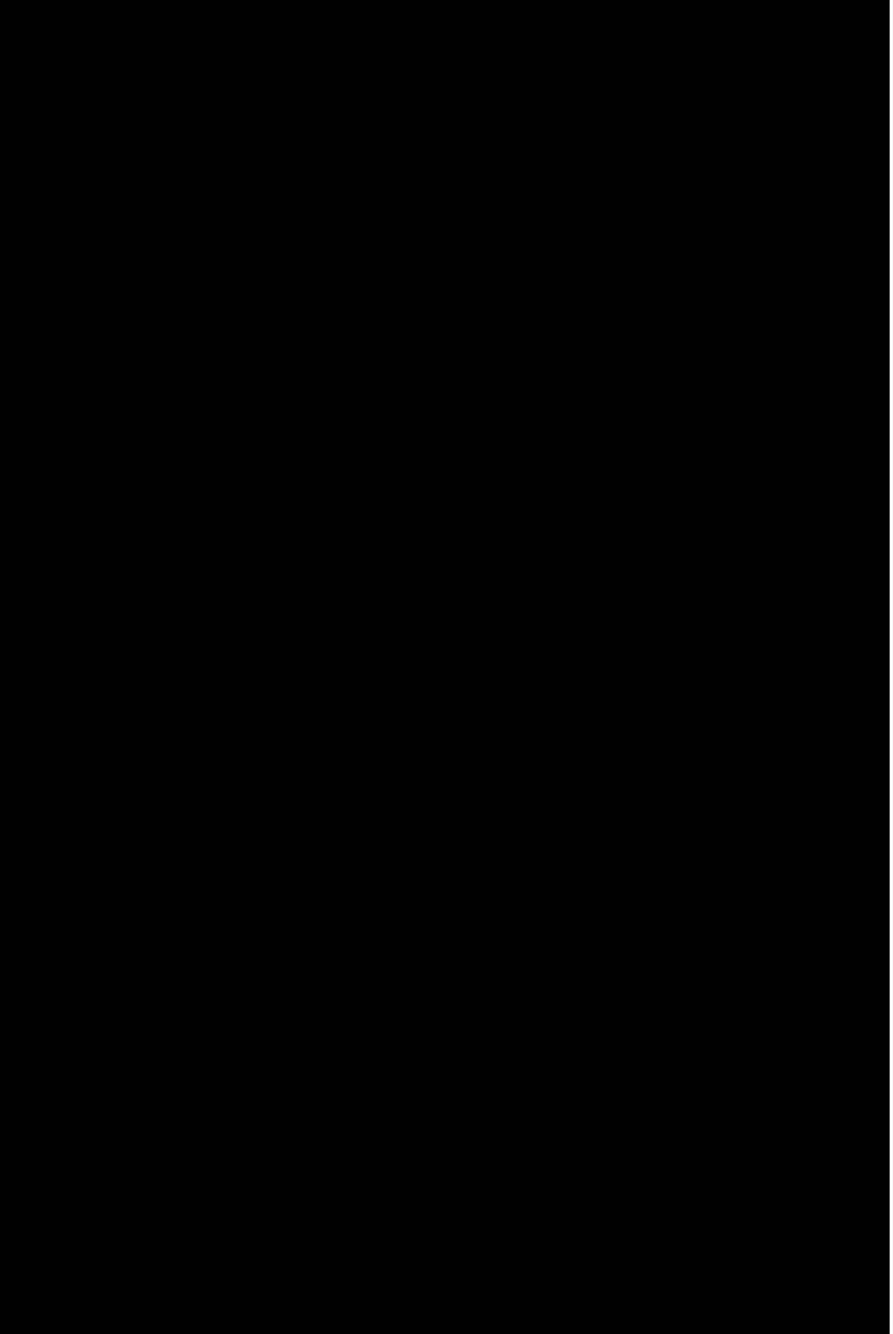}&
                \includegraphics[width=0.7in]{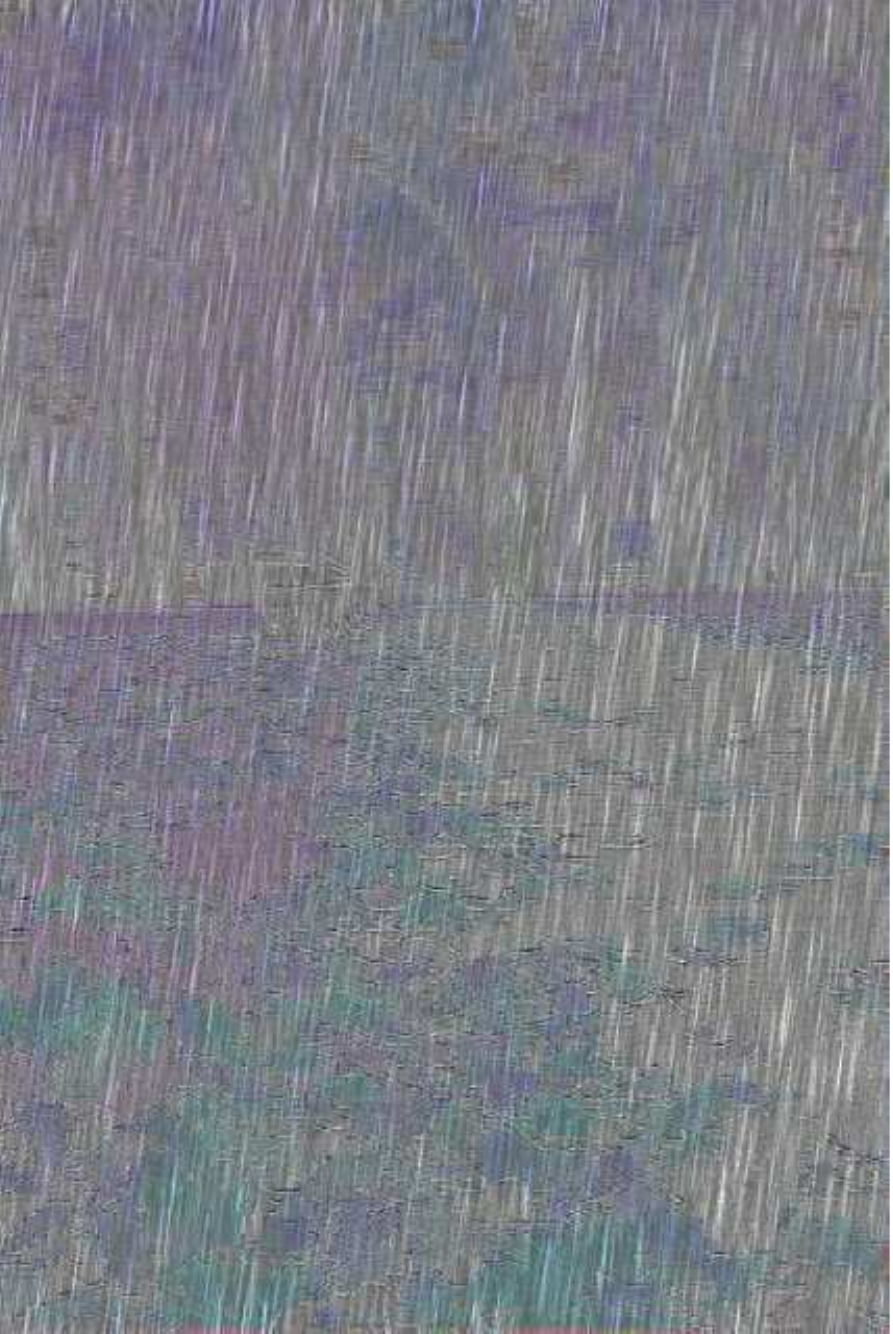}&
                \includegraphics[width=0.7in]{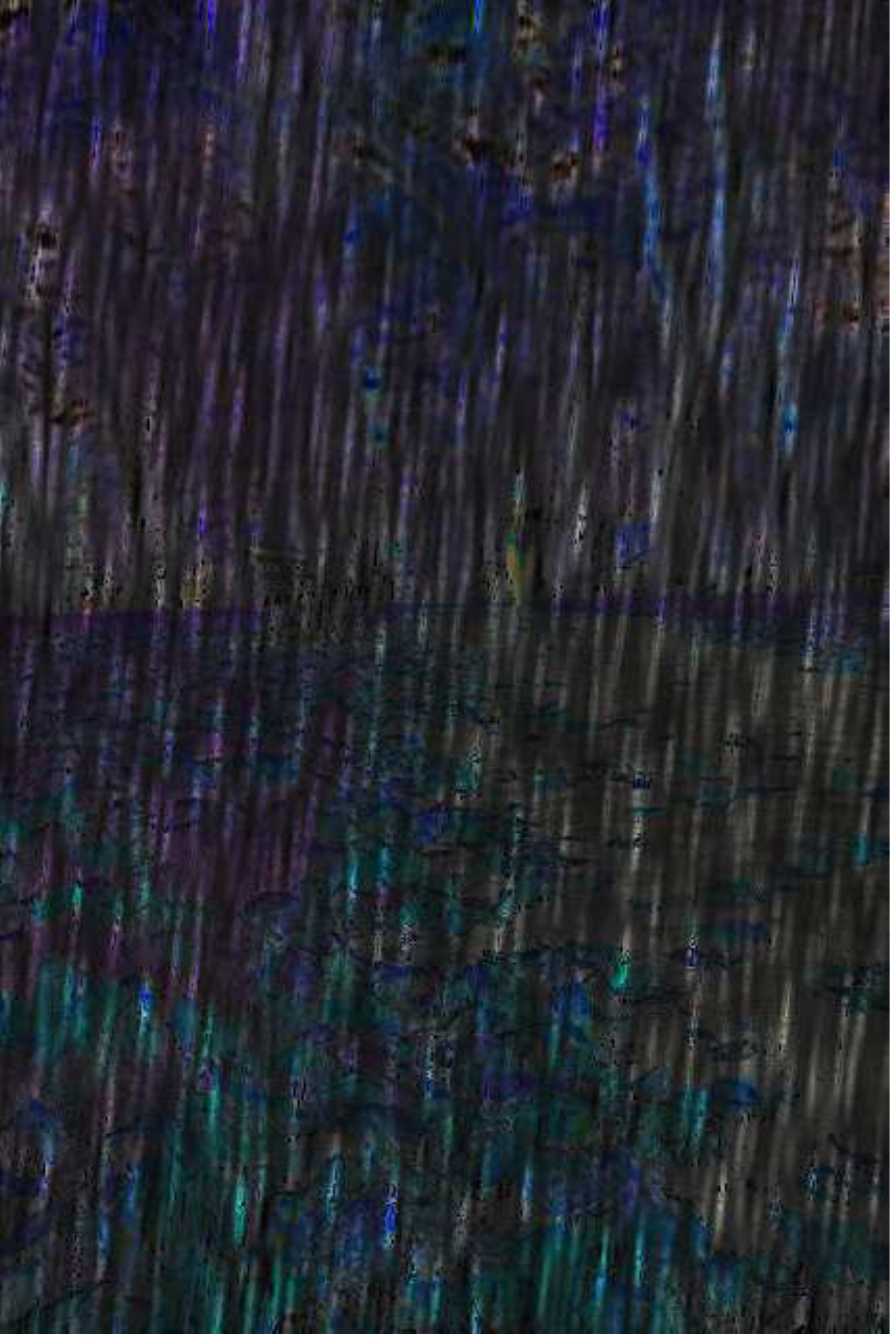}&
                \includegraphics[width=0.7in]{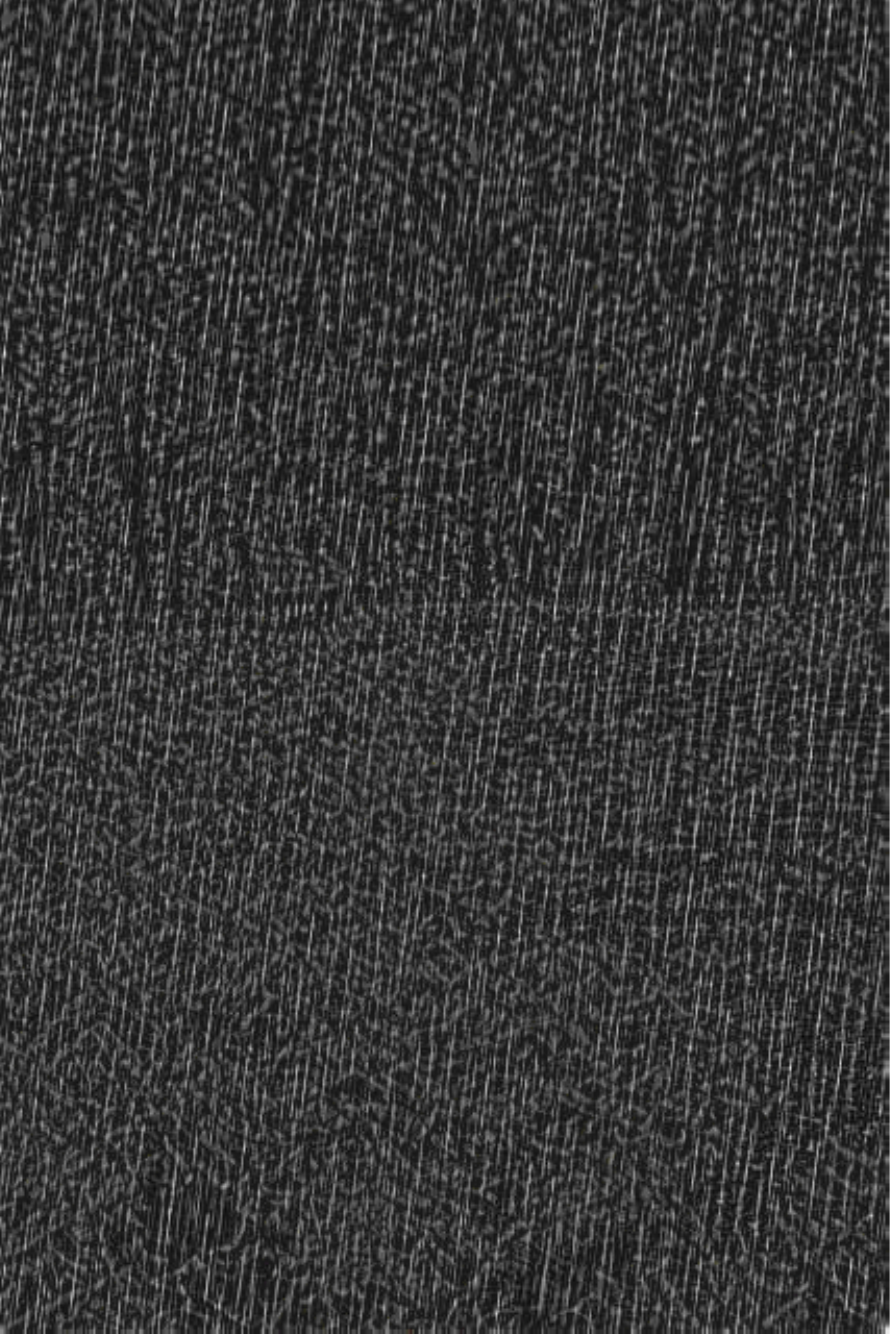}&
                \includegraphics[width=0.7in]{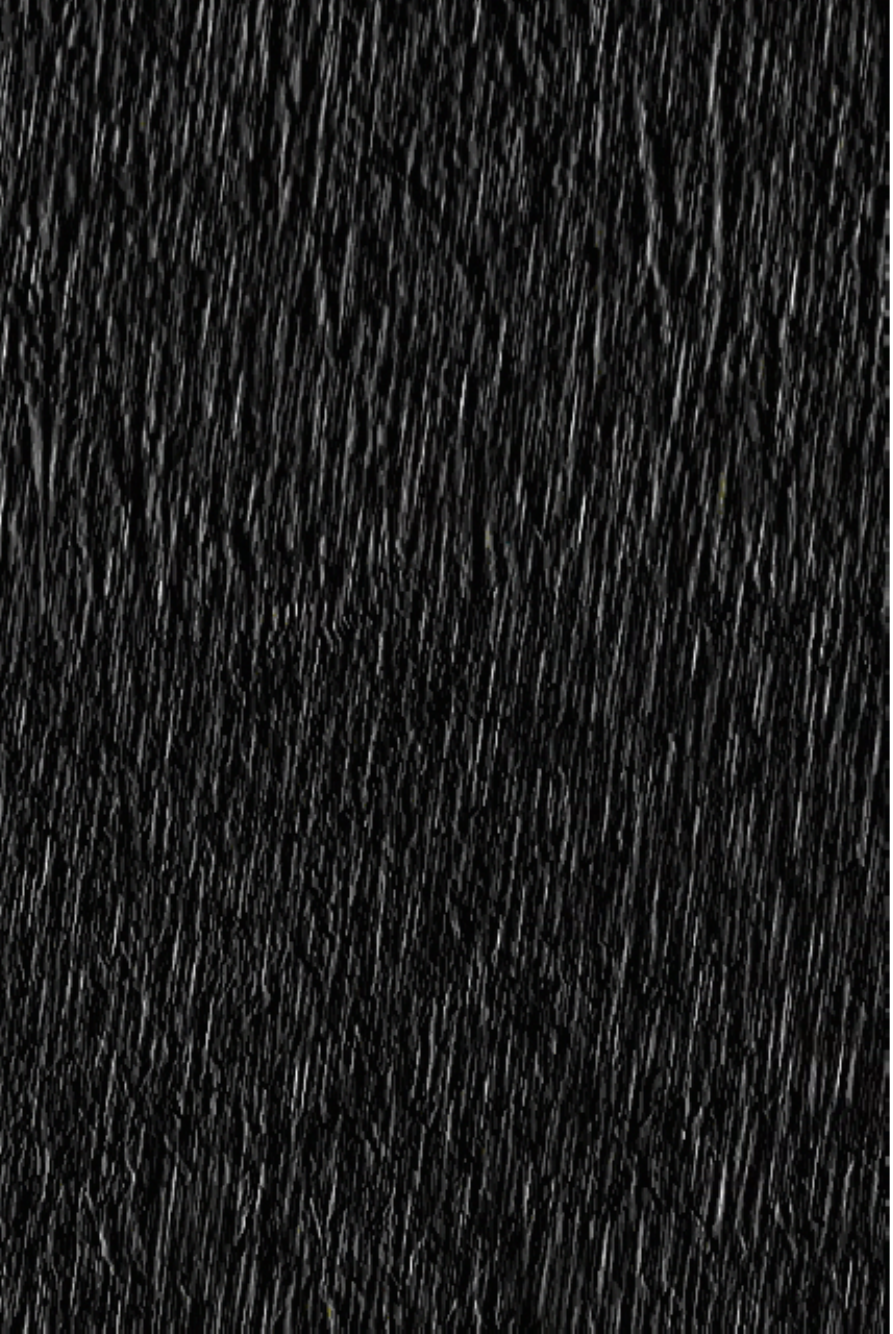}&
                \includegraphics[width=0.7in]{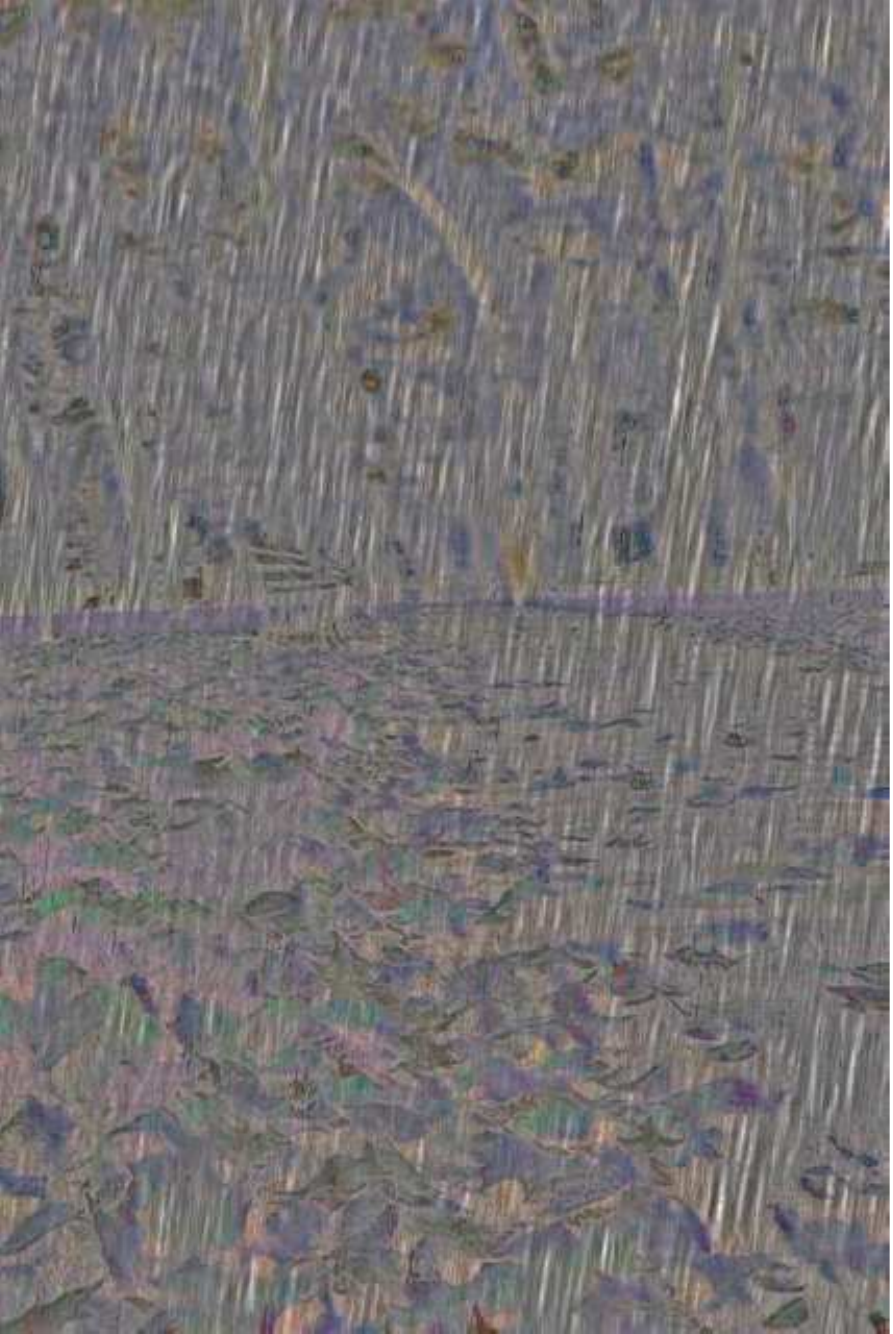}&
                \includegraphics[width=0.7in]{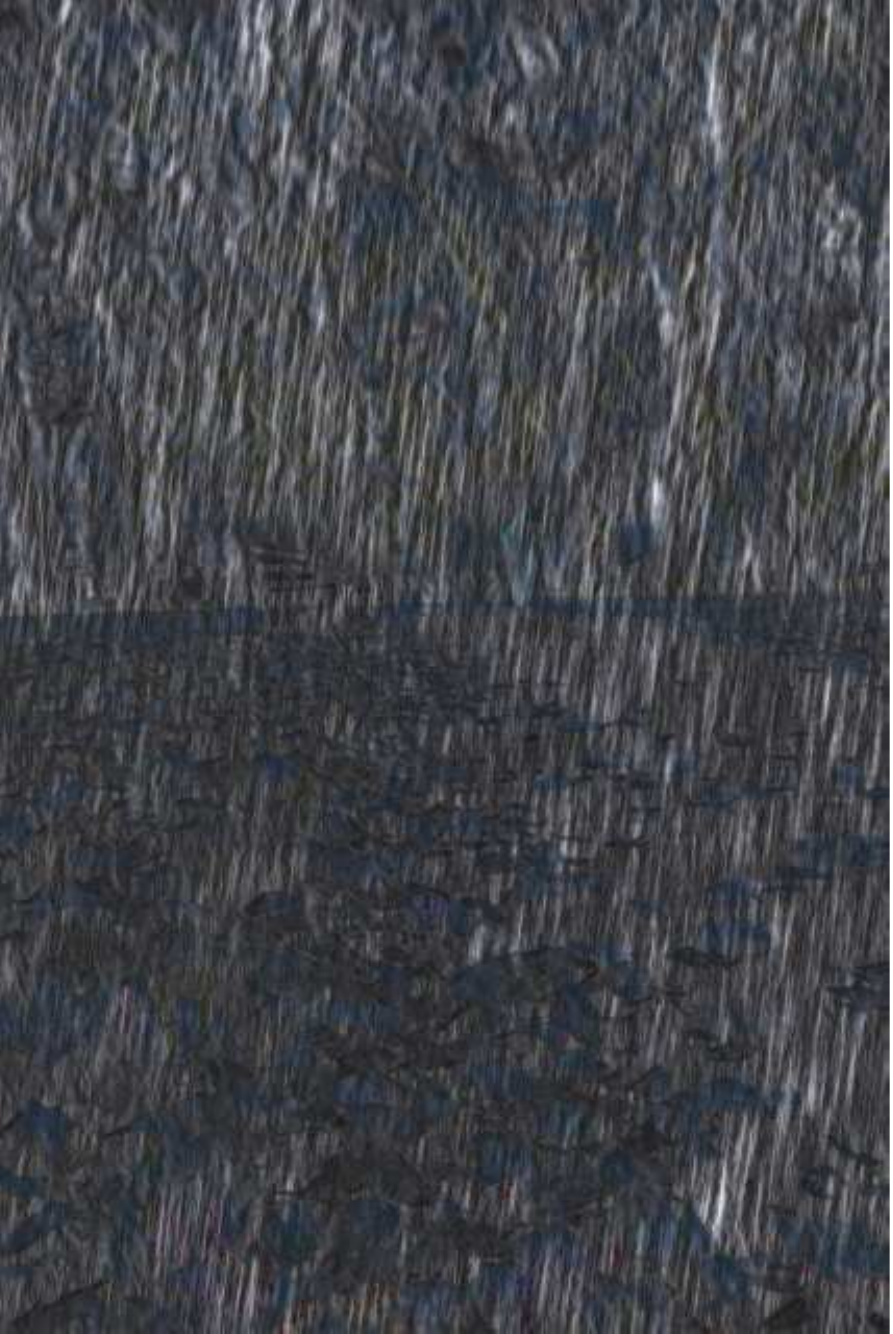}&
                \includegraphics[width=0.7in]{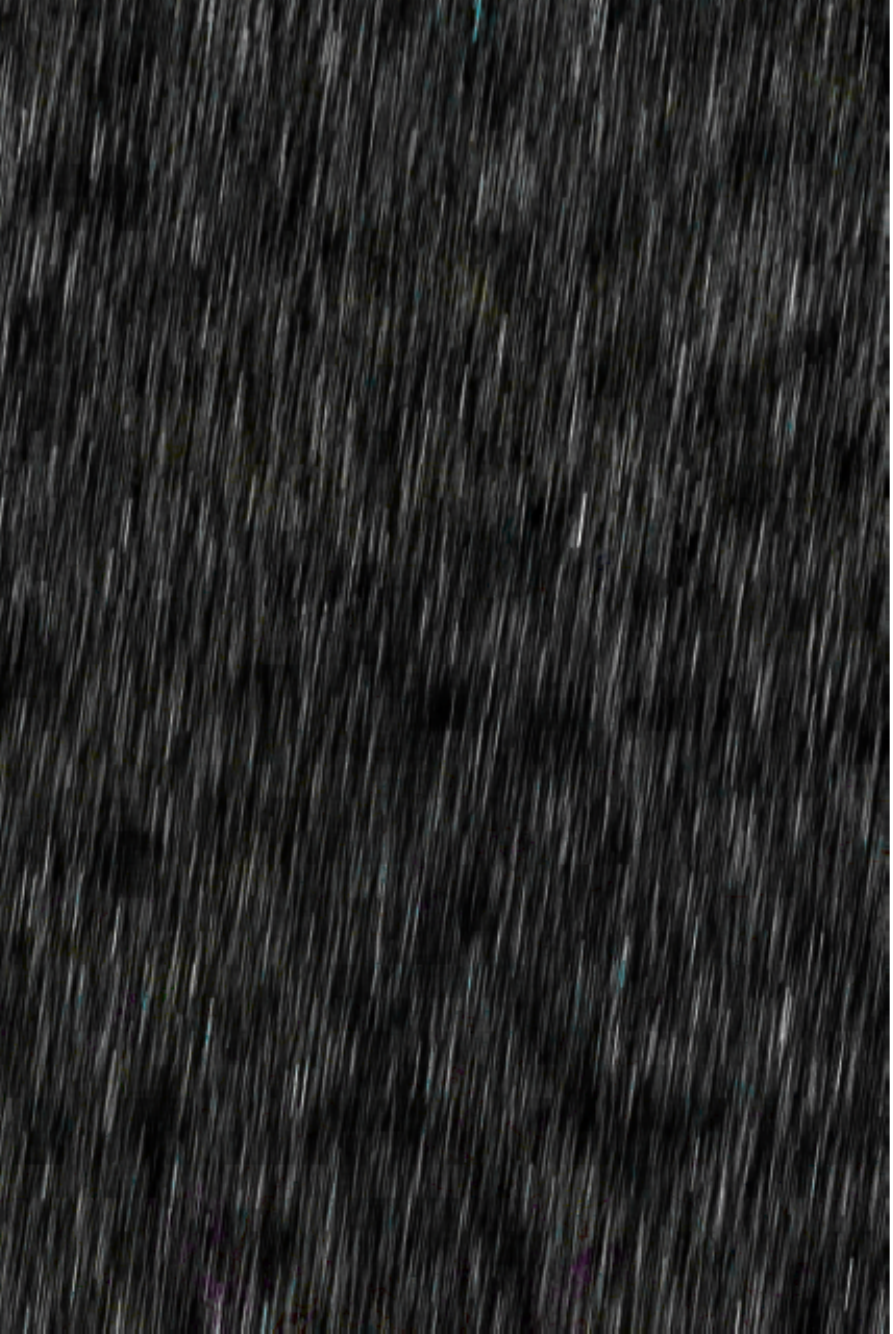}\\
                \vspace{0.5mm}

                \includegraphics[width=0.7in]{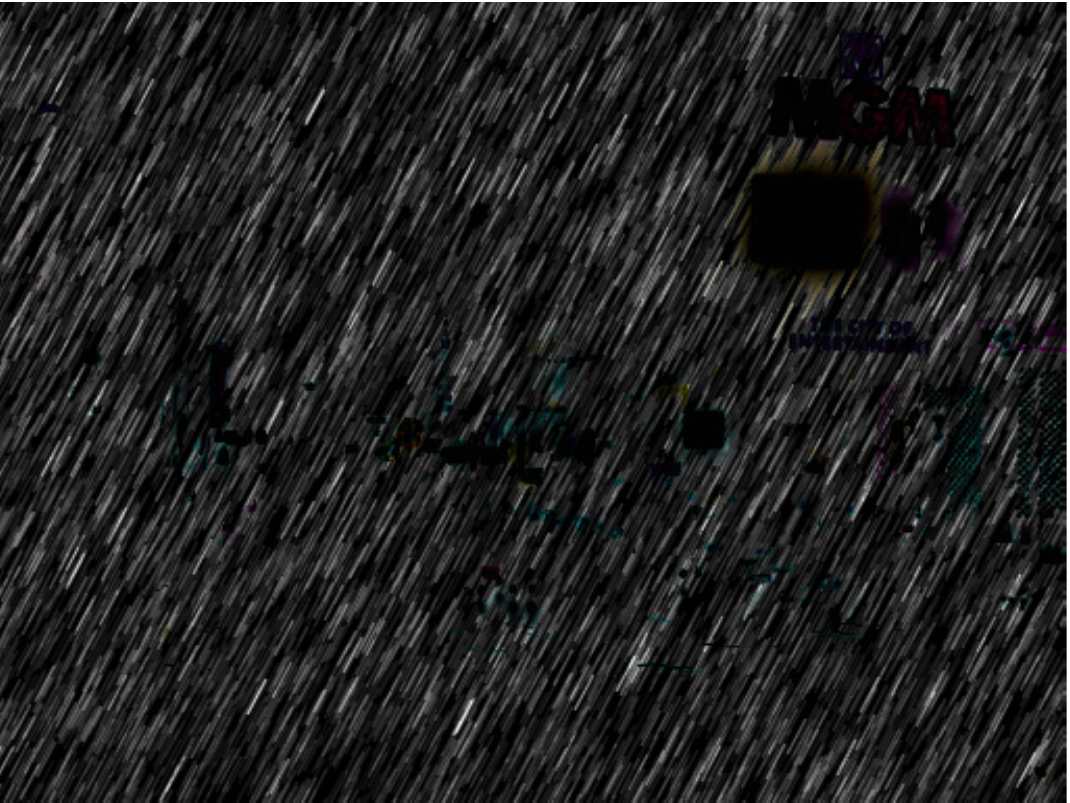}&
                \includegraphics[width=0.7in]{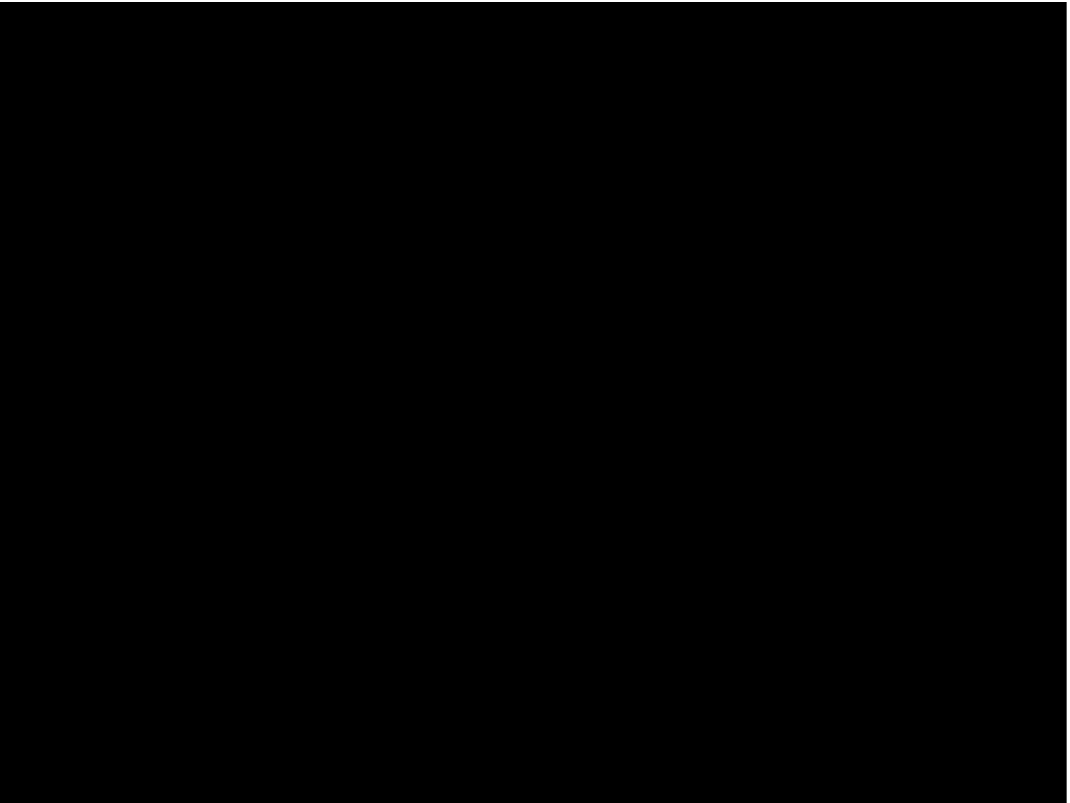}&
                \includegraphics[width=0.7in]{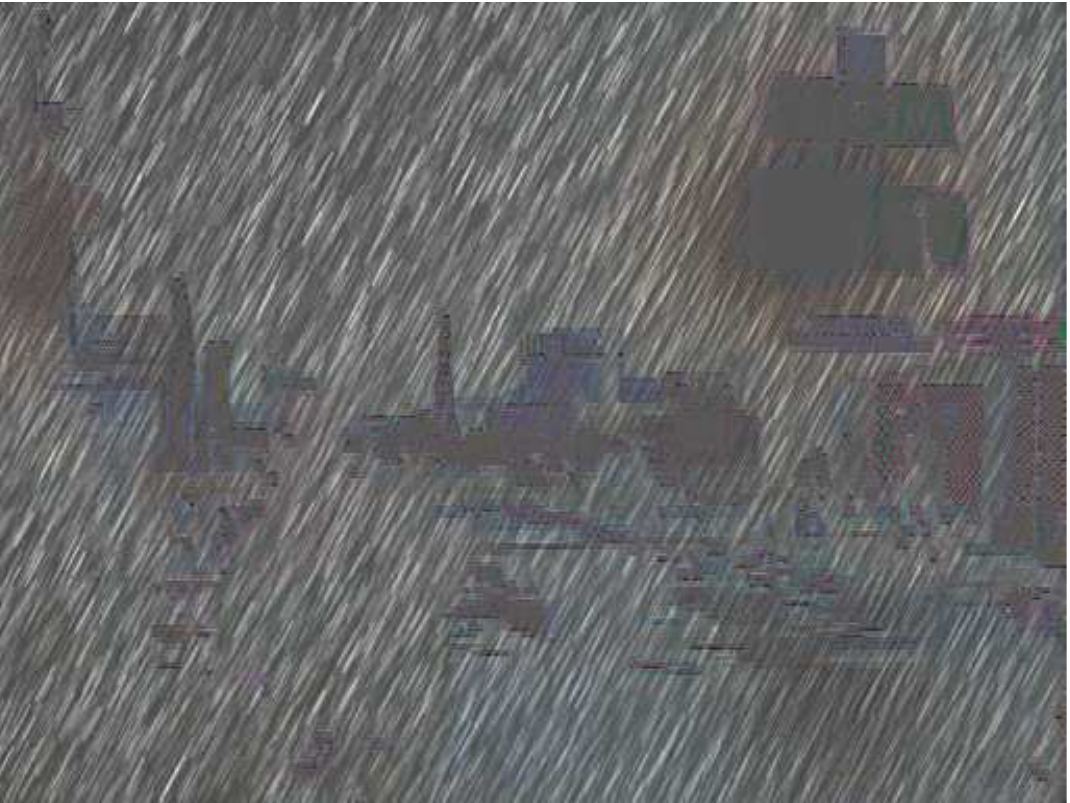}&
                \includegraphics[width=0.7in]{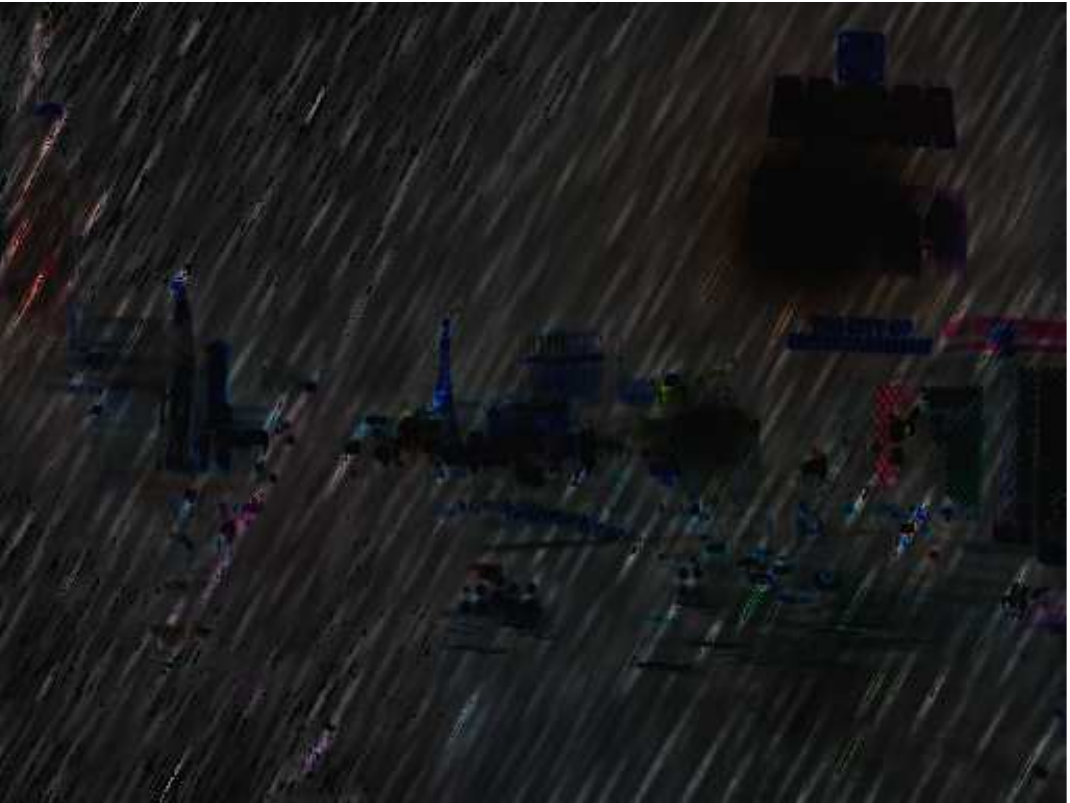}&
                \includegraphics[width=0.7in]{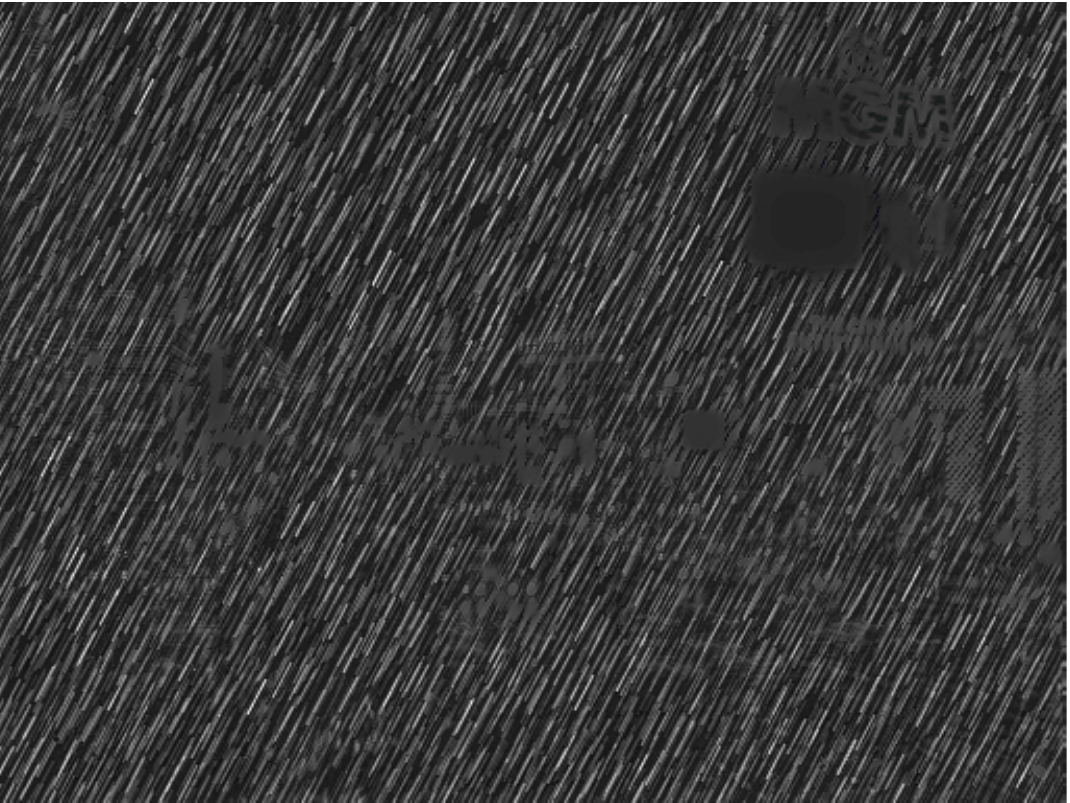}&
                \includegraphics[width=0.7in]{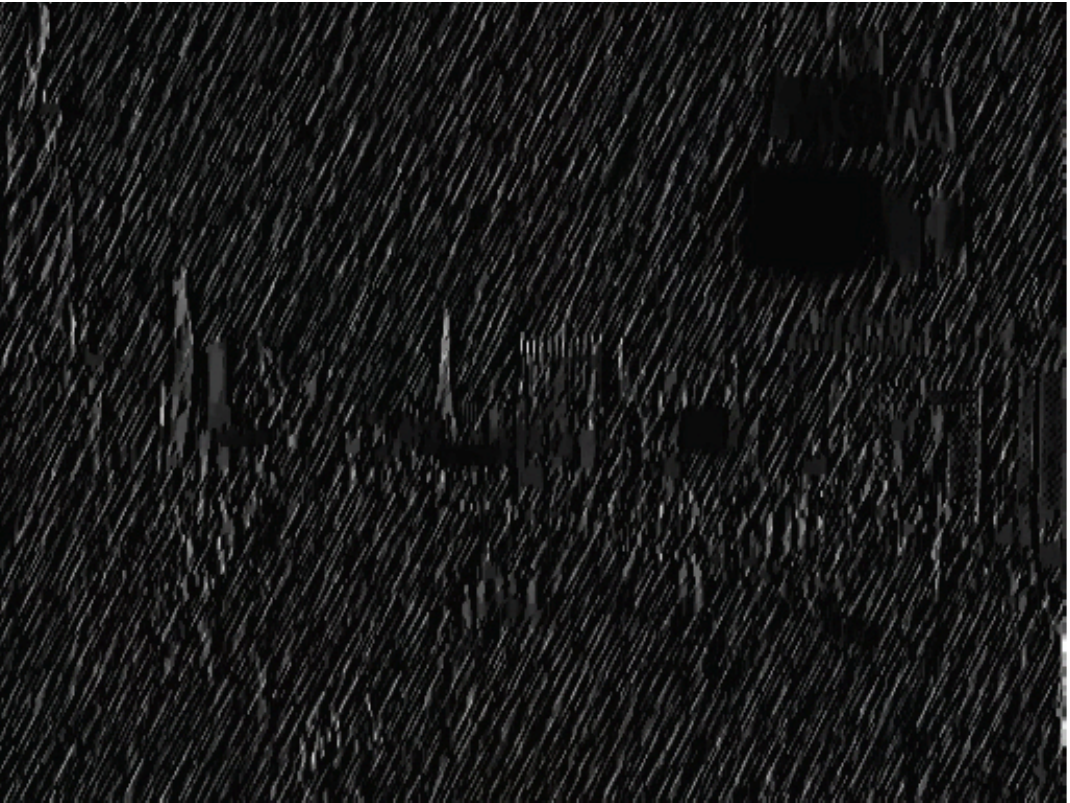}&
                \includegraphics[width=0.7in]{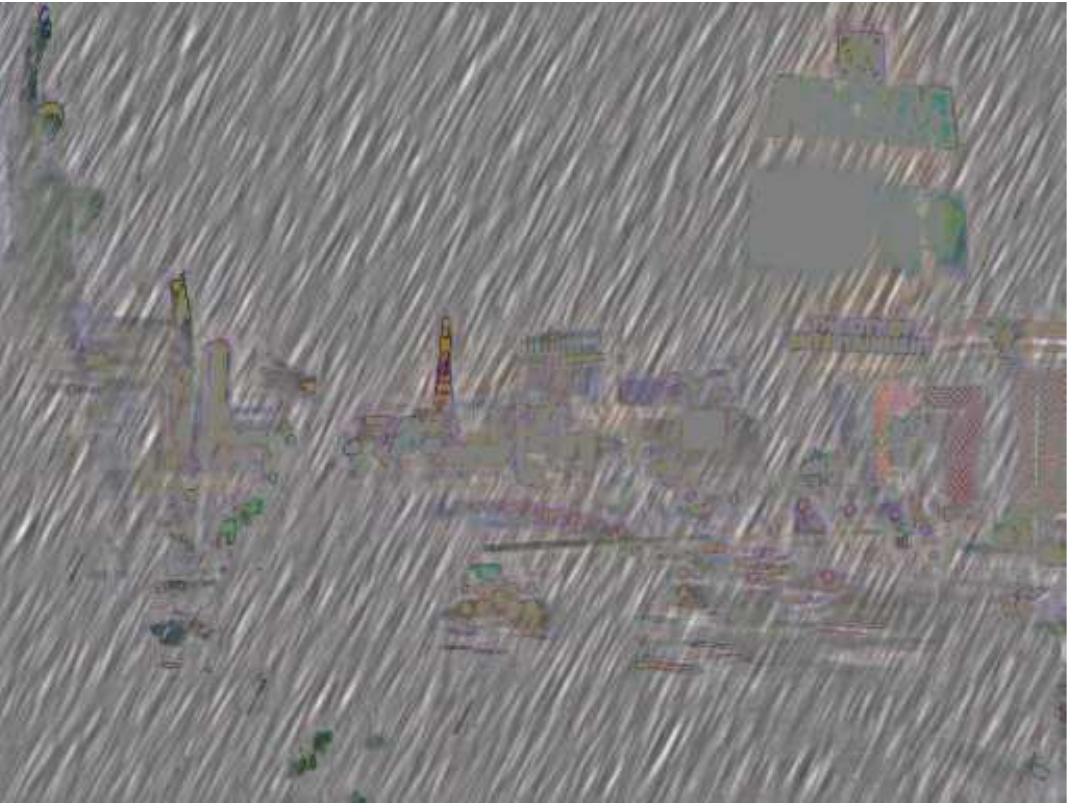}&
                \includegraphics[width=0.7in]{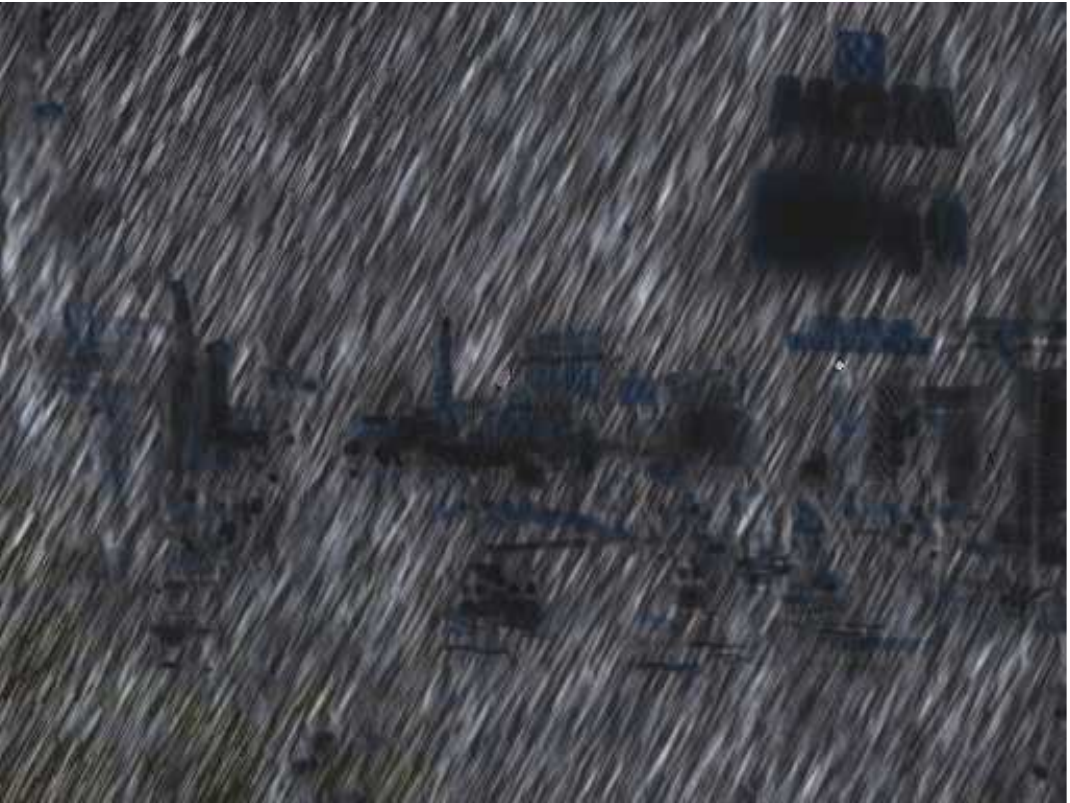}&
                \includegraphics[width=0.7in]{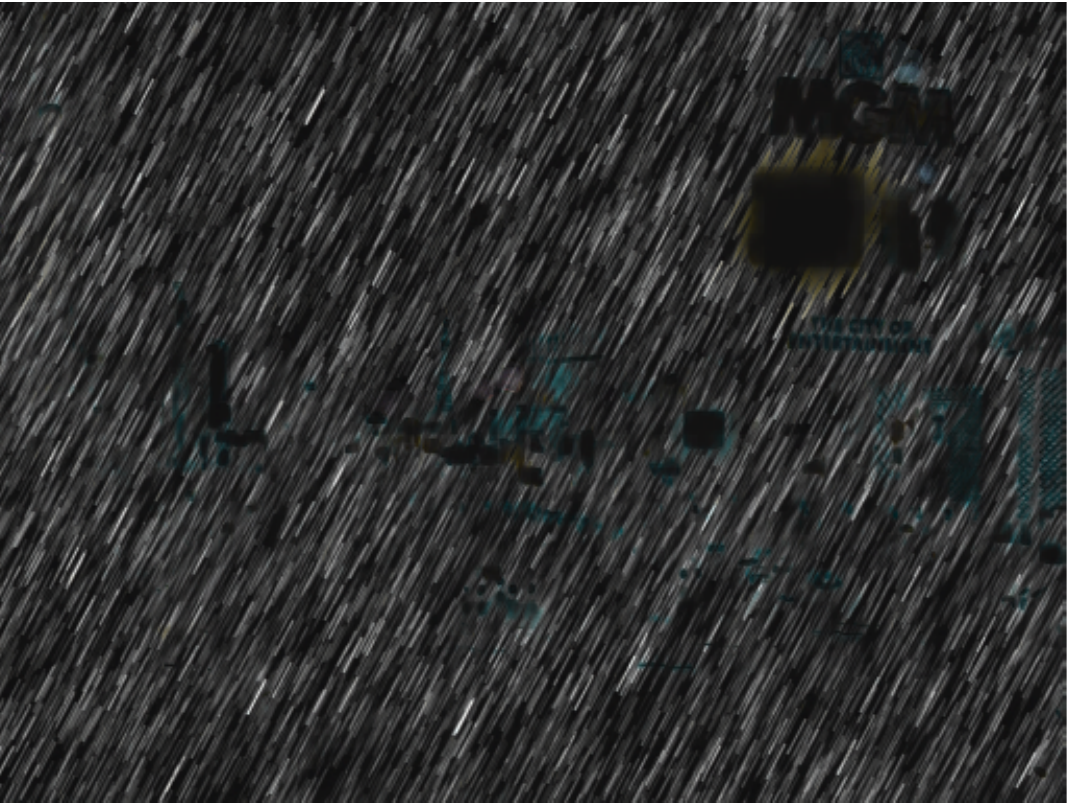}\\
                \vspace{0.5mm}

                \includegraphics[width=0.7in]{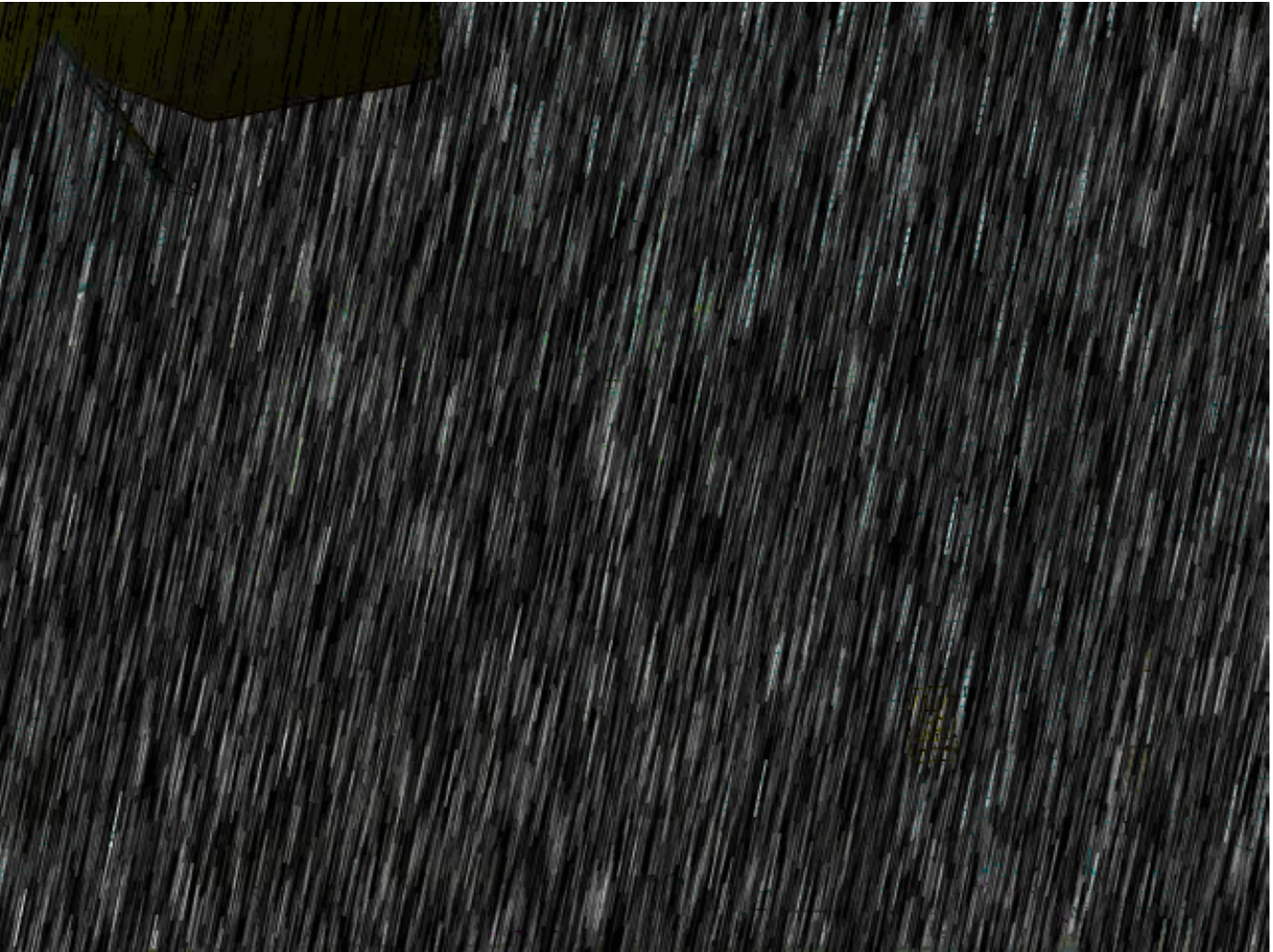}&
                \includegraphics[width=0.7in]{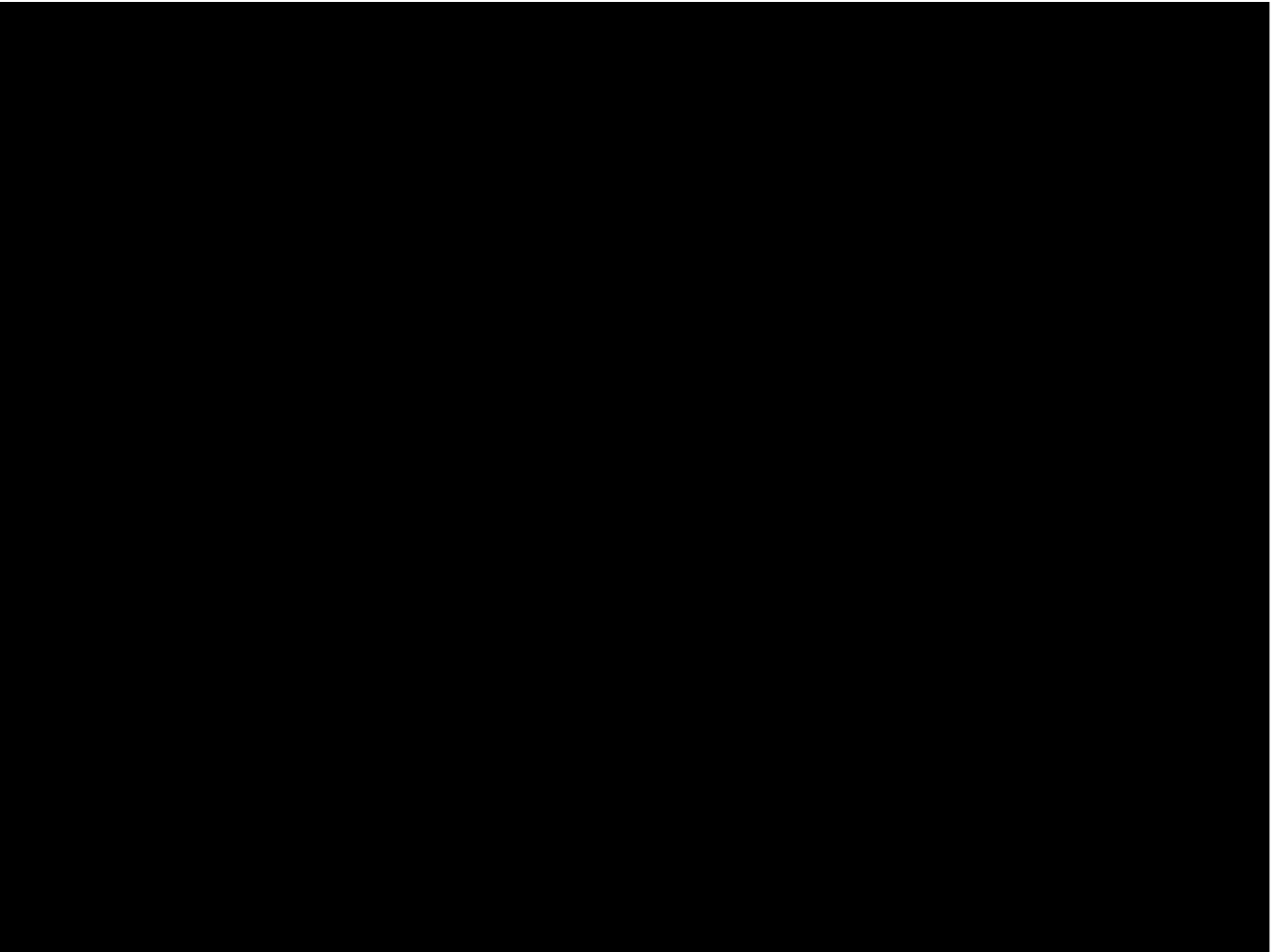}&
                \includegraphics[width=0.7in]{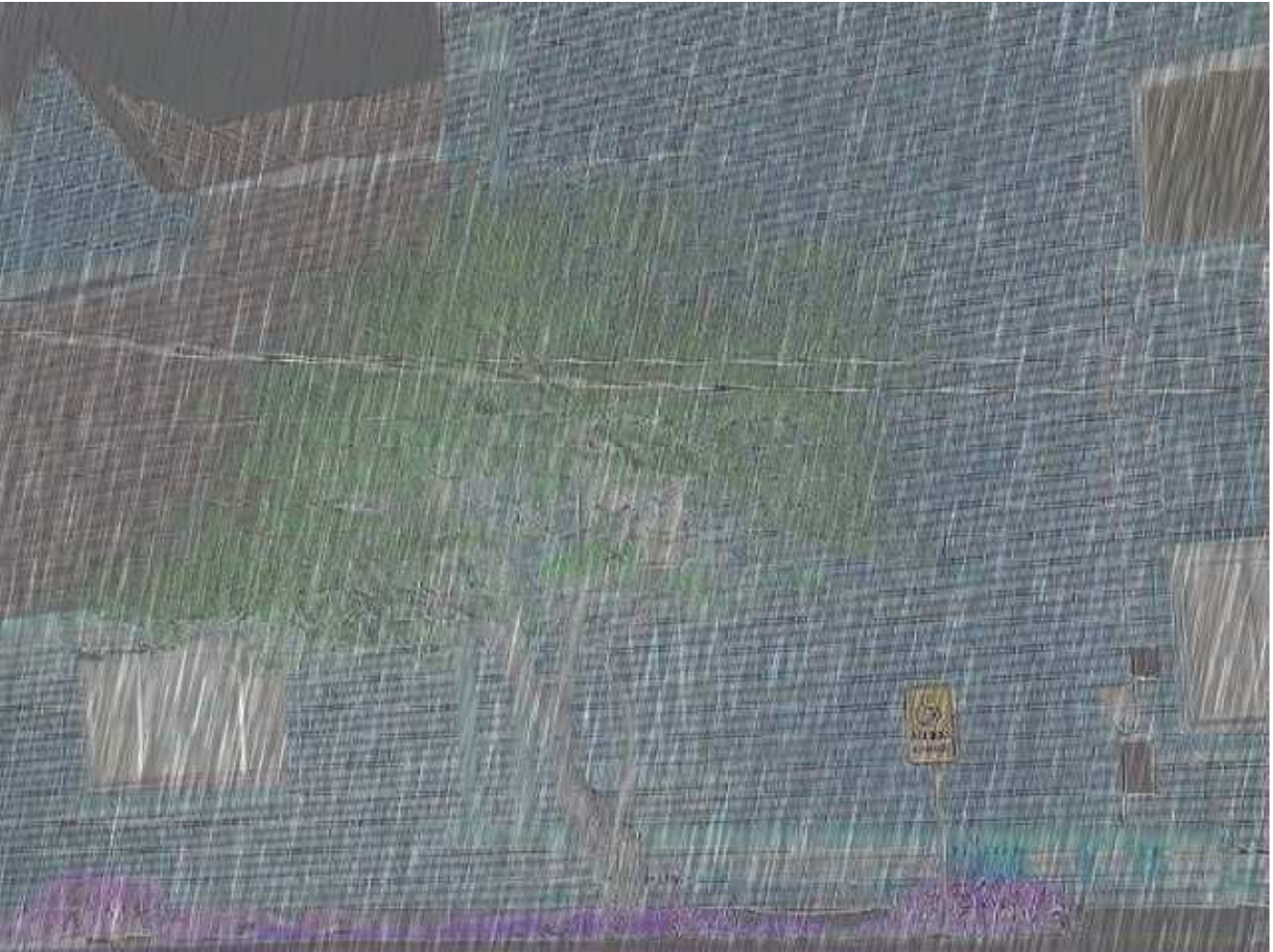}&
                \includegraphics[width=0.7in]{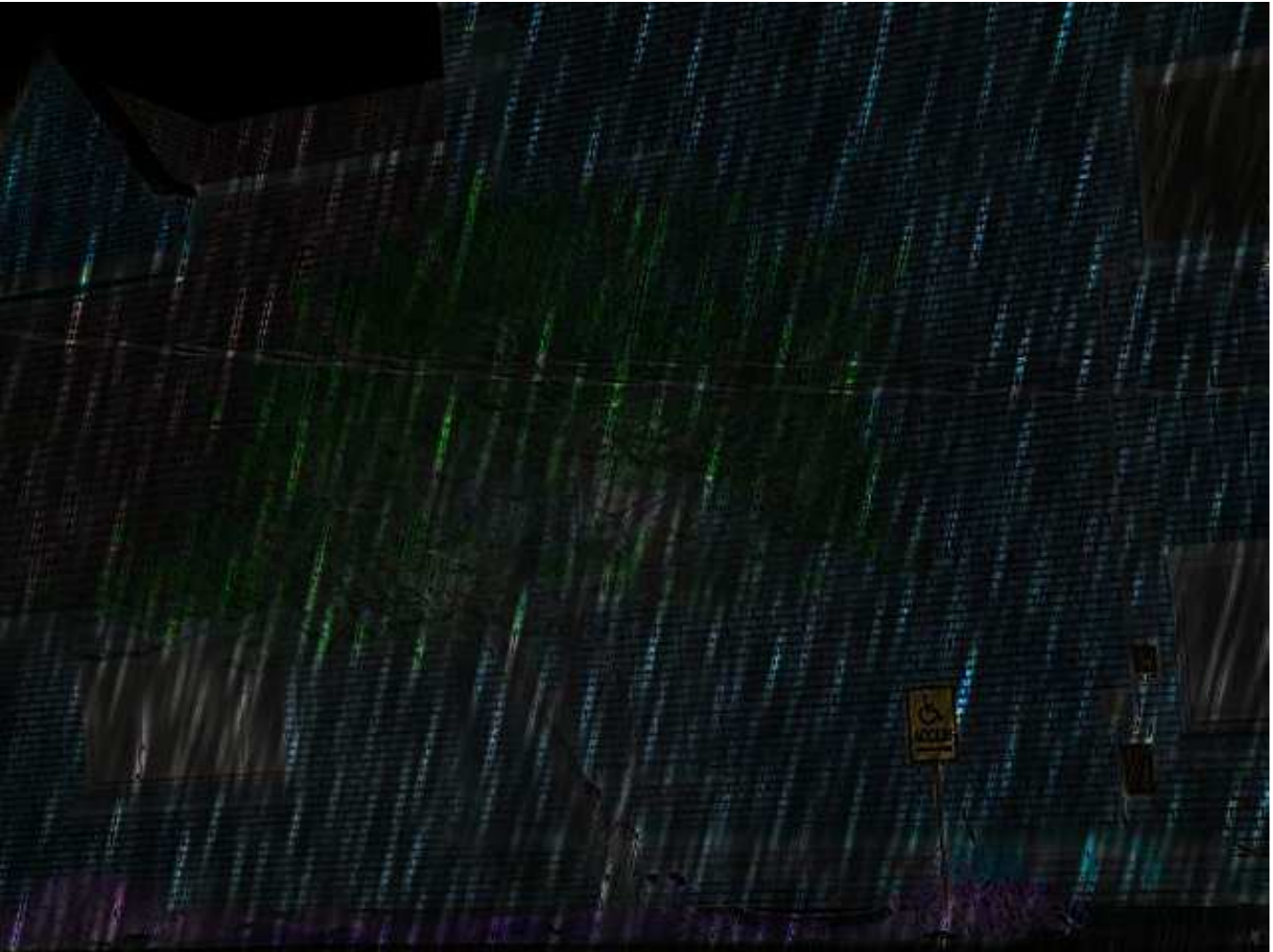}&
                \includegraphics[width=0.7in]{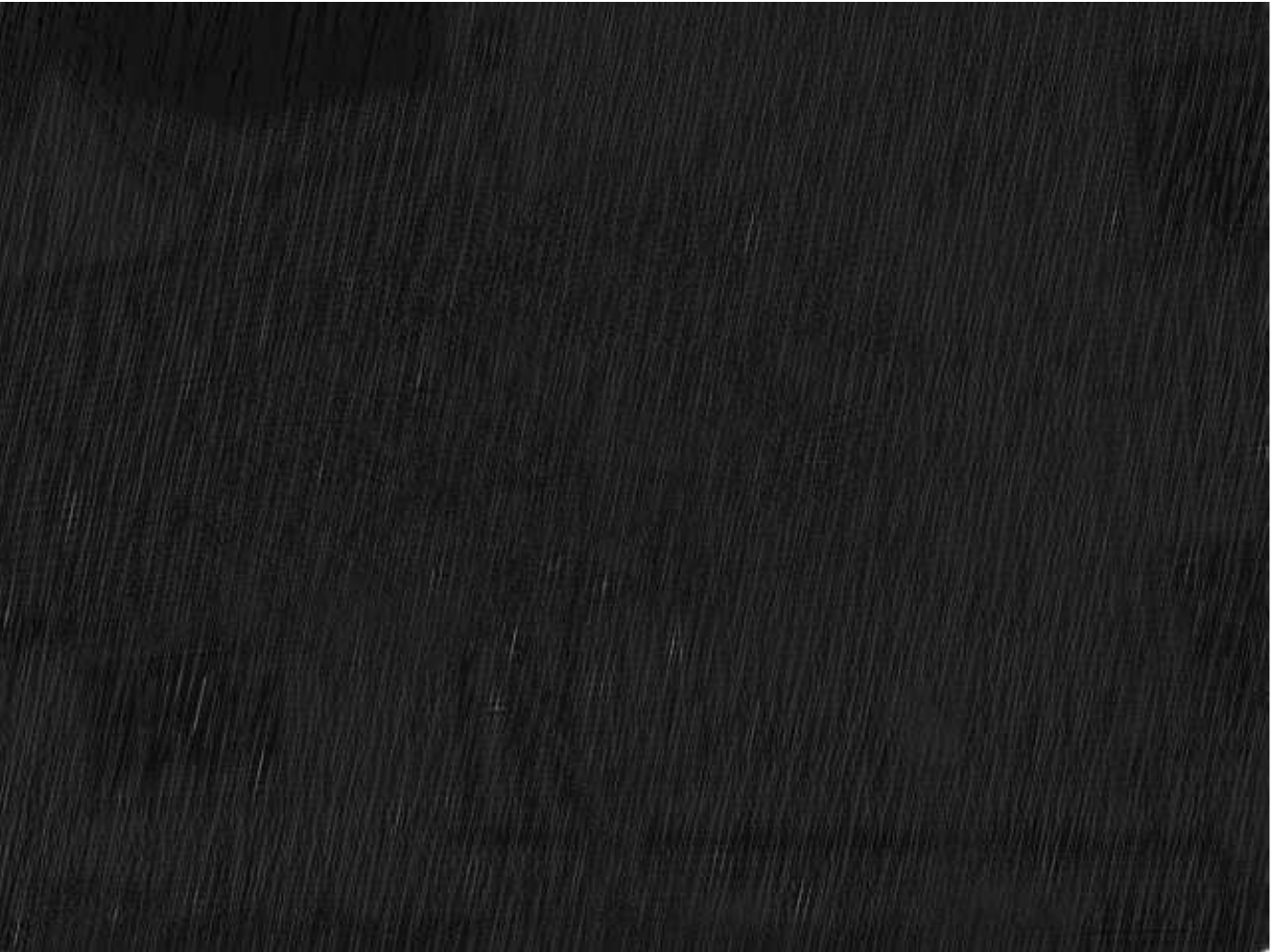}&
                \includegraphics[width=0.7in]{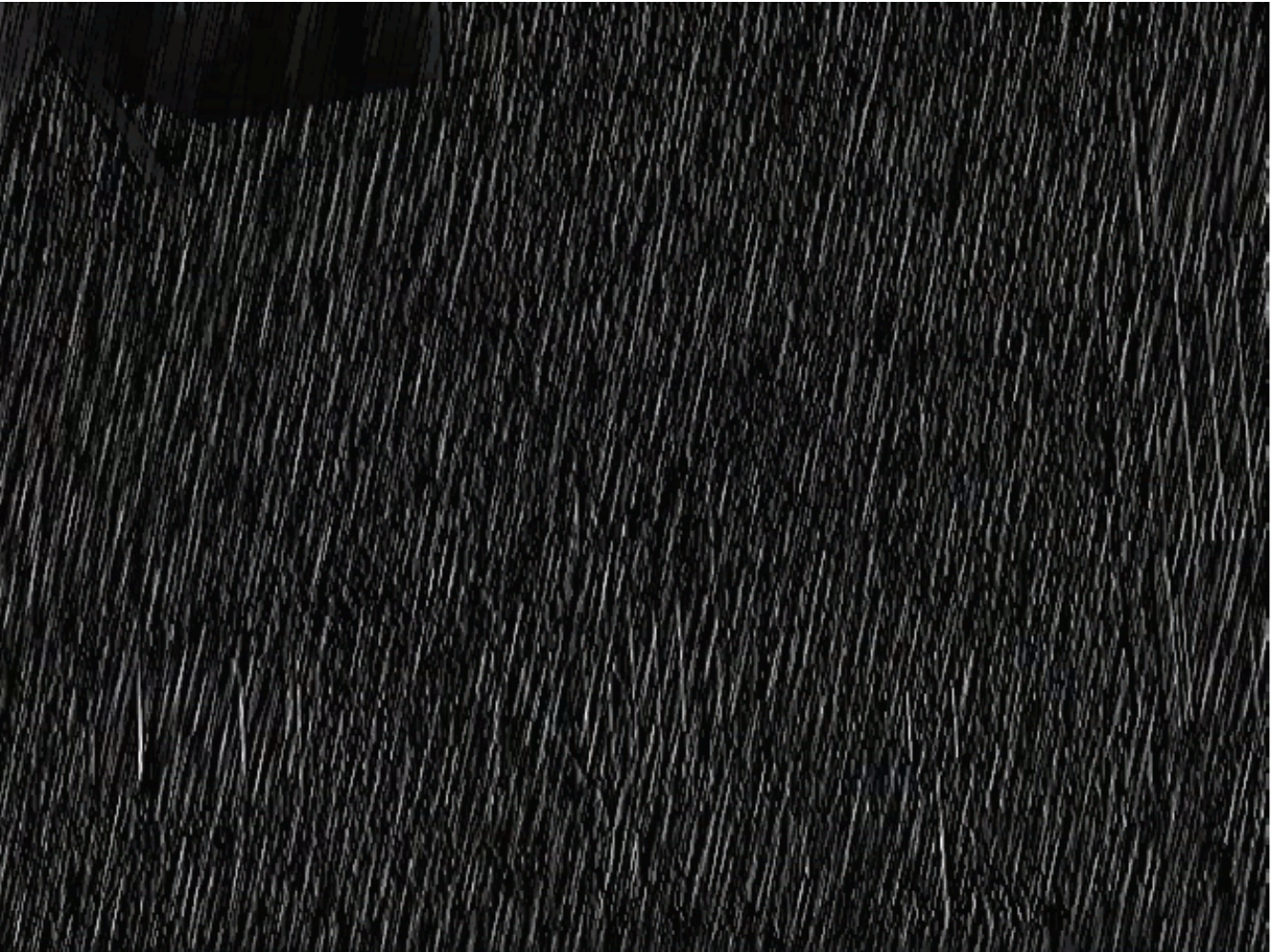}&
                \includegraphics[width=0.7in]{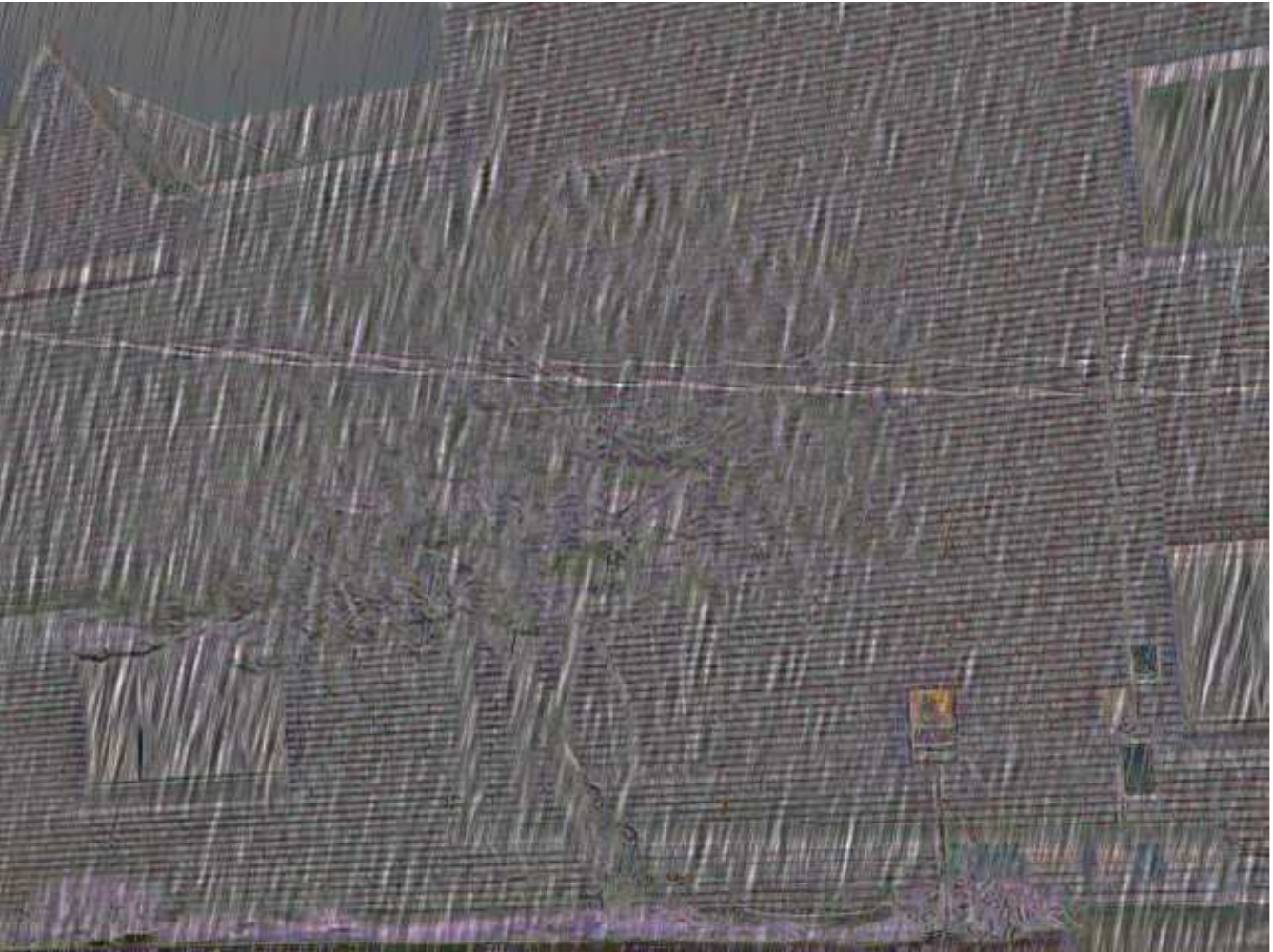}&
                \includegraphics[width=0.7in]{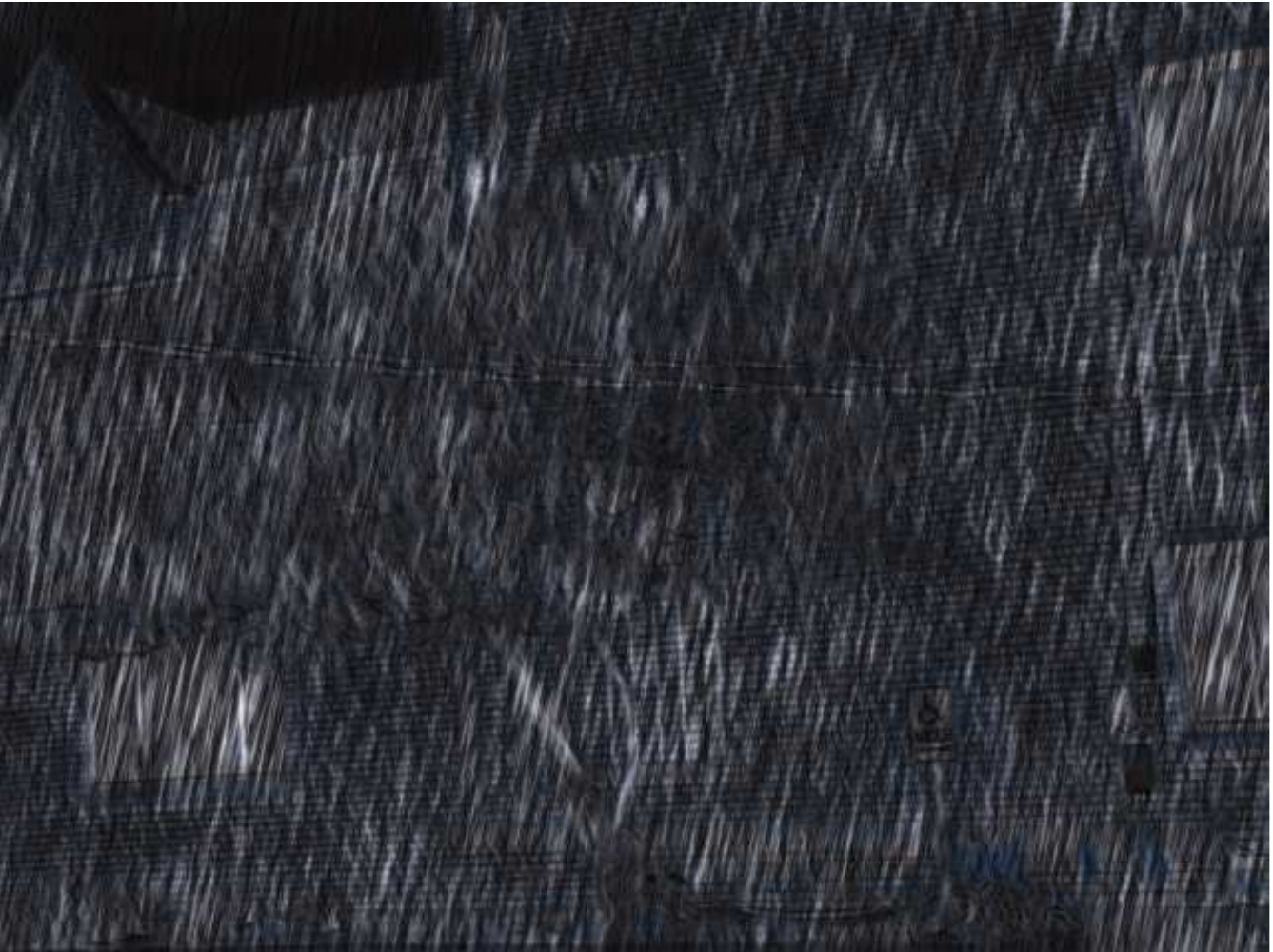}&
                \includegraphics[width=0.7in]{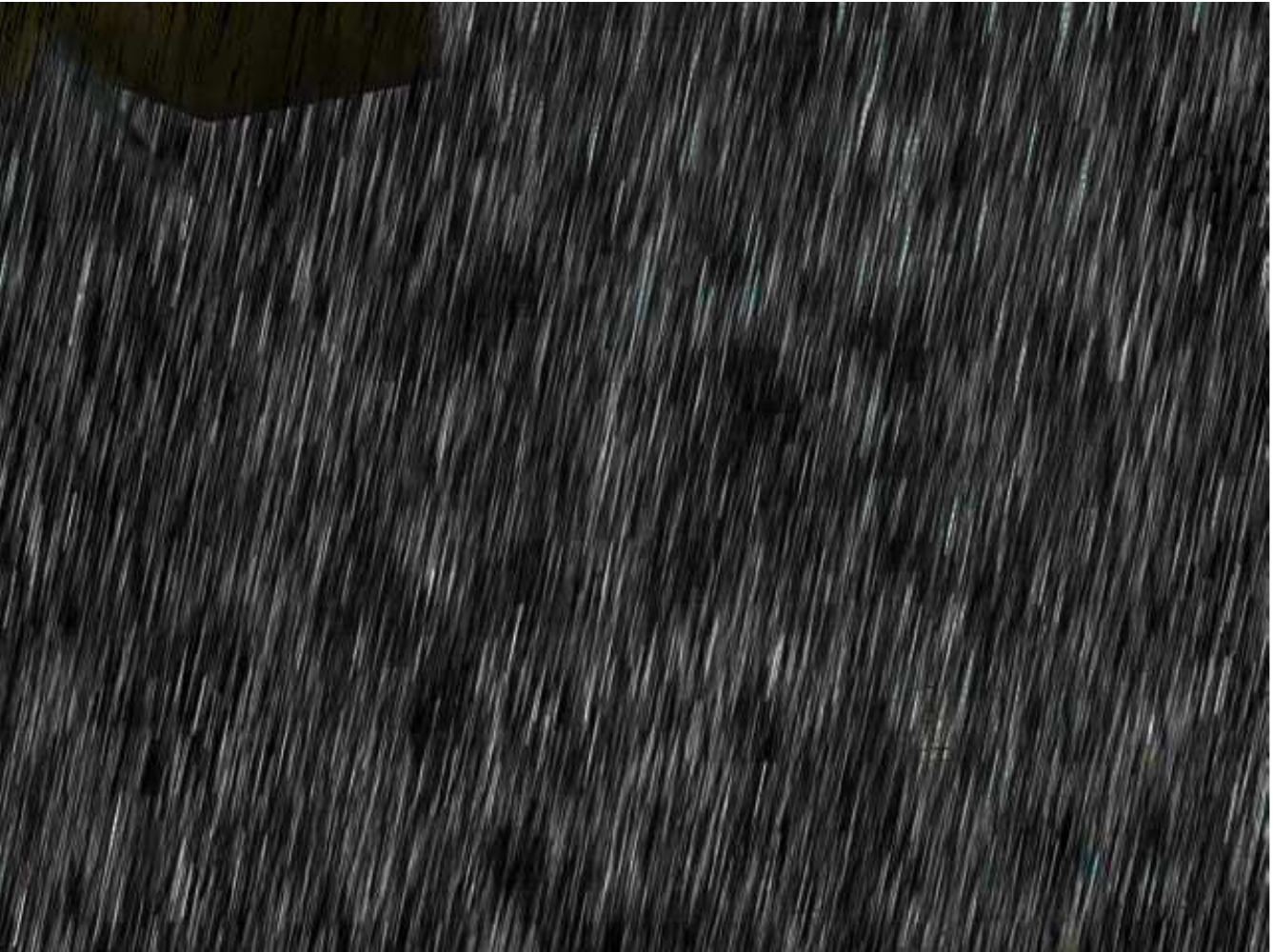}\\
                (a)&
                (b)&
                (c)&
                (d)&
                (e)&
                (f)&
                (g)&
                (h)&
                (i)\\

\end{tabular}
\caption{The rain streak images of the rain streak removal results by different methods on 3 synthetic rain images (road, night, and street) by our simulating method. From left to right: (a) the background, (b) the rainy images, the derain results by (c) DID \cite{zhang2018density}, (d) DSC \cite{luo2015removing}, (e) LP \cite{Li2014Single}, (f) UGSM \cite{Deng2018A}, (g) CNN \cite{fu2017clearing}, (h) DDN \cite{fu2017removing}, and (i) KGCNN.}
\label{synthetic-visual-our-streak}
\end{center}

\end{figure*}

For the second dataset, we generate another 3 synthesized rainy images (road, night, and street) for test.
Some of resulted derain images by different methods are selected to be shown in Fig. \ref{synthetic-visual-our} and Fig. \ref{synthetic-visual-our-streak}.
The visual results also demonstrate that the KGCNN method not only removes rain streaks completely, but also preserves the background information well.
We report the quantitative performance of the derain results obtained by different approaches in Table \ref{synthetic-quant-our}, which shows the superiority of the KGCNN method.

\begin{table}[htb]
\renewcommand\arraystretch{0.9}\setlength{\tabcolsep}{1.5pt}
\caption{Quantitative comparisons of rain streak removal results by DID \cite{zhang2018density}, DSC \cite{luo2015removing}, LP \cite{Li2014Single}, UGSM \cite{Deng2018A}, CNN \cite{fu2017clearing}, DDN \cite{fu2017removing}, and KGCNN on 3 synthetic rainy images by our simulating method.}
\label{synthetic-quant-our}
\begin{center}
\begin{tabular}{c|c|cccccc}
\Xhline{1.2pt}
    Images  &Method &PSNR &SSIM &FSIM &UIQI &GMSD &Time (s)\\
\Xhline{0.8pt}
\multirow{8}[0]{*}{\bf{road}}
                      &rainy    &18.171    &0.809    &0.909    &0.824    &0.150    &-      \\
                      &DID    &22.409    &0.889    &0.927    &1.005    &0.104    &\bf{0.625}      \\
                      &DSC    &22.302    &0.863    &0.926    &\bf{1.164}    &0.122    &197.776      \\
                      &LP     &22.227    &0.889    &0.919    &0.911     &0.122  &298.329\\
                      &UGSM    &22.763    &0.909    &0.933    &0.911    &0.105    &11.703      \\
                      &TIP    &18.045    &0.836    &0.911    &0.830    &0.124    &12.419      \\
                      &CVPR    &23.325    &0.902    &0.921    &0.875    &0.117    &0.843      \\
                      &KGCNN    &\bf{27.851}    &\bf{0.957}    &\bf{0.957}    &0.978    &\bf{0.076}    &9.023      \\
\hline
\multirow{8}[0]{*}{\bf{night}}
                      &rainy    &18.506    &0.539    &0.787    &0.753    &0.261    &-      \\
                      &DID    &25.294    &0.851    &0.924    &0.813    &0.099    &\bf{0.625}      \\
                      &DSC    &22.219    &0.598    &0.835    &0.904    &0.208    &200.172      \\
                      &LP     &23.034   &0.780    &0.872     &0.806    &0.175    &297.194\\
                      &UGSM    &20.626    &0.641    &0.824    &0.784    &0.225    &5.568      \\
                      &TIP    &19.215    &0.648    &0.851    &0.757    &0.193    &9.185      \\
                      &CVPR    &23.781    &0.674    &0.882    &0.846    &0.172    &0.375      \\
                      &KGCNN    &\bf{31.231}    &\bf{0.957}    &\bf{0.970}    &\bf{0.930}    &\bf{0.046}    &8.848      \\
\hline
\multirow{8}[0]{*}{\bf{tree}}
                      &rainy    &18.798    &0.837    &0.882    &0.968    &0.162    &-      \\
                      &DID    &22.359    &0.865    &0.902    &0.975    &0.128    &\bf{0.640}      \\
                      &DSC    &21.519    &0.846    &0.898    &0.992    &0.141    &307.007      \\
                      &LP     &22.393    &0.894    &0.907    &0.987    &0.145    &276.631\\
                      &UGSM    &22.382    &0.913    &0.921    &0.985    &0.111    &8.236      \\
                      &TIP    &18.793    &0.872    &0.903    &0.969    &0.126    &14.075      \\
                      &CVPR    &24.013    &0.900    &0.908    &0.994    &0.121    &0.845      \\
                      &KGCNN    &\bf{28.031}    &\bf{0.966}    &\bf{0.954}    &\bf{0.997}    &\bf{0.073}    &8.751      \\
\hline
\Xhline{1.2pt}
\end{tabular}
\end{center}
\end{table}

\begin{figure*}[!htb]
\renewcommand\arraystretch{0.8}\setlength{\tabcolsep}{1.8pt}
\begin{center}
\begin{tabular}{ccccccccc}
                ~\includegraphics[width=0.7in]{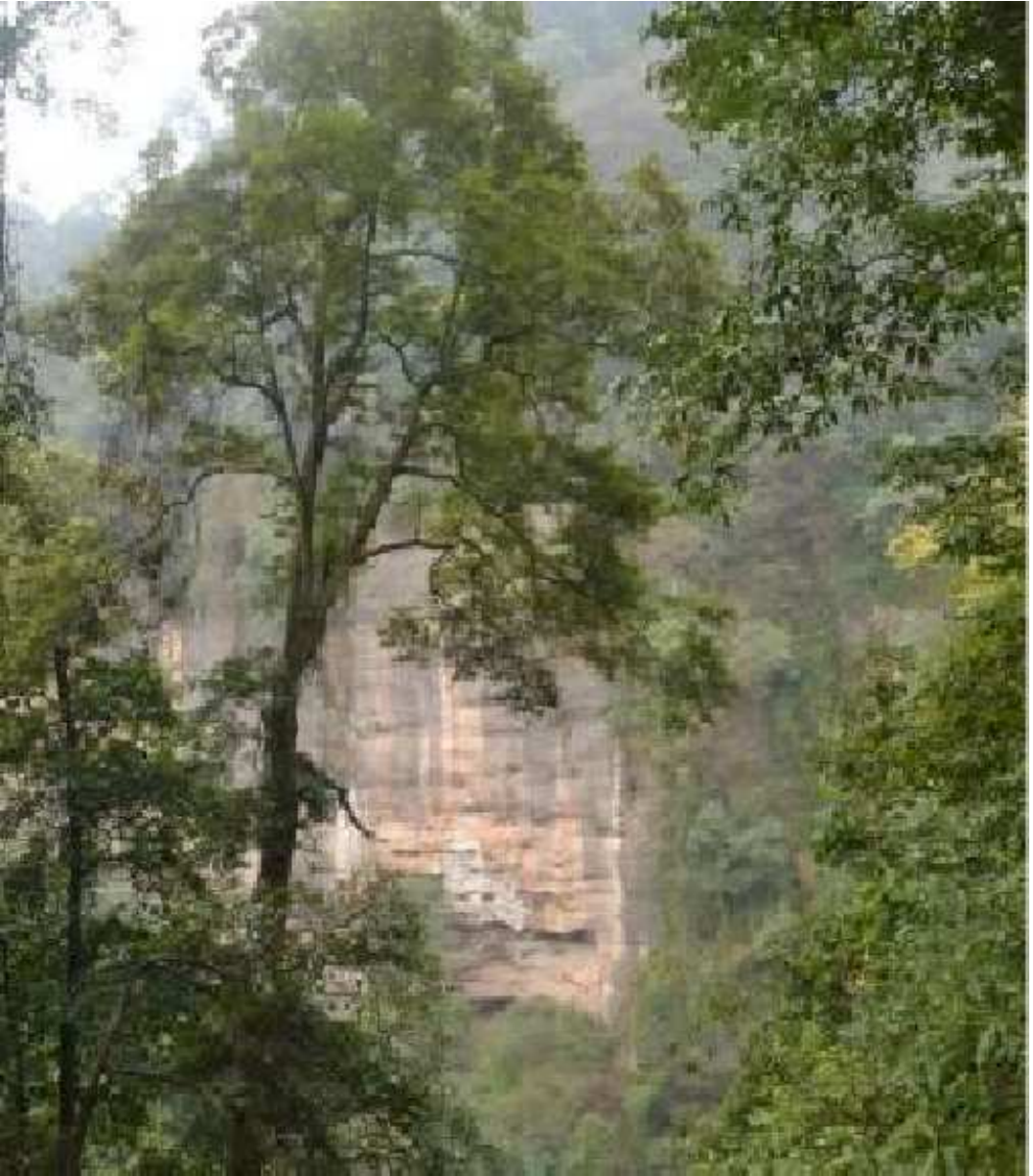}&
                \includegraphics[width=0.7in]{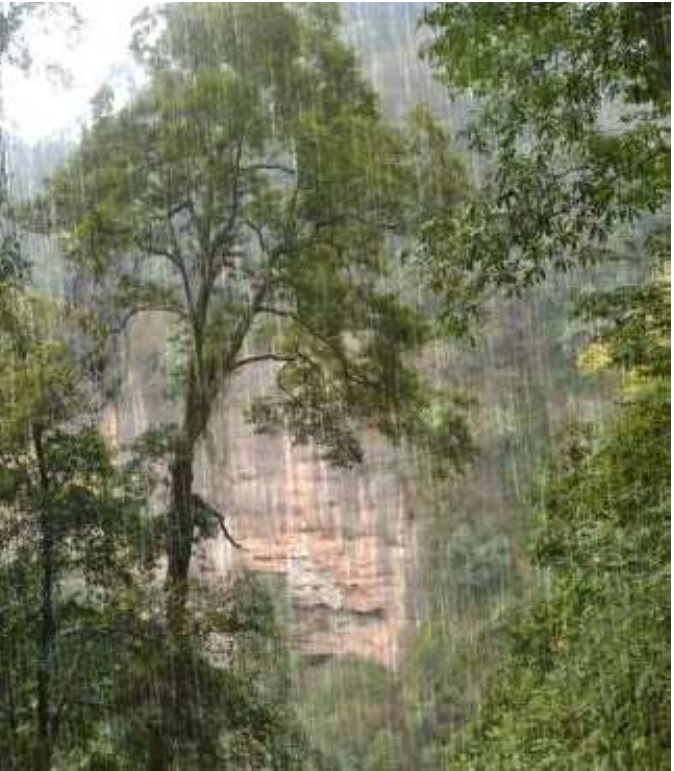}&
                \includegraphics[width=0.7in]{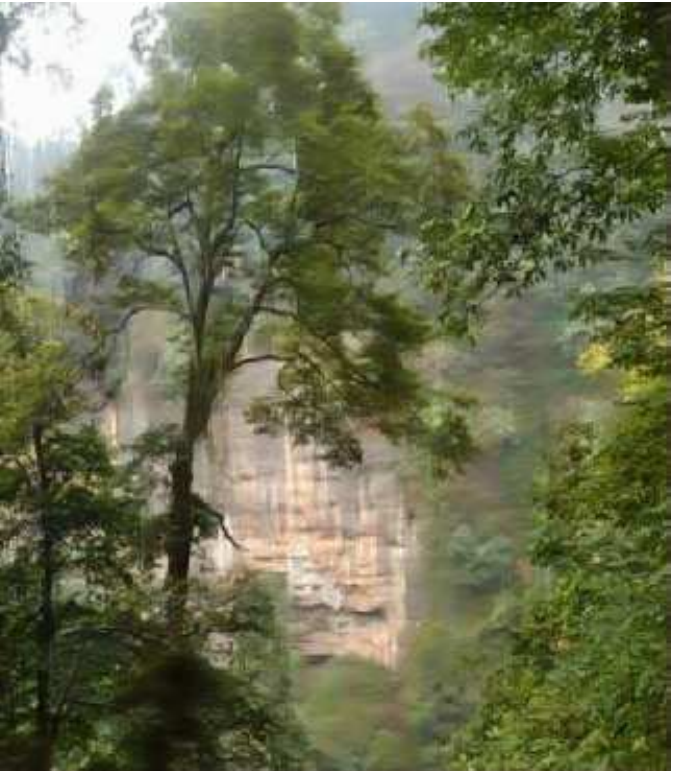}&
                \includegraphics[width=0.7in]{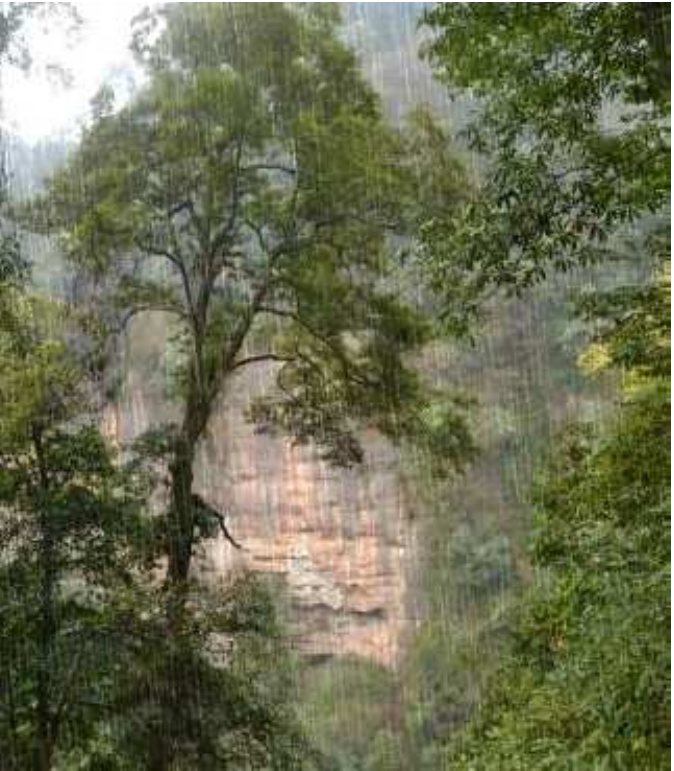}&
                \includegraphics[width=0.7in]{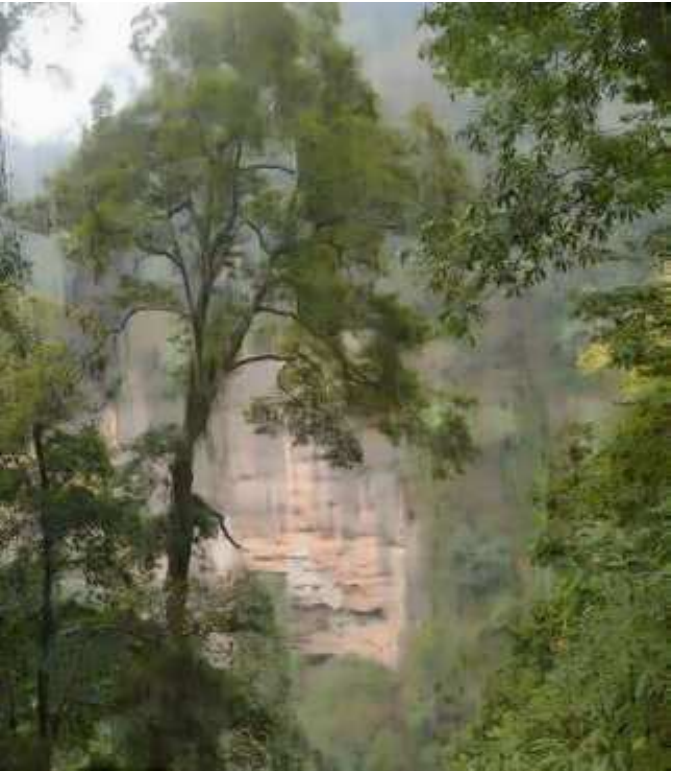}&
                \includegraphics[width=0.7in]{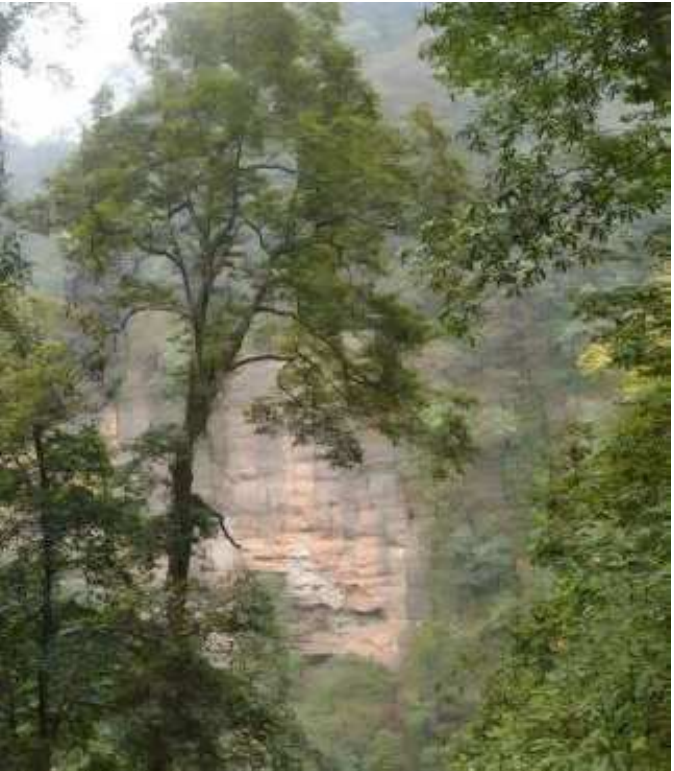}&
                \includegraphics[width=0.7in]{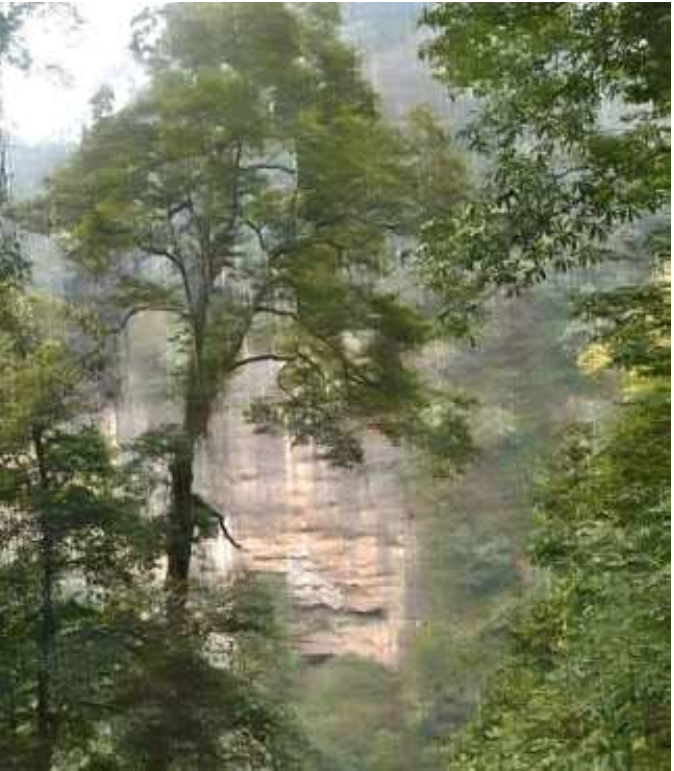}&
                \includegraphics[width=0.7in]{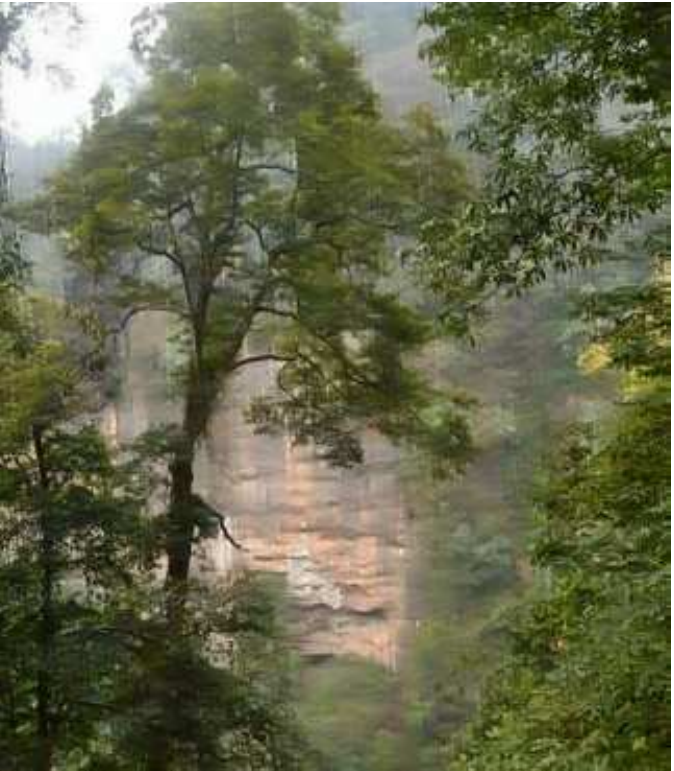}&
                \includegraphics[width=0.7in]{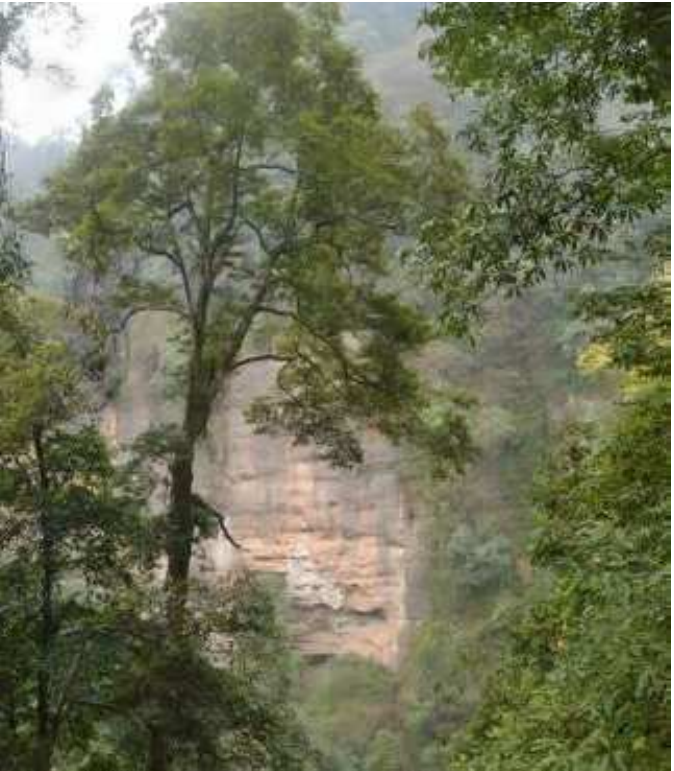}\\
                \vspace{0.5mm}

                \includegraphics[width=0.7in]{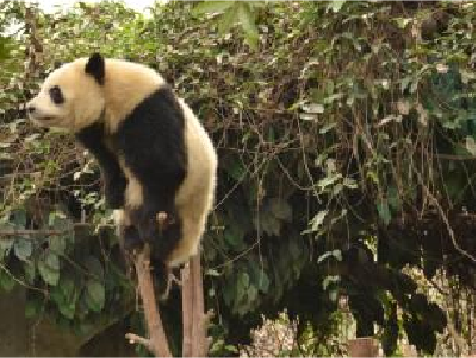}&
                \includegraphics[width=0.7in]{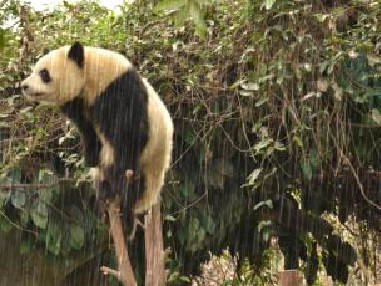}&
                \includegraphics[width=0.7in]{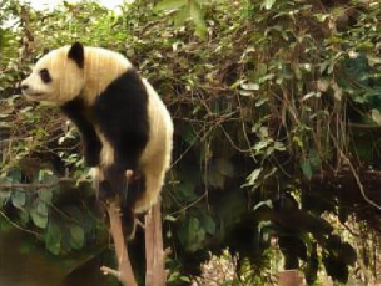}&
                \includegraphics[width=0.7in]{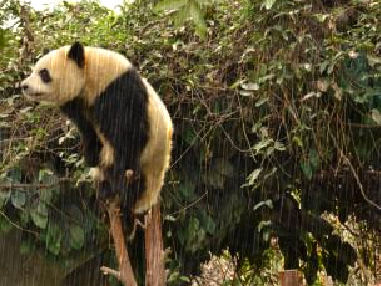}&
                \includegraphics[width=0.7in]{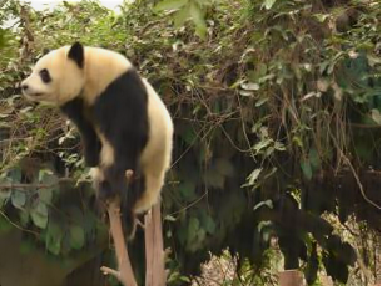}&
                \includegraphics[width=0.7in]{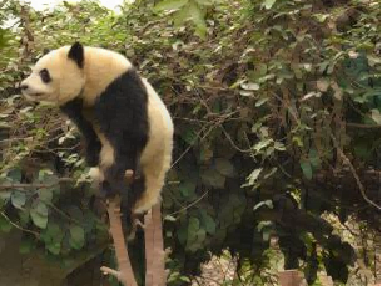}&
                \includegraphics[width=0.7in]{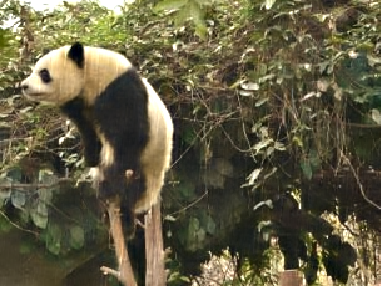}&
                \includegraphics[width=0.7in]{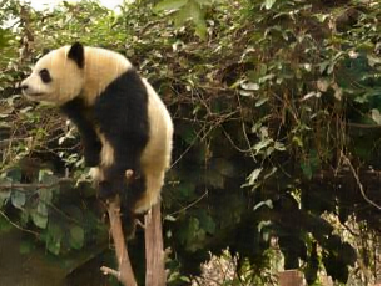}&
                \includegraphics[width=0.7in]{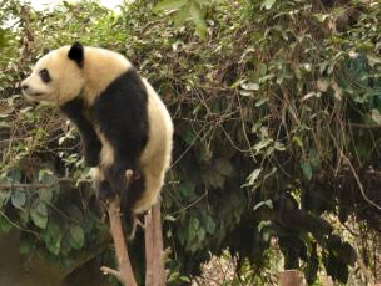}\\
                \vspace{0.5mm}

                \includegraphics[width=0.7in]{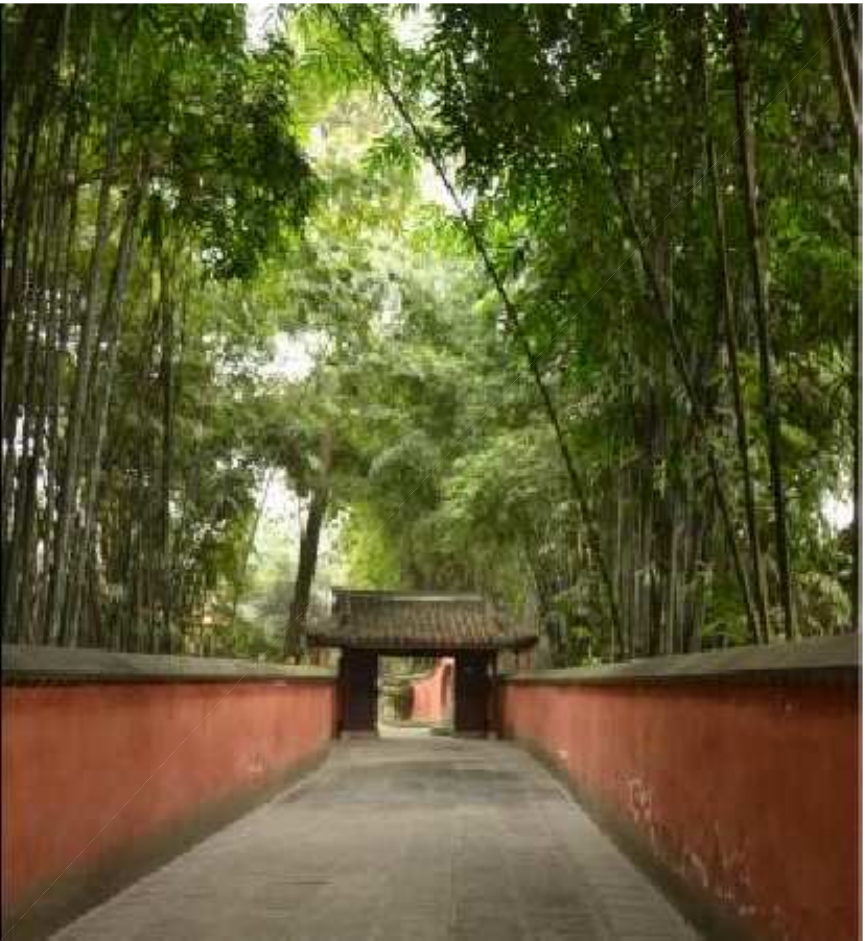}&
                \includegraphics[width=0.7in]{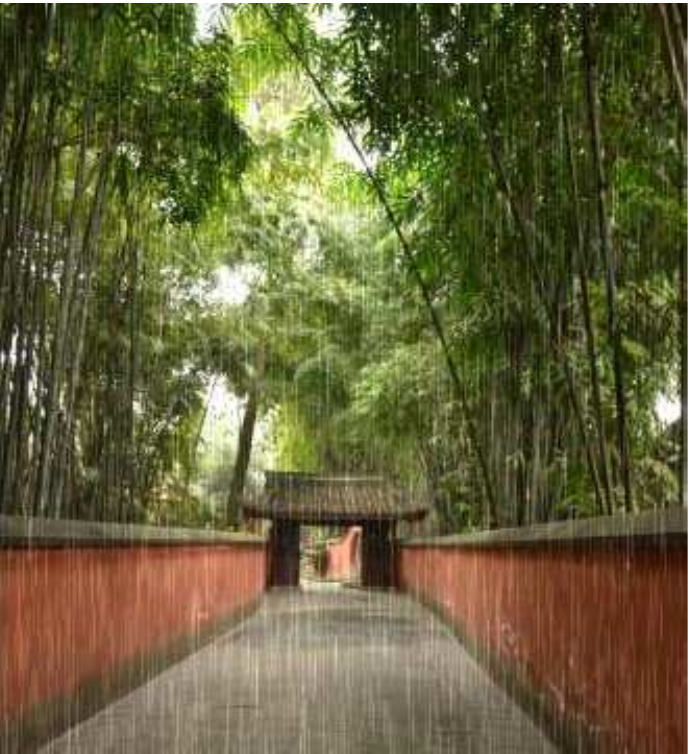}&
                \includegraphics[width=0.7in]{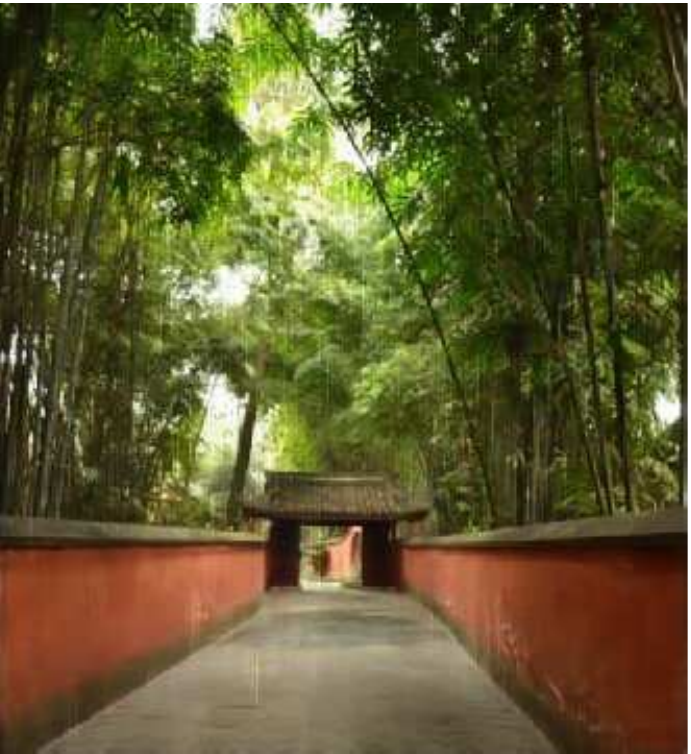}&
                \includegraphics[width=0.7in]{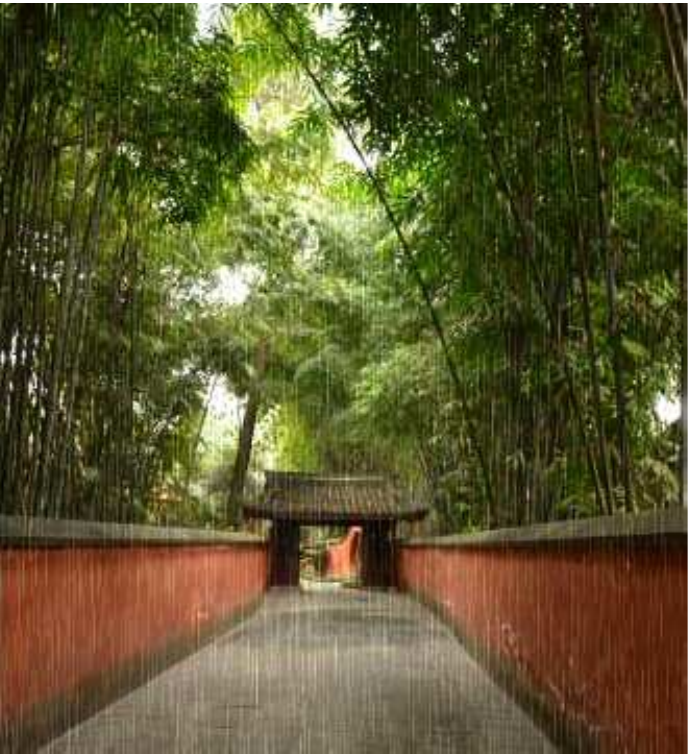}&
                \includegraphics[width=0.7in]{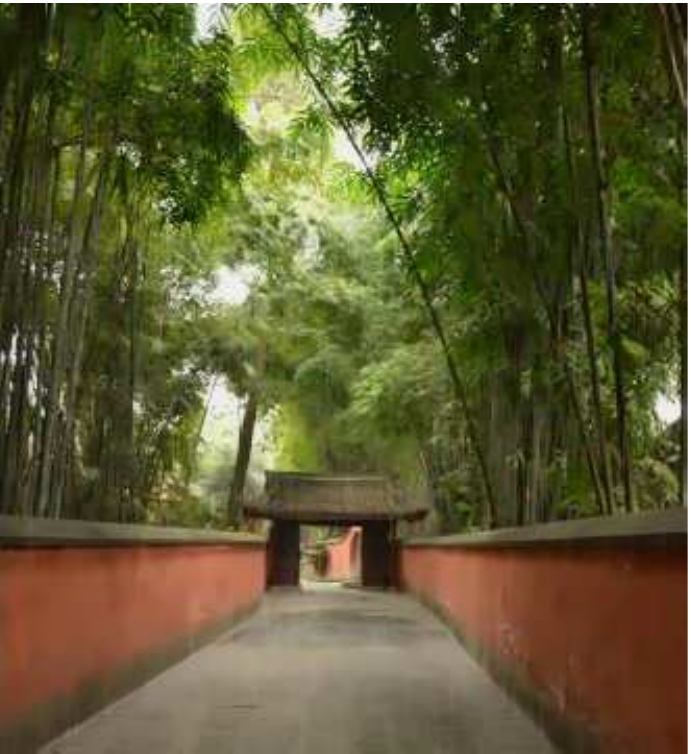}&
                \includegraphics[width=0.7in]{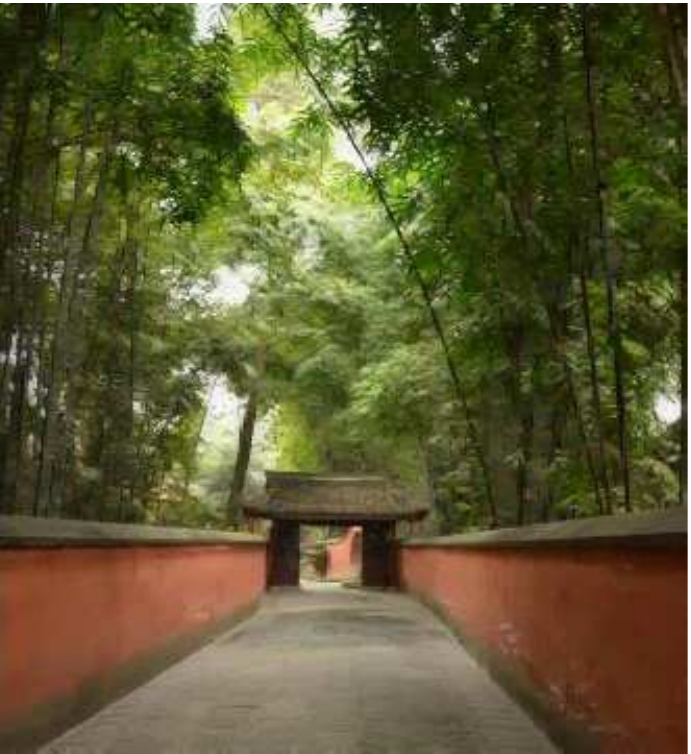}&
                \includegraphics[width=0.7in]{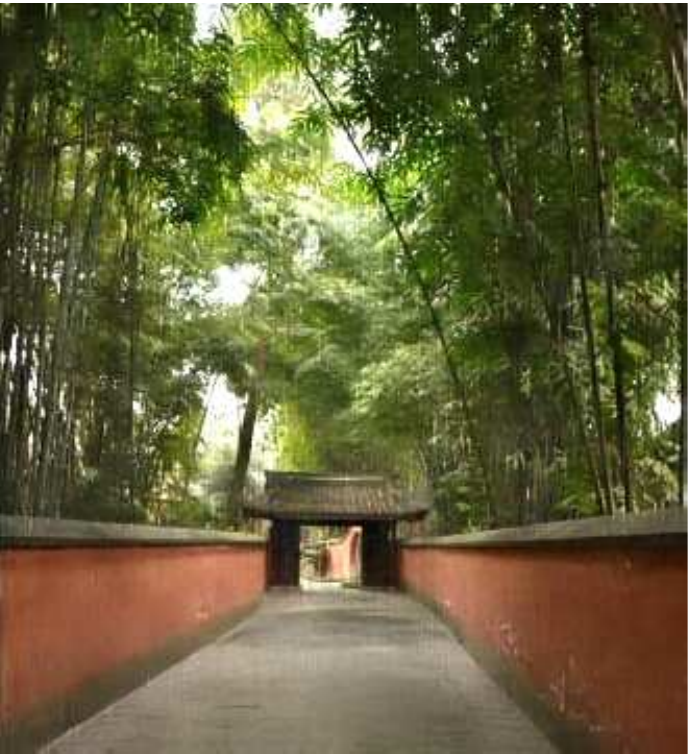}&
                \includegraphics[width=0.7in]{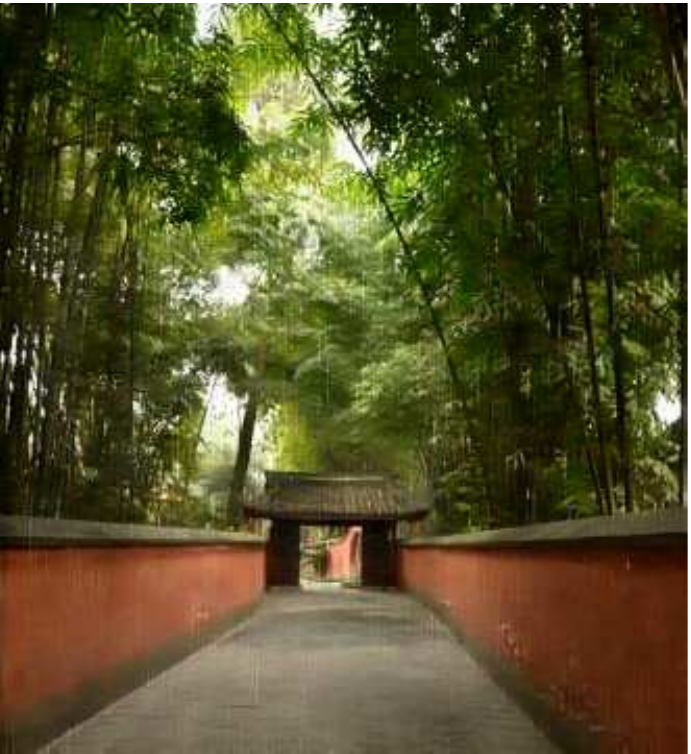}&
                \includegraphics[width=0.7in]{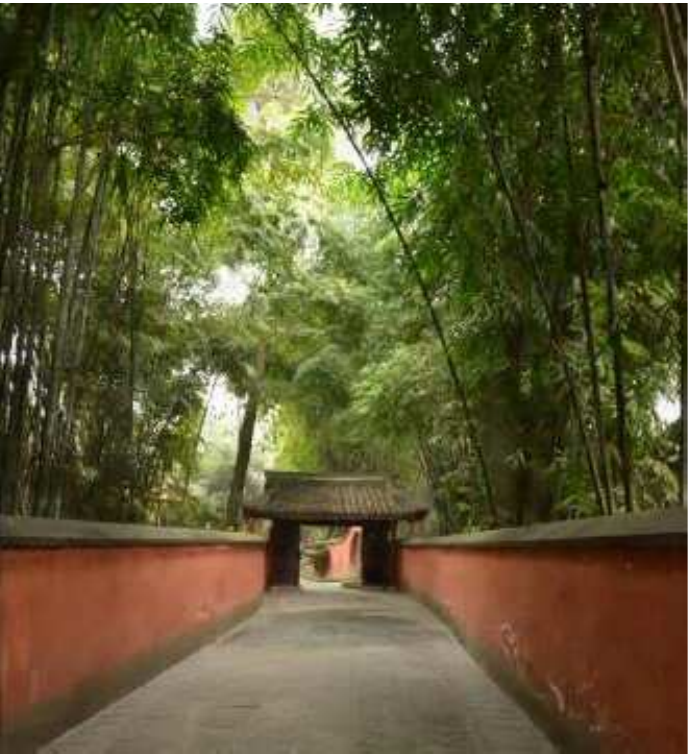}\\
                (a)&
                (b)&
                (c)&
                (d)&
                (e)&
                (f)&
                (g)&
                (h)&
                (i)\\
\end{tabular}

\caption{Rain streak removal results by different methods on 3 synthetic rainy images (tree, panda, and bamboo) selected from \cite{Deng2018A}. From left to right: (a) the background, (b) the rainy images, the derain results by (c) DID \cite{zhang2018density}, (d) DSC \cite{luo2015removing}, (e) LP \cite{Li2014Single}, (f) UGSM \cite{Deng2018A}, (g) CNN \cite{fu2017clearing}, (h) DDN \cite{fu2017removing}, and (i) KGCNN.}
\label{synthetic-visual-UGSM}
\end{center}
\end{figure*}

\begin{figure*}[!htb]
\renewcommand\arraystretch{0.8}\setlength{\tabcolsep}{1.8pt}
\begin{center}
\begin{tabular}{ccccccccc}
                ~\includegraphics[width=0.7in]{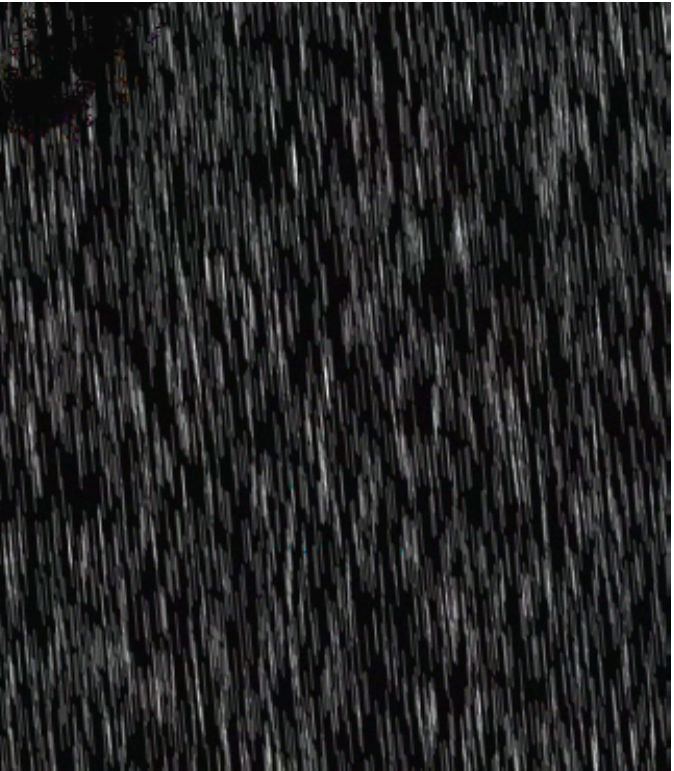}&
                \includegraphics[width=0.7in]{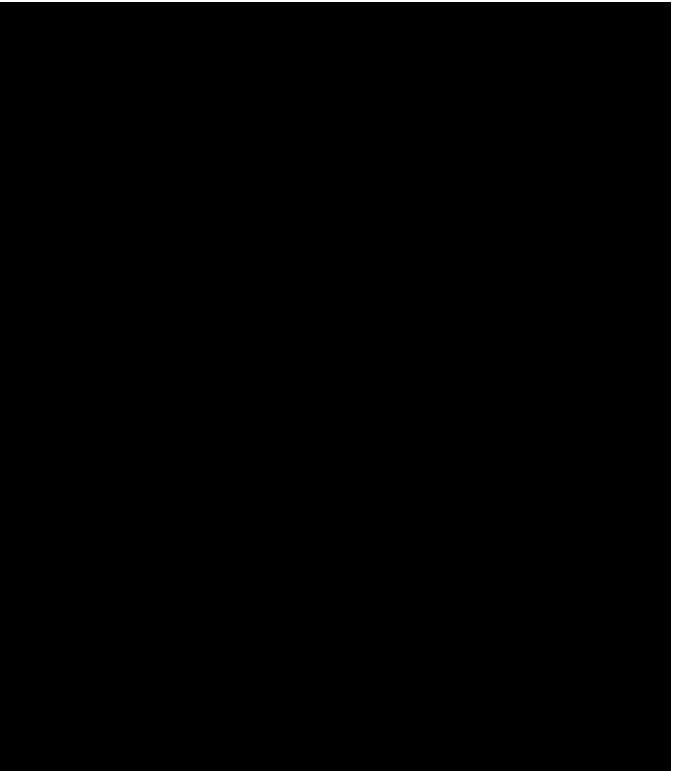}&
                \includegraphics[width=0.7in]{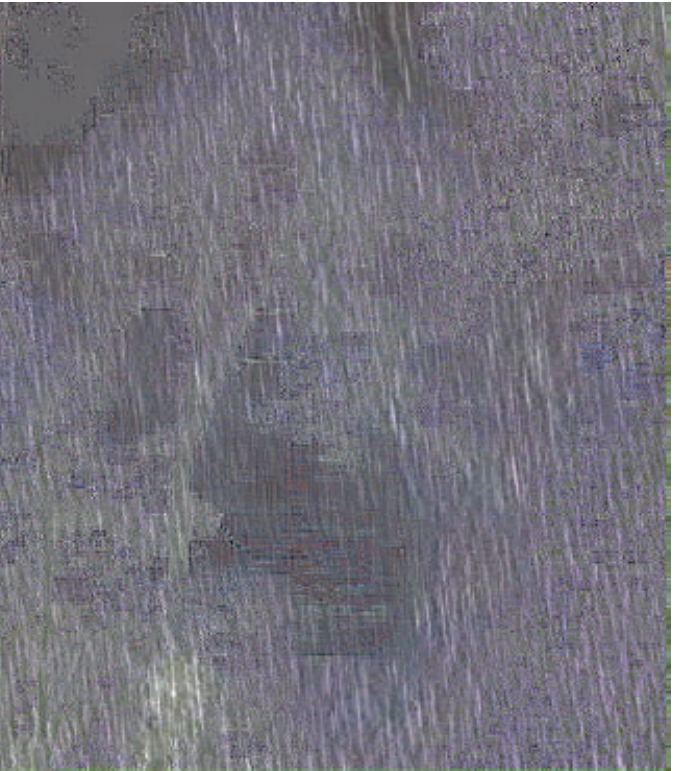}&
                \includegraphics[width=0.7in]{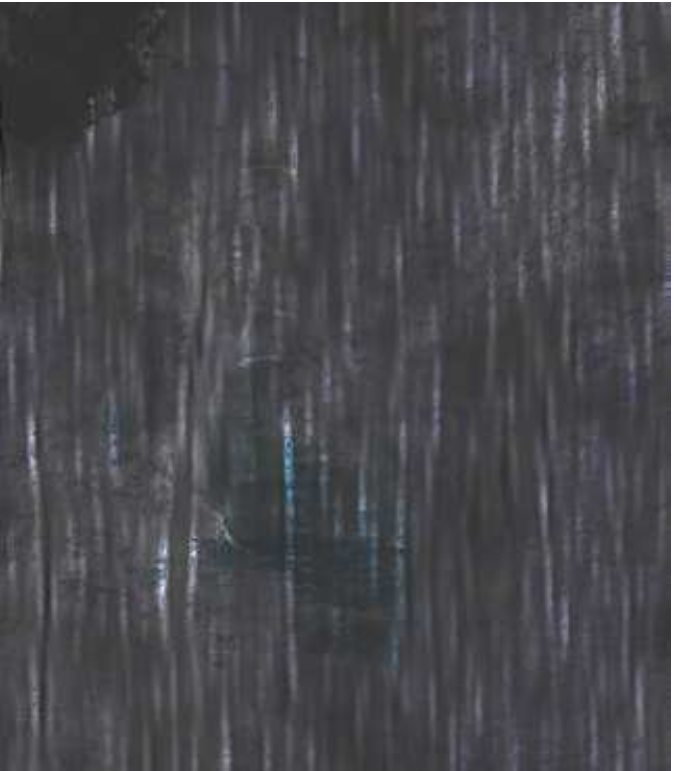}&
                \includegraphics[width=0.7in]{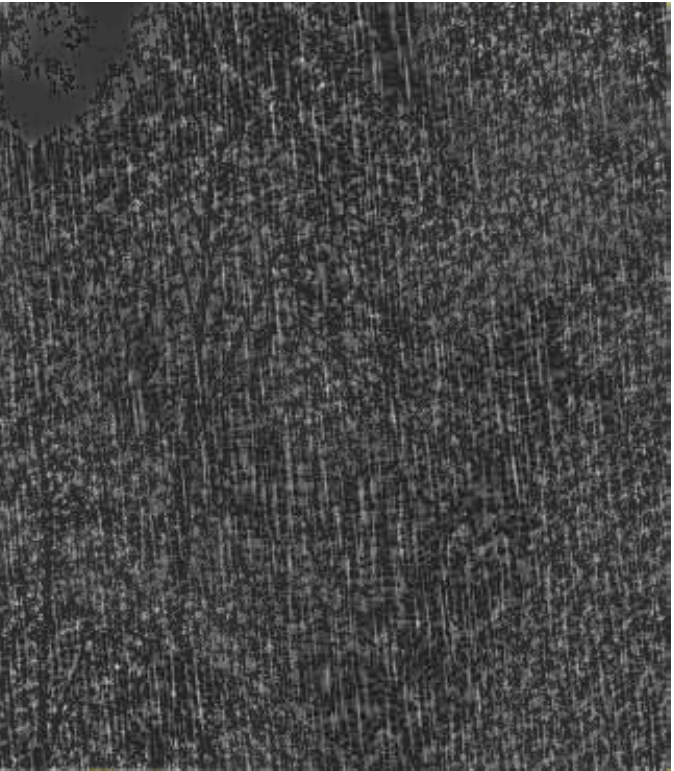}&
                \includegraphics[width=0.7in]{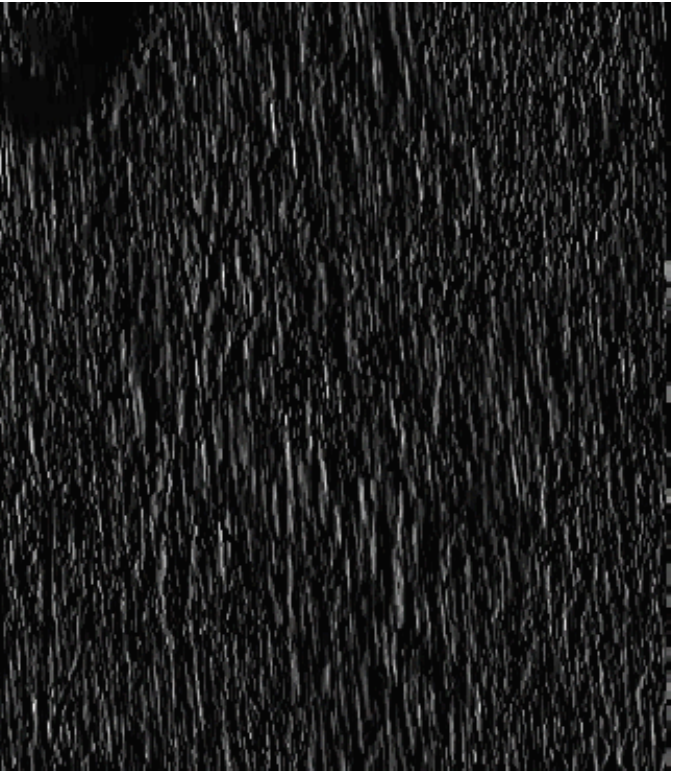}&
                \includegraphics[width=0.7in]{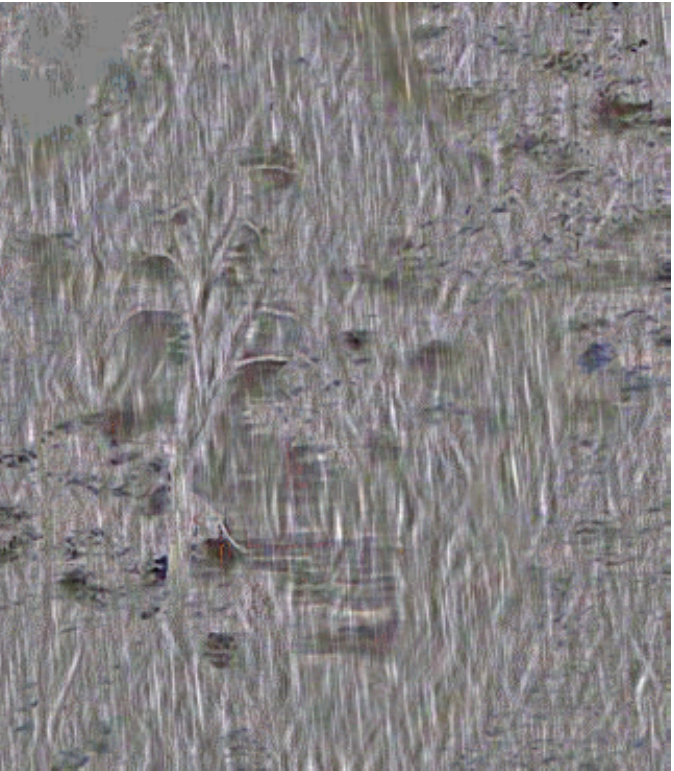}&
                \includegraphics[width=0.7in]{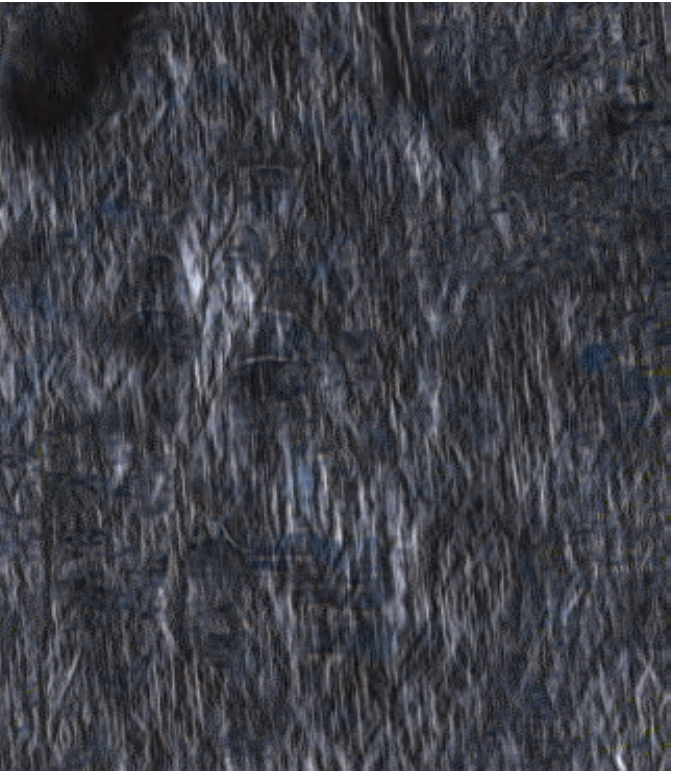}&
                \includegraphics[width=0.7in]{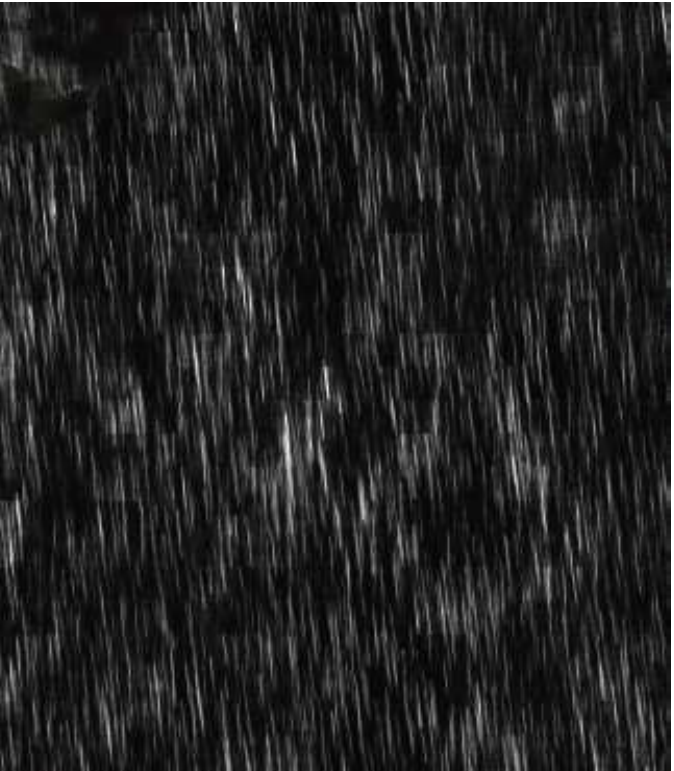}\\
                \vspace{0.5mm}

                \includegraphics[width=0.7in]{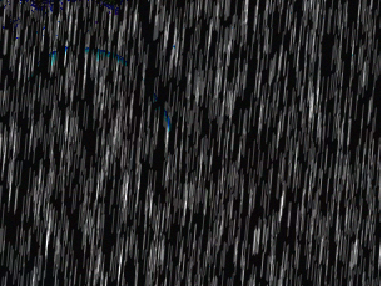}&
                \includegraphics[width=0.7in]{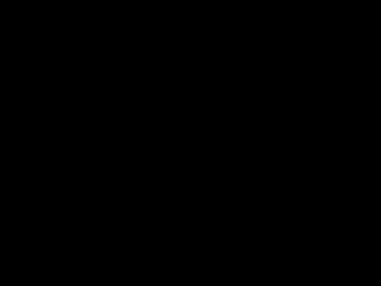}&
                \includegraphics[width=0.7in]{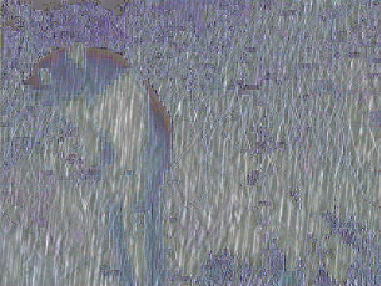}&
                \includegraphics[width=0.7in]{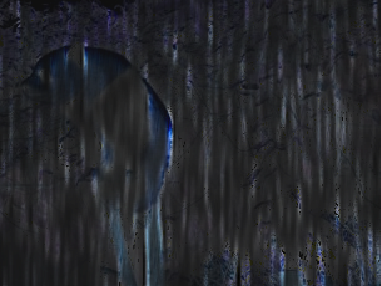}&
                \includegraphics[width=0.7in]{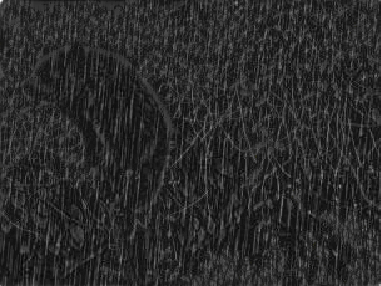}&
                \includegraphics[width=0.7in]{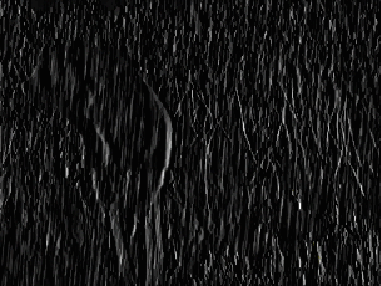}&
                \includegraphics[width=0.7in]{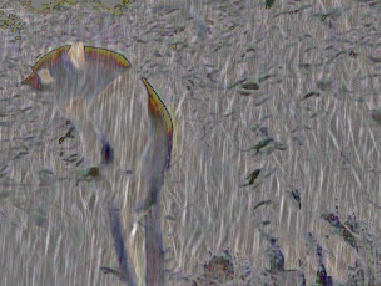}&
                \includegraphics[width=0.7in]{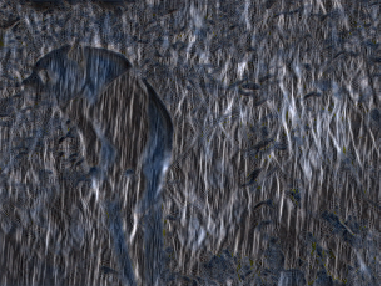}&
                \includegraphics[width=0.7in]{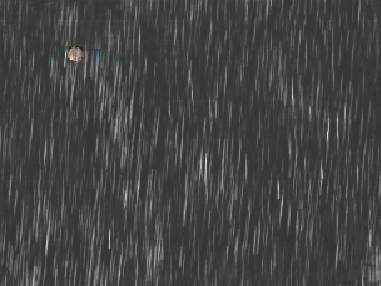}\\
                \vspace{0.5mm}

                \includegraphics[width=0.7in]{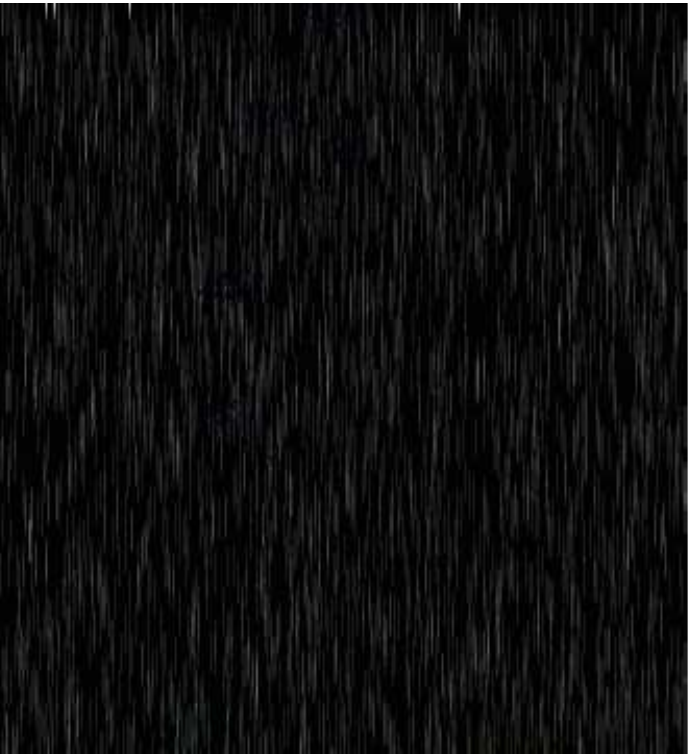}&
                \includegraphics[width=0.7in]{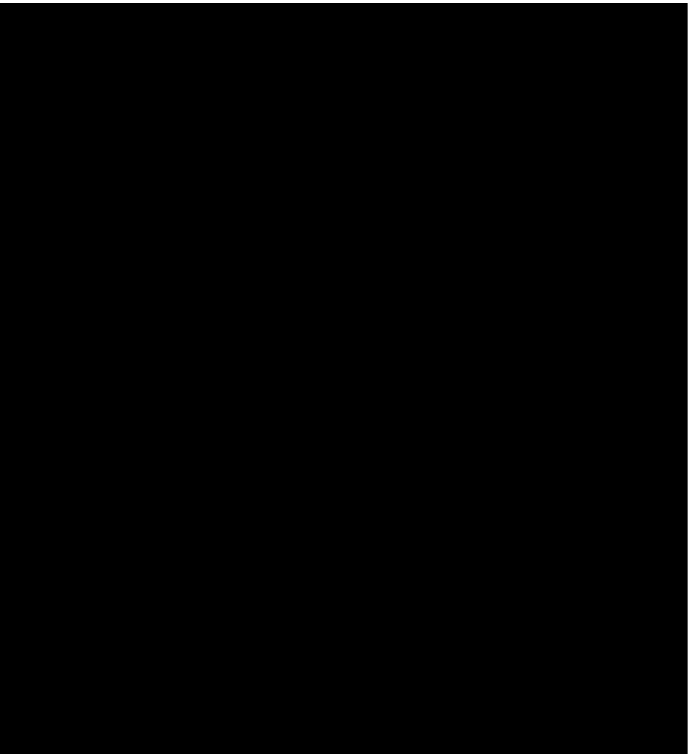}&
                \includegraphics[width=0.7in]{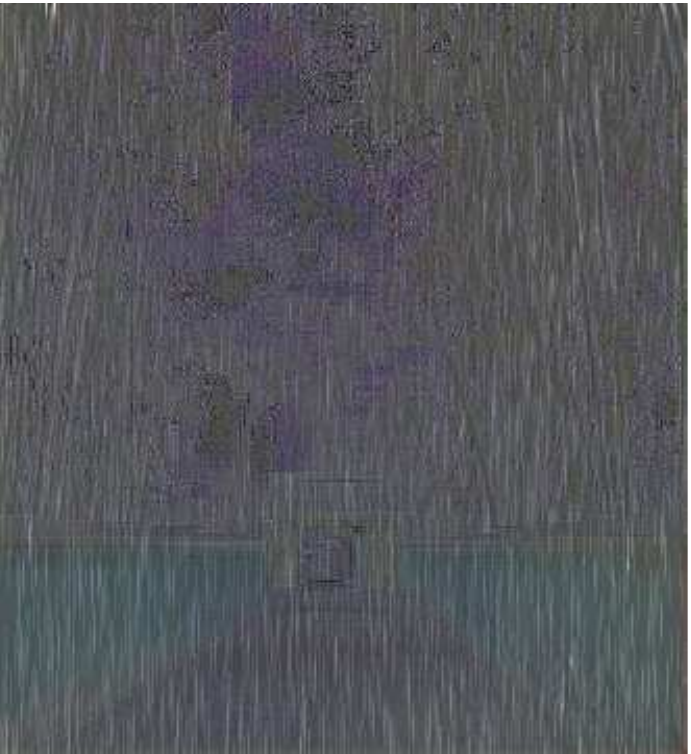}&
                \includegraphics[width=0.7in]{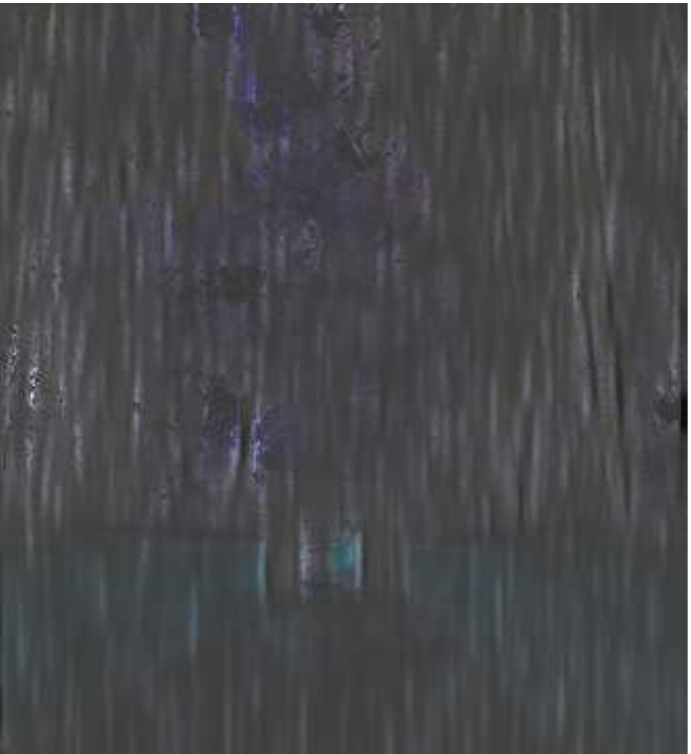}&
                \includegraphics[width=0.7in]{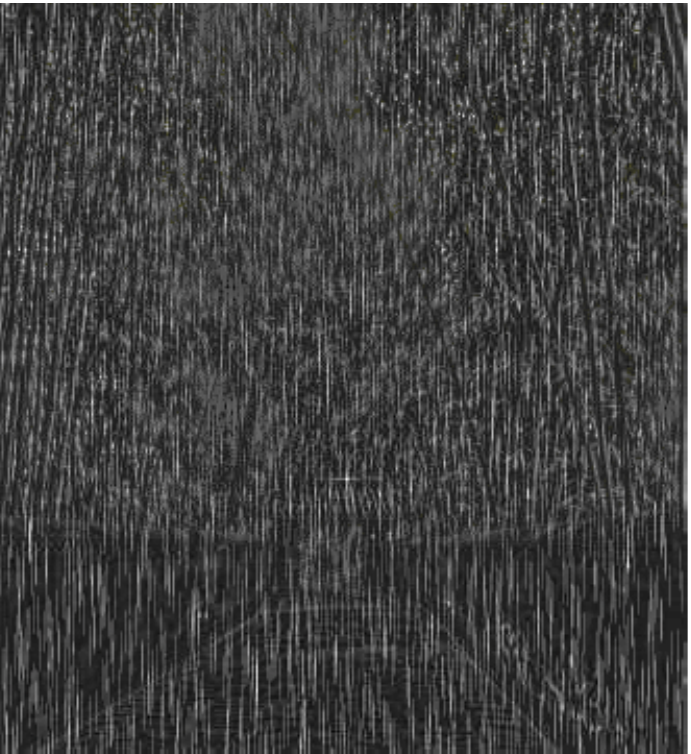}&
                \includegraphics[width=0.7in]{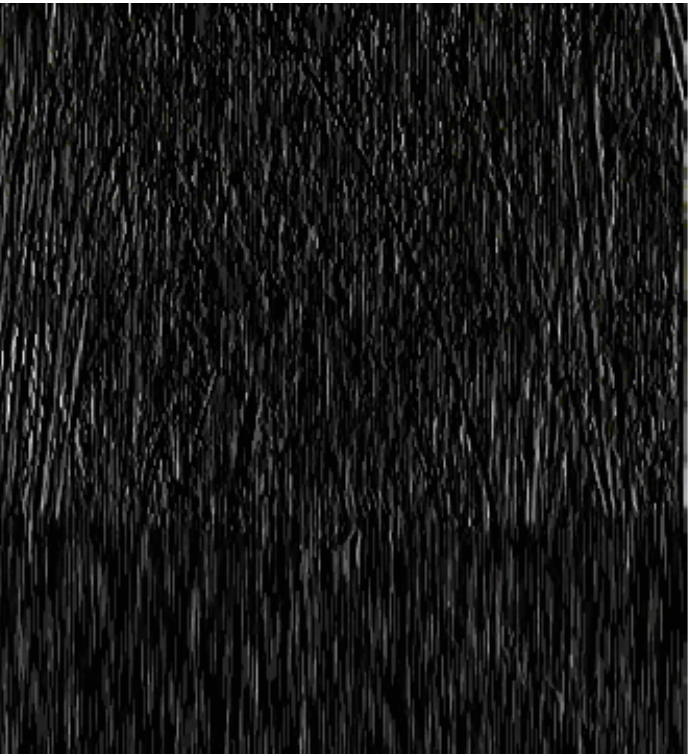}&
                \includegraphics[width=0.7in]{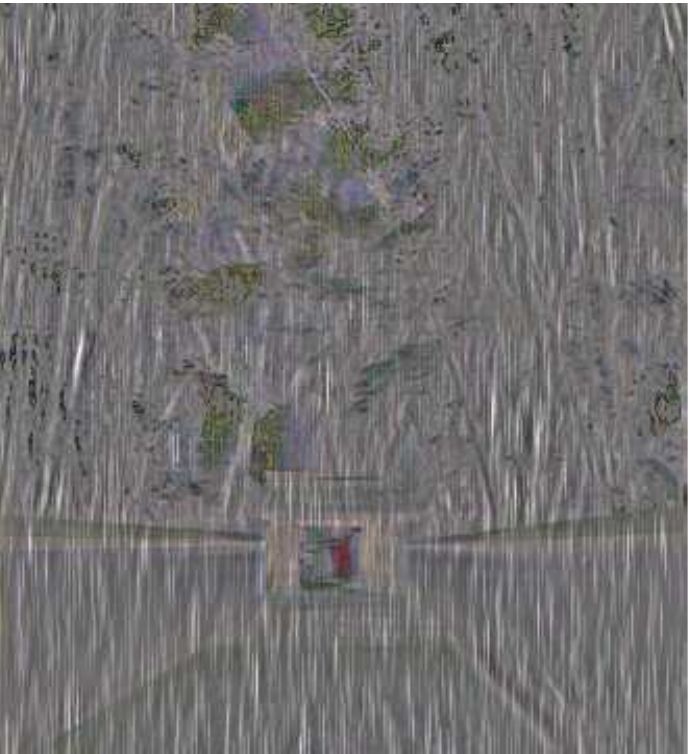}&
                \includegraphics[width=0.7in]{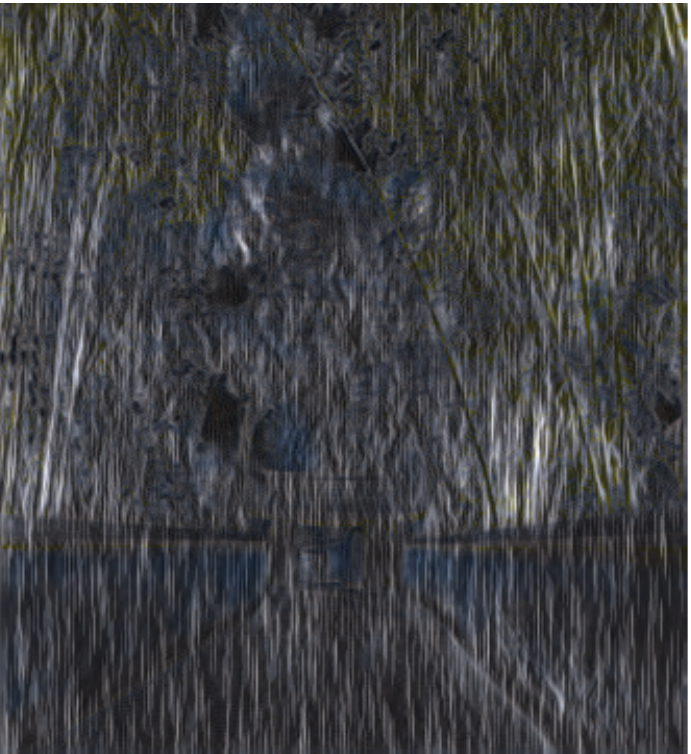}&
                \includegraphics[width=0.7in]{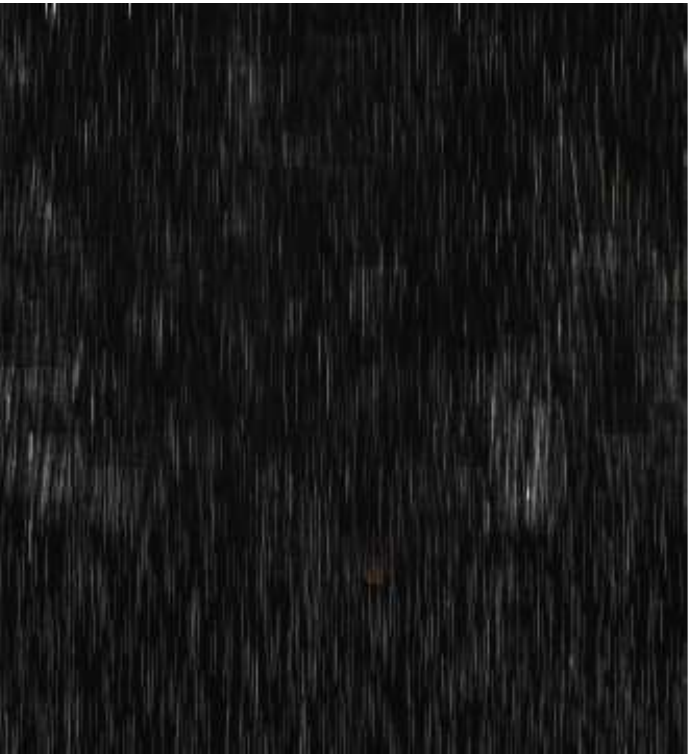}\\
                (a)&
                (b)&
                (c)&
                (d)&
                (e)&
                (f)&
                (g)&
                (h)&
                (i)\\

\end{tabular}

\caption{The rain streak images of the rain streak removal results by different methods on 3 synthetic rainy images (tree, panda, and bamboo) selected from \cite{Deng2018A}. From left to right: (a) the background, (b) the rainy images, the derain results by (c) DID \cite{zhang2018density}, (d) DSC \cite{luo2015removing}, (e) LP \cite{Li2014Single}, (f) UGSM \cite{Deng2018A}, (g) CNN \cite{fu2017clearing}, (h) DDN \cite{fu2017removing}, and (i) KGCNN.}
\label{synthetic-visual-UGSM-streak}
\end{center}
\end{figure*}

UGSM performs quite competitively for Rain12.
Therefore, it is necessary to take more test images from UGSM (tree, panda, and bamboo) to compare the rain removal ability of the proposed method and UGSM method.
In addition, based on the code provided by the authors in \cite{Deng2018A}, we also change the rain streaks' angles of these images from UGSM to generate three new synthesize images (tree2, panda2, and bamboo2) for testing.
Fig. \ref{synthetic-visual-UGSM} and Fig. \ref{synthetic-visual-UGSM-streak} respectively present the visual and rain streak results of these images from \cite{Deng2018A}, which indicates the superiority of the proposed method.

\begin{figure*}[!htb]
\renewcommand\arraystretch{0.8}\setlength{\tabcolsep}{1.8pt}
\begin{center}
\begin{tabular}{ccccccccc}
                ~\includegraphics[width=0.7in]{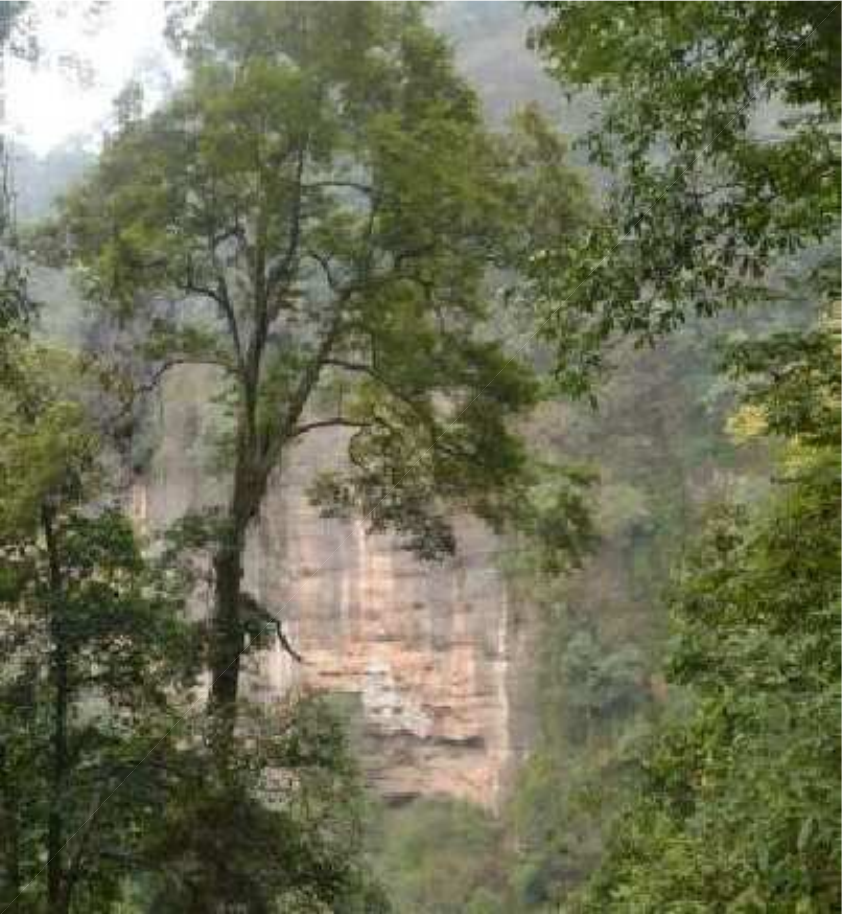}&
                \includegraphics[width=0.7in]{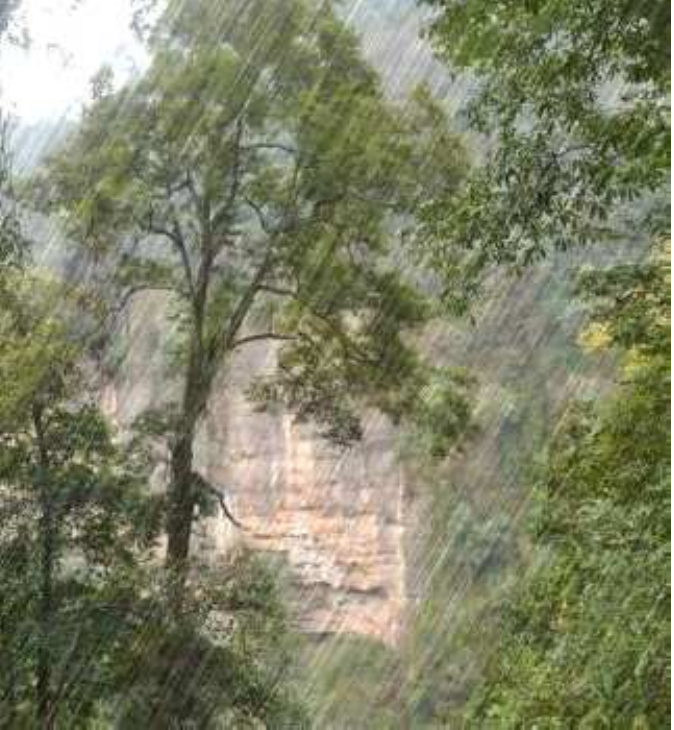}&
                \includegraphics[width=0.7in]{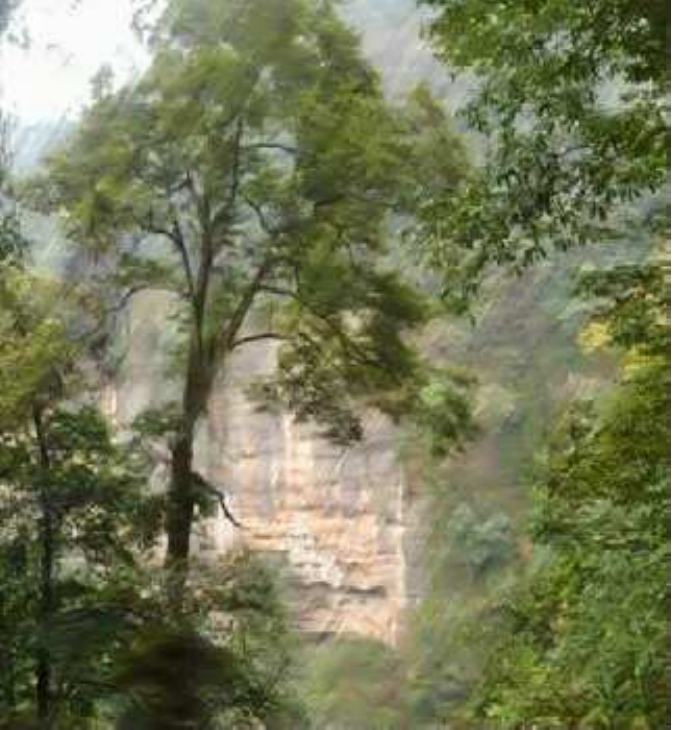}&
                \includegraphics[width=0.7in]{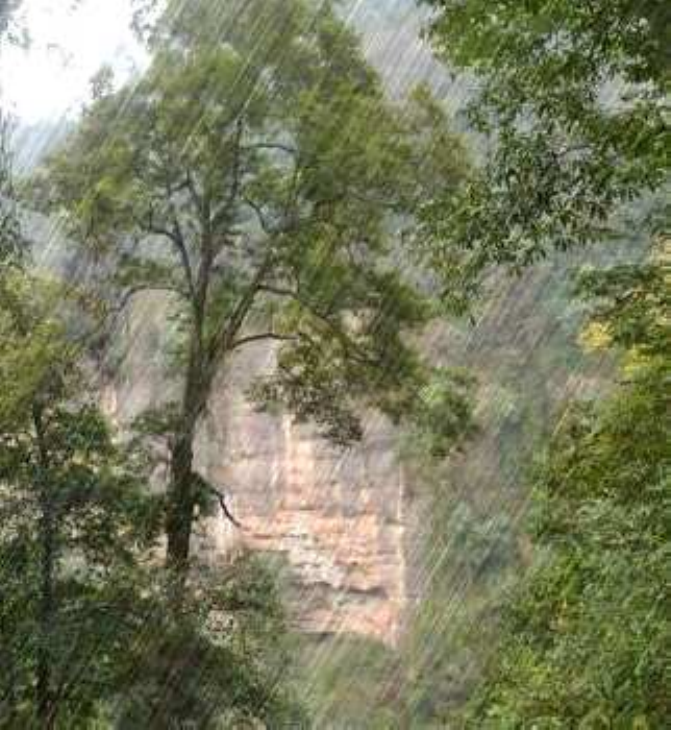}&
                \includegraphics[width=0.7in]{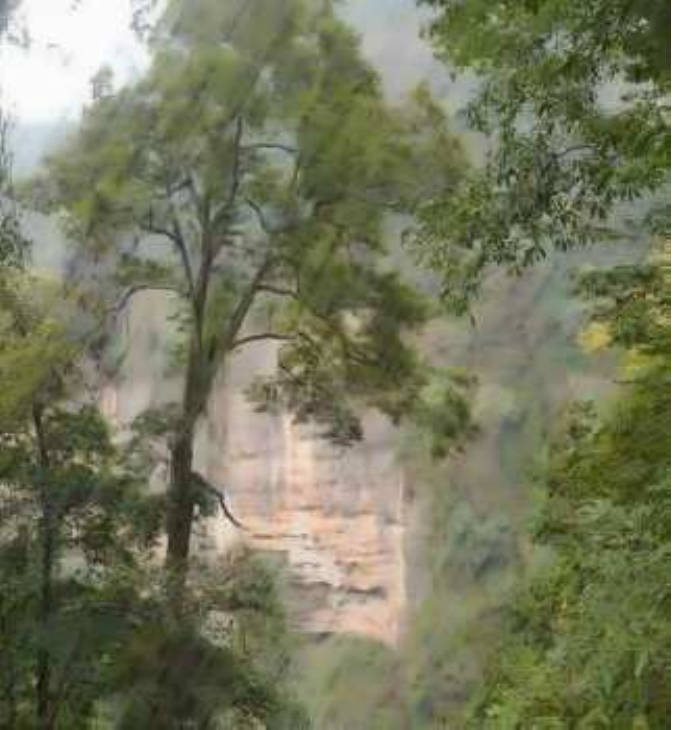}&
                \includegraphics[width=0.7in]{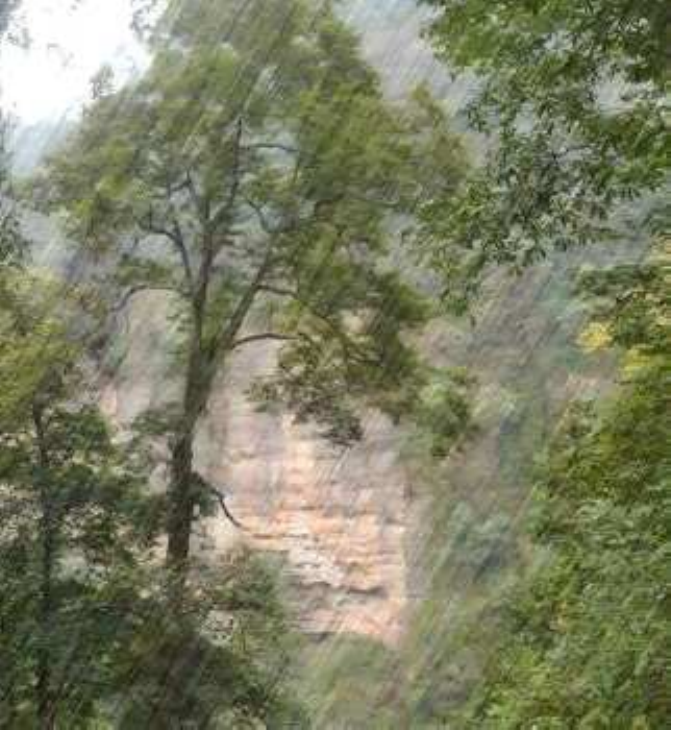}&
                \includegraphics[width=0.7in]{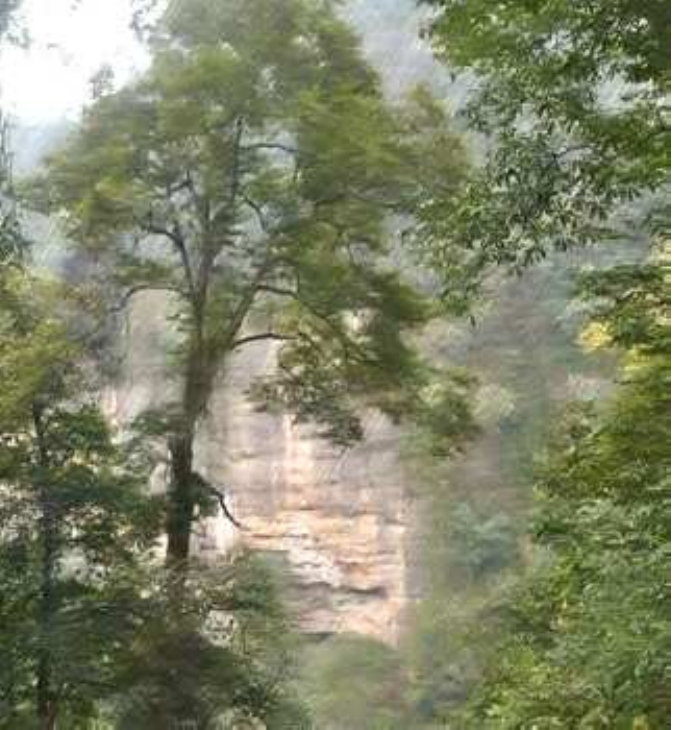}&
                \includegraphics[width=0.7in]{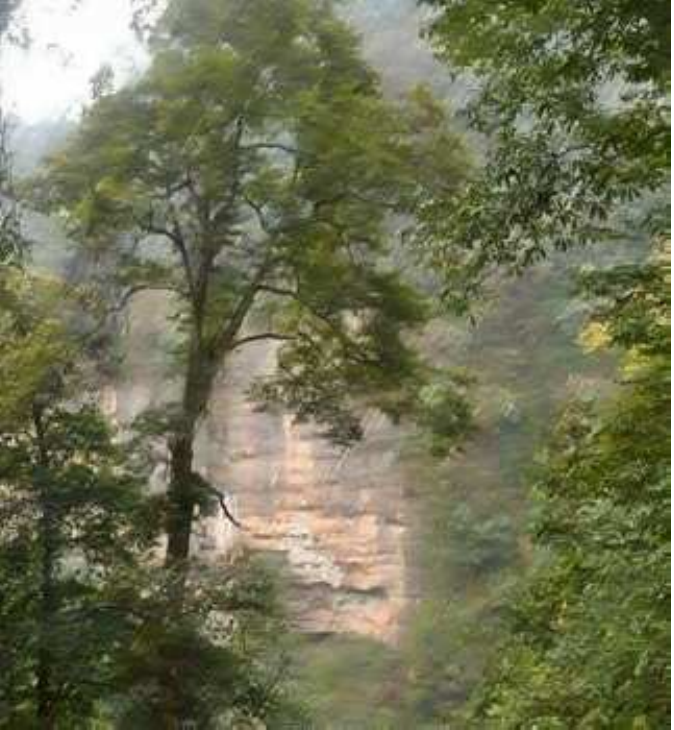}&
                \includegraphics[width=0.7in]{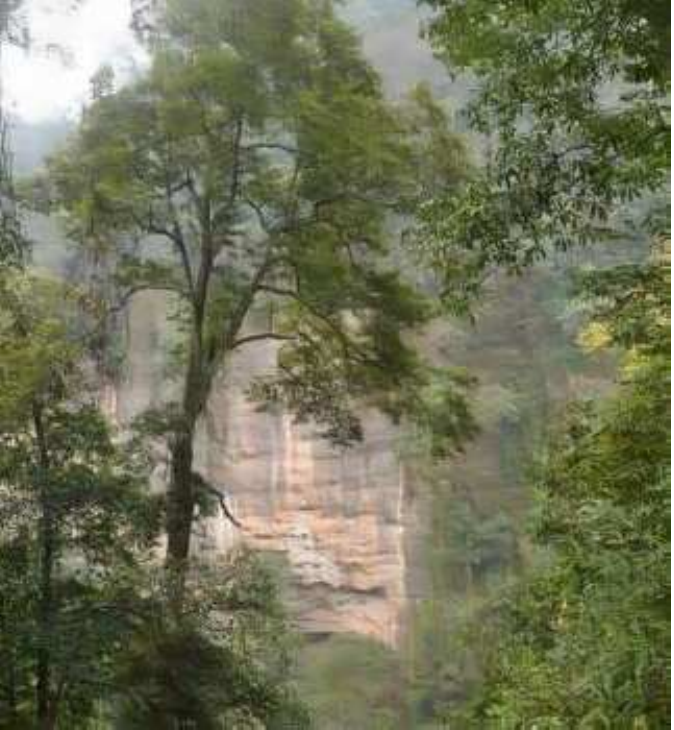}\\
                \vspace{0.5mm}

                \includegraphics[width=0.7in]{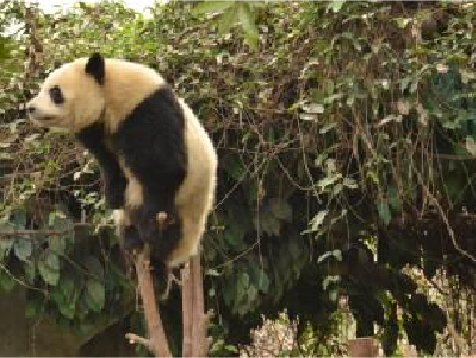}&
                \includegraphics[width=0.7in]{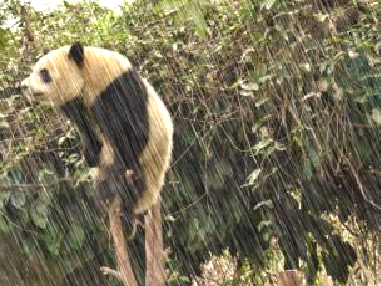}&
                \includegraphics[width=0.7in]{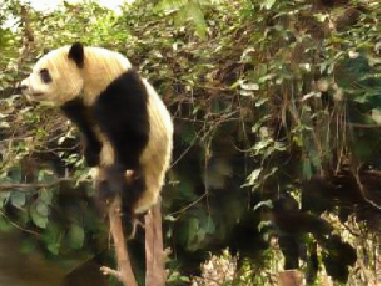}&
                \includegraphics[width=0.7in]{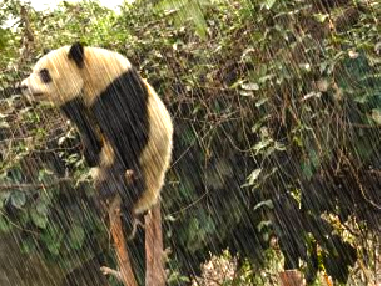}&
                \includegraphics[width=0.7in]{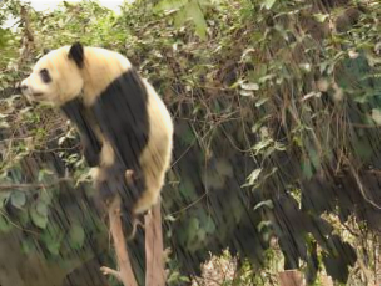}&
                \includegraphics[width=0.7in]{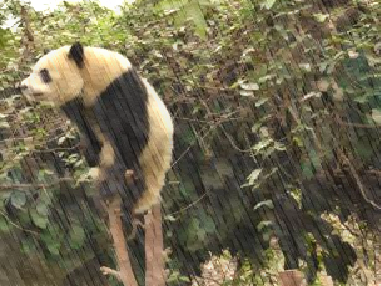}&
                \includegraphics[width=0.7in]{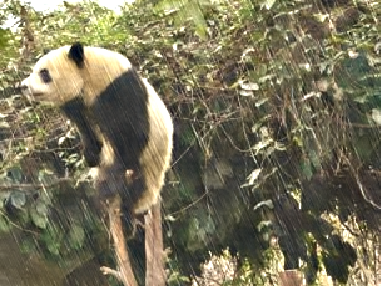}&
                \includegraphics[width=0.7in]{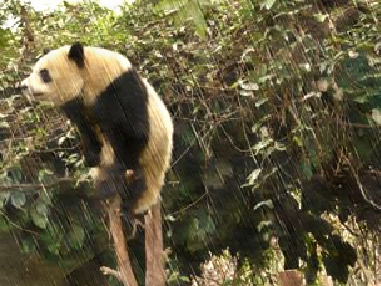}&
                \includegraphics[width=0.7in]{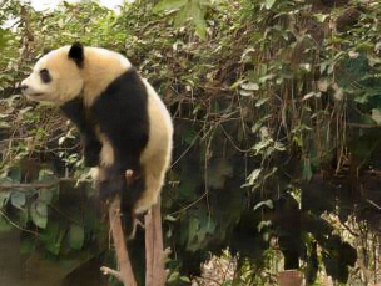}\\
                \vspace{0.5mm}

                \includegraphics[width=0.7in]{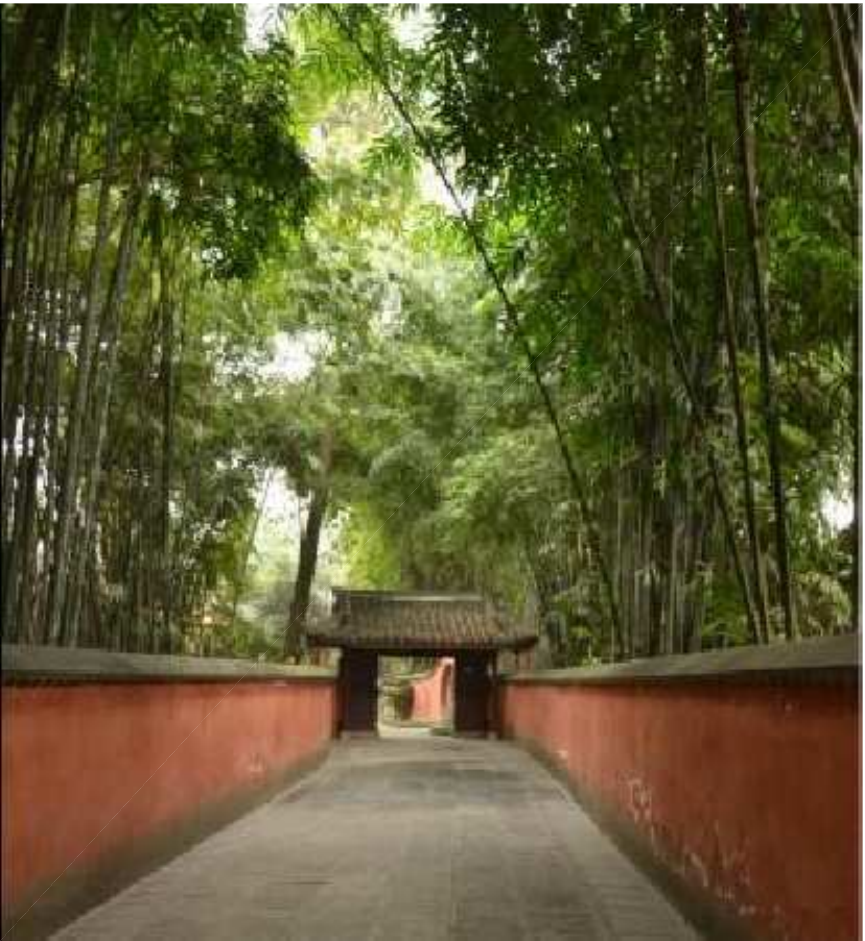}&
                \includegraphics[width=0.7in]{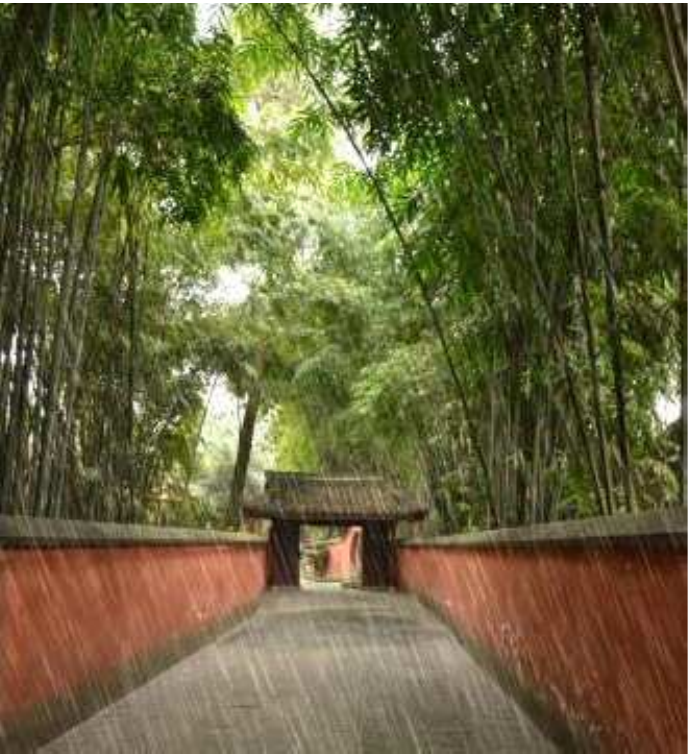}&
                \includegraphics[width=0.7in]{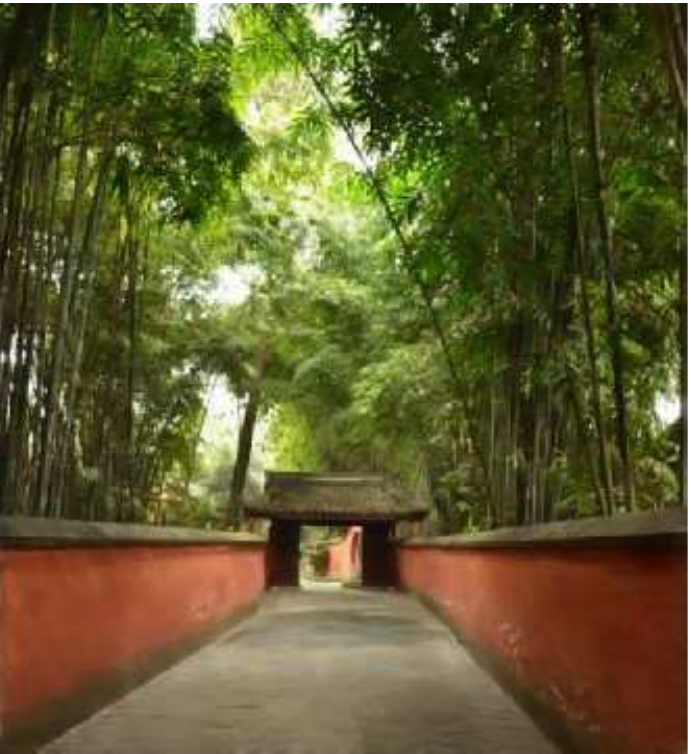}&
                \includegraphics[width=0.7in]{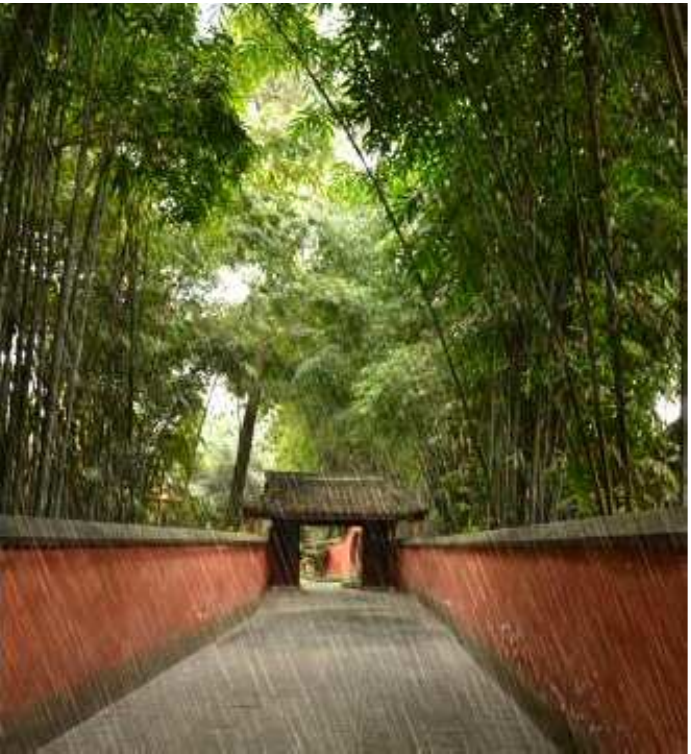}&
                \includegraphics[width=0.7in]{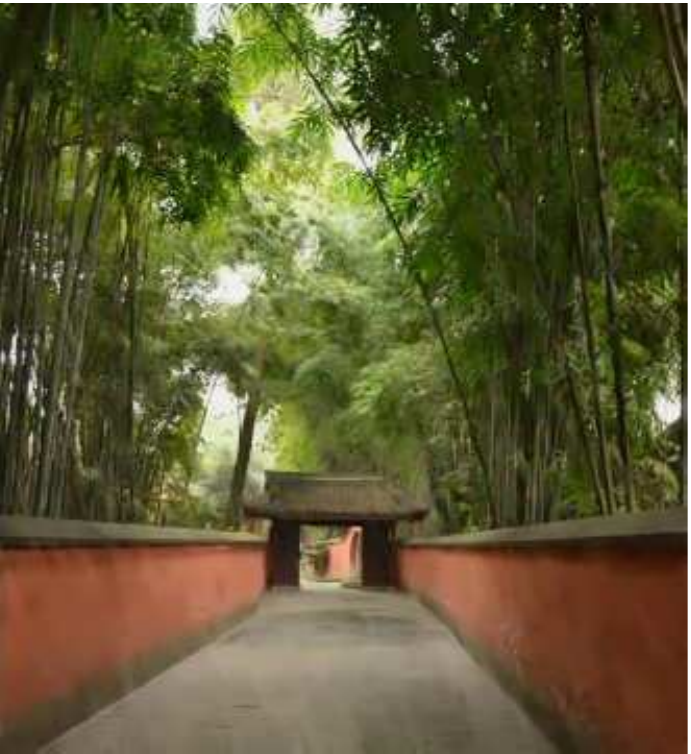}&
                \includegraphics[width=0.7in]{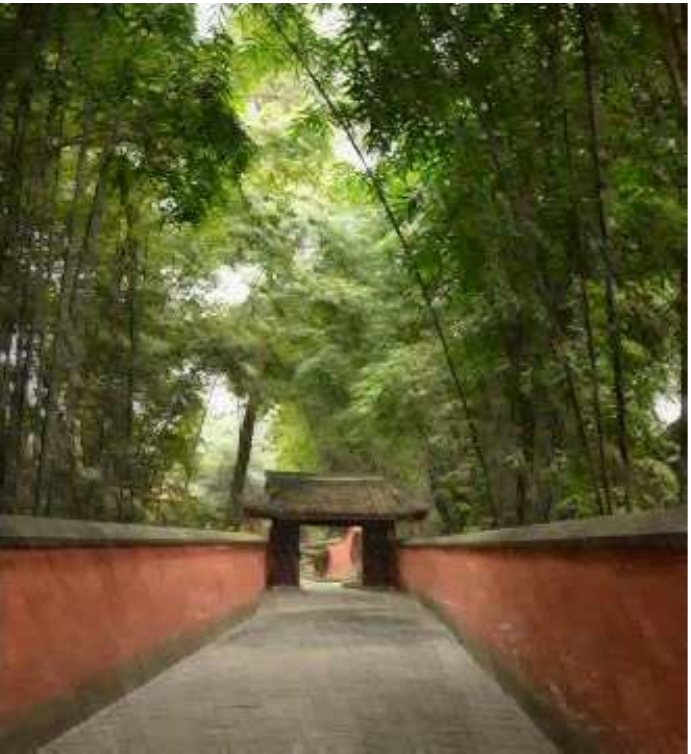}&
                \includegraphics[width=0.7in]{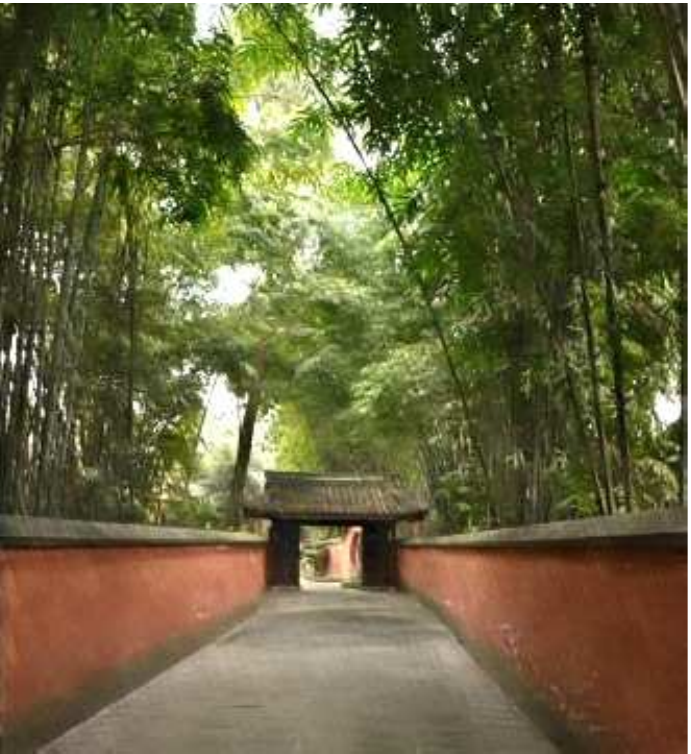}&
                \includegraphics[width=0.7in]{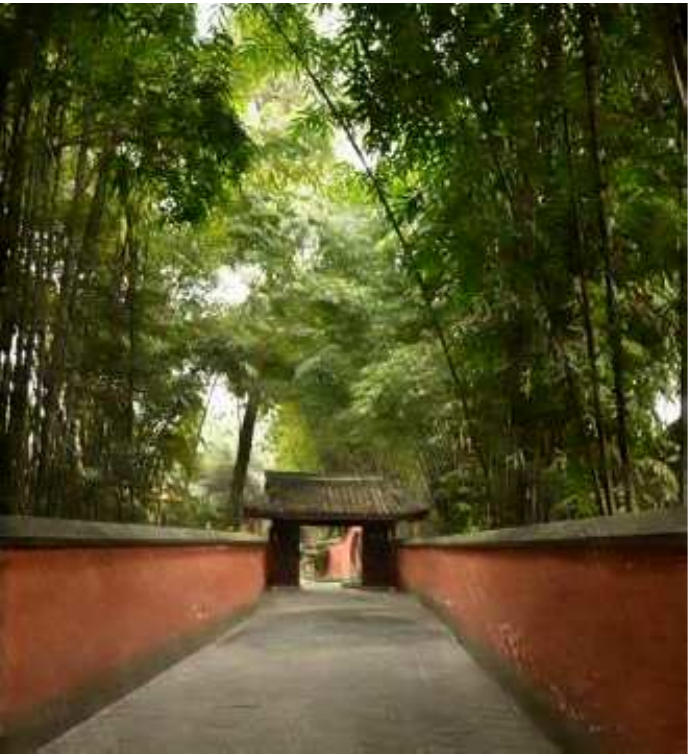}&
                \includegraphics[width=0.7in]{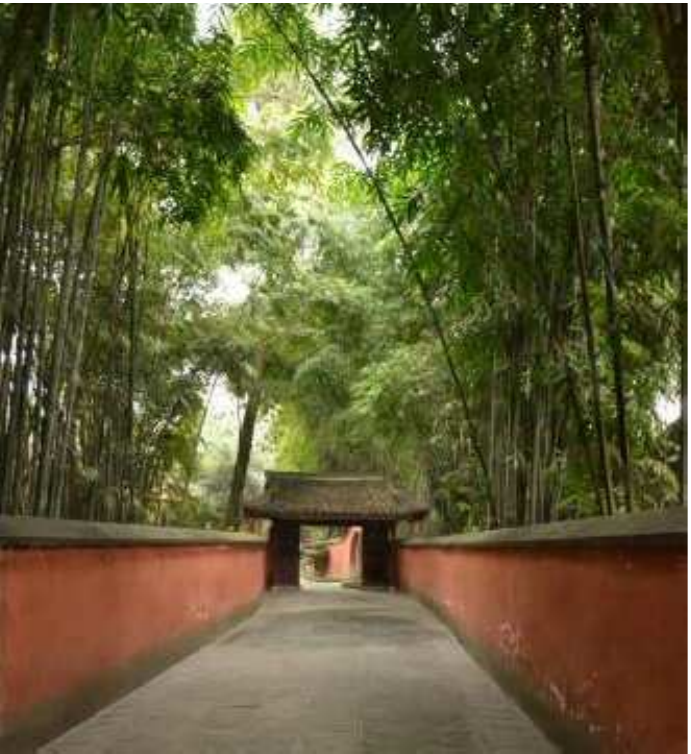}\\
                (a)&
                (b)&
                (c)&
                (d)&
                (e)&
                (f)&
                (g)&
                (h)&
                (i)\\

\end{tabular}

\caption{Rain streak removal results by different methods on 3 new synthetic rainy images (tree2, panda2, and bamboo2). From left to right: (a) the background, (b) the rainy images, the derain results by (c) DID \cite{zhang2018density}, (d) DSC \cite{luo2015removing}, (e) LP \cite{Li2014Single}, (f) UGSM \cite{Deng2018A}, (g) CNN \cite{fu2017clearing}, (h) DDN \cite{fu2017removing}, and (i) KGCNN.}
\label{synthetic-visual-UGSM2}
\end{center}
\end{figure*}

\begin{figure*}[!htb]
\renewcommand\arraystretch{0.8}\setlength{\tabcolsep}{1.8pt}
\begin{center}
\begin{tabular}{ccccccccc}
                ~\includegraphics[width=0.7in]{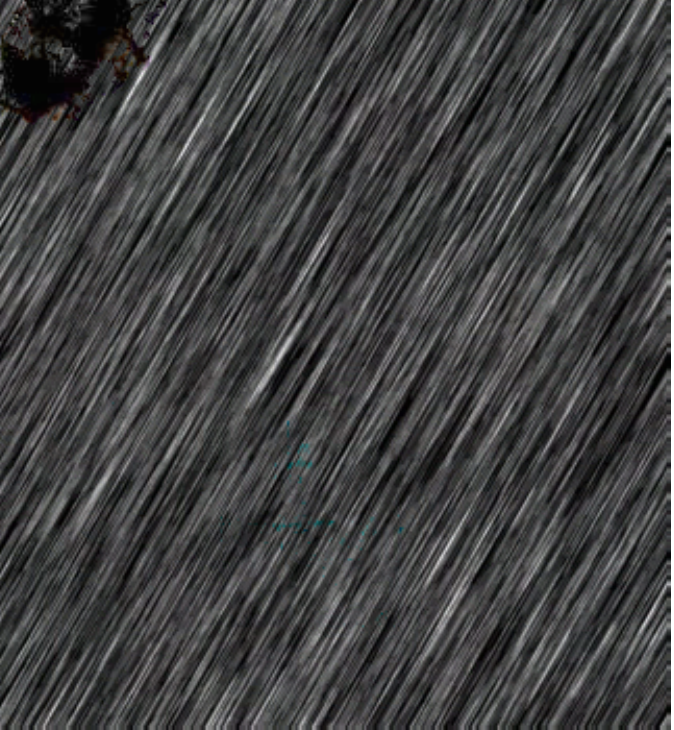}&
                \includegraphics[width=0.7in]{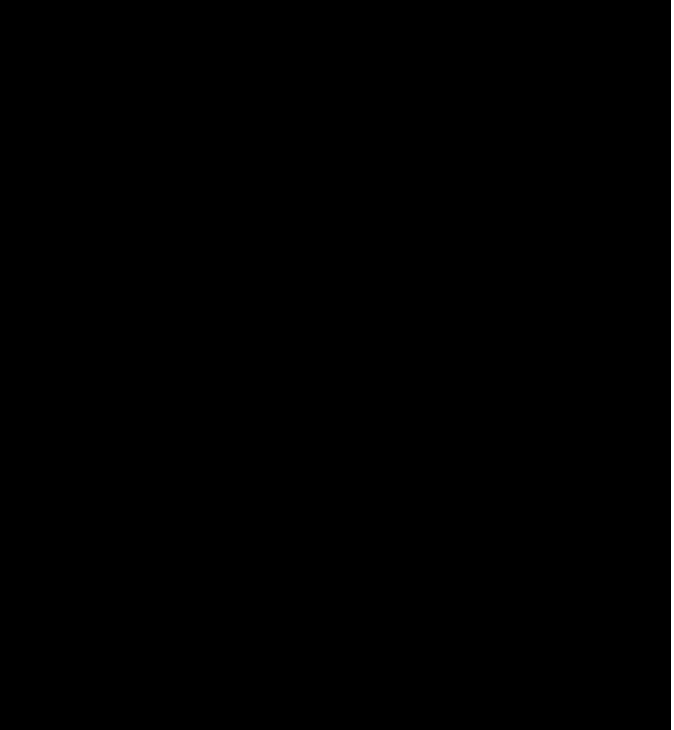}&
                \includegraphics[width=0.7in]{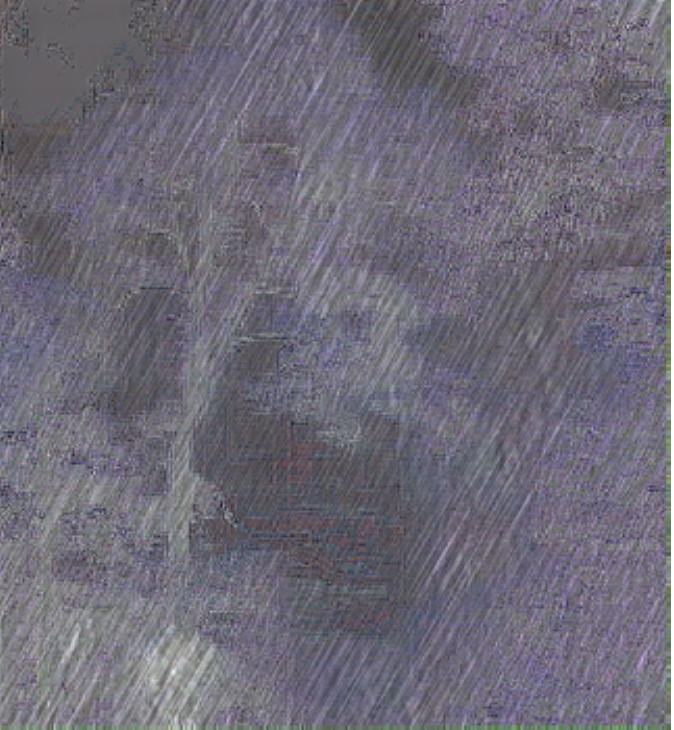}&
                \includegraphics[width=0.7in]{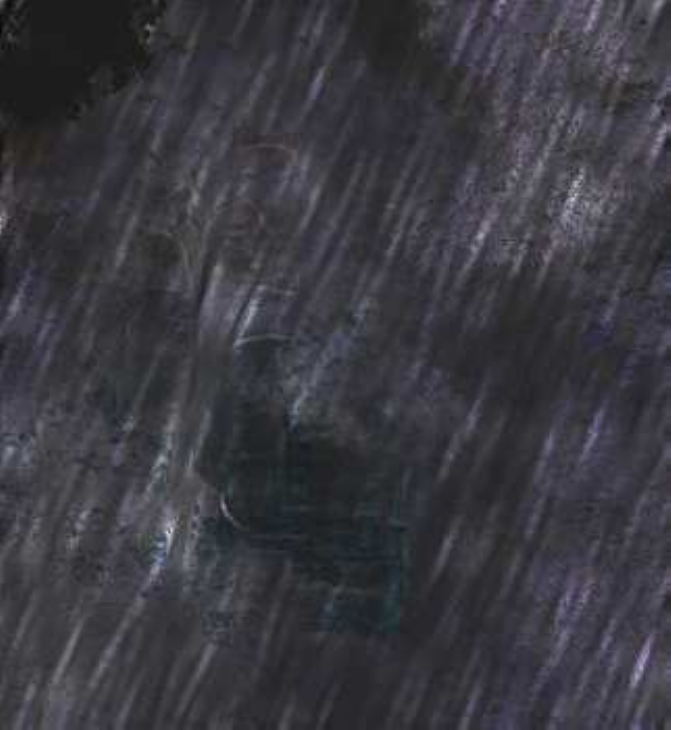}&
                \includegraphics[width=0.7in]{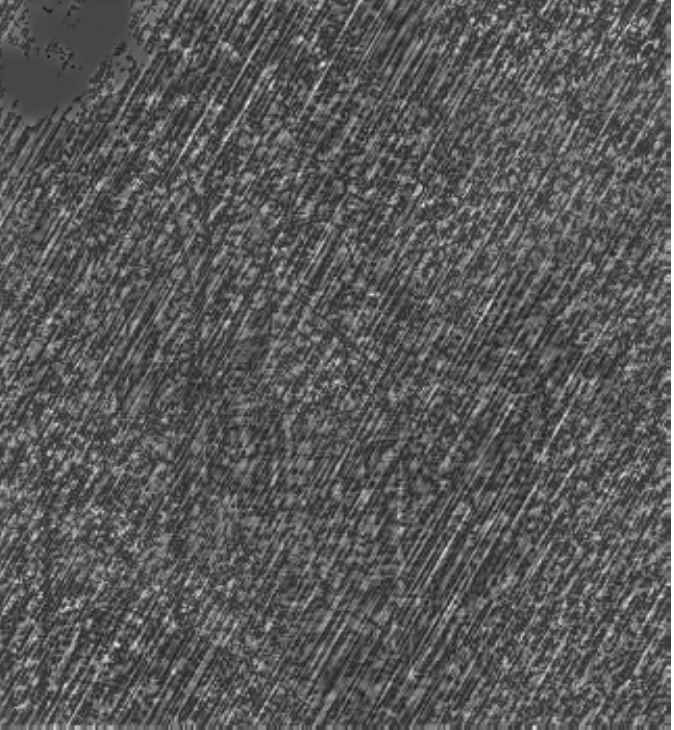}&
                \includegraphics[width=0.7in]{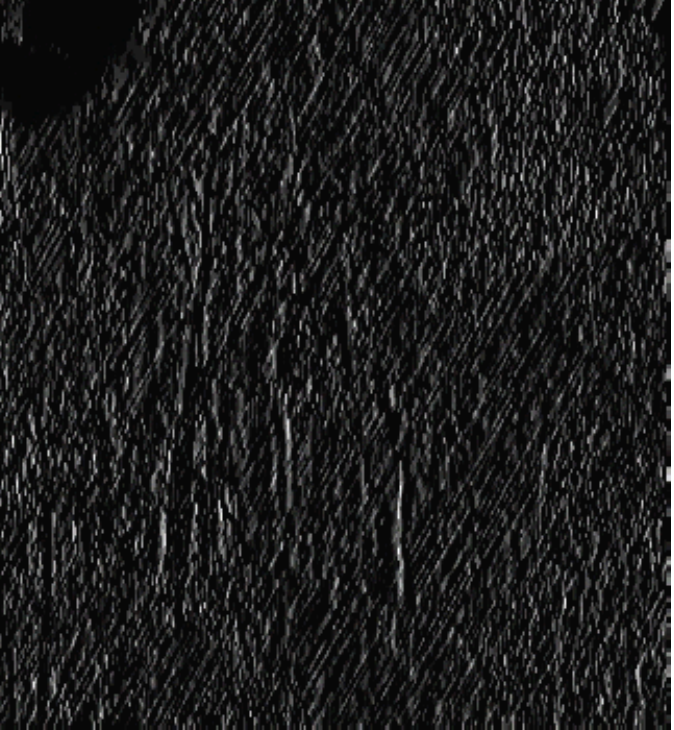}&
                \includegraphics[width=0.7in]{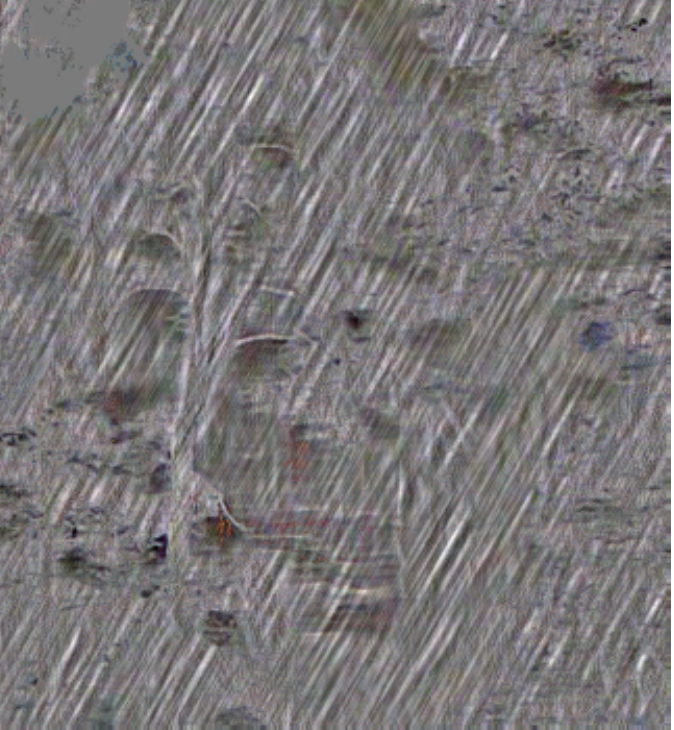}&
                \includegraphics[width=0.7in]{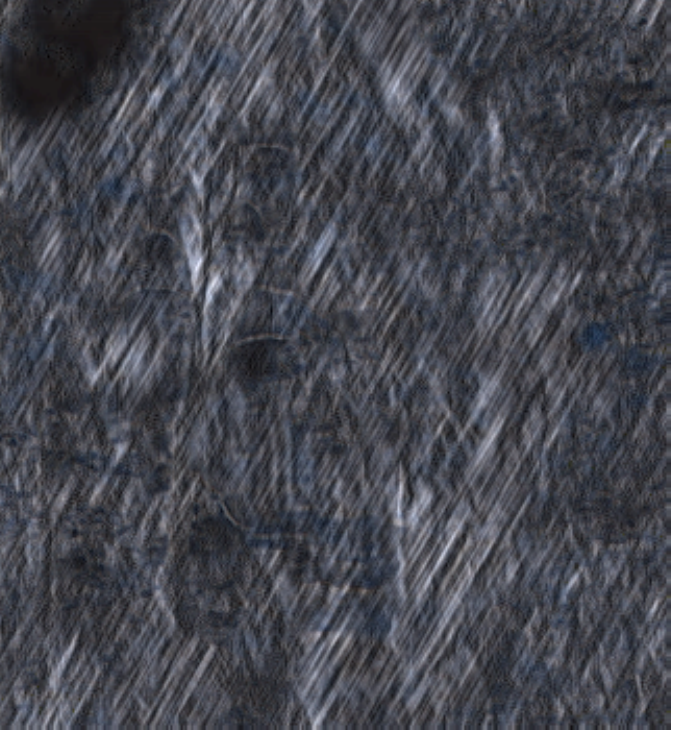}&
                \includegraphics[width=0.7in]{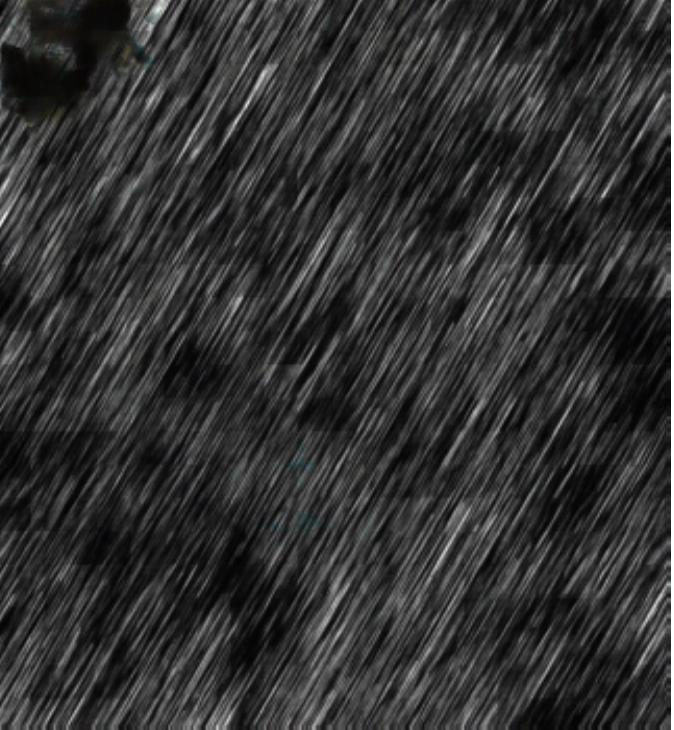}\\
                \vspace{0.5mm}

                \includegraphics[width=0.7in]{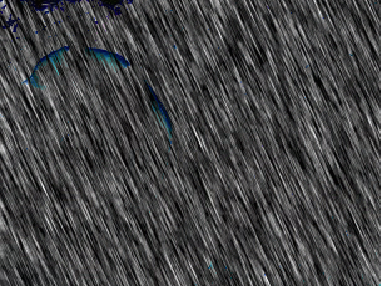}&
                \includegraphics[width=0.7in]{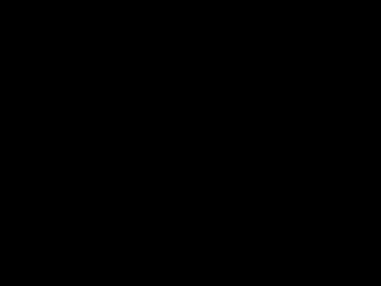}&
                \includegraphics[width=0.7in]{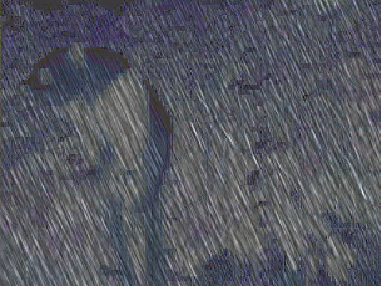}&
                \includegraphics[width=0.7in]{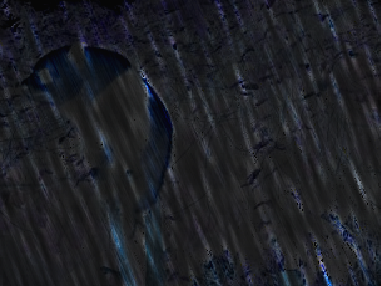}&
                \includegraphics[width=0.7in]{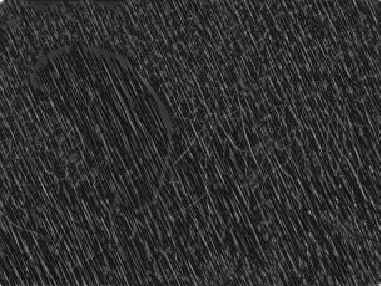}&
                \includegraphics[width=0.7in]{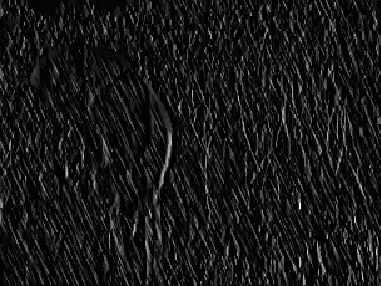}&
                \includegraphics[width=0.7in]{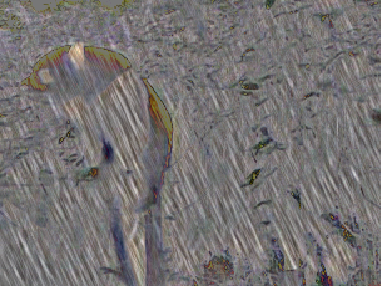}&
                \includegraphics[width=0.7in]{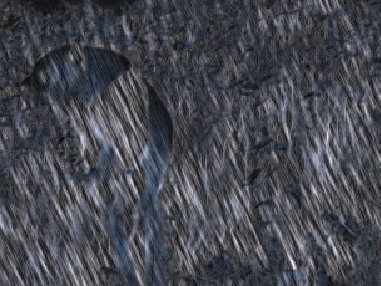}&
                \includegraphics[width=0.7in]{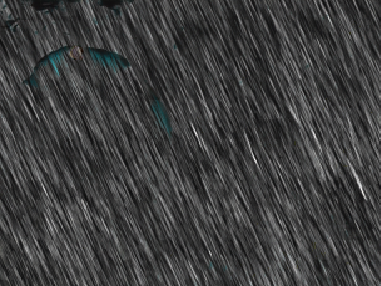}\\
                \vspace{0.5mm}

                \includegraphics[width=0.7in]{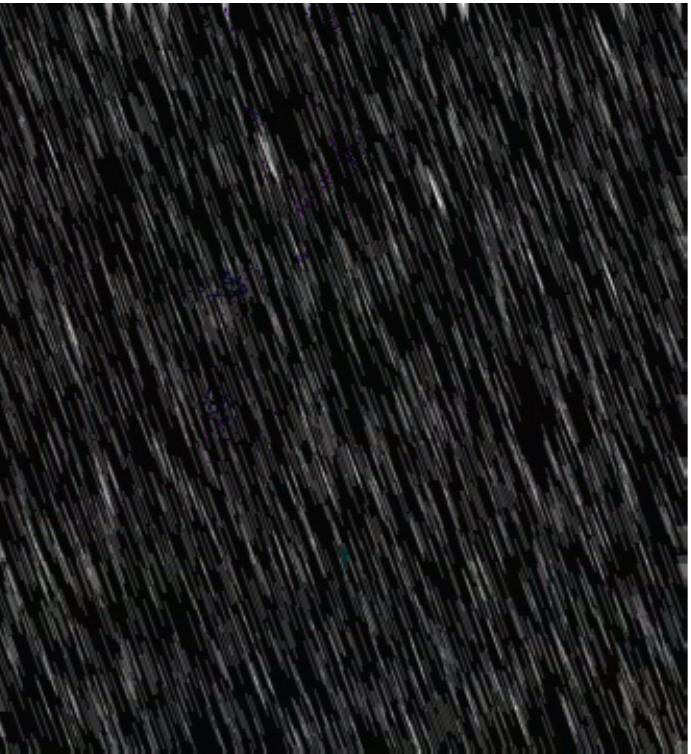}&
                \includegraphics[width=0.7in]{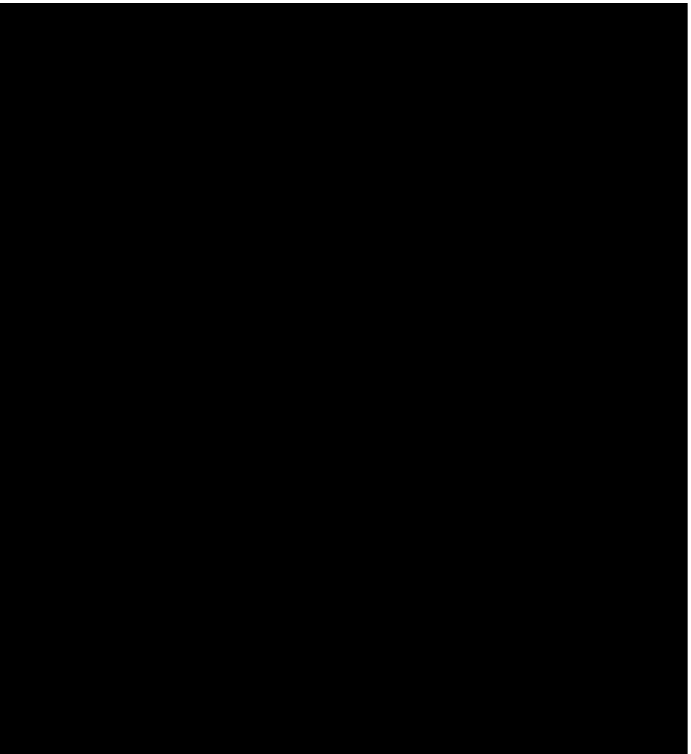}&
                \includegraphics[width=0.7in]{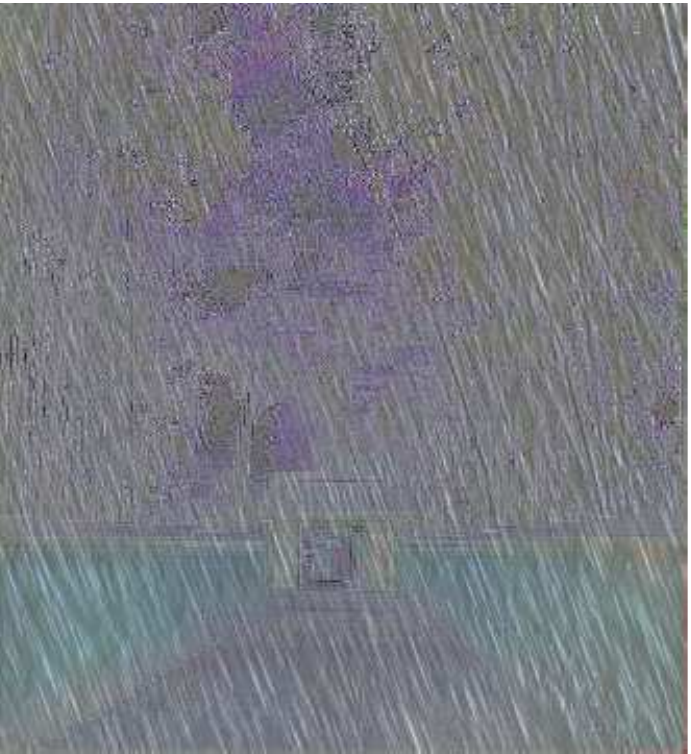}&
                \includegraphics[width=0.7in]{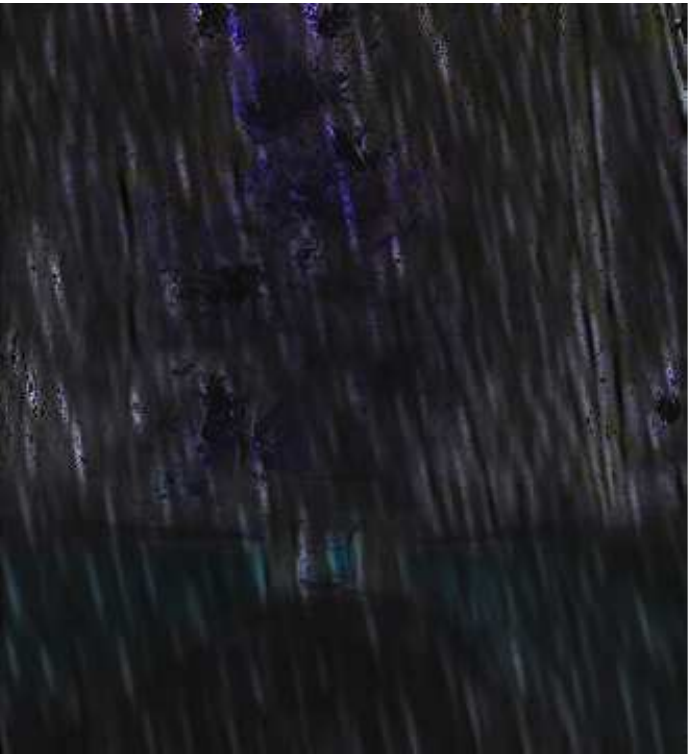}&
                \includegraphics[width=0.7in]{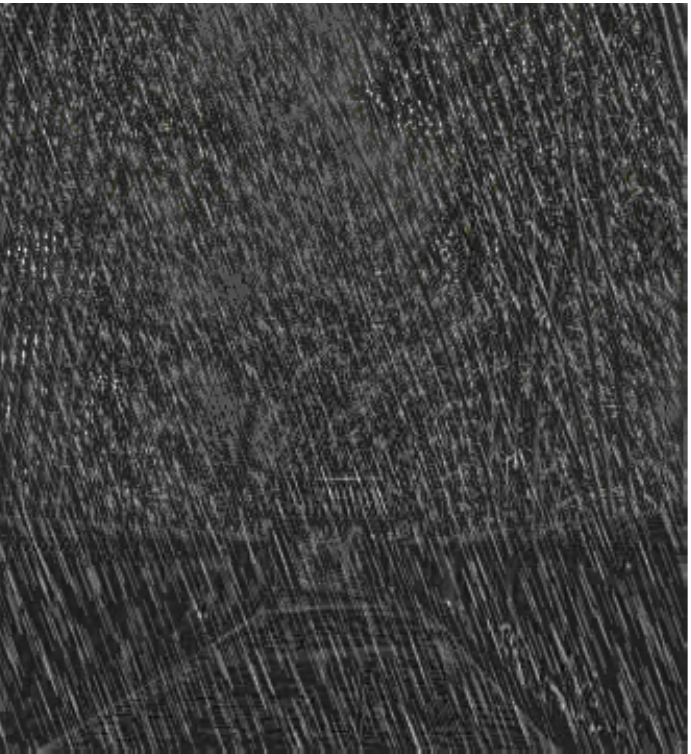}&
                \includegraphics[width=0.7in]{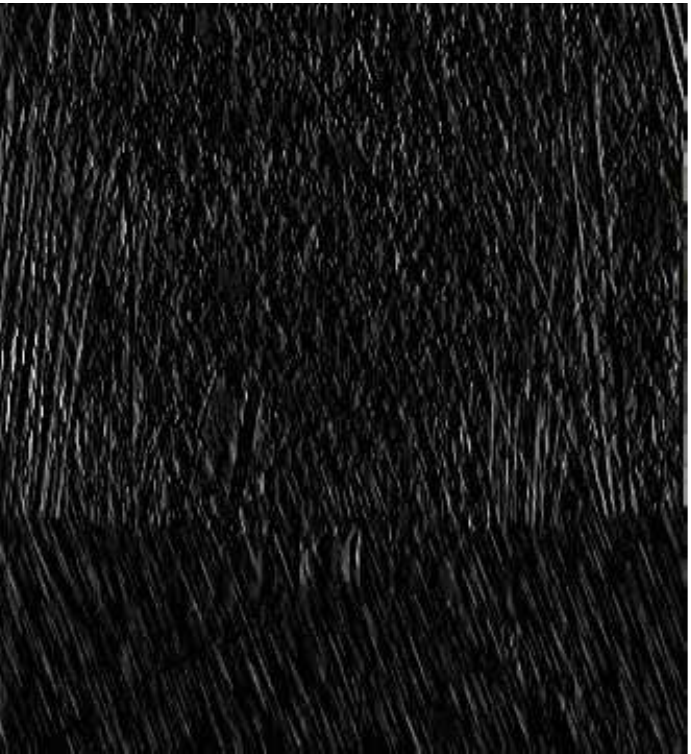}&
                \includegraphics[width=0.7in]{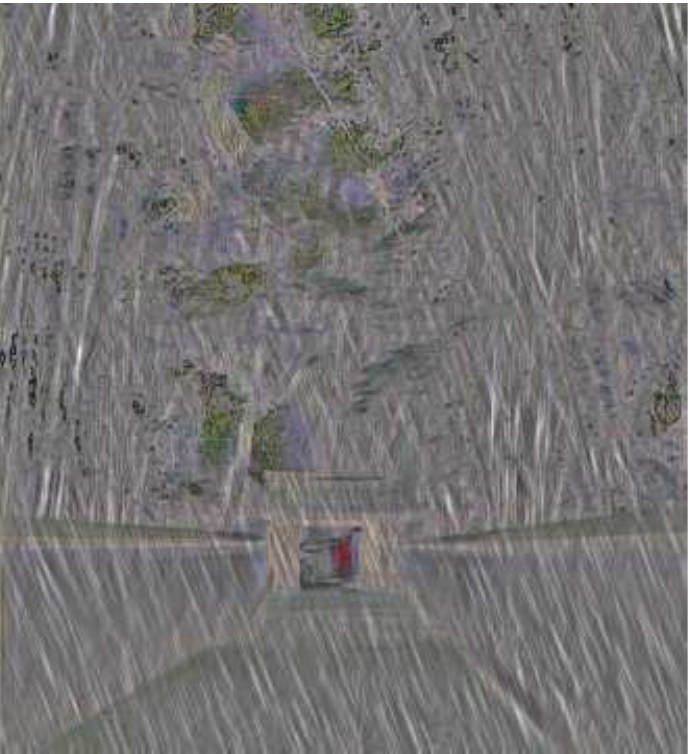}&
                \includegraphics[width=0.7in]{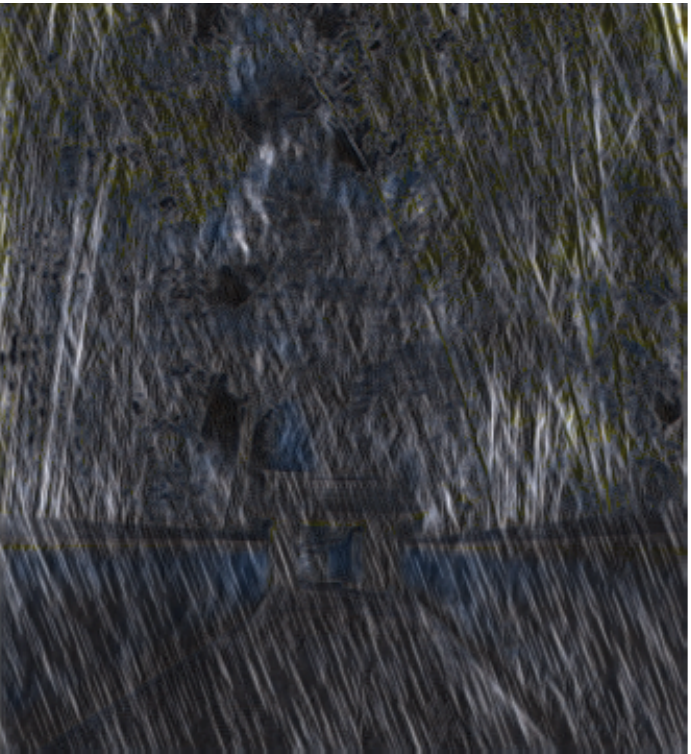}&
                \includegraphics[width=0.7in]{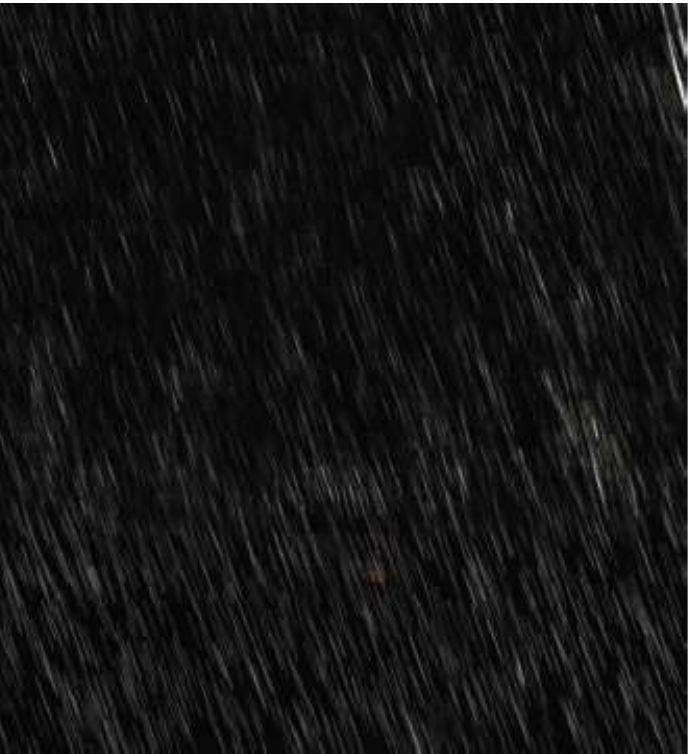}\\
                (a)&
                (b)&
                (c)&
                (d)&
                (e)&
                (f)&
                (g)&
                (h)&
                (i)\\

\end{tabular}

\caption{The rain streak images of the rain streak removal results by different methods on 3 new synthetic rainy images (tree2, panda2, and bamboo2). From left to right: (a) the background, (b) the rainy images, the derain results by (c) DID \cite{zhang2018density}, (d) DSC \cite{luo2015removing}, (e) LP \cite{Li2014Single}, (f) UGSM \cite{Deng2018A}, (g) CNN \cite{fu2017clearing}, (h) DDN \cite{fu2017removing}, and (i) KGCNN.}
\label{synthetic-visual-UGSM-streak2}
\end{center}
\end{figure*}

In Fig. \ref{synthetic-visual-UGSM2} and Fig. \ref{synthetic-visual-UGSM-streak2}, rain streaks with very large angles are added to background to formulate rainy images.
Noting that the UGSM is based on directional priors, which is quite effective to the case of vertical rain streaks, but less effective for the case of oblique rain streaks.
Therefore, from Fig. \ref{synthetic-visual-UGSM2} and Fig. \ref{synthetic-visual-UGSM-streak2}, we can know that the proposed KGCNN method performs significantly better than UGSM method, both visually and quantitatively.
Moreover, the KGCNN method also exhibits better ability of rain streak removal, compared with other state-of-the-art methods.
Table \ref{synthetic-quant-UGSM} also demonstrate the effectiveness of our method from the perspective of quantitative results.

\begin{table*}[htb]
\renewcommand\arraystretch{1.2}\setlength{\tabcolsep}{2.4pt}
\caption{Quantitative comparisons of rain streak removal results by DID \cite{zhang2018density}, DSC \cite{luo2015removing}, LP \cite{Li2014Single}, UGSM \cite{Deng2018A}, CNN \cite{fu2017clearing}, DDN \cite{fu2017removing}, and KGCNN on 3 synthetic rainy images selected in \cite{Deng2018A}.}
\begin{center}
 \label{synthetic-quant-UGSM}
\begin{tabular}{c|c|cccccc|cccccc}
 \Xhline{1.2pt}
        ~ &Rain type  &\multicolumn{6}{c|}{original images} &\multicolumn{6}{c}{images with large angle} \\
        \hline
        Images  &Method &PSNR &SSIM &FSIM &UIQI &GMSD &Time (s) &PSNR &SSIM &FSIM &UIQI &GMSD &Time (s) \\
\Xhline{0.8pt}
\multirow{8}[0]{*}{\bf{tree}}
                      &rainy    &27.375    &0.934    &0.942    &0.989    &0.076    &-     &21.475 &0.895 &0.926 &0.952 &0.078    &-  \\
                      &DID    &27.826    &0.926    &0.937    &0.982    &0.060    &\bf{0.625} &25.076 &0.916 &0.929 &0.989 &0.069  &\bf{0.615} \\
                      &DSC    &29.646    &0.940    &0.944    &0.996    &0.064    &87.726 &24.466 &0.908 &0.924 &0.982 &0.073   &89.726\\
                       &LP &29.435 &0.927 &0.912 &0.997 &0.076 &226.091  &25.497 &0.909 &0.899 &0.983 &0.081 &236.091\\
                      &UGSM    &30.659    &0.954    &0.948    &0.998    &0.056    &0.979 &22.842 &0.906 &0.927 &0.964 &0.074   &0.979\\
                      &CNN    &26.532    &0.942    &0.941    &0.989    &0.065    &6.139 &21.201 &0.915 &0.934 &0.952 &0.068    &6.339 \\
                      &DDN    &26.944    &0.944    &0.945    &0.977    &0.066    &0.812 &28.640 &0.936 &0.935 &\bf{0.996} &0.069    &0.812\\
                      &KGCNN    &\bf{32.269}    &\bf{0.971}    &\bf{0.966}    &\bf{0.998}    &\bf{0.049}    &9.402 &\bf{29.393} &\bf{0.960} &\bf{0.956} &0.993 &\bf{0.057}    &10.422 \\

\hline
\multirow{8}[0]{*}{\bf{panda}}
                      &rainy    &27.102    &0.920    &0.946    &0.978    &0.083    &-    &17.569 &0.750 &0.871 &0.836 &0.140    &-   \\
                      &DID    &27.120    &0.923    &0.946    &0.952    &0.066    &0.625 &24.077 &0.891 &0.914 &0.960 &0.094  &0.665\\
                      &DSC    &26.568    &0.914    &0.940    &0.968    &0.077    &70.239 &21.688 &0.790 &0.867 &0.959 &0.124    &70.639\\
                      &LP &29.250 &0.938 &0.943 &0.994 &0.073 &133.709 &21.197 &0.857 &0.904 &0.908 &0.105 &143.709 \\
                      &UGSM    &27.823    &0.925    &0.926    &0.994    &0.082    &2.505 &19.621 &0.798 &0.876 &0.885 &0.118    &2.750\\
                      &CNN    &24.838    &0.927    &0.937    &0.976    &0.070    &3.421  &17.145 &0.791 &0.883 &0.832 &0.105   &4.442 \\
                      &DDN    &25.693    &0.921    &0.947    &0.934    &0.077    &\bf{0.562}  &23.332 &0.868 &0.904 &0.972 &0.099   &\bf{0.562} \\
                      &KGCNN    &\bf{30.958}    &\bf{0.964}    &\bf{0.967}    &\bf{0.994}    &\bf{0.055}    &6.997 &\bf{27.130} &\bf{0.925} &\bf{0.937} &\bf{0.977} &\bf{0.089}    &7.907 \\
\hline
\multirow{8}[0]{*}{\bf{bamboo}}
                      &rainy    &26.091    &0.923    &0.930    &0.966    &0.102    &-   &26.997 &0.944 &0.938 &0.967 &0.090    &-     \\
                      &DID    &27.355    &0.939    &0.930    &0.964    &0.071    &0.625 &27.987 &0.949 &0.938 &0.990 &0.062    &0.635 \\
                      &DSC    &26.426    &0.923    &0.925    &0.964    &0.086    &120.959 &27.374 &0.942 &0.933 &0.959 &0.080   &124.459\\
                       &LP &29.337 &0.954 &0.933 &0.953 &0.071 &218.103 &29.594 &0.960 &0.936 &1.032 &0.069 &213.103 \\
                      &UGSM    &27.647    &0.936    &0.919    &\bf{0.991}    &0.086    &2.794 &27.514 &0.933 &0.916 &0.990 &0.088    &2.679 \\
                      &CNN    &25.761    &0.941    &0.926    &0.975    &0.081    &5.499 &25.877 &0.946 &0.932 &0.966 &0.075   &5.499\\
                      &DDN    &24.385    &0.905    &0.926    &0.843    &0.097    &\bf{0.531} &25.429 &0.924 &0.935 &0.898 &0.087   &\bf{0.521}\\
                      &KGCNN    &\bf{29.587}    &\bf{0.963}    &\bf{0.955}    &0.962    &\bf{0.070}    &9.058 &\bf{31.303} &\bf{0.976} &\bf{0.965} &\bf{1.113} &\bf{0.056}   &9.445\\
\hline
 \Xhline{1.2pt}
\end{tabular}
\end{center}
\end{table*}

\subsection{Real-world data}
\begin{figure*}[!htb]
\renewcommand\arraystretch{0.8}\setlength{\tabcolsep}{1.8pt}
\begin{center}
\begin{tabular}{cccccccc}
                ~\includegraphics[width=0.8in]{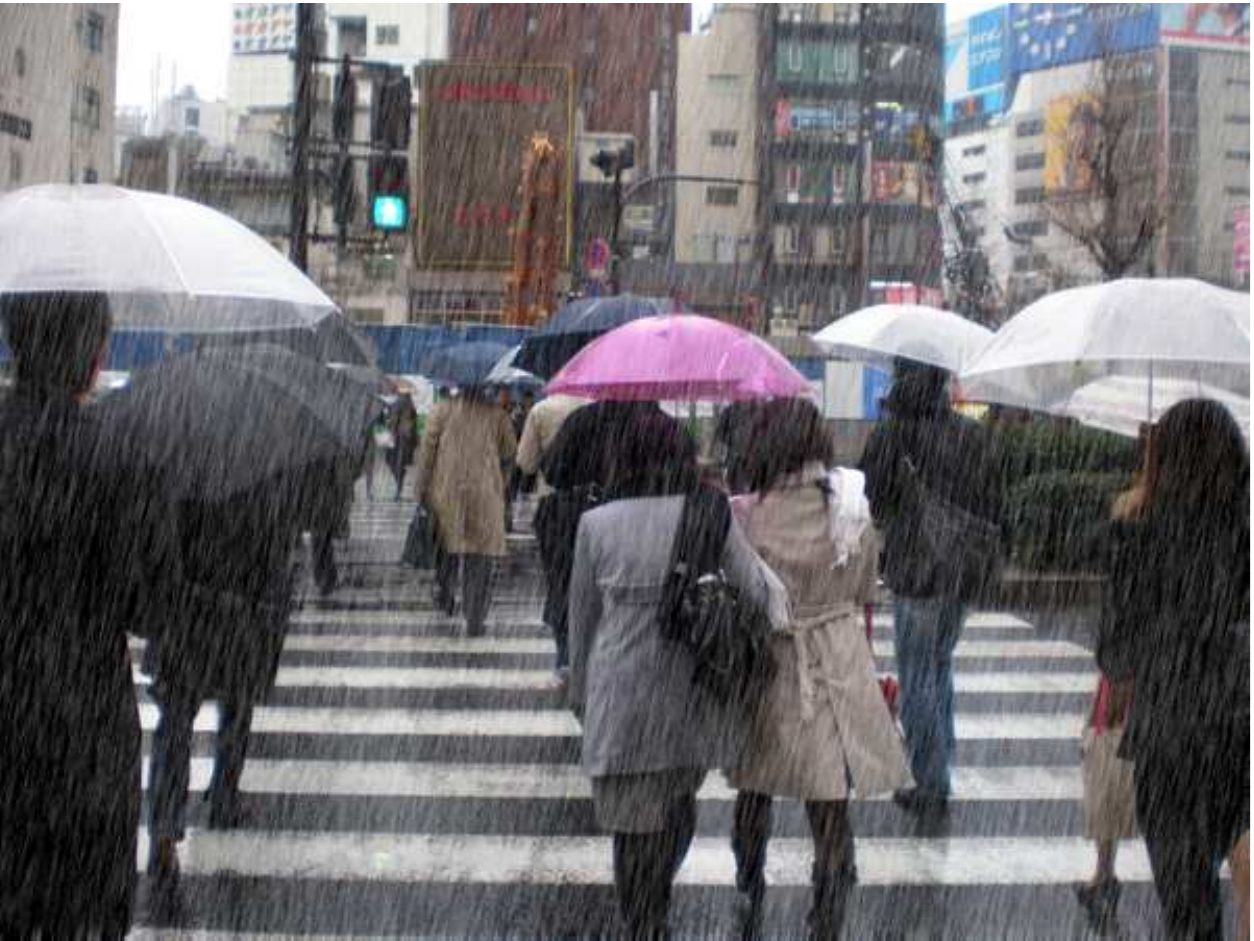}&
                \includegraphics[width=0.8in]{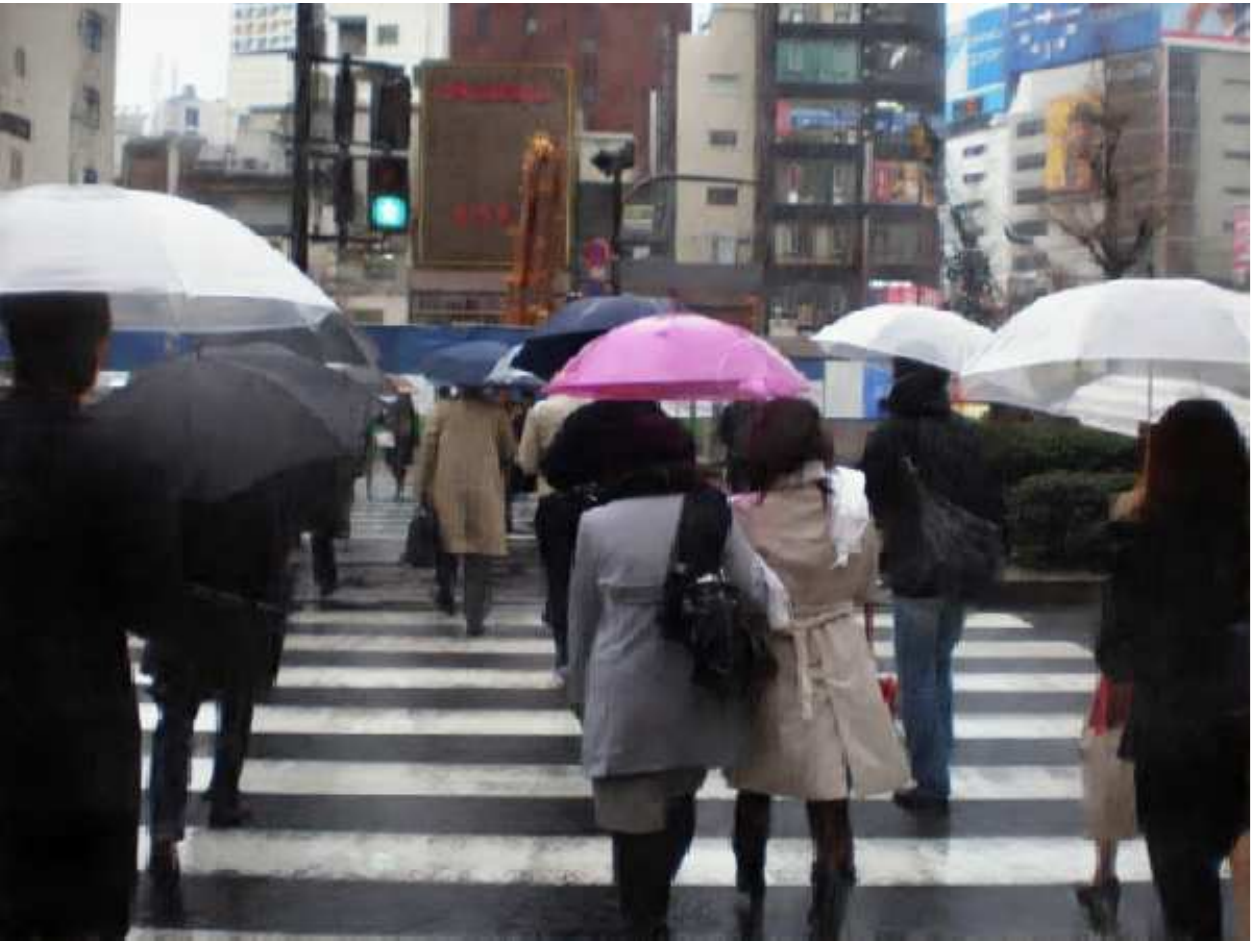}&
                \includegraphics[width=0.8in]{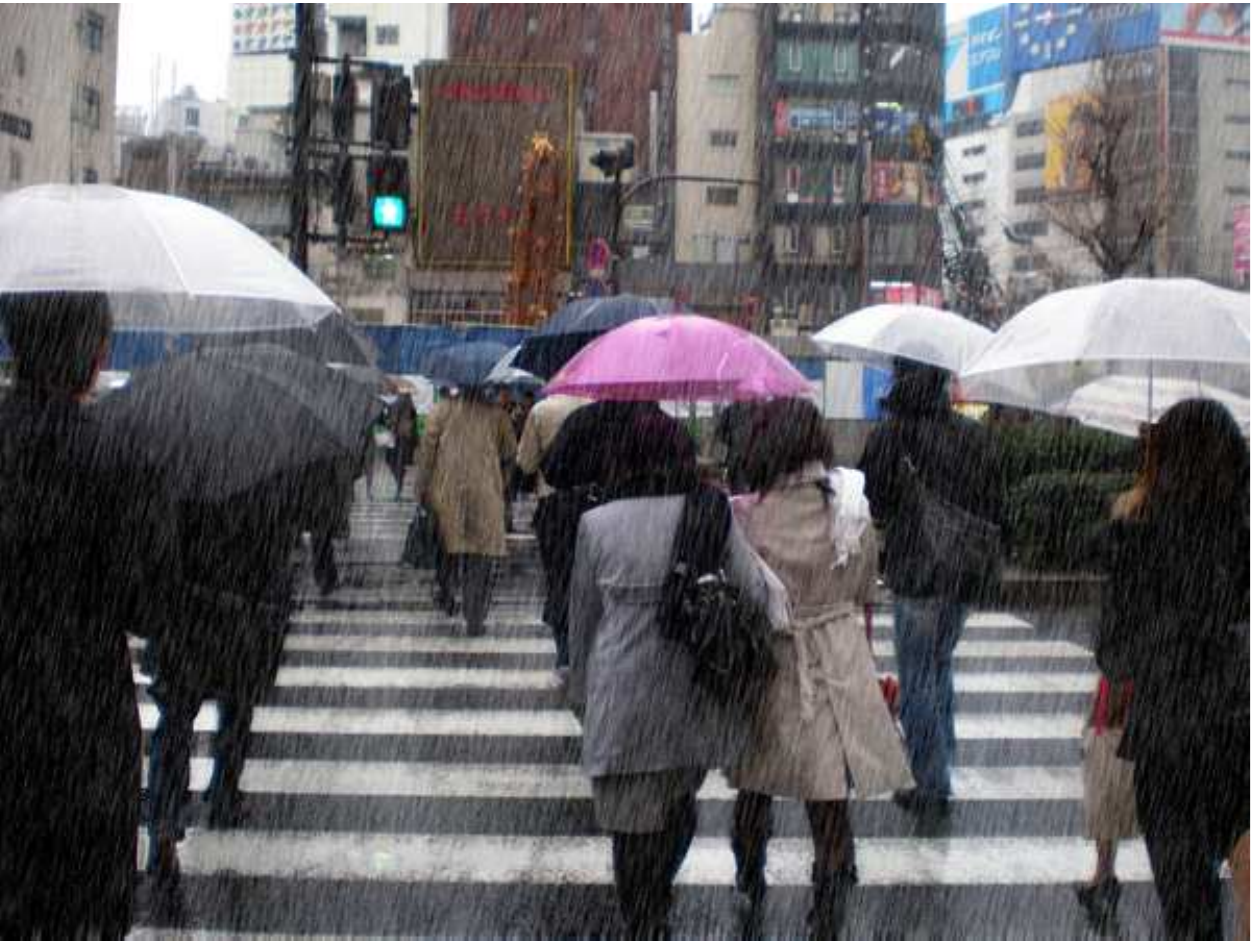}&
                \includegraphics[width=0.8in]{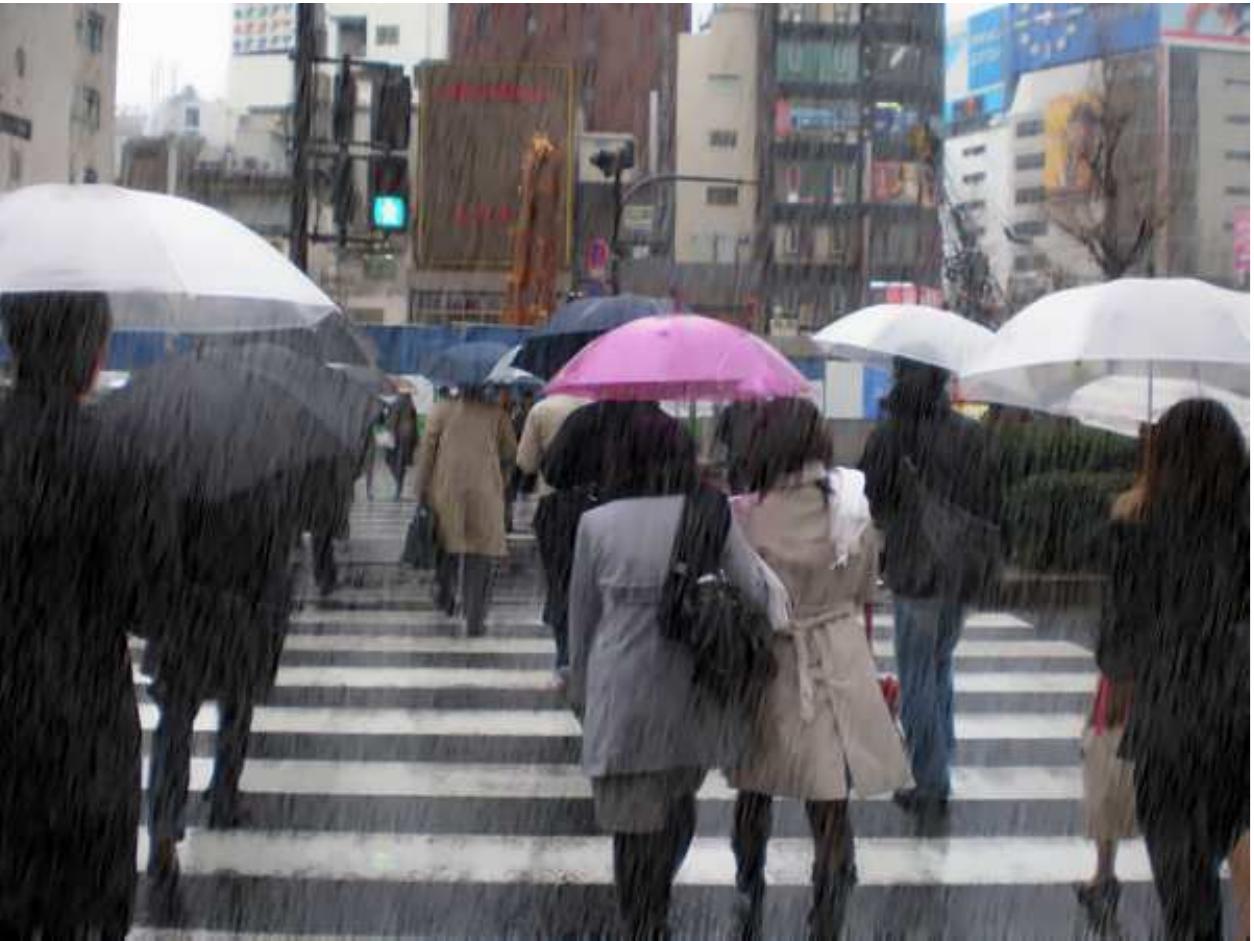}&
                \includegraphics[width=0.8in]{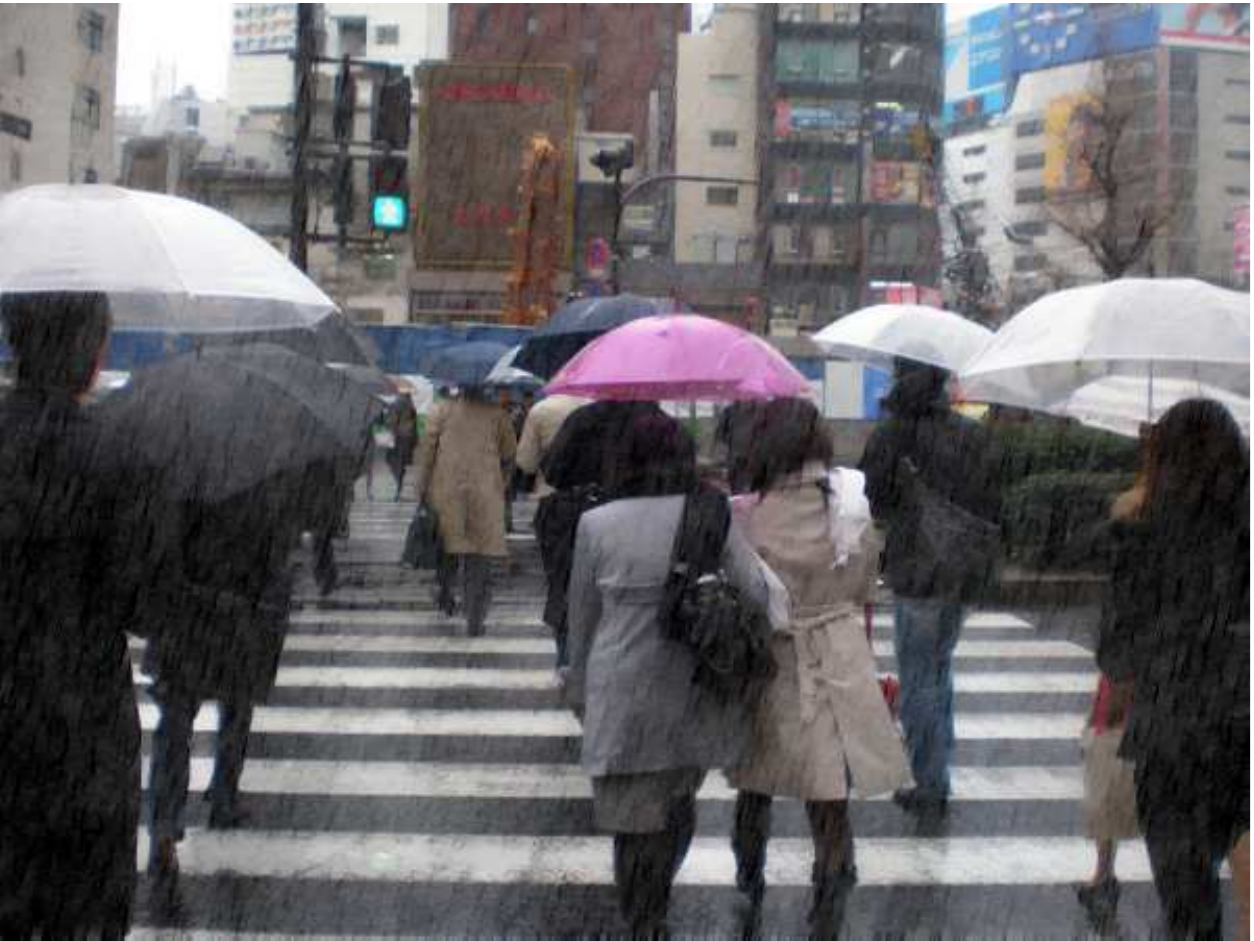}&
                \includegraphics[width=0.8in]{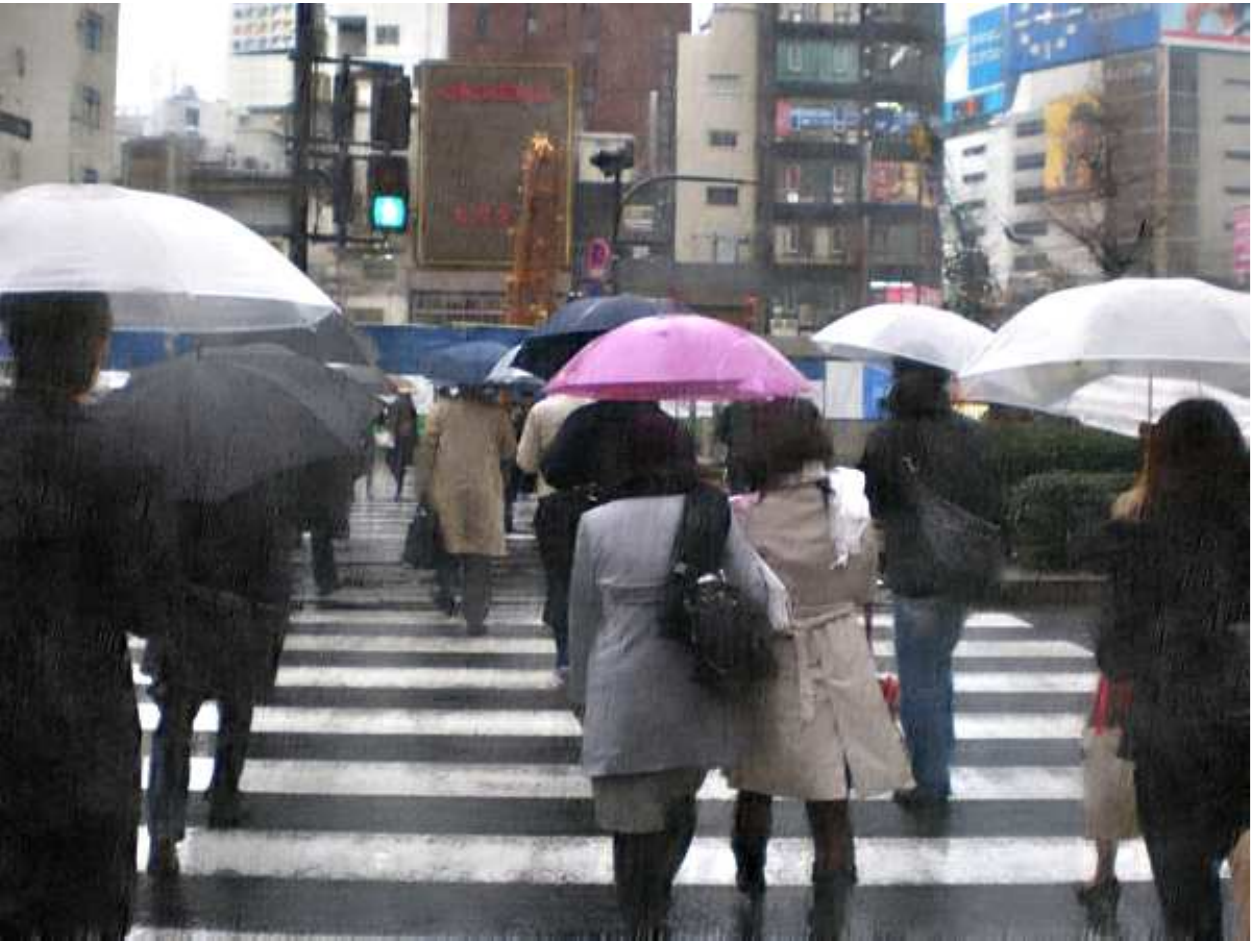}&
                \includegraphics[width=0.8in]{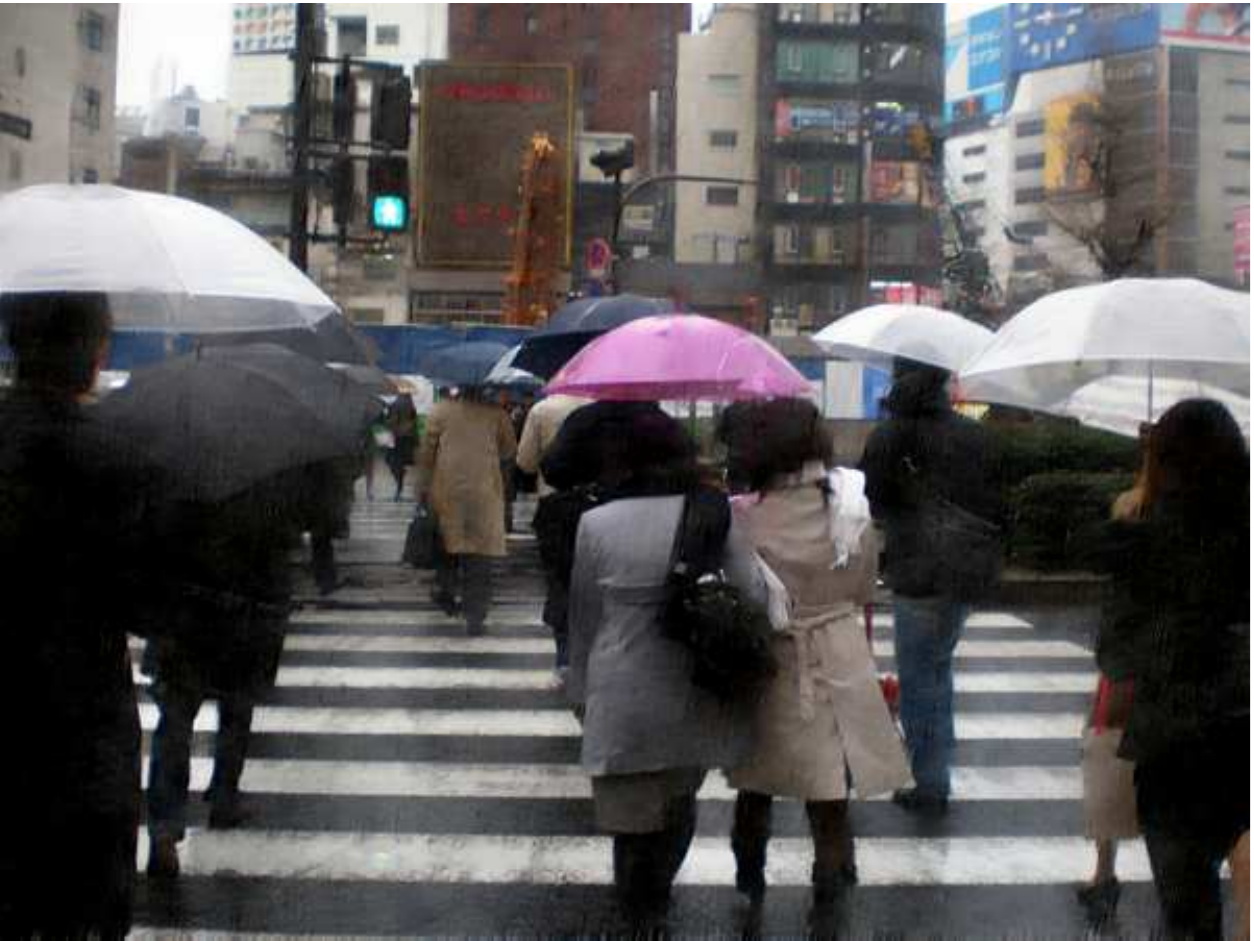}&
                \includegraphics[width=0.8in]{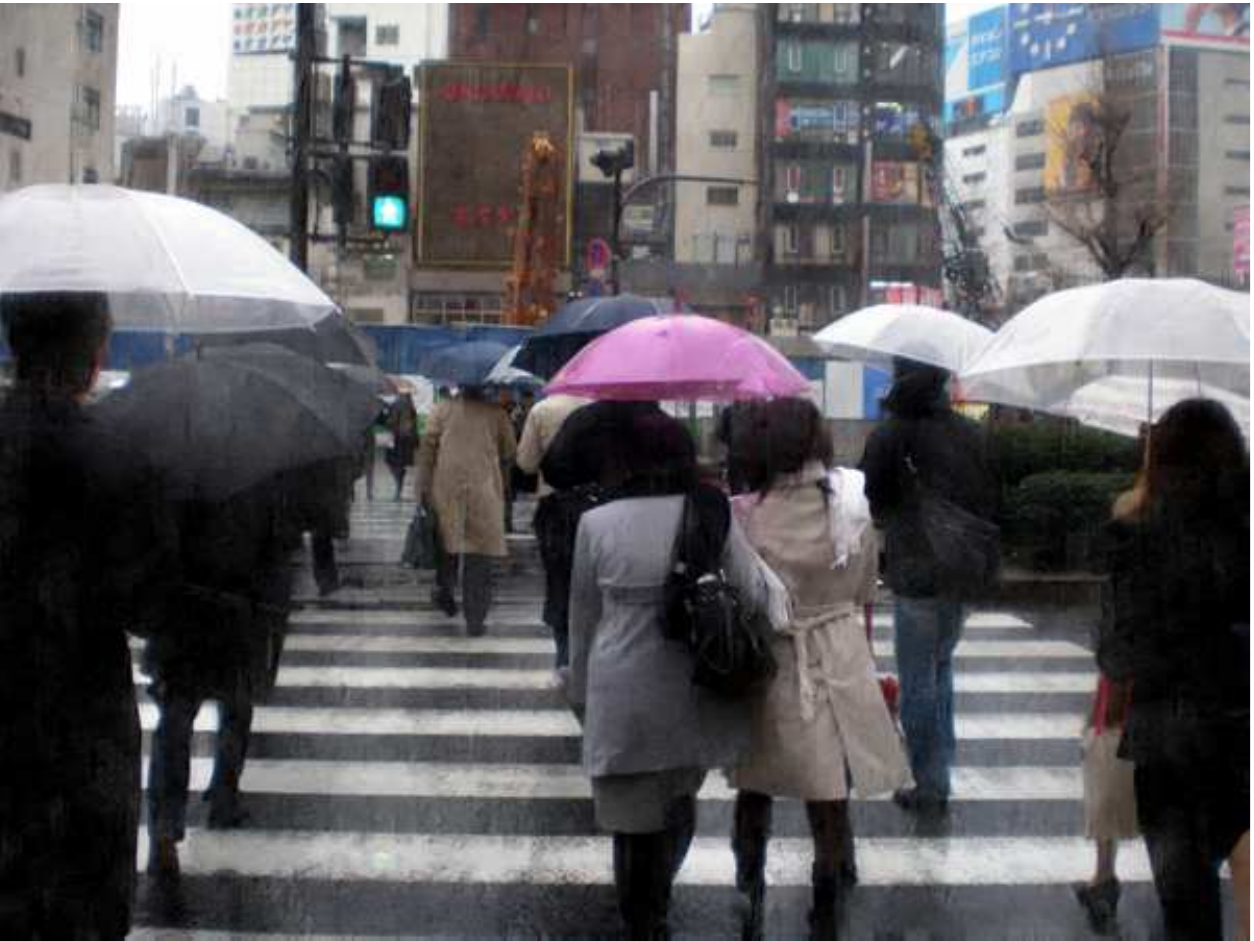}\\
                \vspace{0.5mm}

                \includegraphics[width=0.8in]{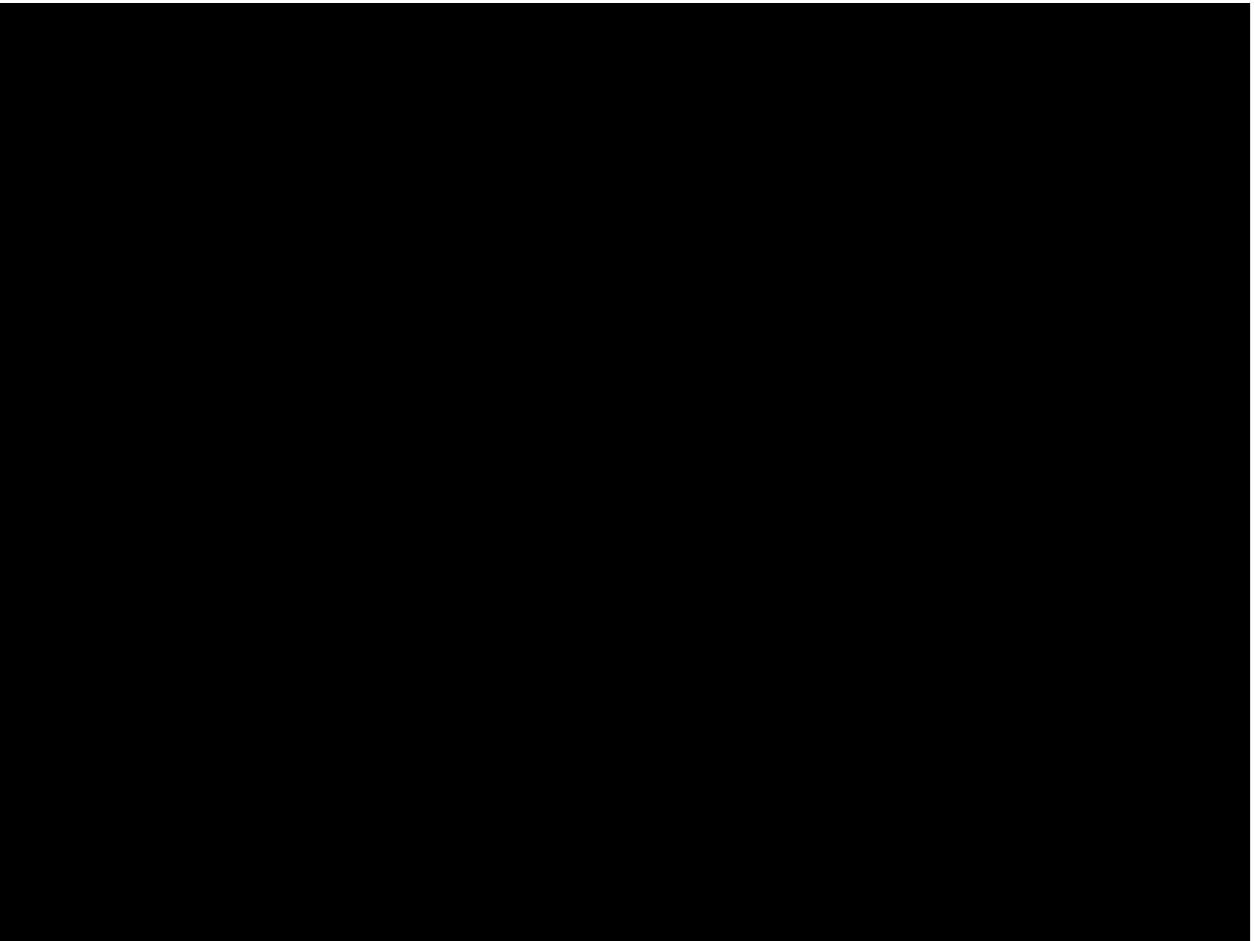}&
                \includegraphics[width=0.8in]{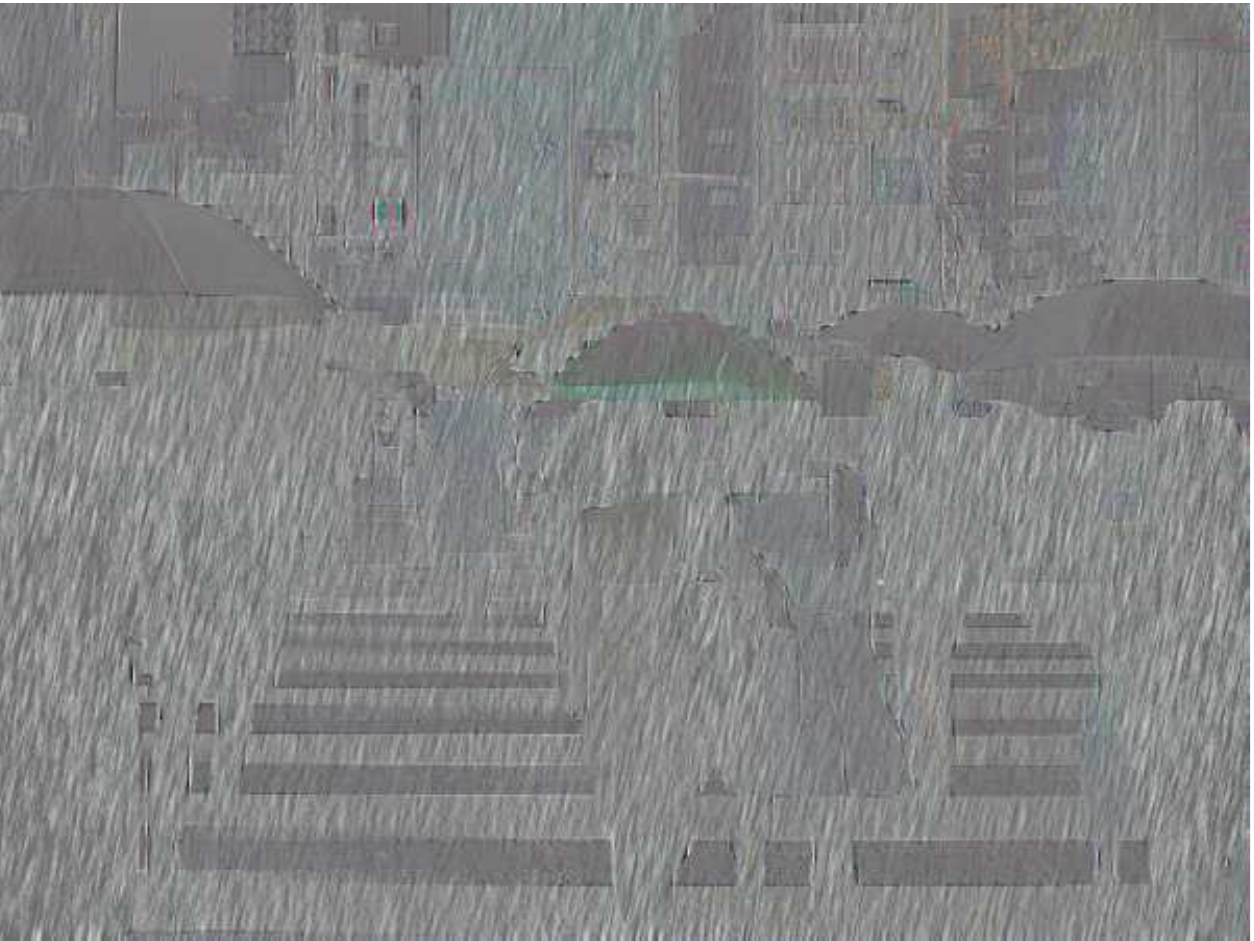}&
                \includegraphics[width=0.8in]{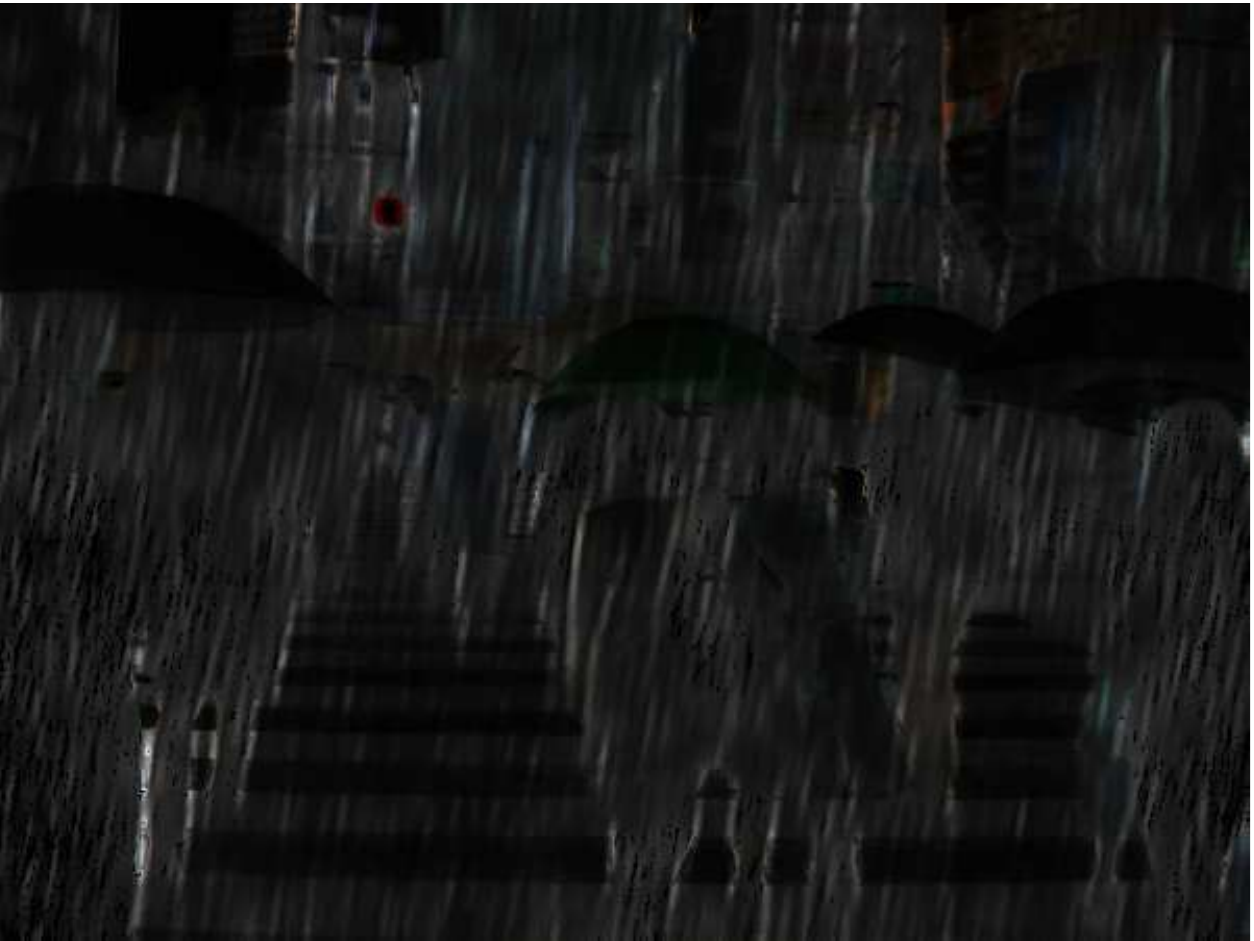}&
                \includegraphics[width=0.8in]{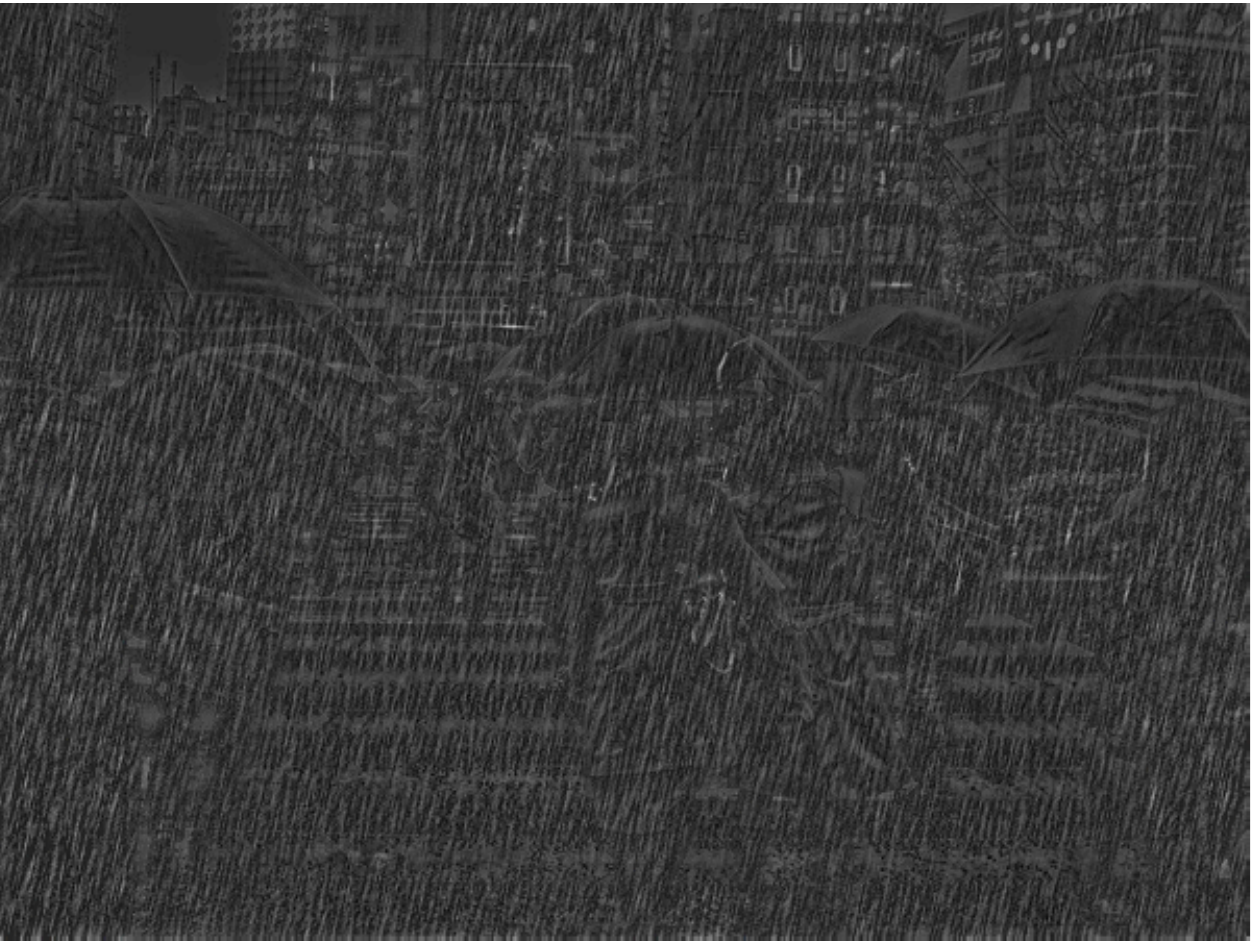}&
                \includegraphics[width=0.8in]{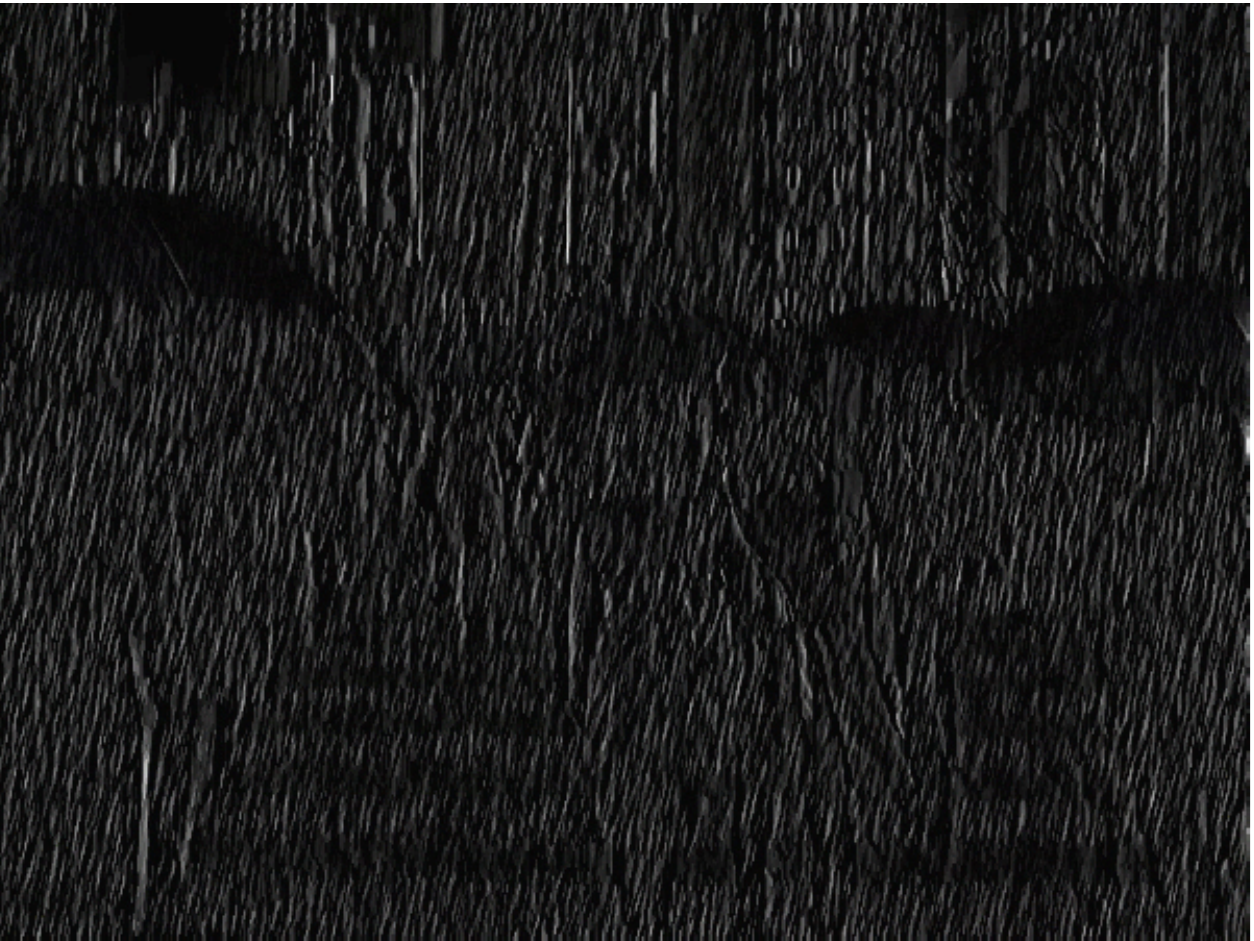}&
                \includegraphics[width=0.8in]{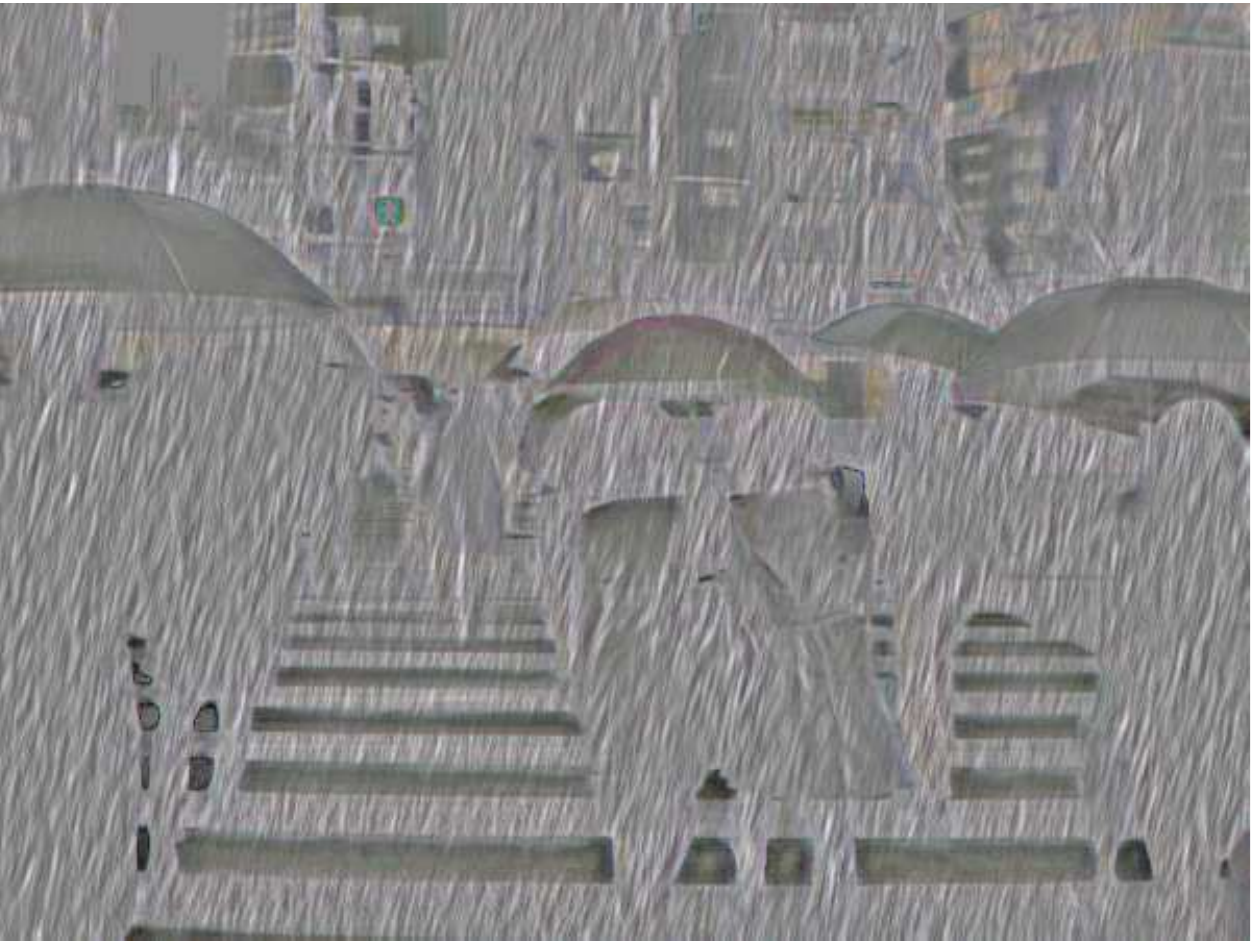}&
                \includegraphics[width=0.8in]{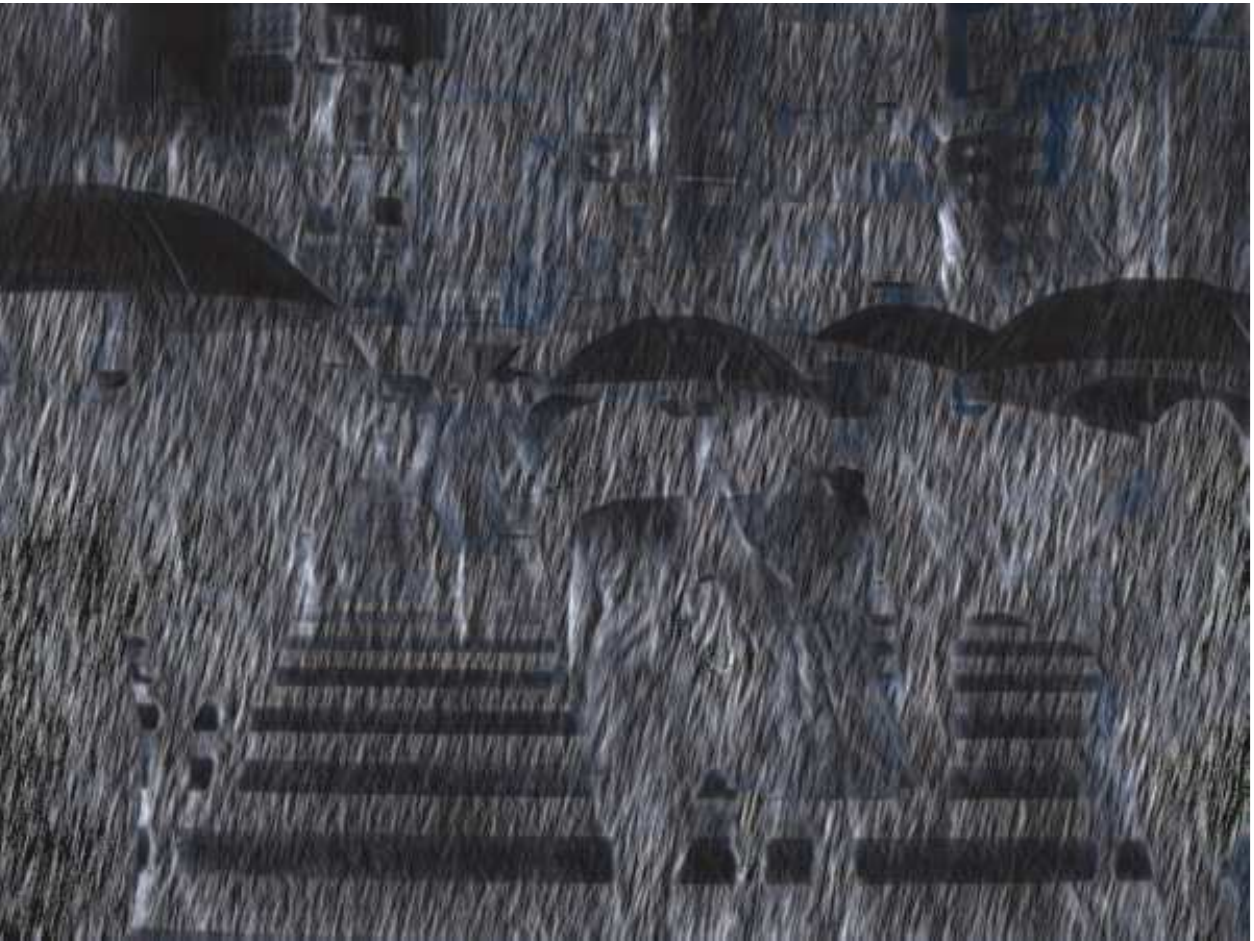}&
                \includegraphics[width=0.8in]{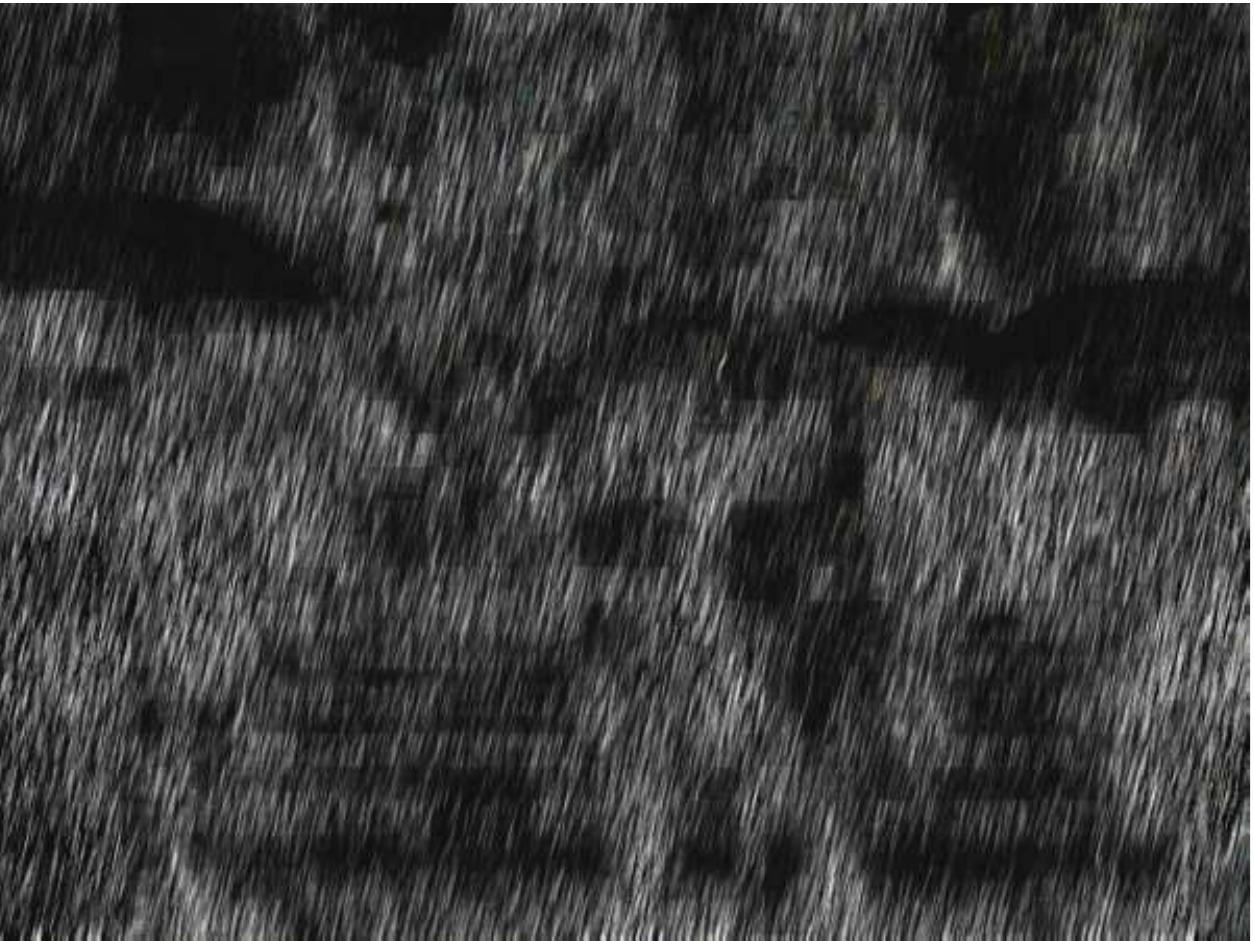}\\
                (a)&
                (b)&
                (c)&
                (d)&
                (e)&
                (f)&
                (g)&
                (h)\\

\end{tabular}

\caption{Rain streak removal results and rain streak images by different methods on real rainy images. From left to right: (a) the rainy images, the results by (b) DID \cite{zhang2018density}, (c) DSC \cite{luo2015removing}, (d) LP \cite{Li2014Single}, (e) UGSM \cite{Deng2018A}, (f) CNN \cite{fu2017clearing}, (g) DDN \cite{fu2017removing}, and (h) KGCNN.
}
\label{real-visual1}
\end{center}
\end{figure*}

For real-world data, since the ground truth images are unknown, we do not give the quantitative comparisons and only evaluate the performance of different methods visually, including the derain images and the rain streak images.
From Fig. \ref{real-visual1}, the methods of DID, DDN, and KGCNN exhibit similar visual results and remove rain streaks completely, while other approaches fail to remove all rain streaks.
In addition, from the rain streak images in Fig. \ref{real-visual1}, the methods of DID and DDN fail to separate the rain streaks and background texture well and leave some background texture to rain streaks, which demonstrates that the networks of DID and DDN could not distinguish background texture and the rain streaks well.
Moreover, the DID method also changes image contrast significantly.

%

\begin{figure*}[!htb]
\renewcommand\arraystretch{0.8}\setlength{\tabcolsep}{1.8pt}
\begin{center}
\begin{tabular}{cccccccc}

                ~\includegraphics[width=0.8in]{rain/real/rainy/11-eps-converted-to.pdf}&
                \includegraphics[width=0.8in]{rain/real/DID/11-eps-converted-to.pdf}&
                \includegraphics[width=0.8in]{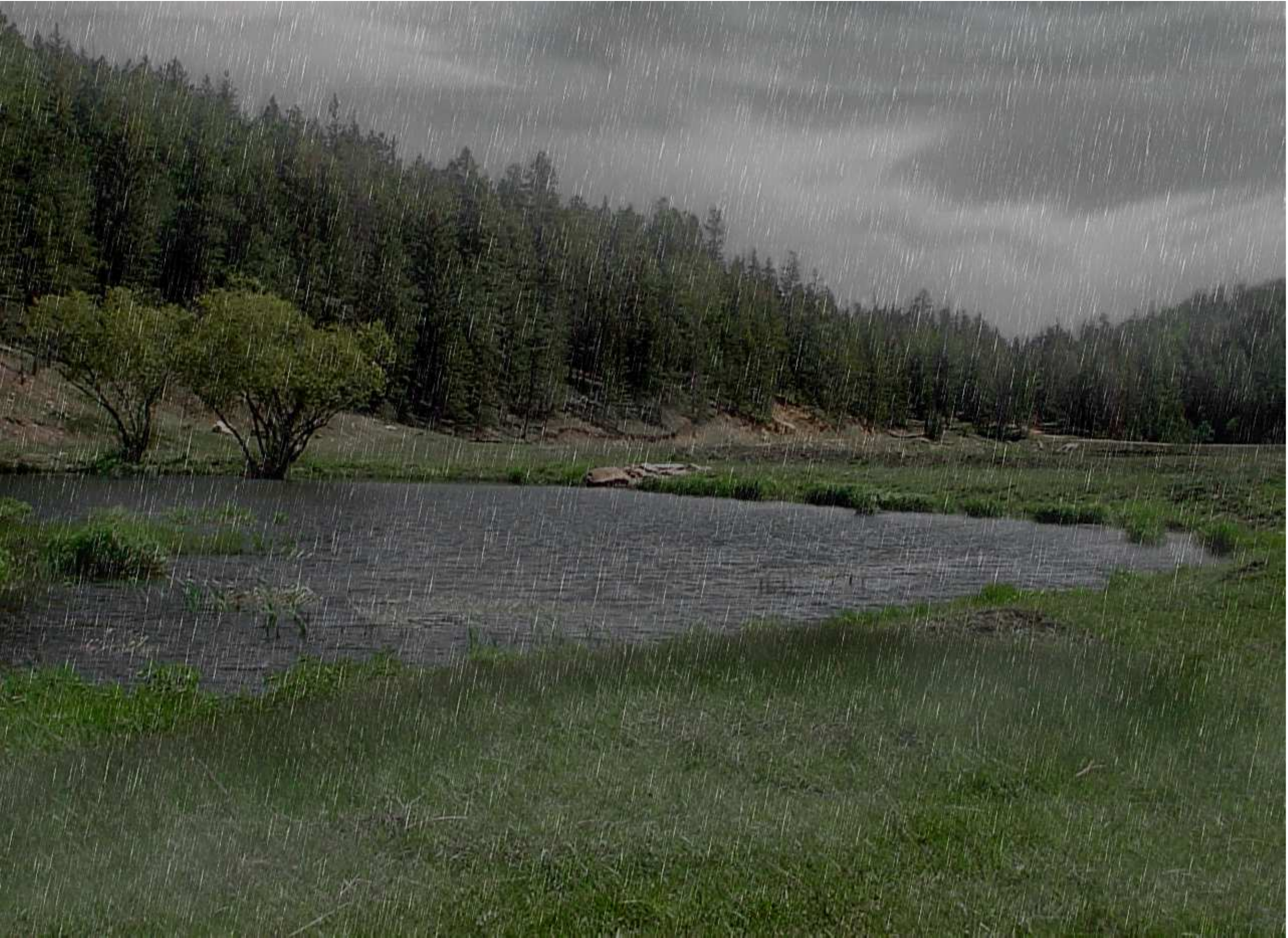}&
                \includegraphics[width=0.8in]{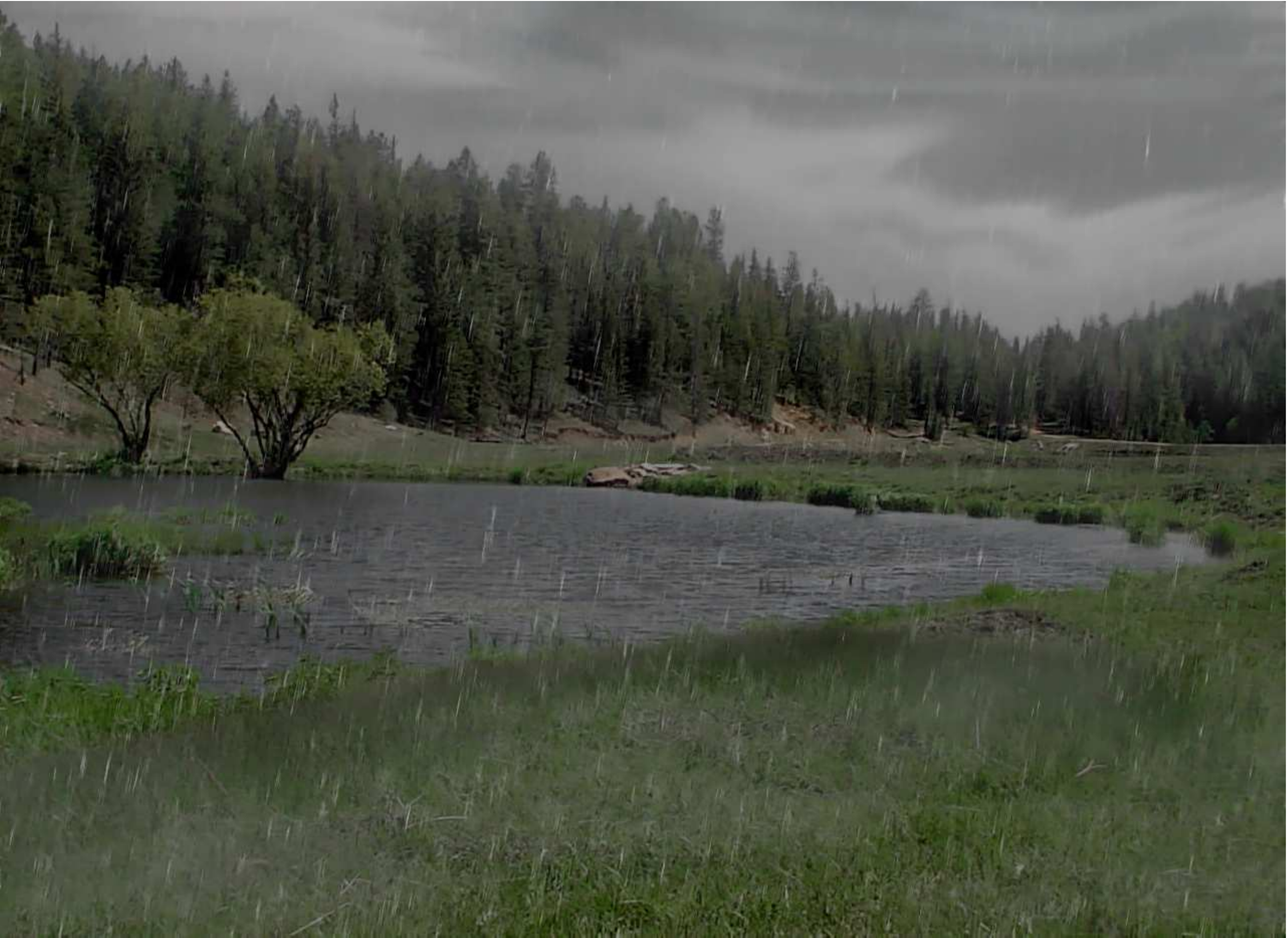}&
                \includegraphics[width=0.8in]{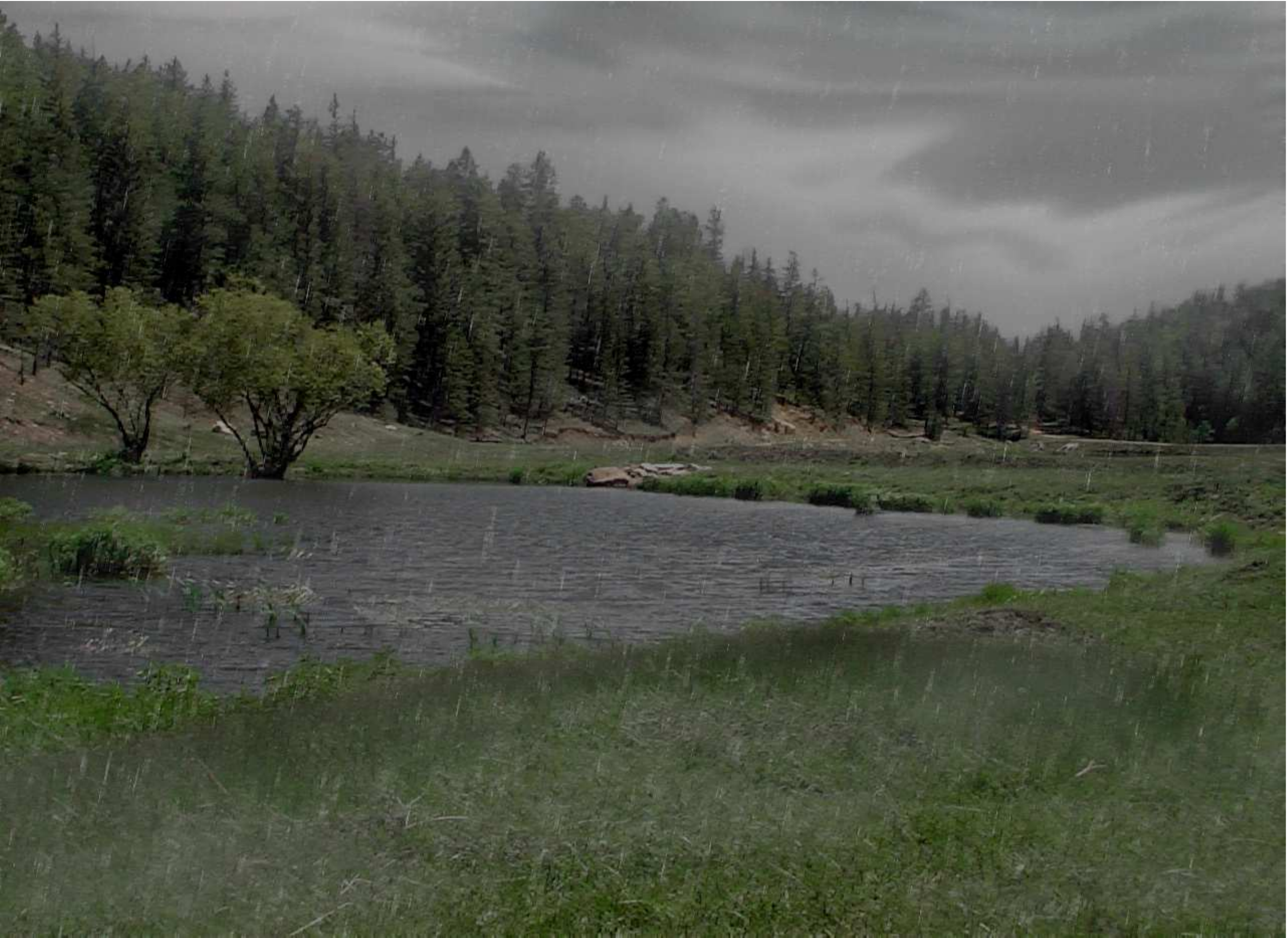}&
                \includegraphics[width=0.8in]{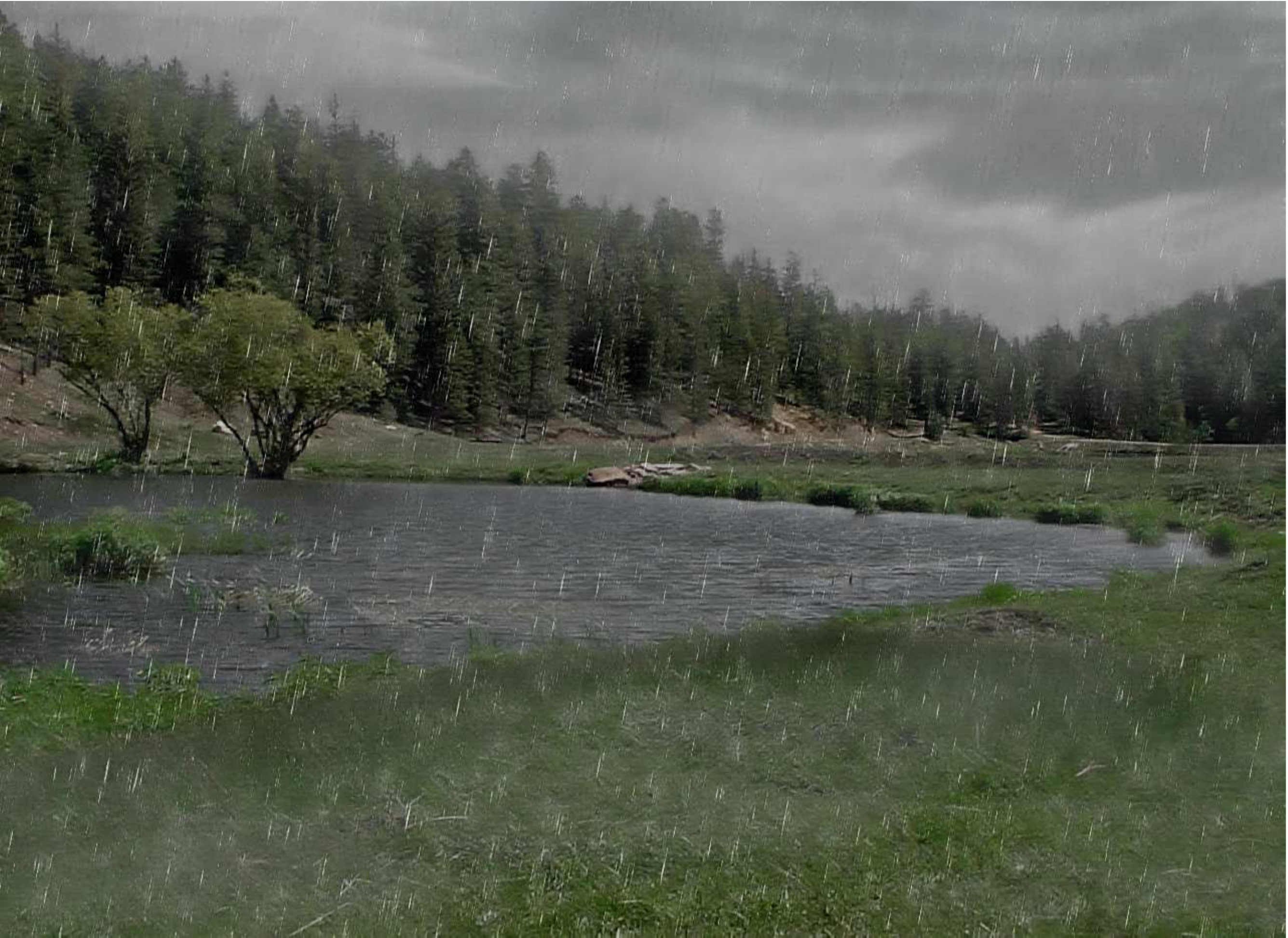}&
                \includegraphics[width=0.8in]{rain/real/CVPR/11-eps-converted-to.pdf}&
                \includegraphics[width=0.8in]{rain/real/our/11-eps-converted-to.pdf}\\
                \vspace{0.5mm}

                \includegraphics[width=0.8in]{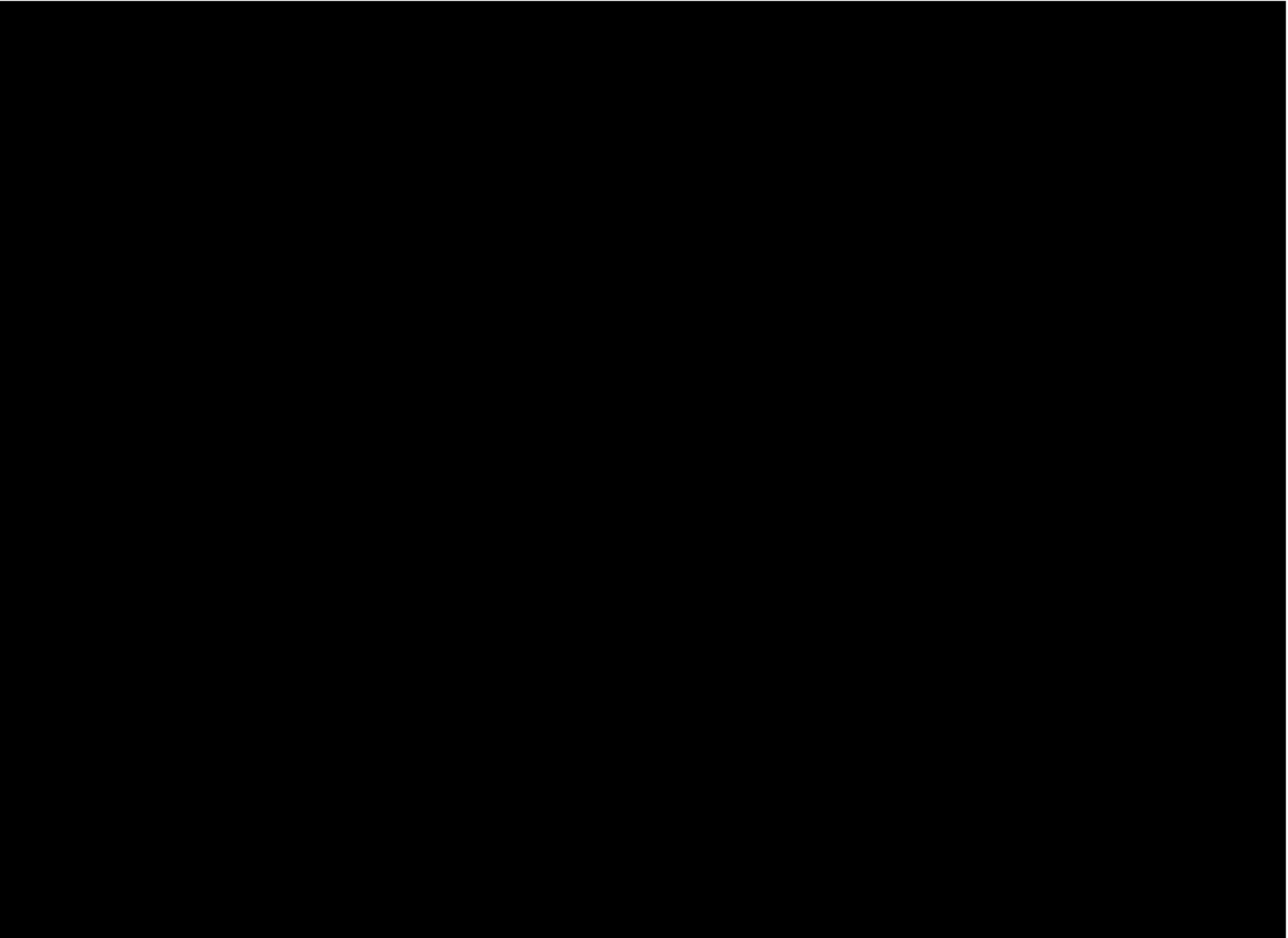}&
                \includegraphics[width=0.8in]{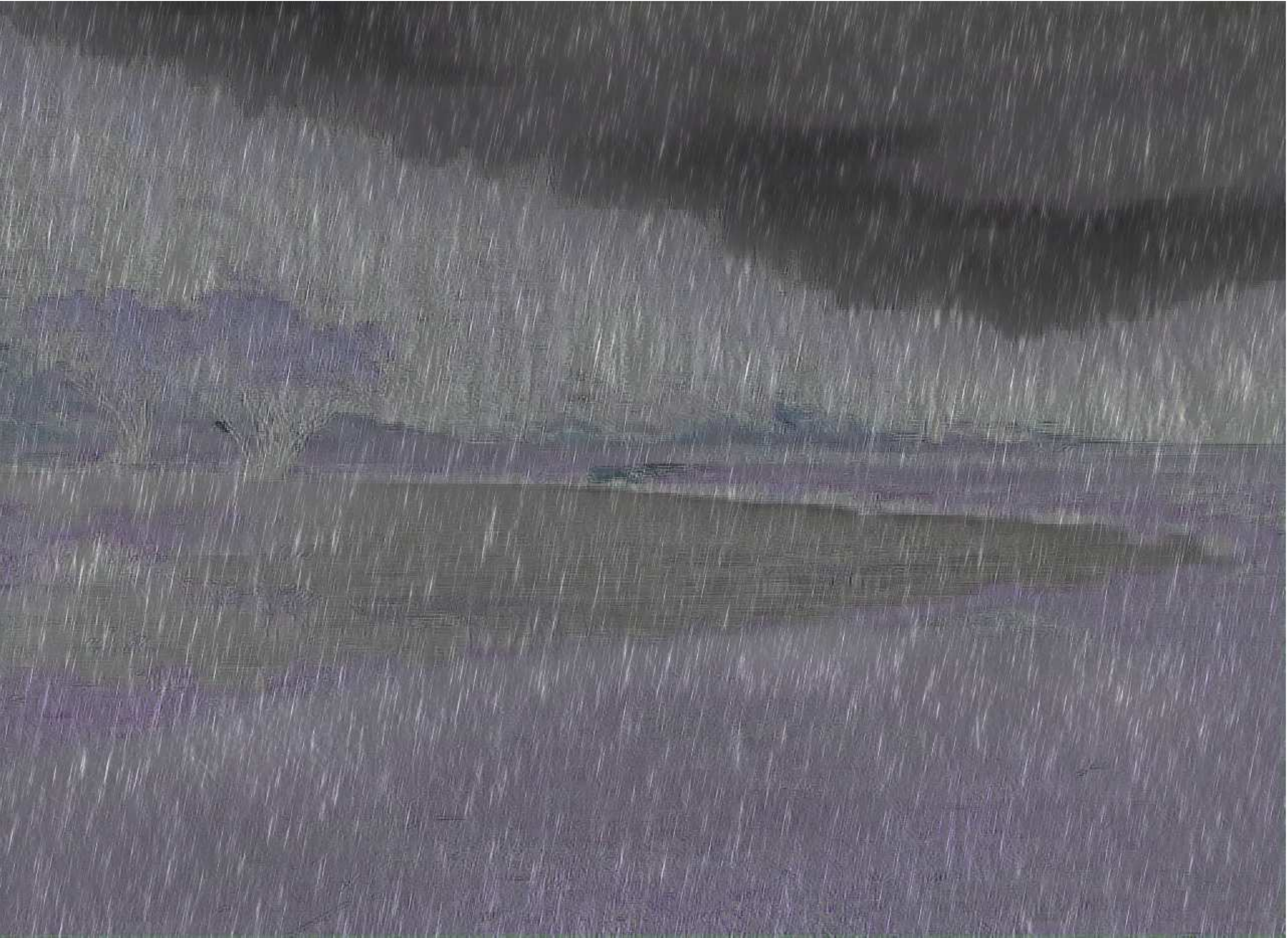}&
                \includegraphics[width=0.8in]{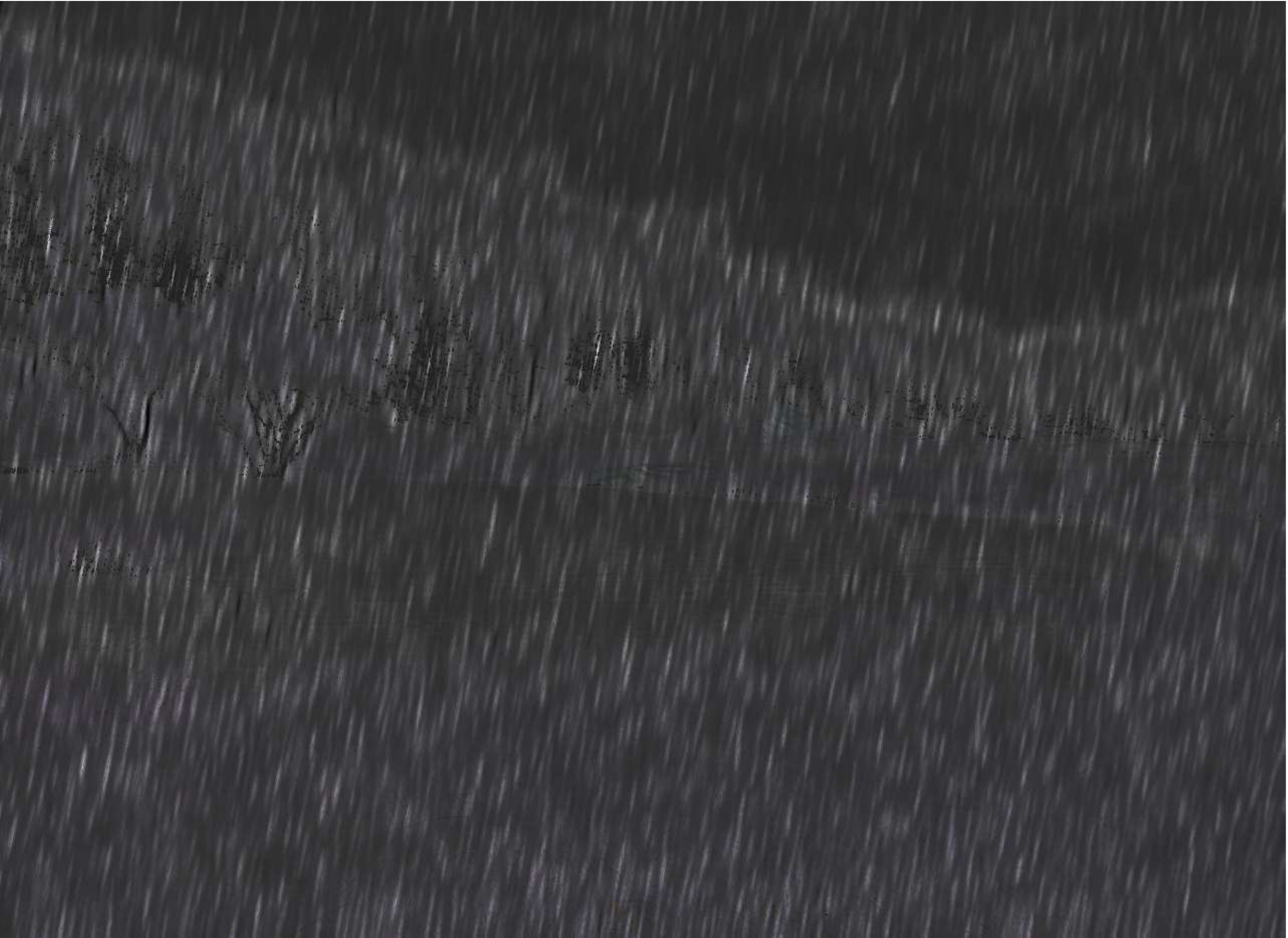}&
                \includegraphics[width=0.8in]{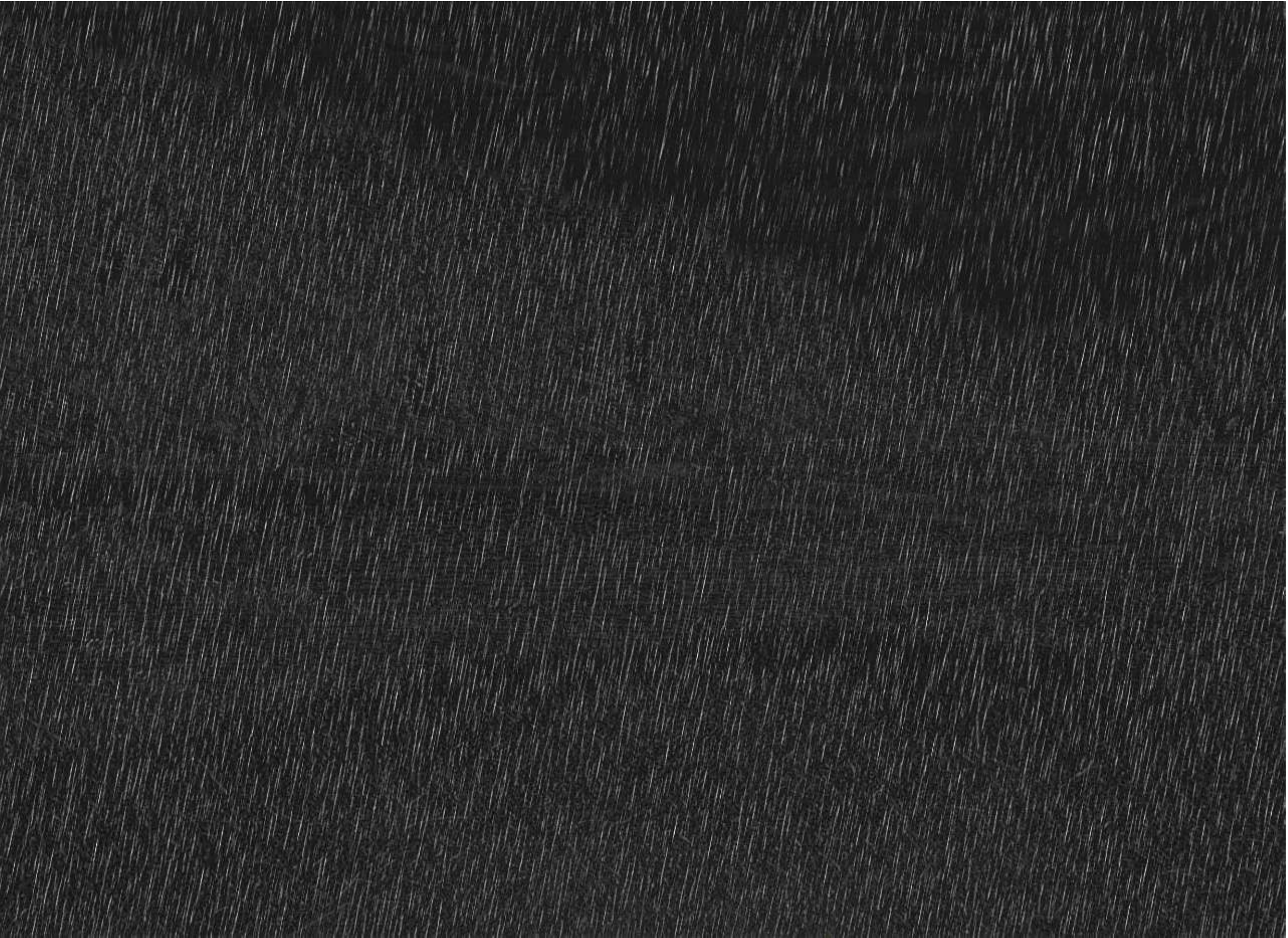}&
                \includegraphics[width=0.8in]{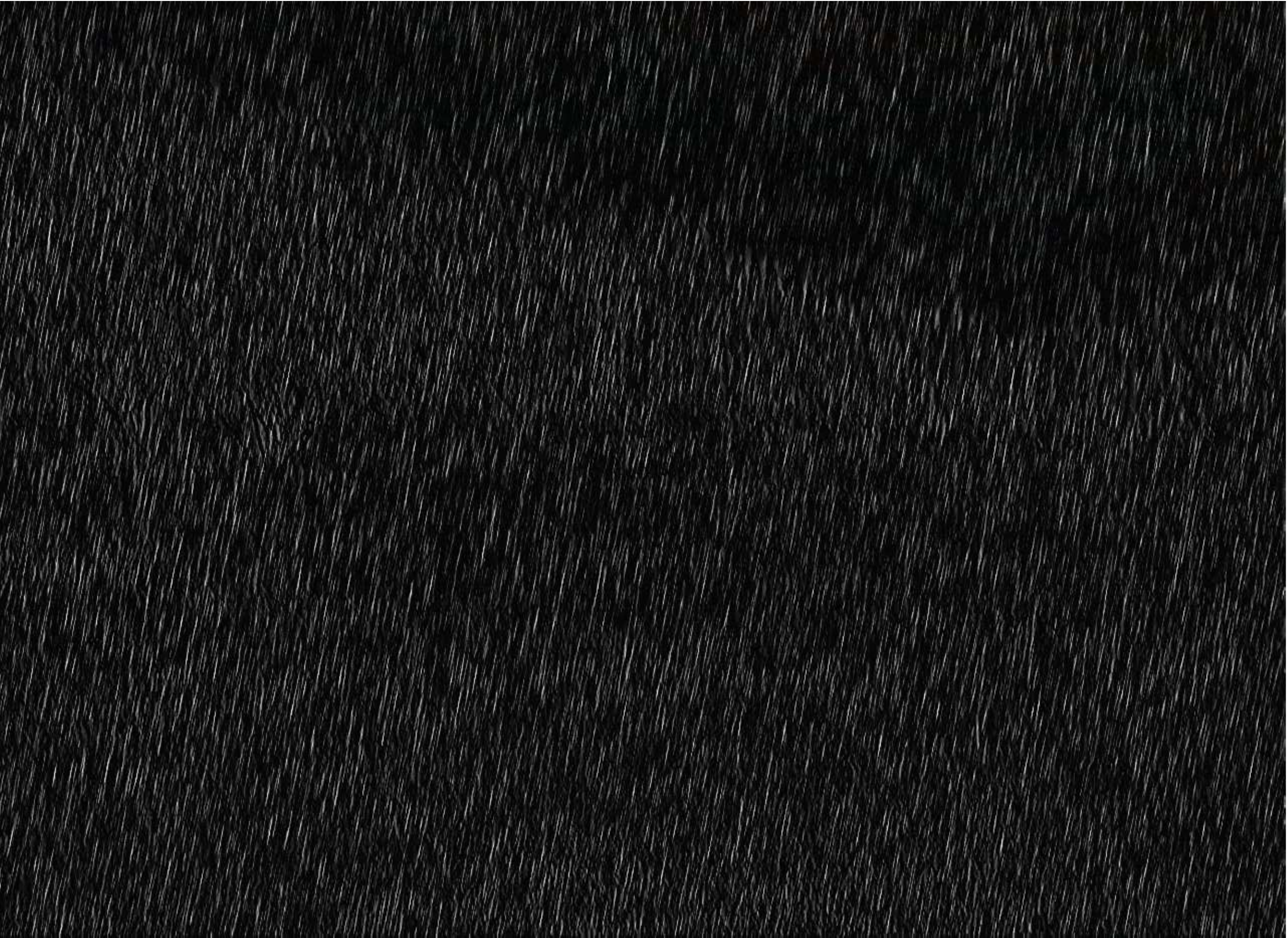}&
                \includegraphics[width=0.8in]{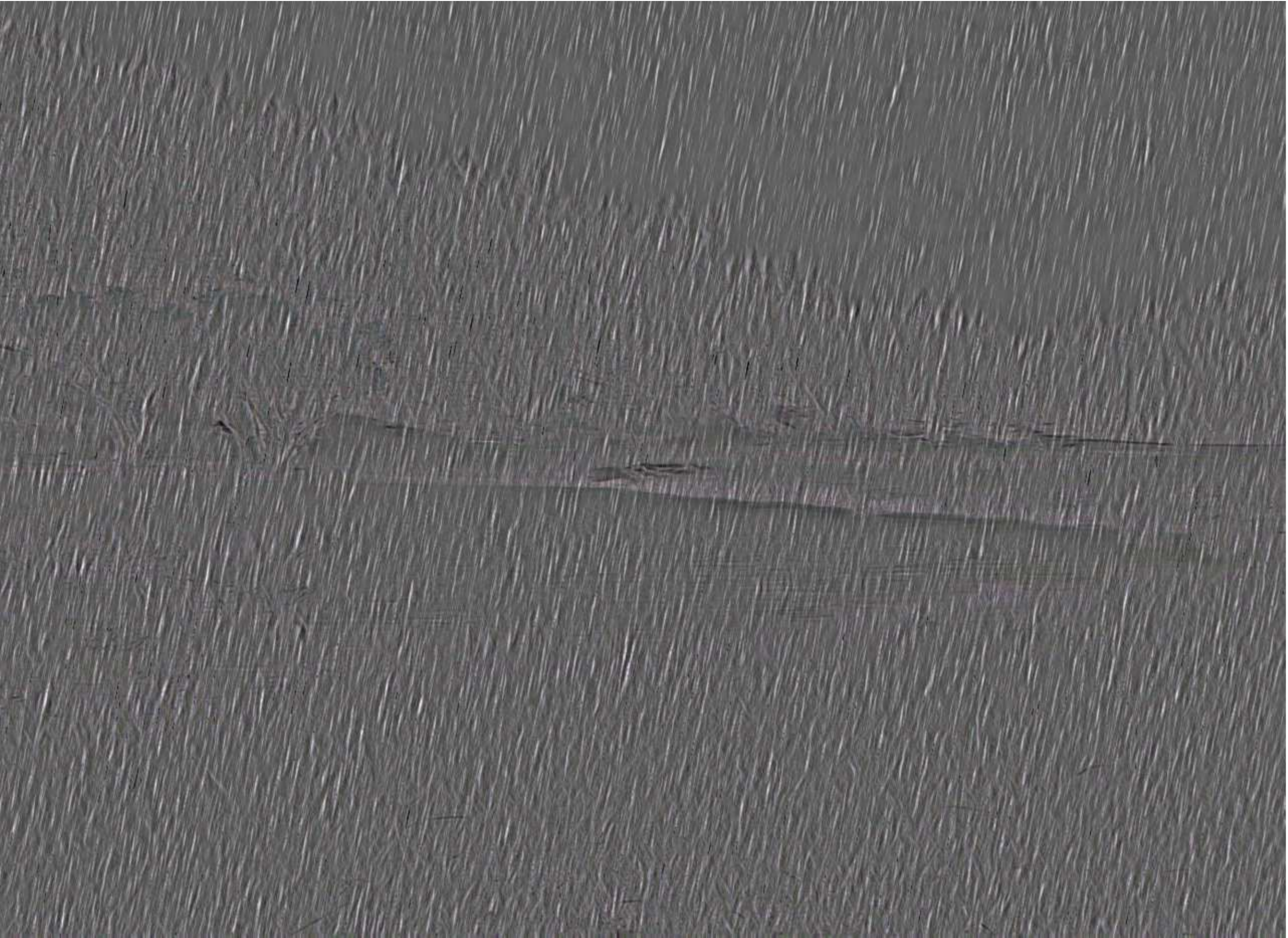}&
                \includegraphics[width=0.8in]{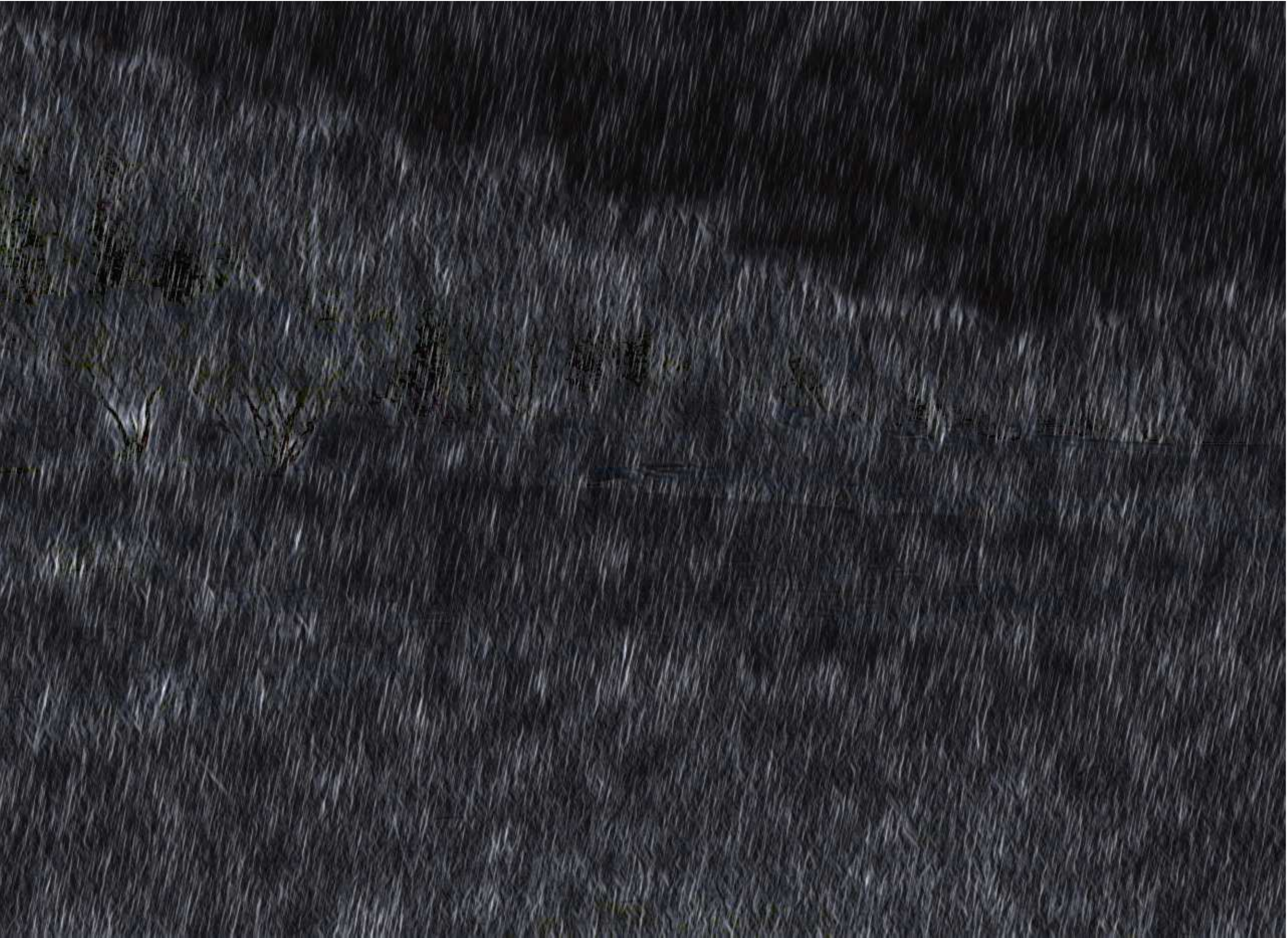}&
                \includegraphics[width=0.8in]{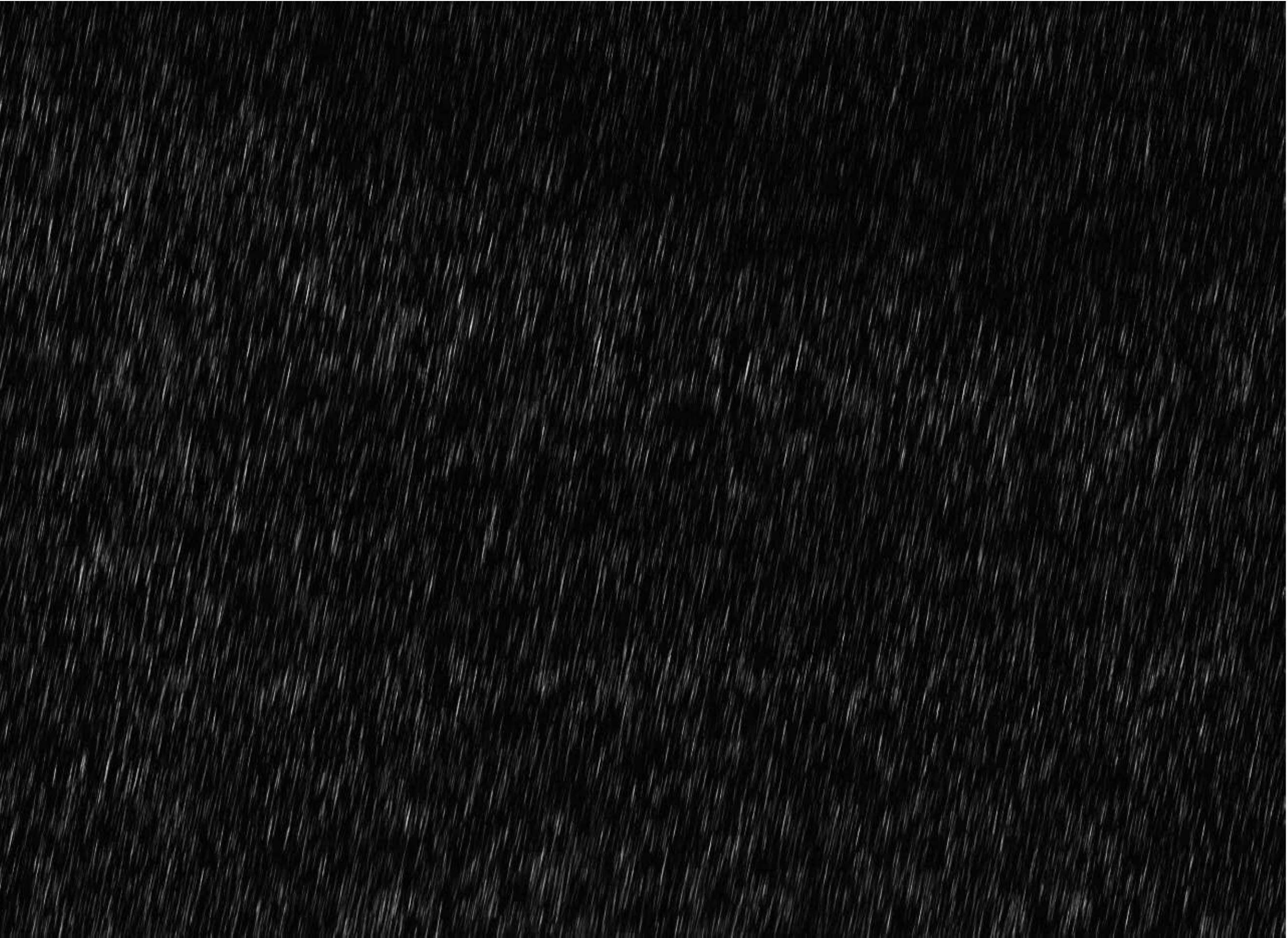}\\

                (a)&
                (b)&
                (c)&
                (d)&
                (e)&
                (f)&
                (g)&
                (h)\\

\end{tabular}
\caption{Rain streak removal results and rain streak images by different methods on real rainy images. From left to right: (a) the rainy images, the results by (b) DID \cite{zhang2018density}, (c) DSC \cite{luo2015removing}, (d) LP \cite{Li2014Single}, (e) UGSM \cite{Deng2018A}, (f) CNN \cite{fu2017clearing}, (g) DDN \cite{fu2017removing}, and (h) KGCNN.}
\label{real-visual2}
\end{center}
\end{figure*}

\begin{figure*}[!htb]
\renewcommand\arraystretch{0.8}\setlength{\tabcolsep}{1.8pt}
\begin{center}
\begin{tabular}{cccccccc}
                ~\includegraphics[width=0.8in]{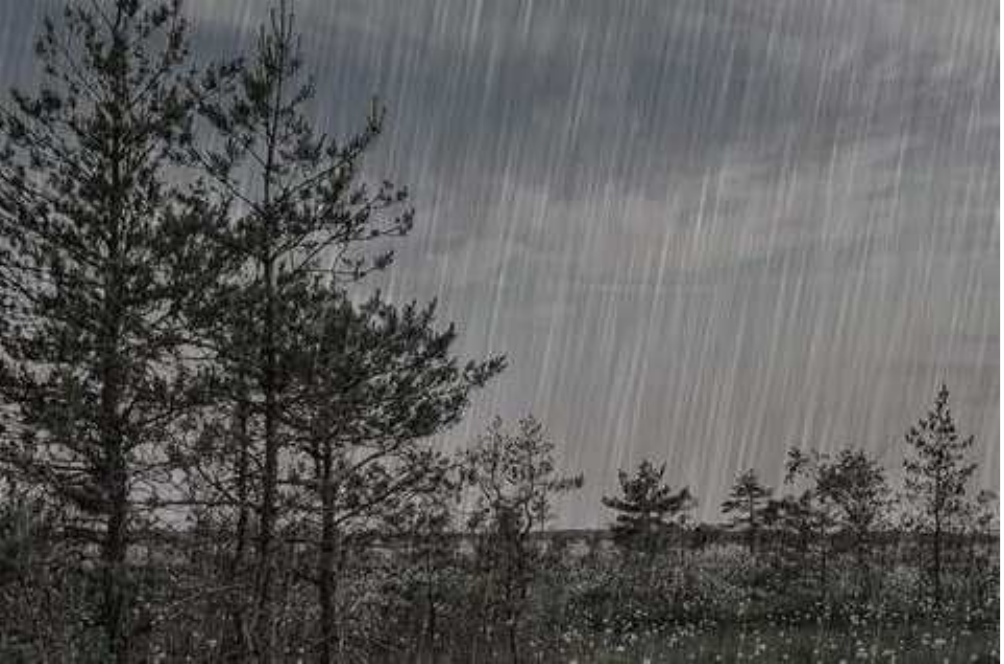}&
                \includegraphics[width=0.8in]{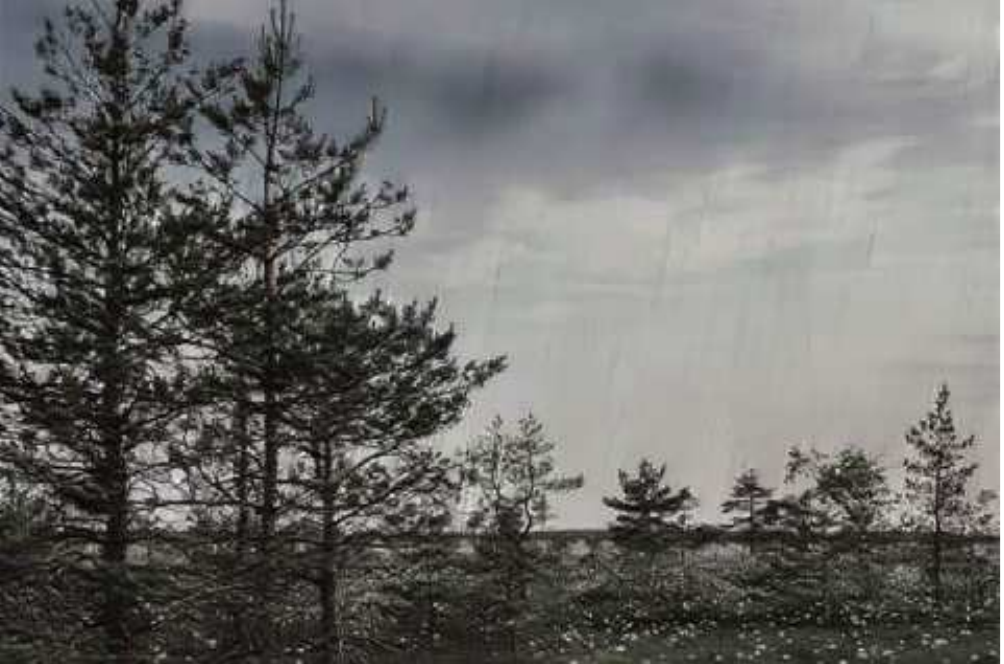}&
                \includegraphics[width=0.8in]{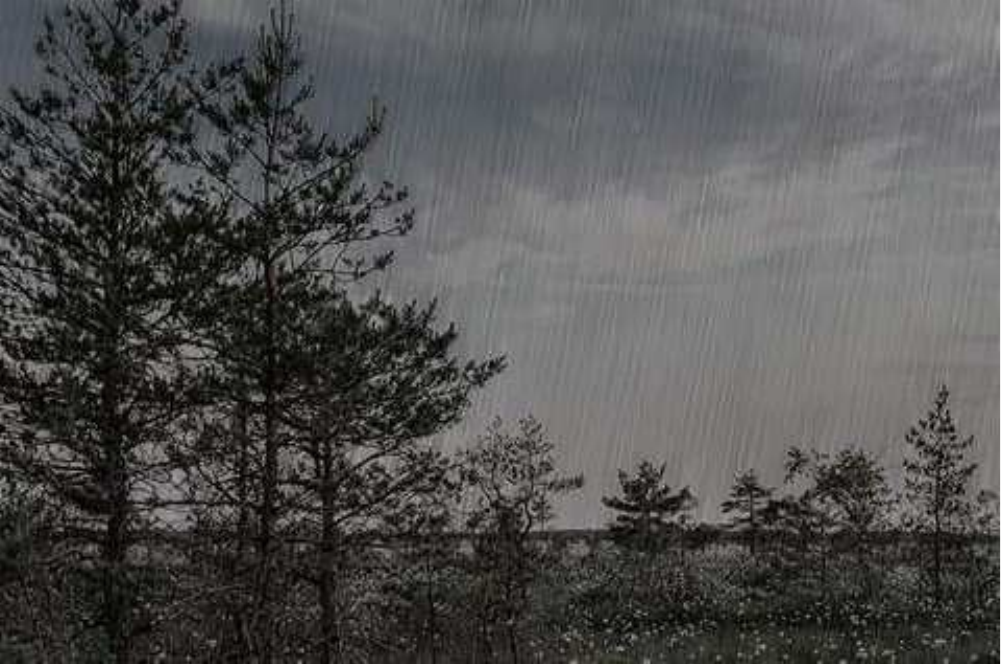}&
                \includegraphics[width=0.8in]{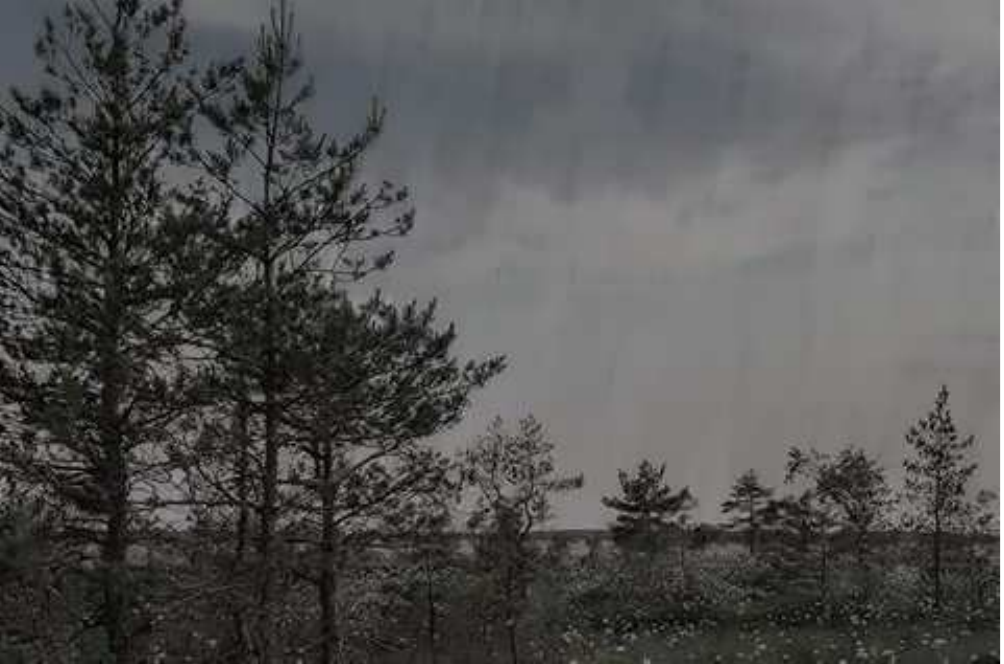}&
                \includegraphics[width=0.8in]{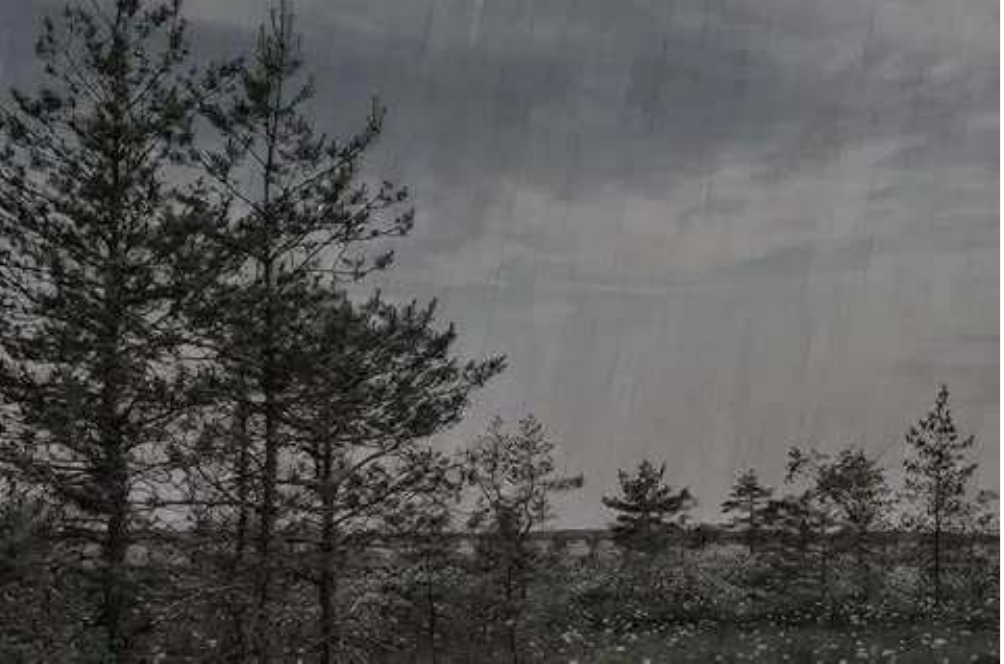}&
                \includegraphics[width=0.8in]{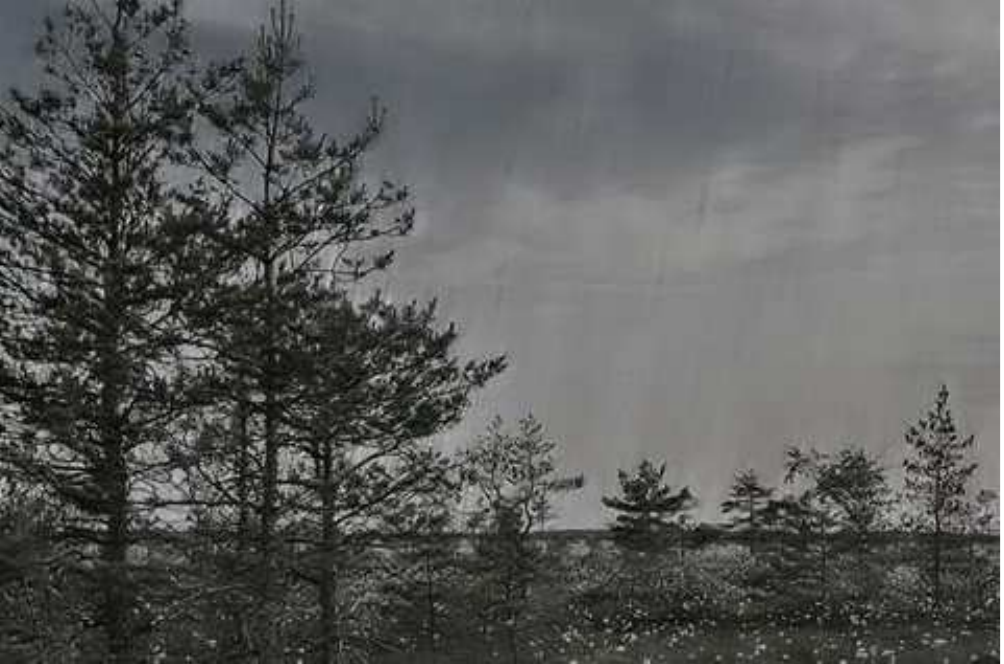}&
                \includegraphics[width=0.8in]{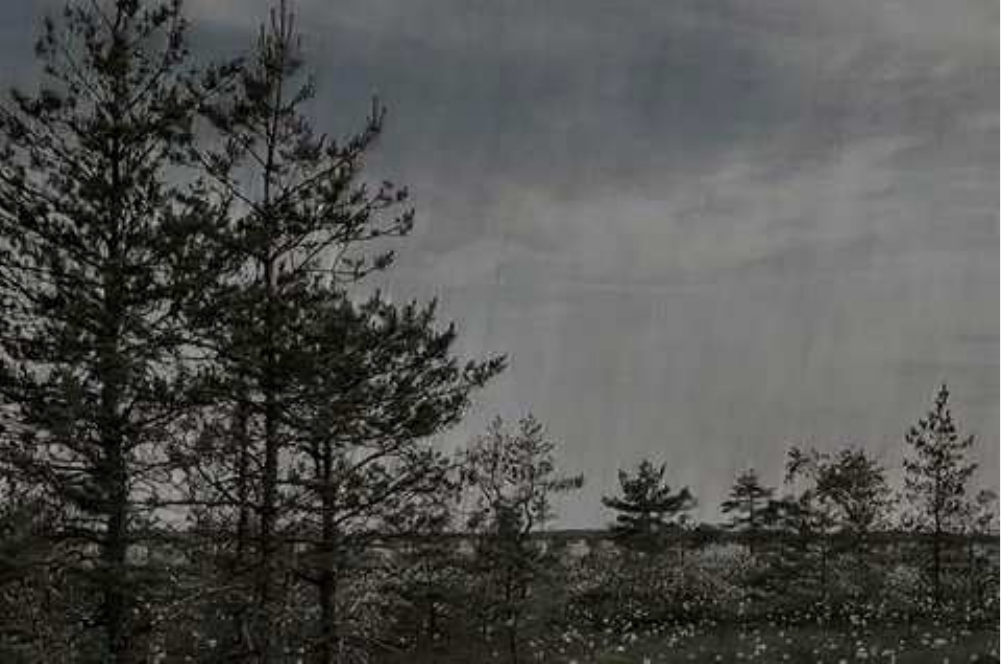}&
                \includegraphics[width=0.8in]{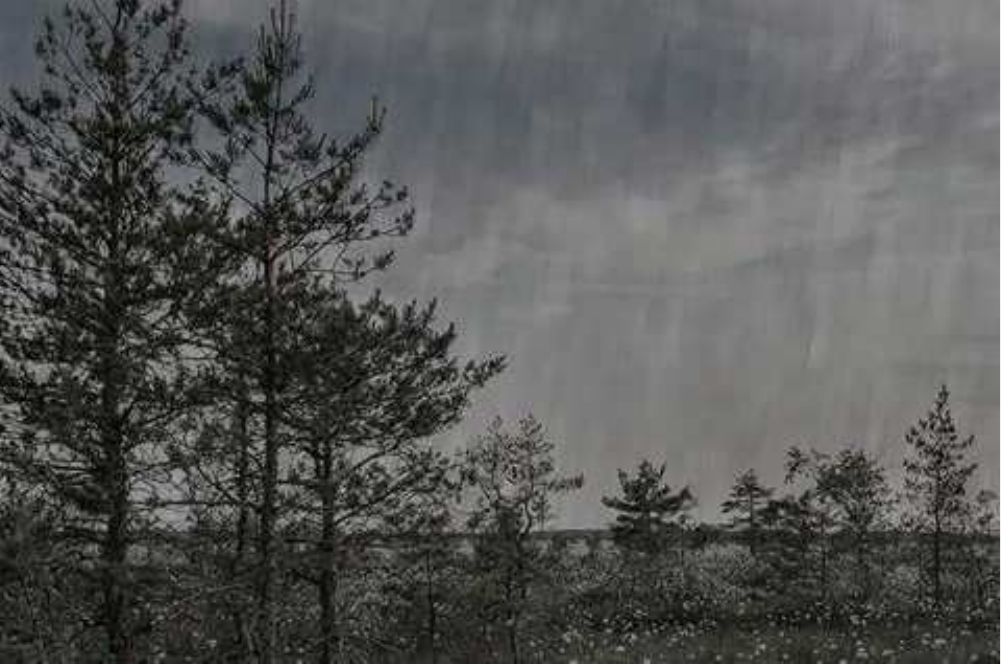}\\
                \vspace{0.5mm}

                \includegraphics[width=0.8in]{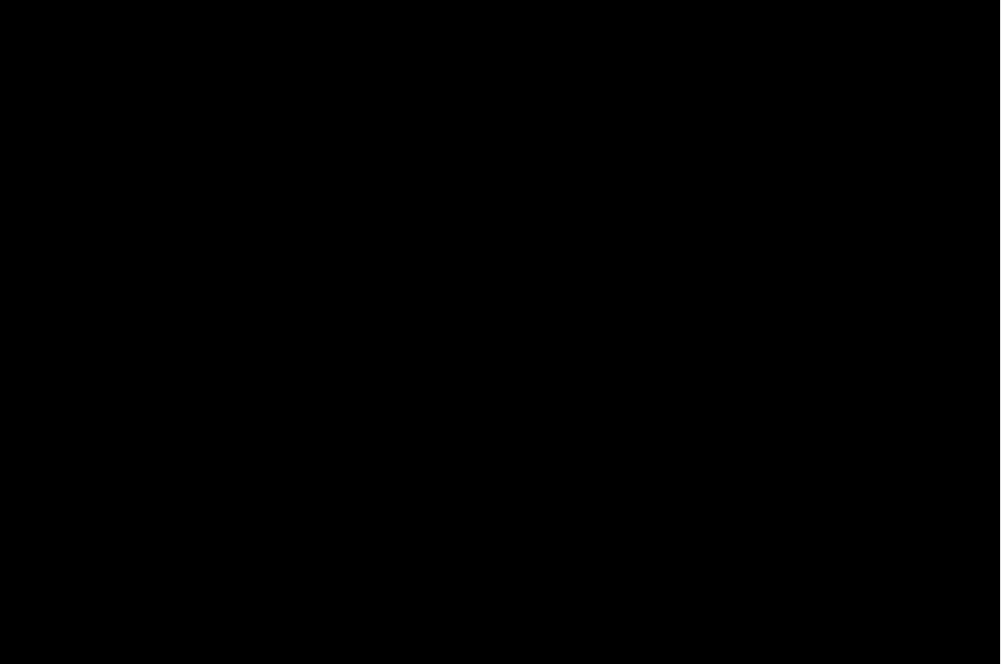}&
                \includegraphics[width=0.8in]{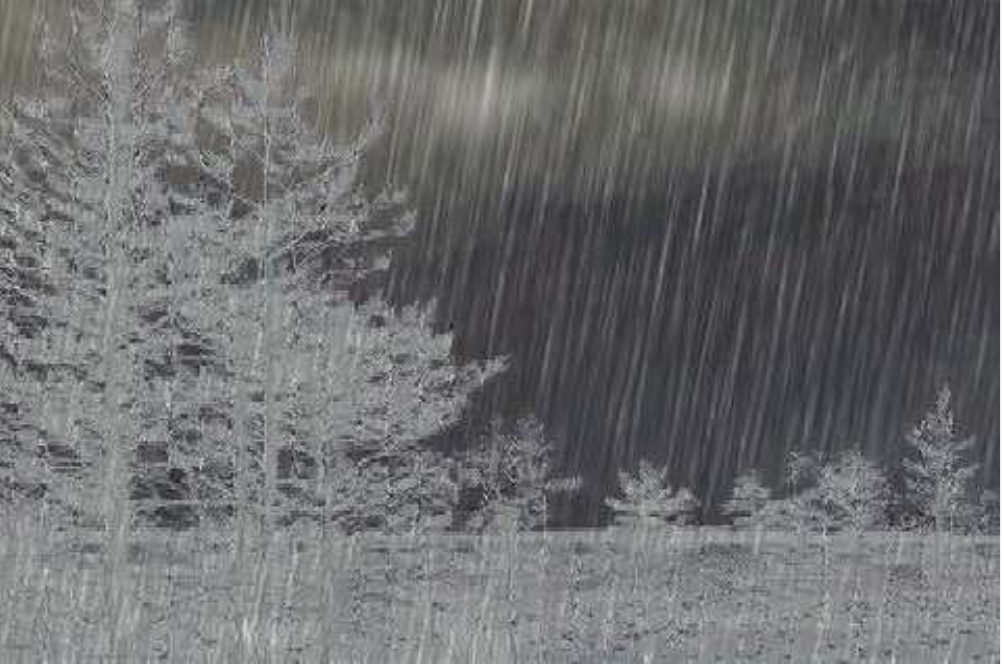}&
                \includegraphics[width=0.8in]{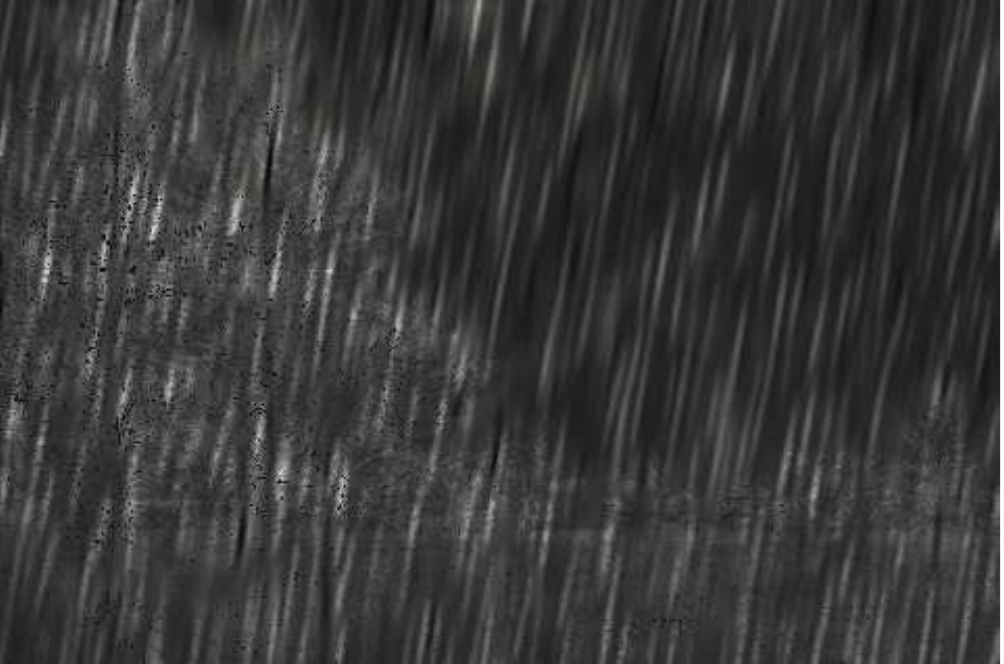}&
                \includegraphics[width=0.8in]{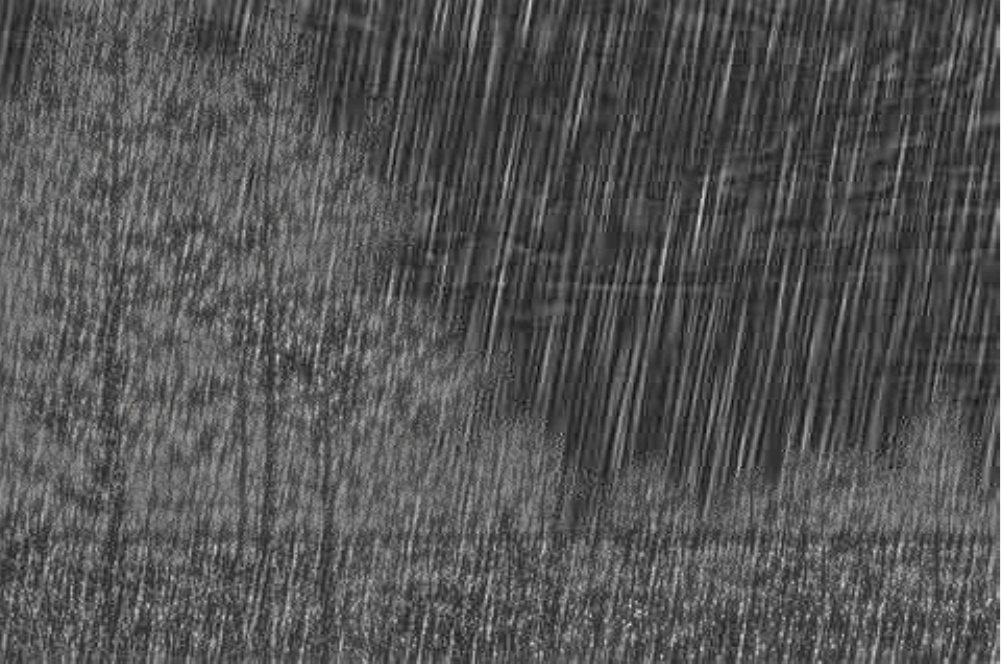}&
                \includegraphics[width=0.8in]{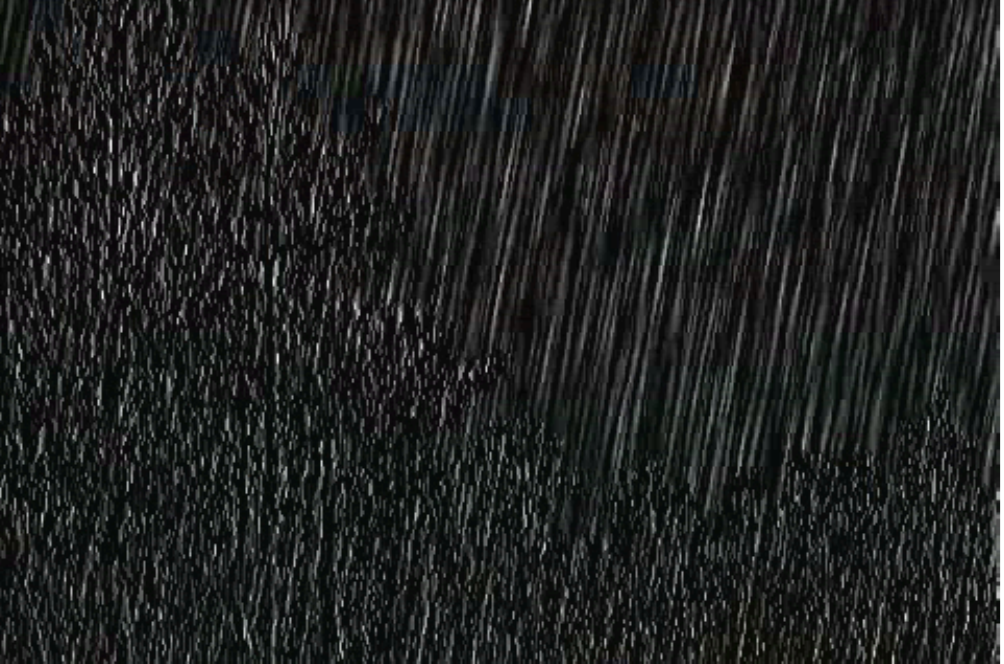}&
                \includegraphics[width=0.8in]{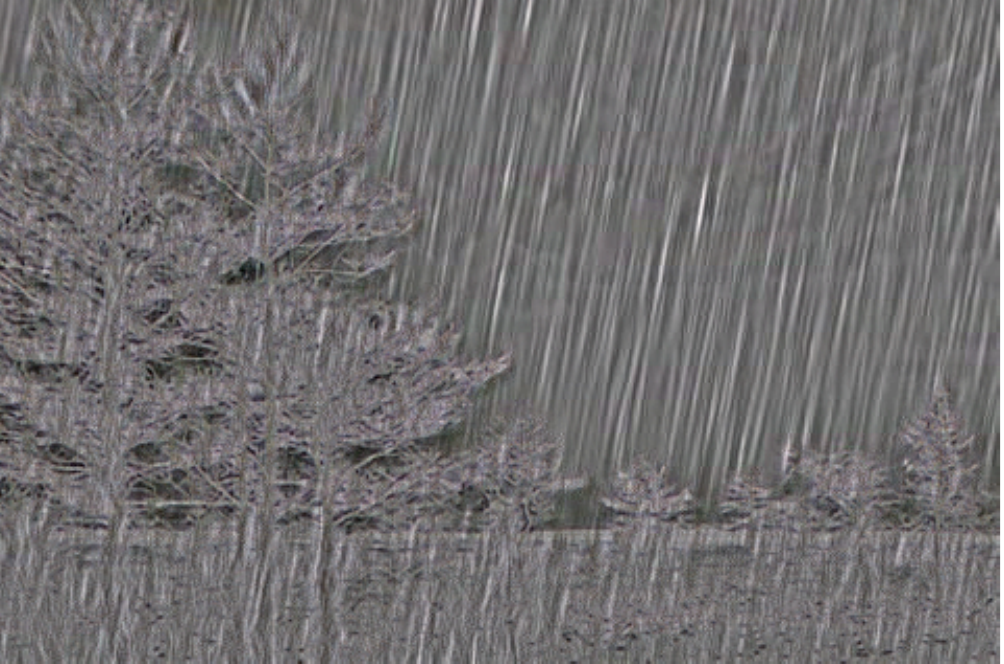}&
                \includegraphics[width=0.8in]{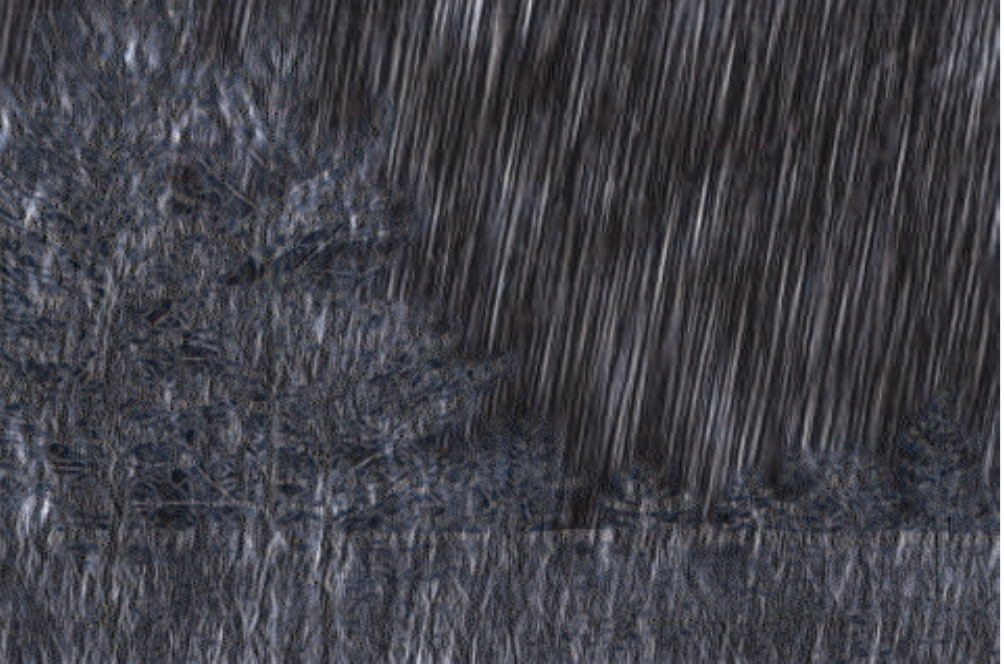}&
                \includegraphics[width=0.8in]{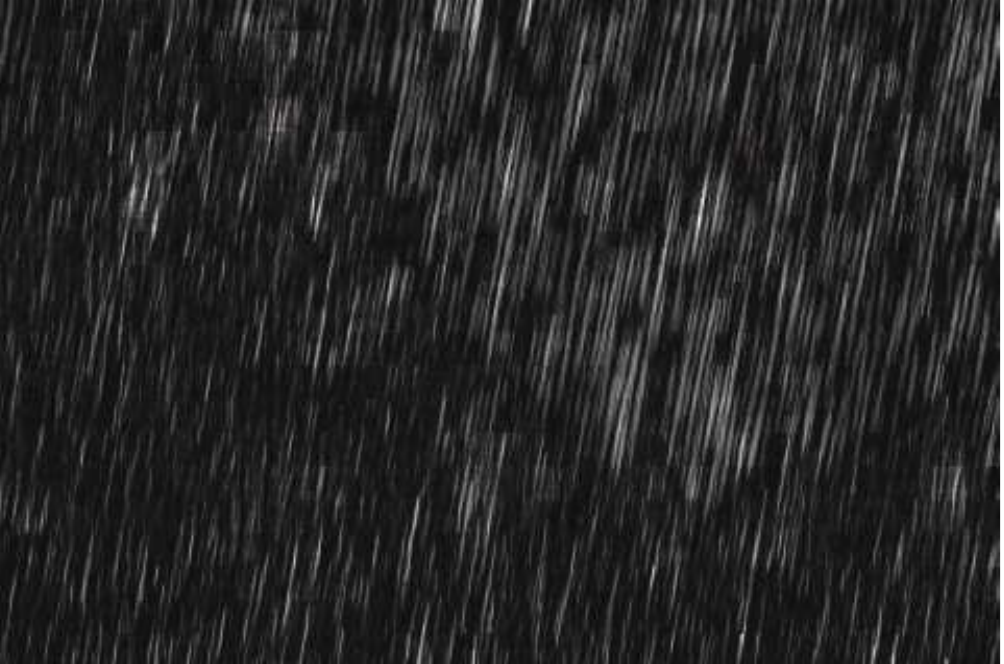}\\
                \vspace{0.5mm}

                \includegraphics[width=0.8in]{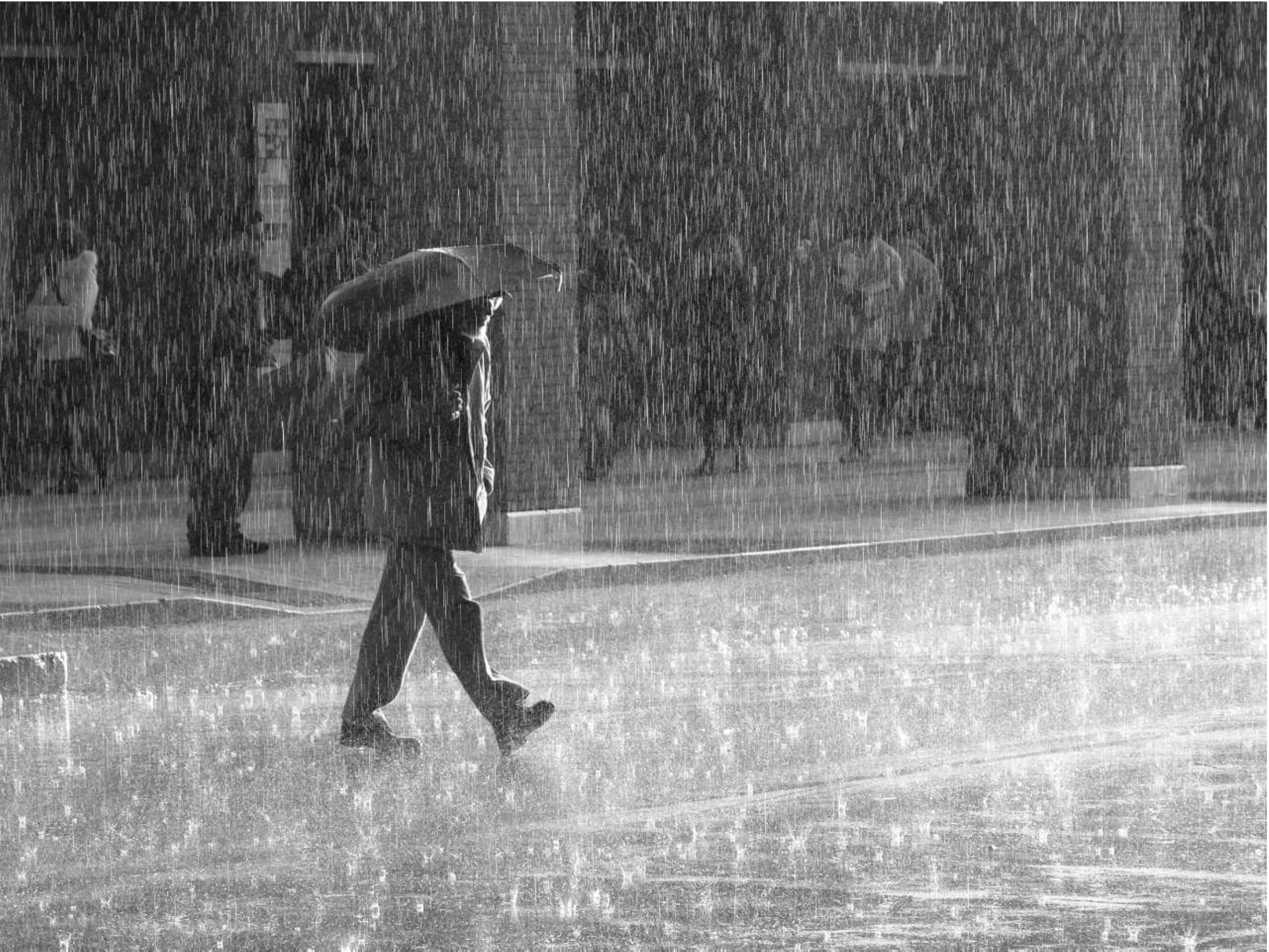}&
                \includegraphics[width=0.8in]{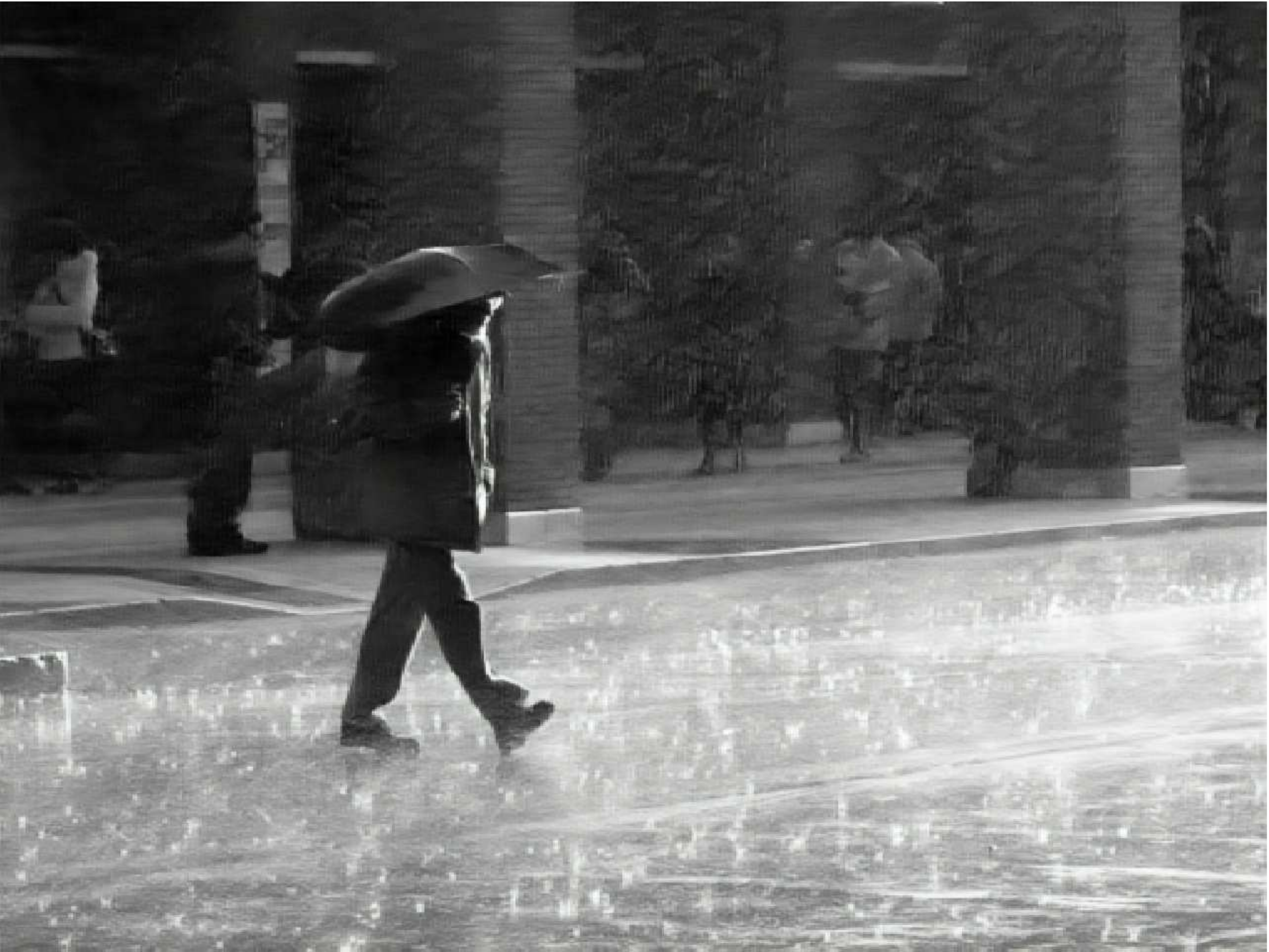}&
                \includegraphics[width=0.8in]{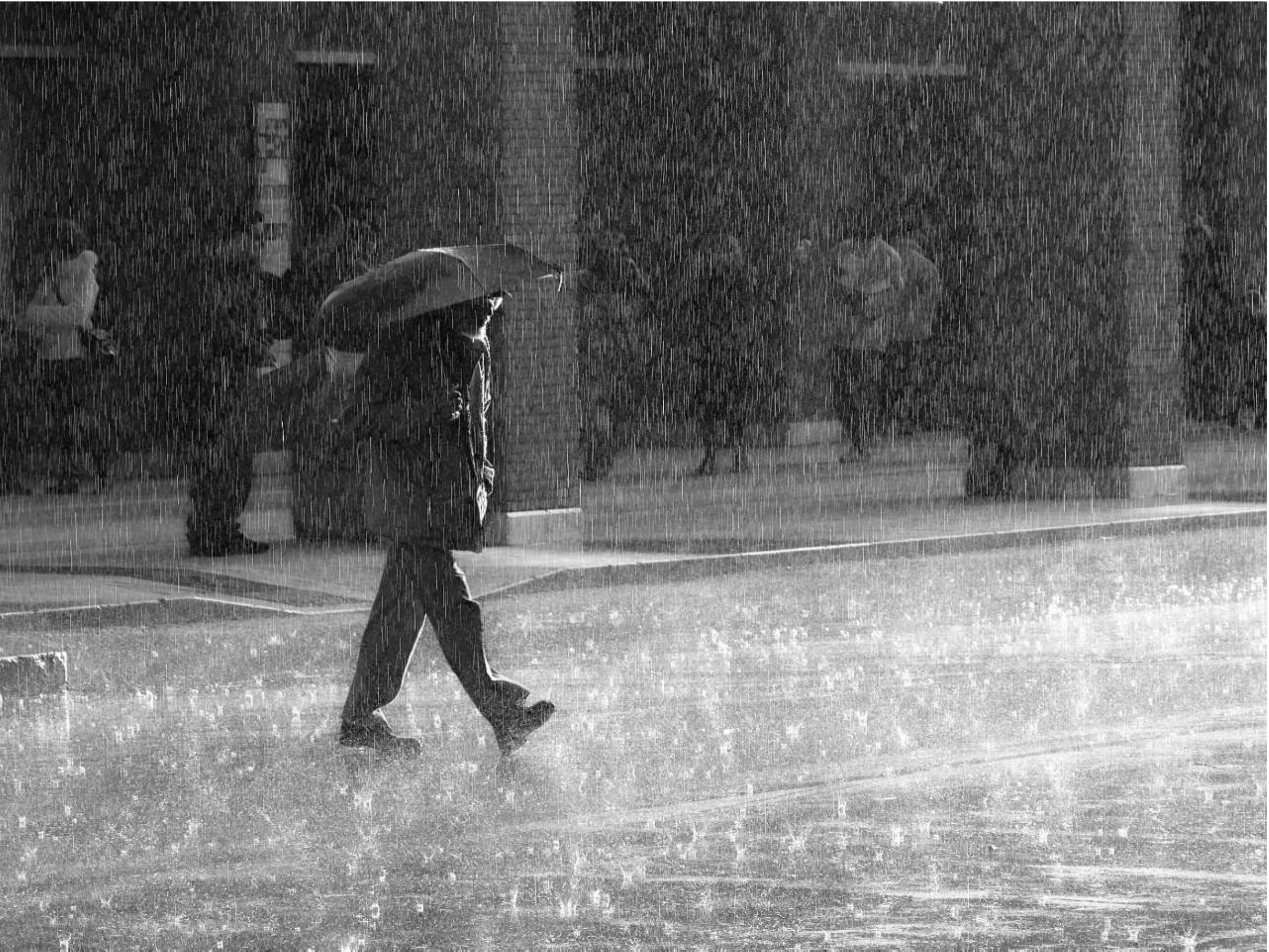}&
                \includegraphics[width=0.8in]{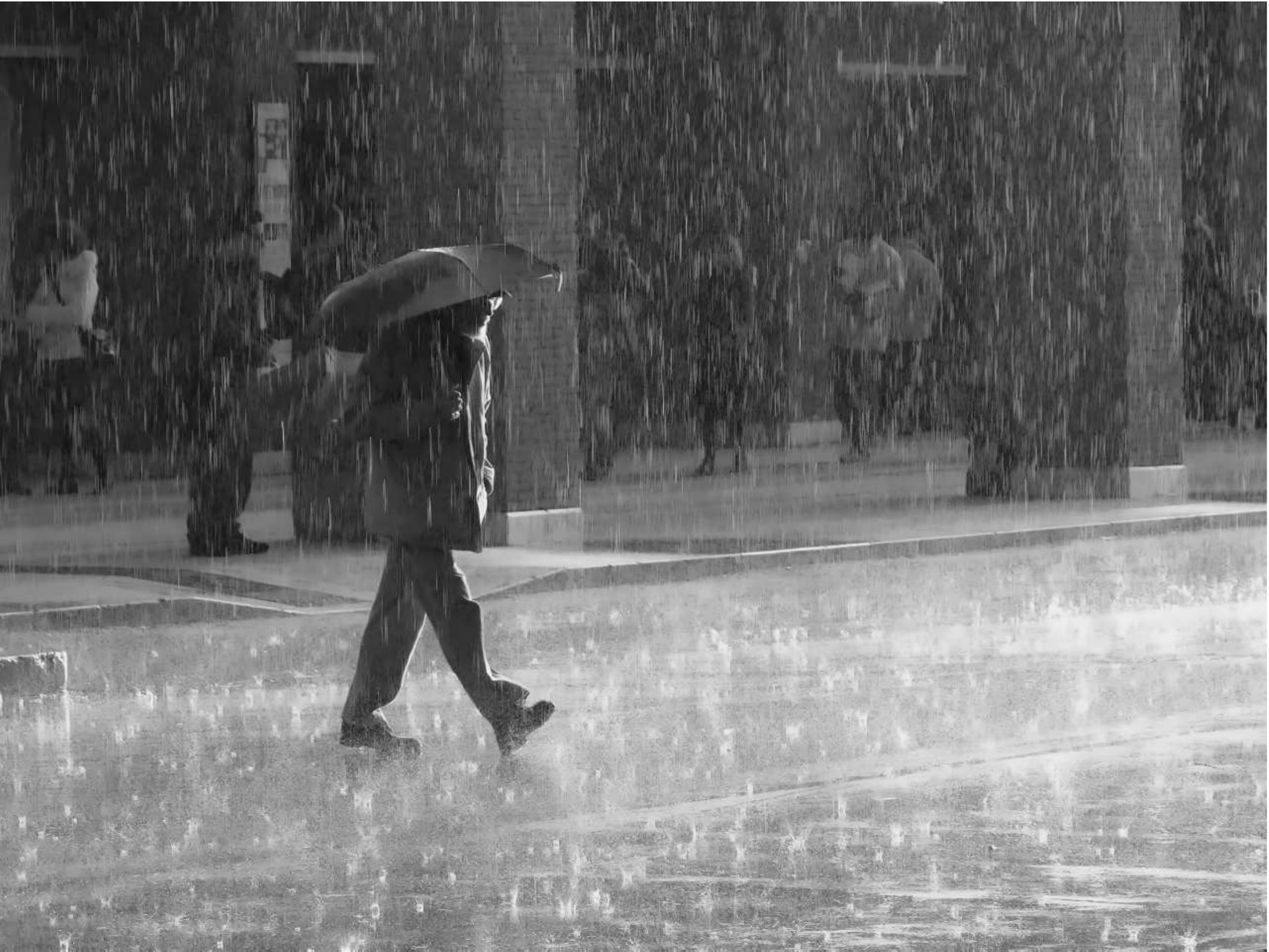}&
                \includegraphics[width=0.8in]{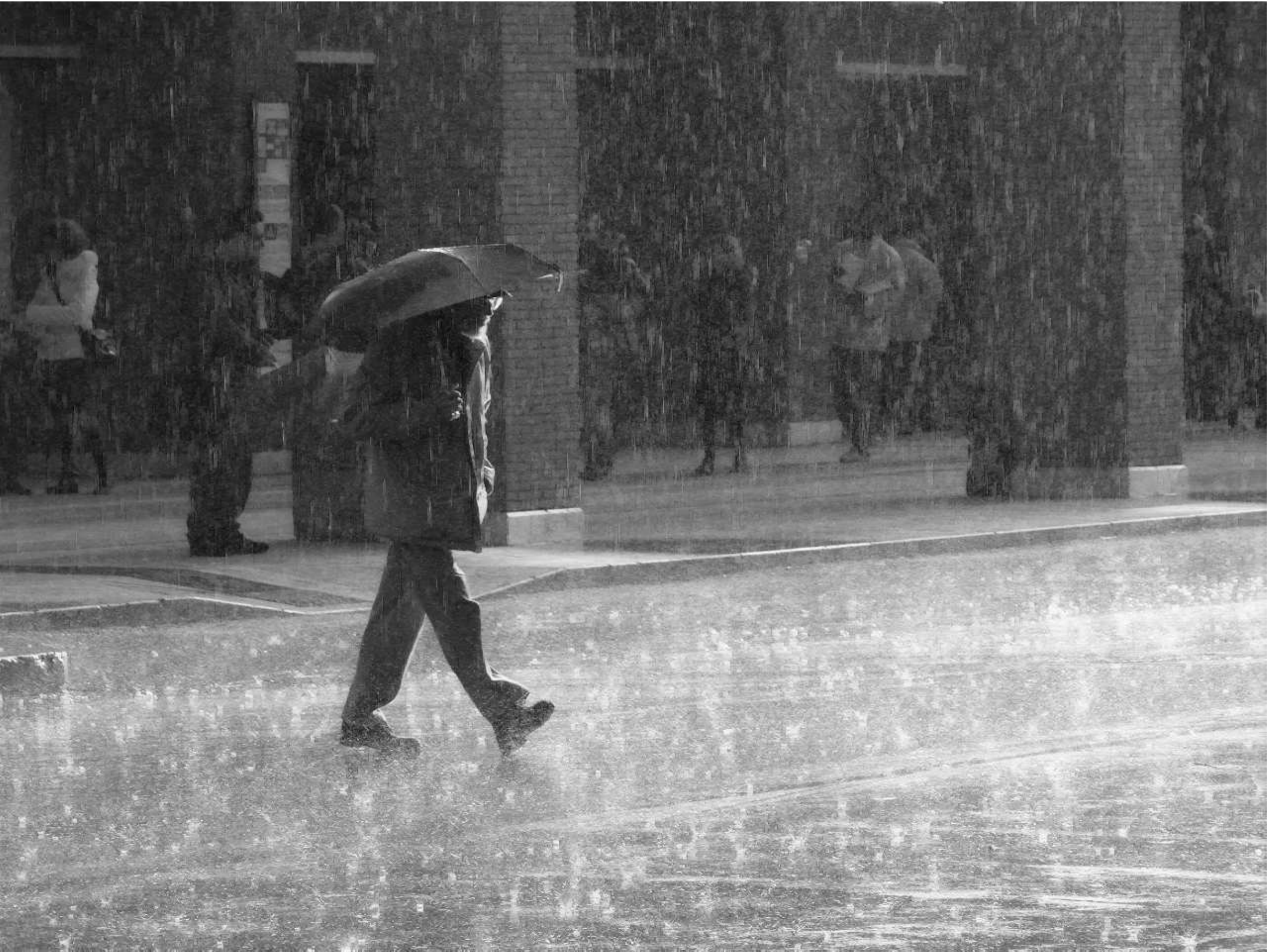}&
                \includegraphics[width=0.8in]{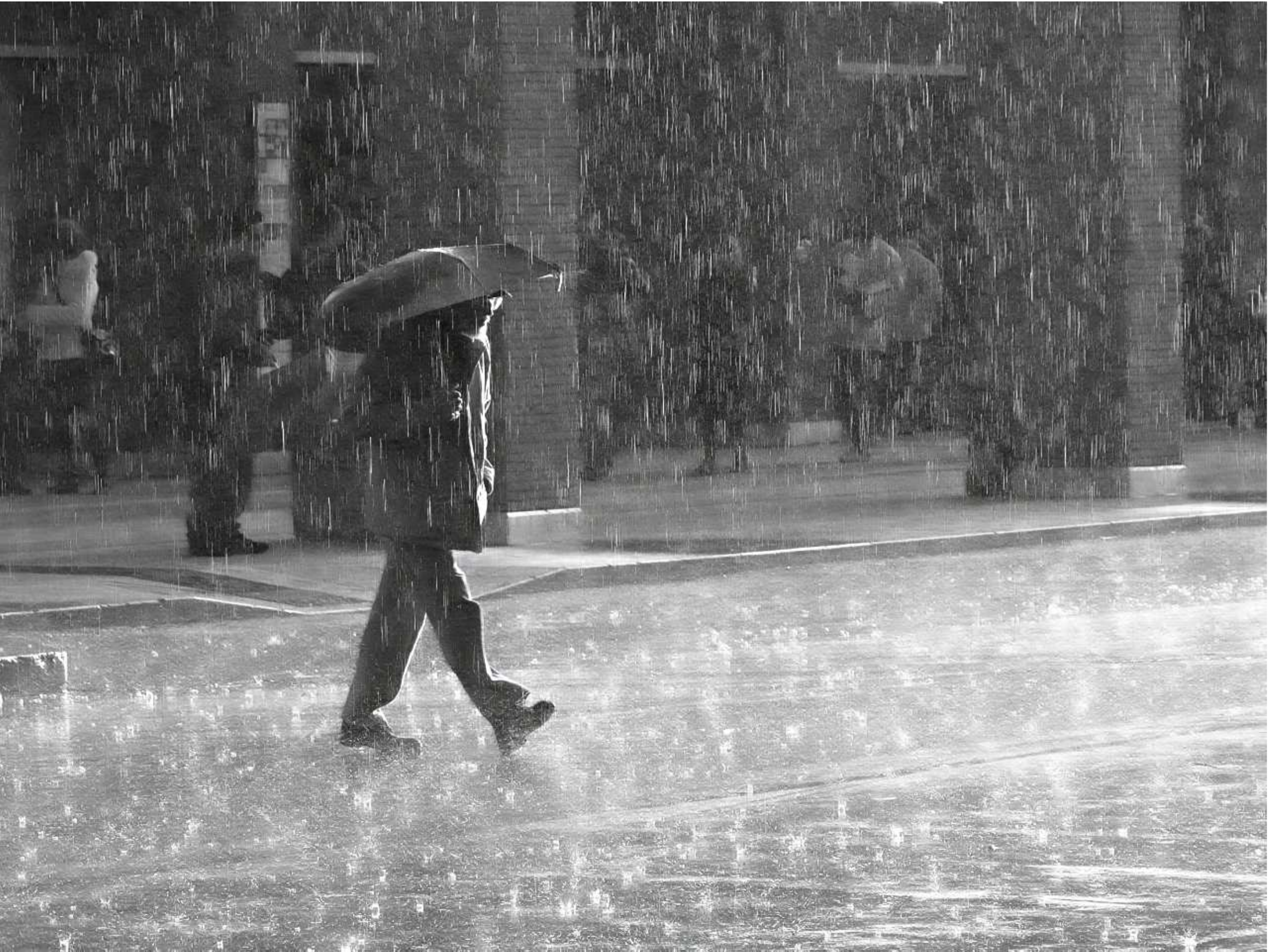}&
                \includegraphics[width=0.8in]{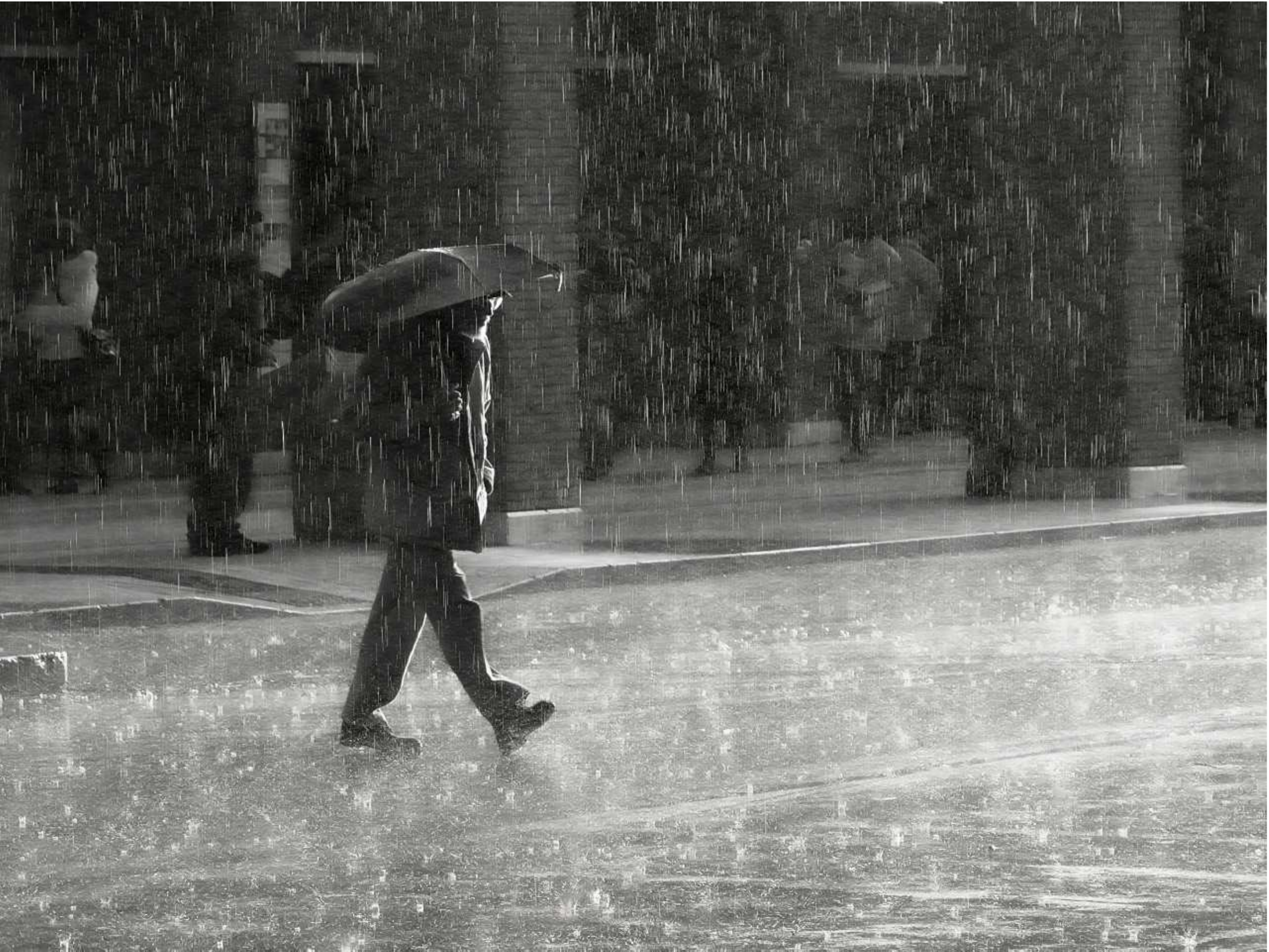}&
                \includegraphics[width=0.8in]{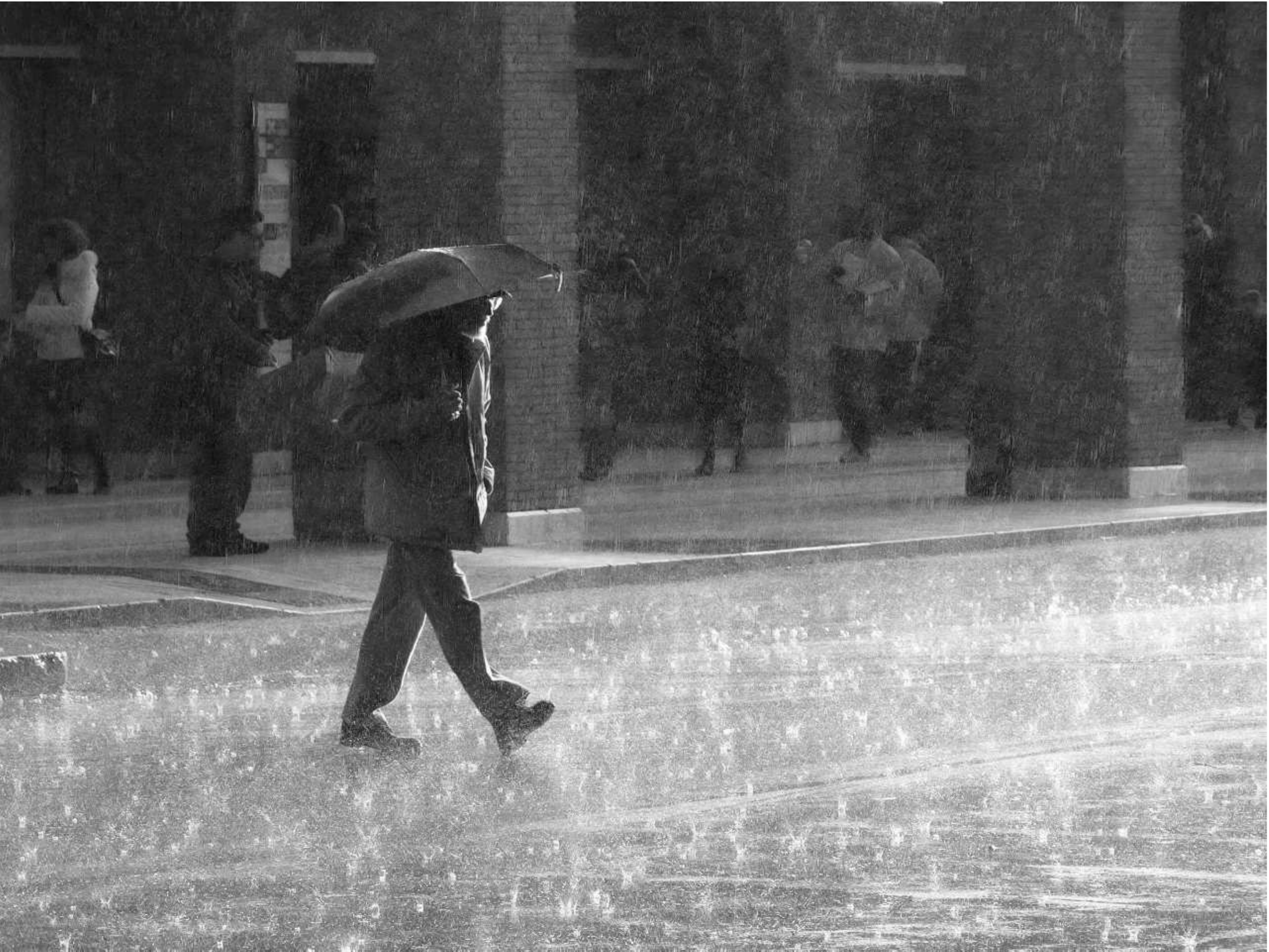}\\
                \vspace{0.5mm}

                \includegraphics[width=0.8in]{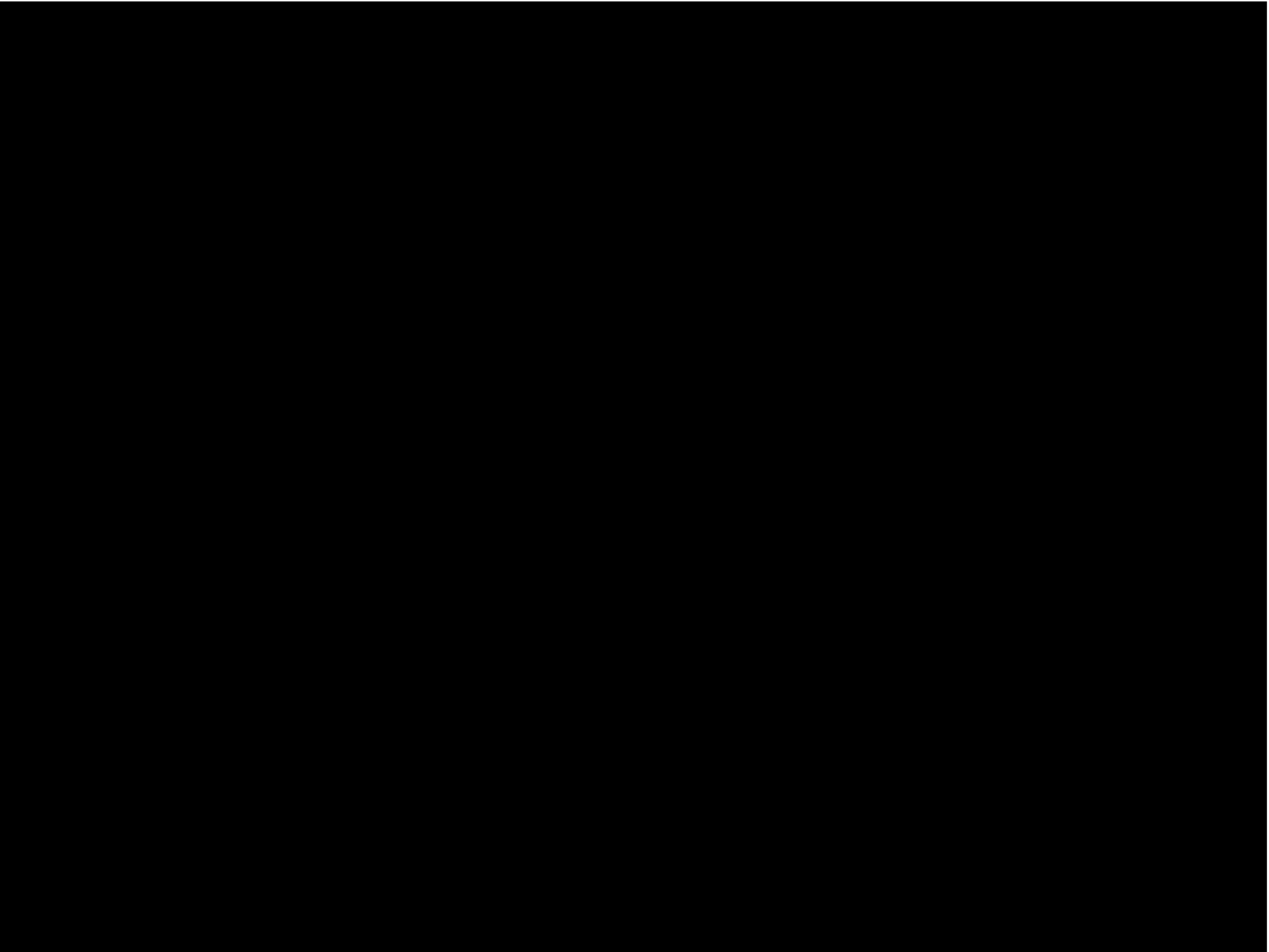}&
                \includegraphics[width=0.8in]{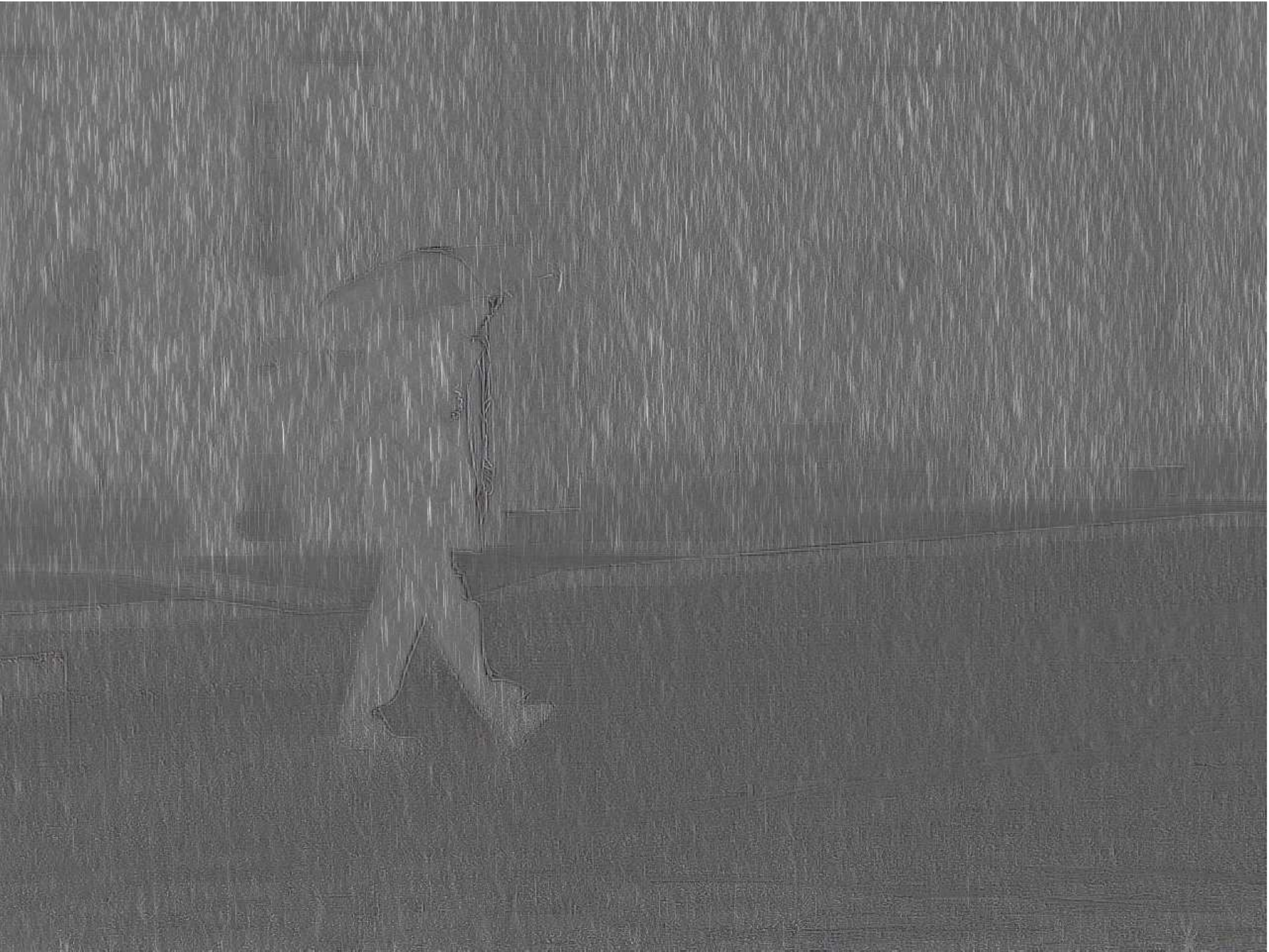}&
                \includegraphics[width=0.8in]{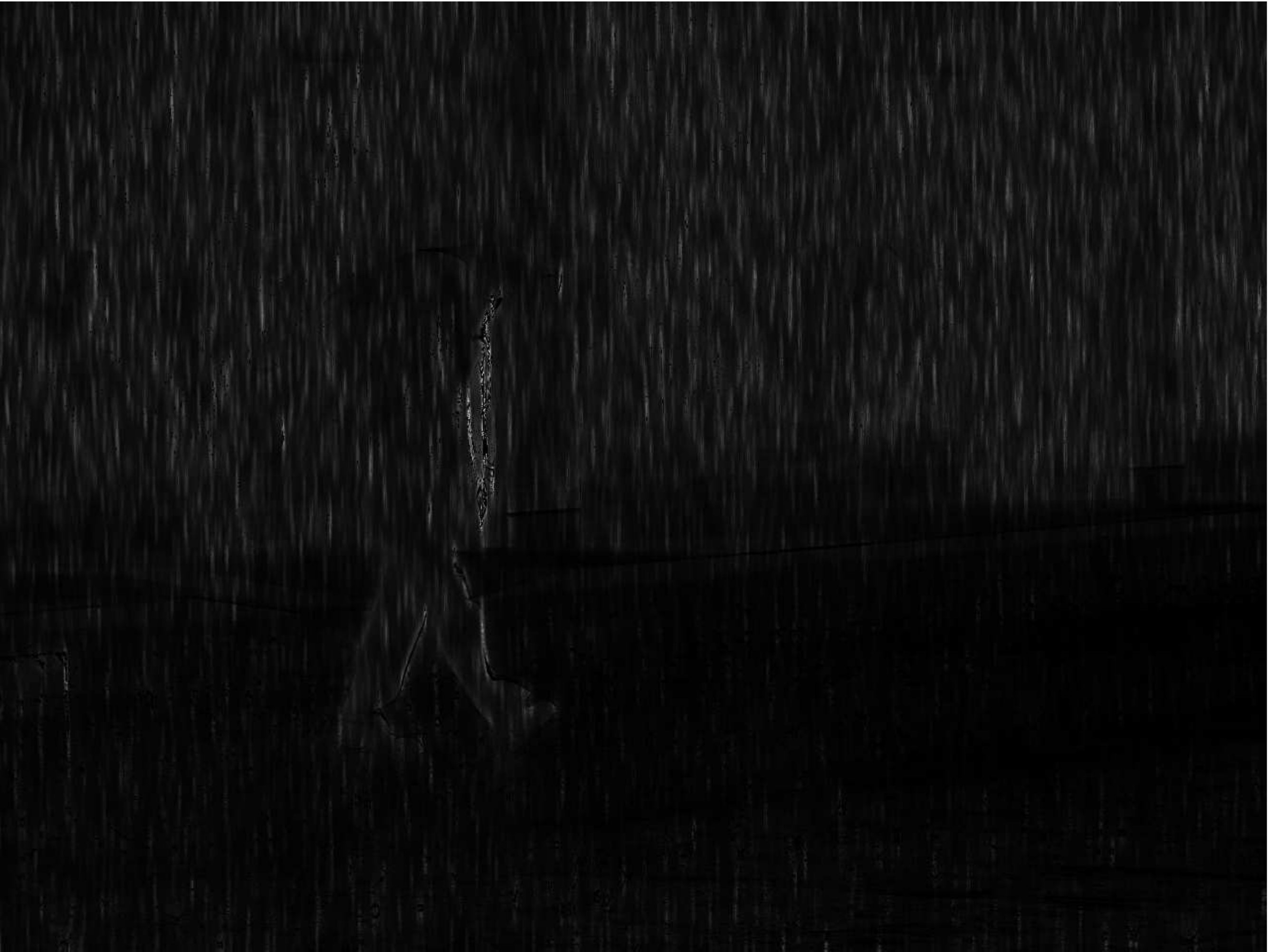}&
                \includegraphics[width=0.8in]{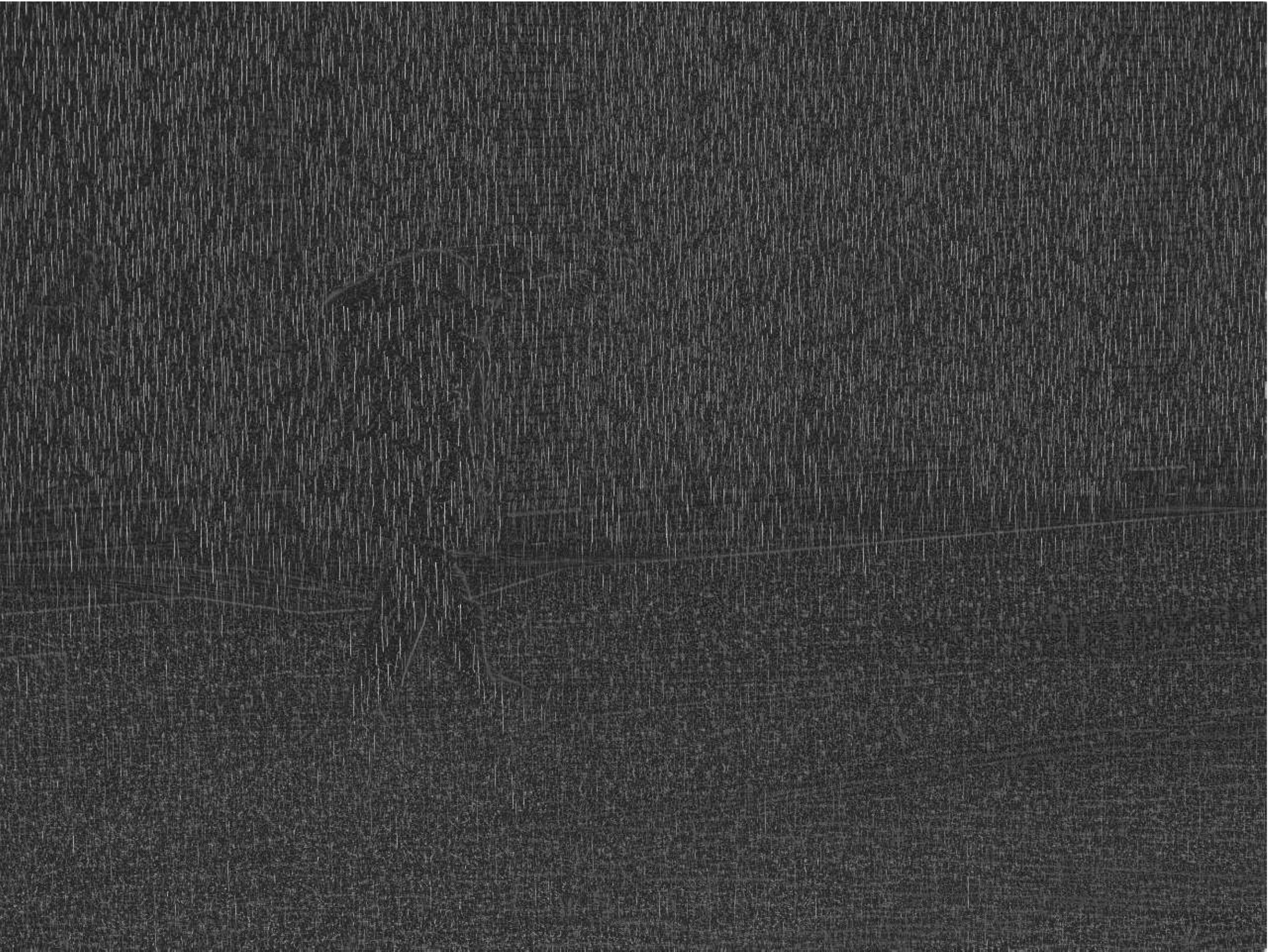}&
                \includegraphics[width=0.8in]{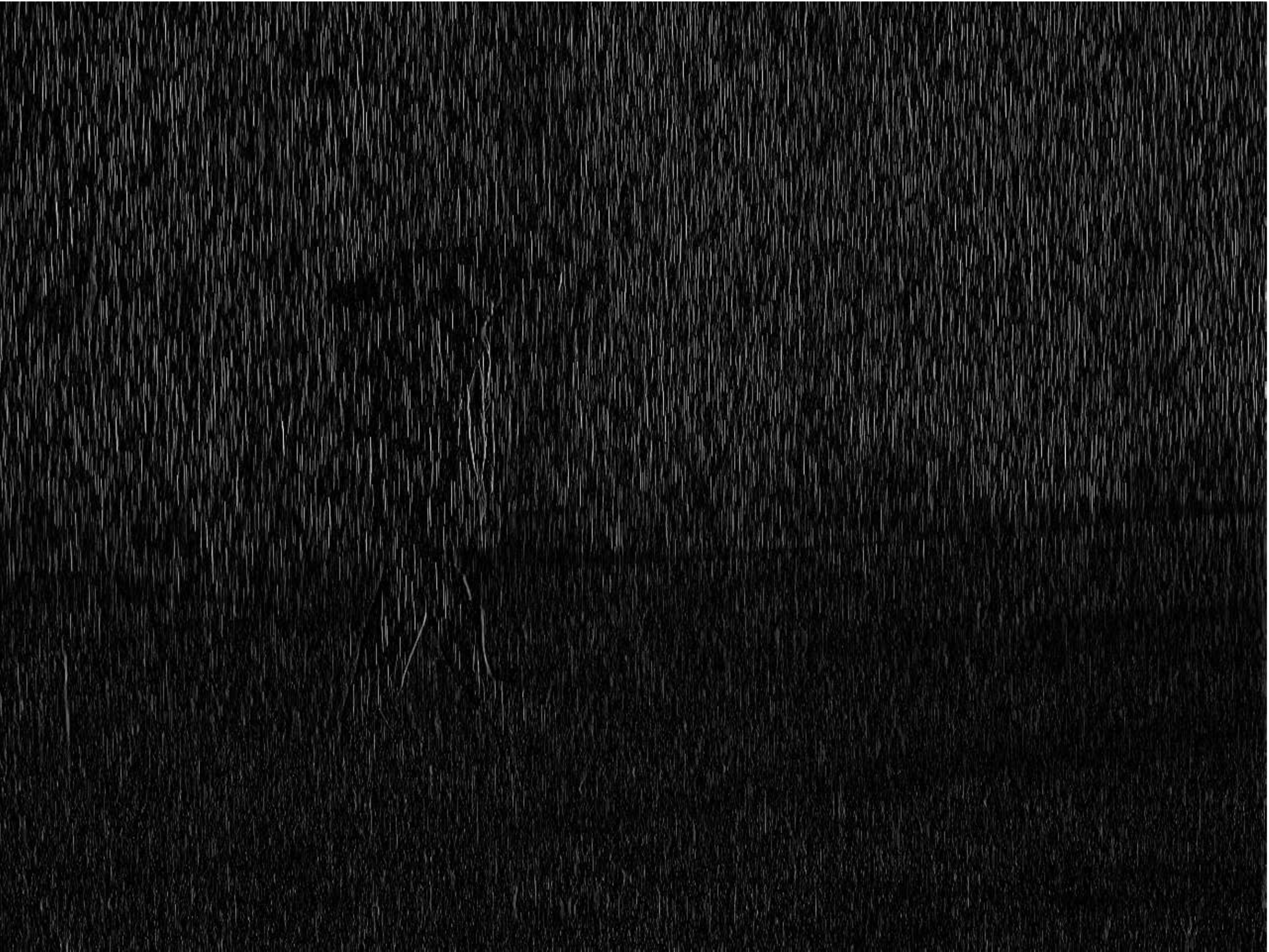}&
                \includegraphics[width=0.8in]{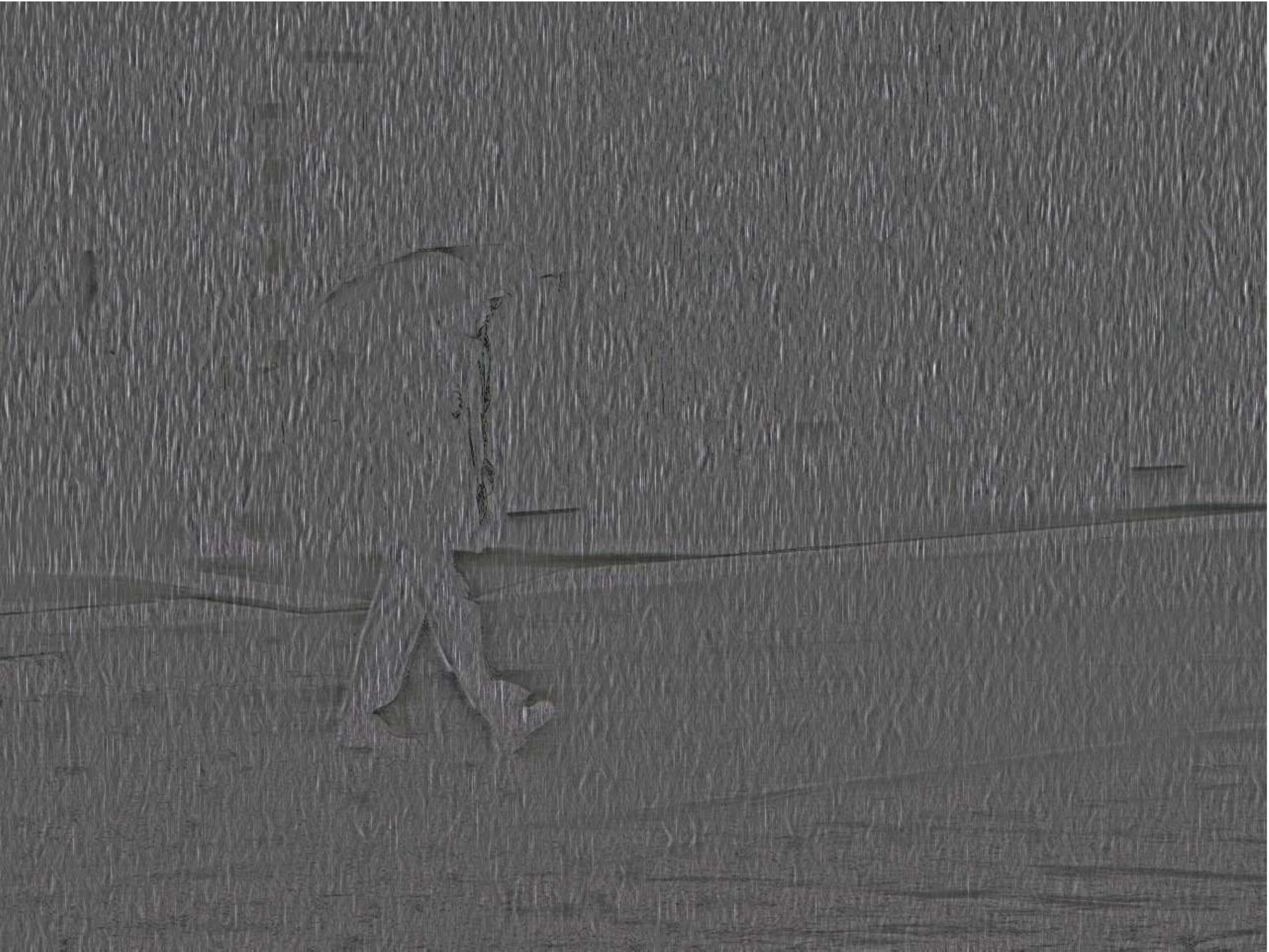}&
                \includegraphics[width=0.8in]{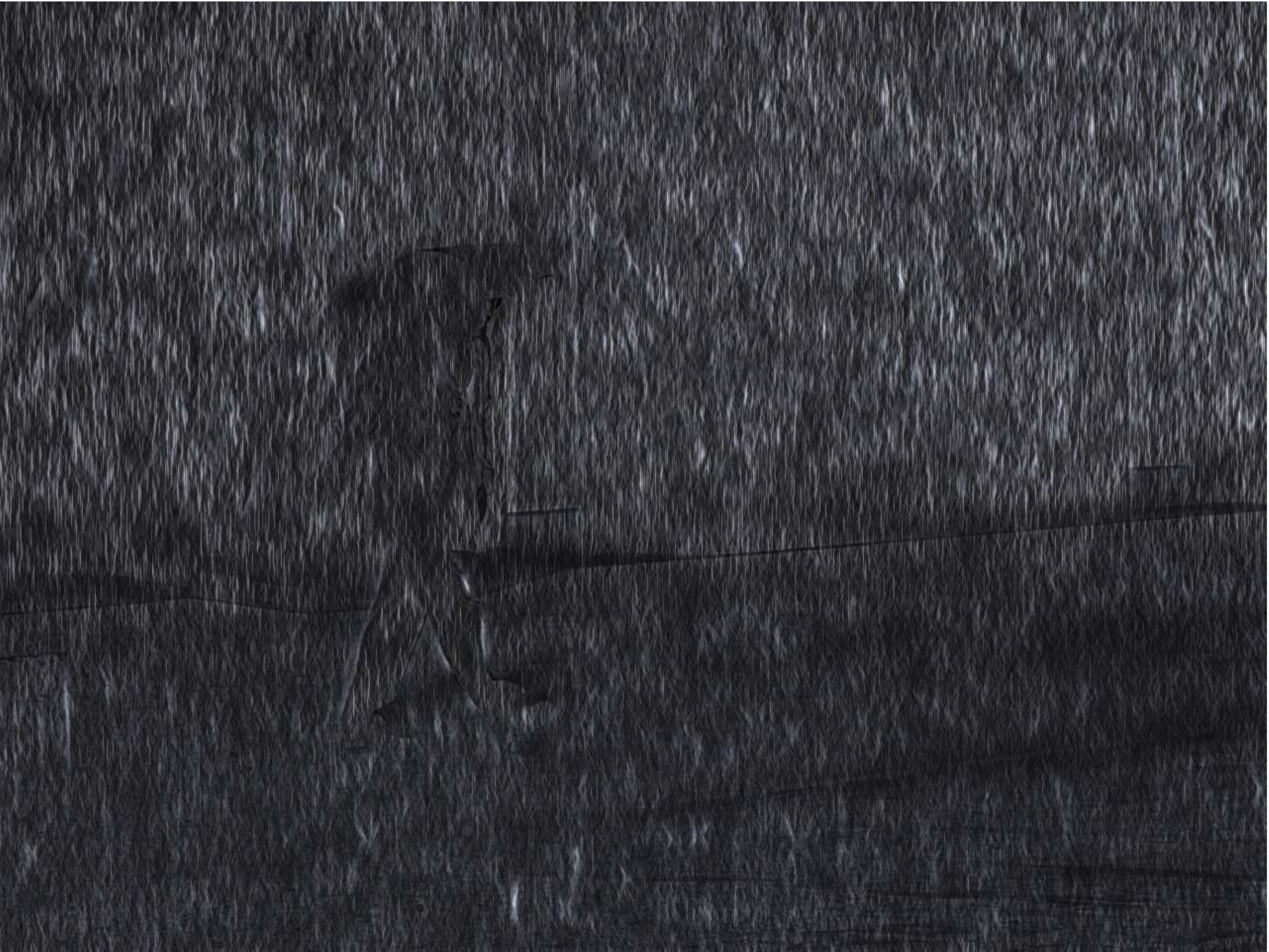}&
                \includegraphics[width=0.8in]{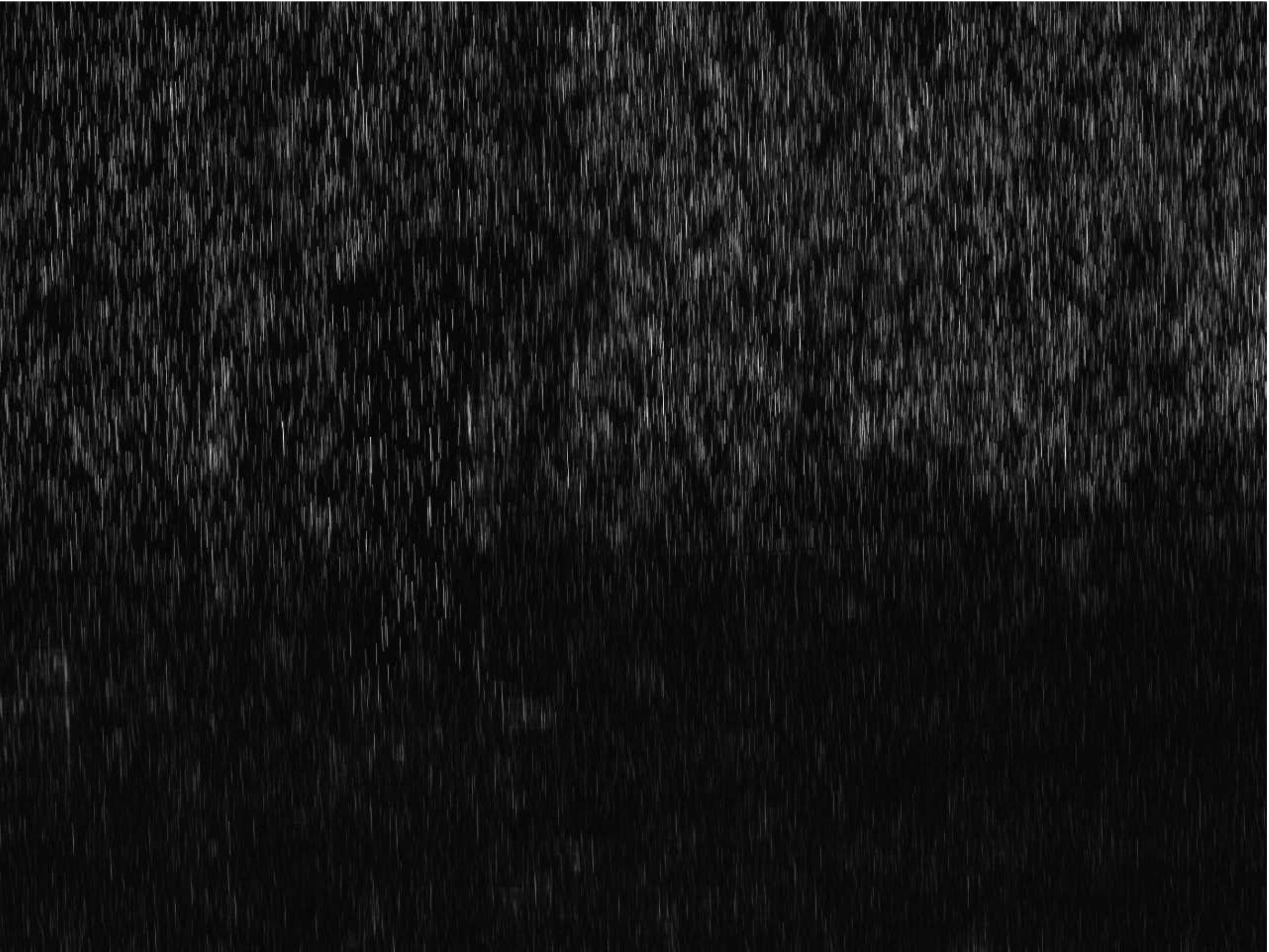}\\
                \vspace{0.5mm}

                \includegraphics[width=0.8in]{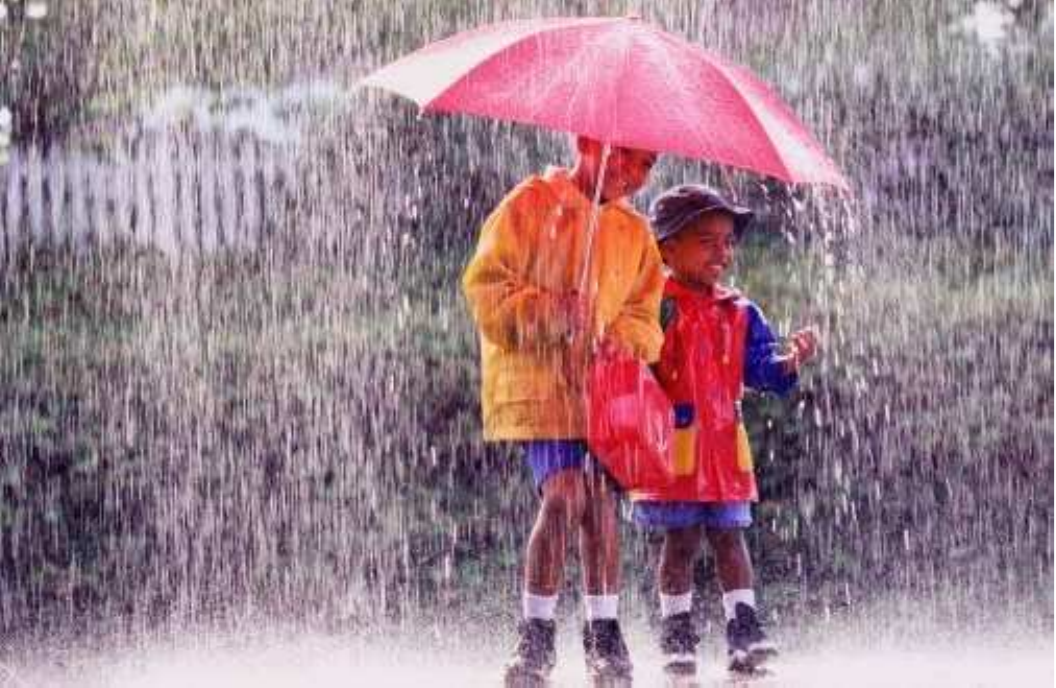}&
                \includegraphics[width=0.8in]{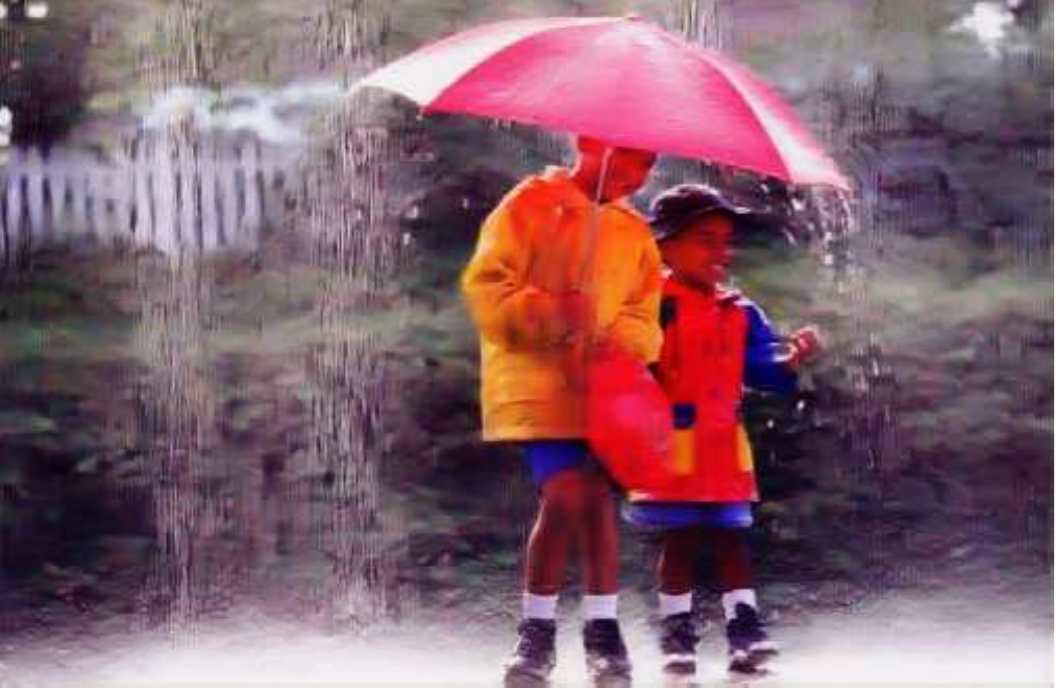}&
                \includegraphics[width=0.8in]{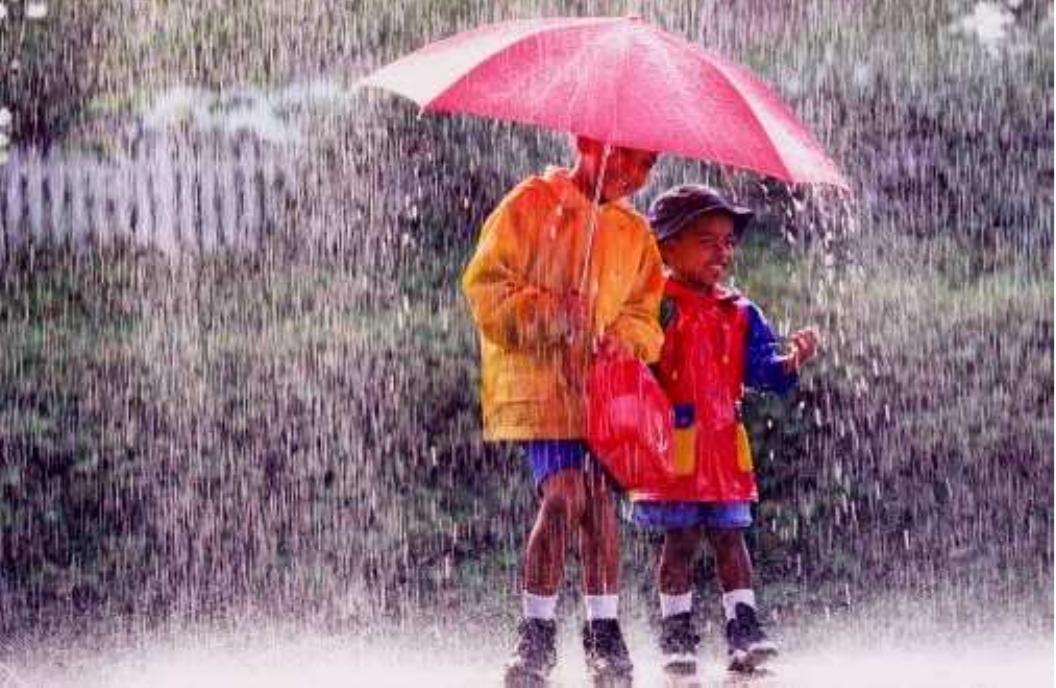}&
                \includegraphics[width=0.8in]{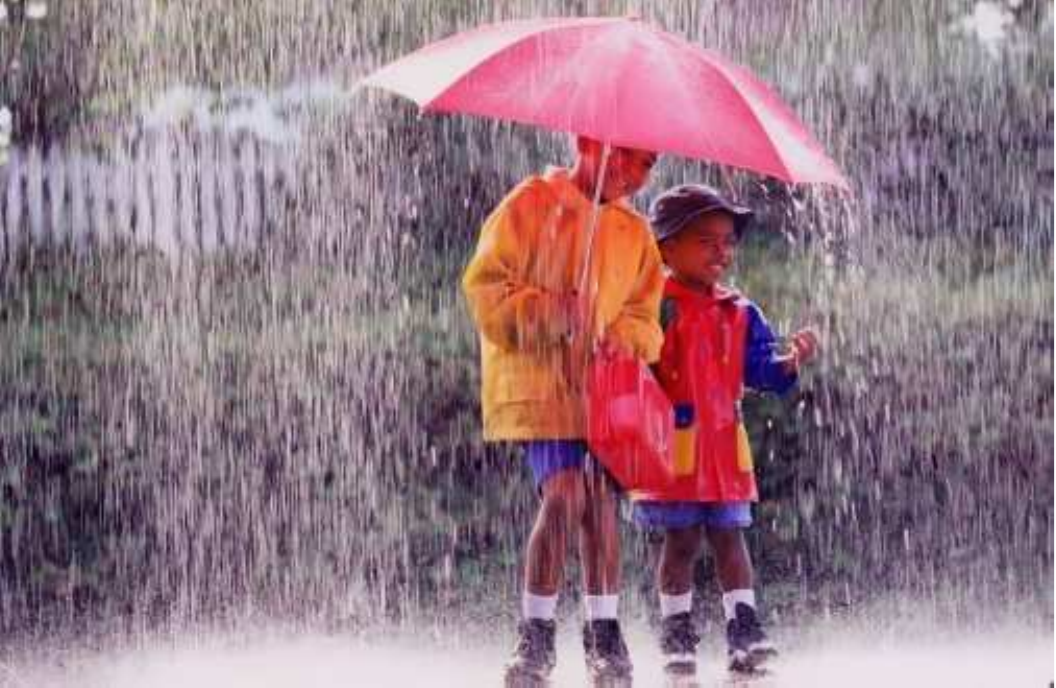}&
                \includegraphics[width=0.8in]{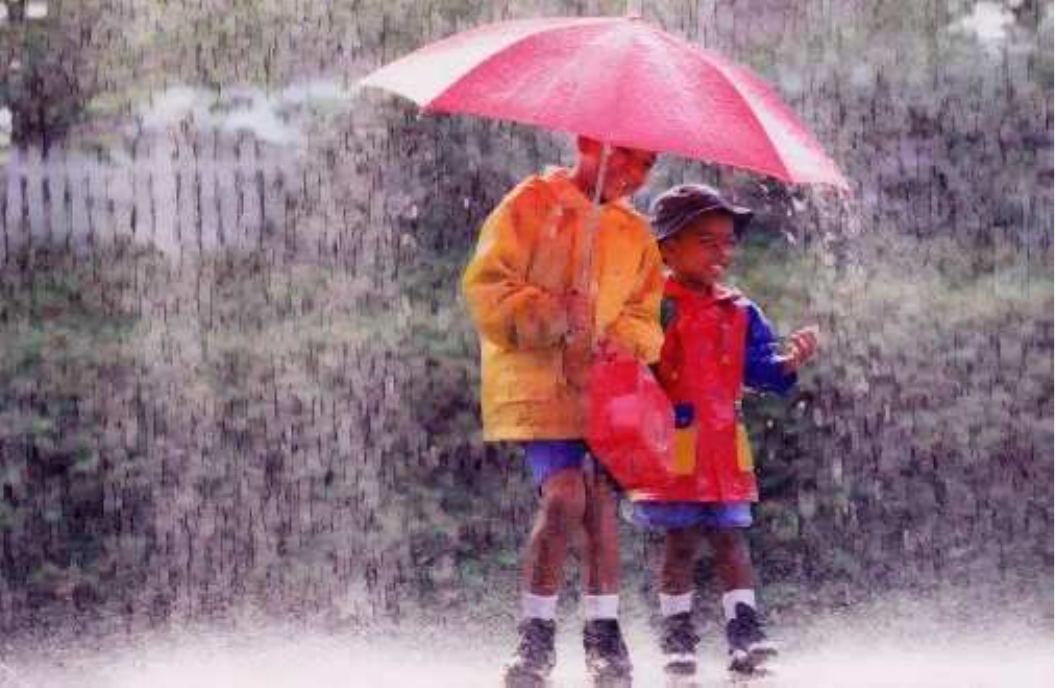}&
                \includegraphics[width=0.8in]{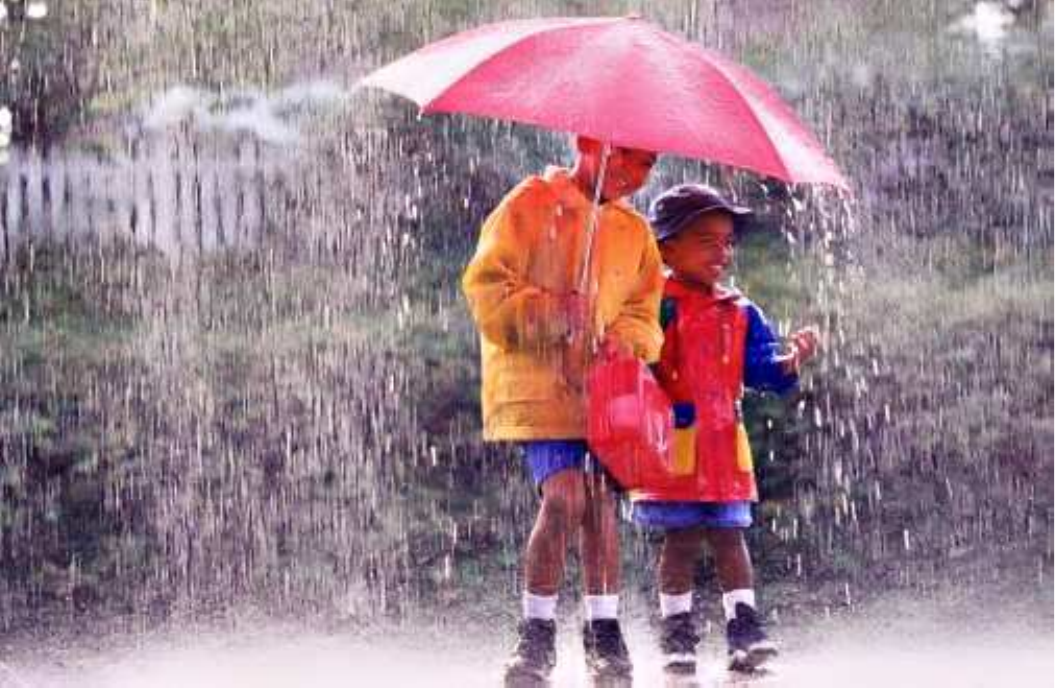}&
                \includegraphics[width=0.8in]{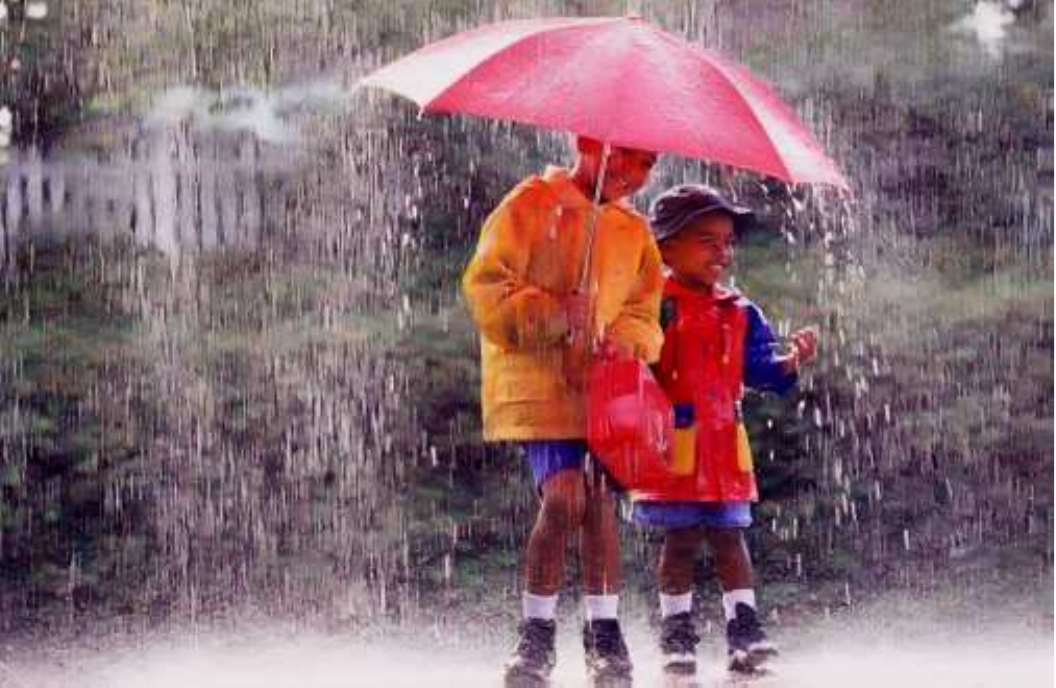}&
                \includegraphics[width=0.8in]{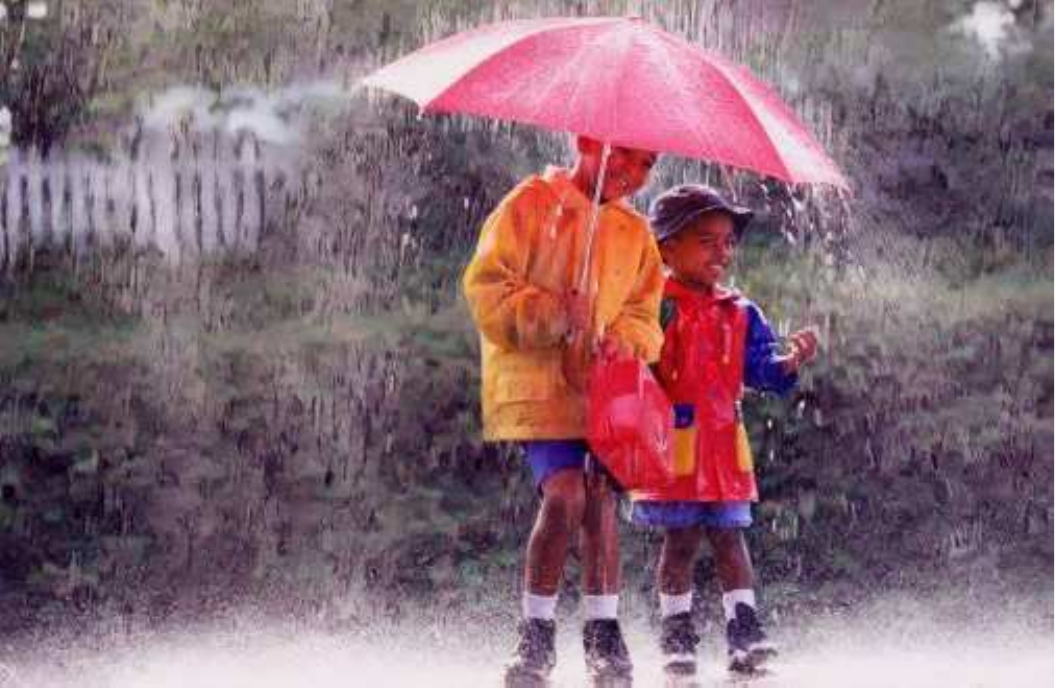}\\
                \vspace{0.5mm}

                \includegraphics[width=0.8in]{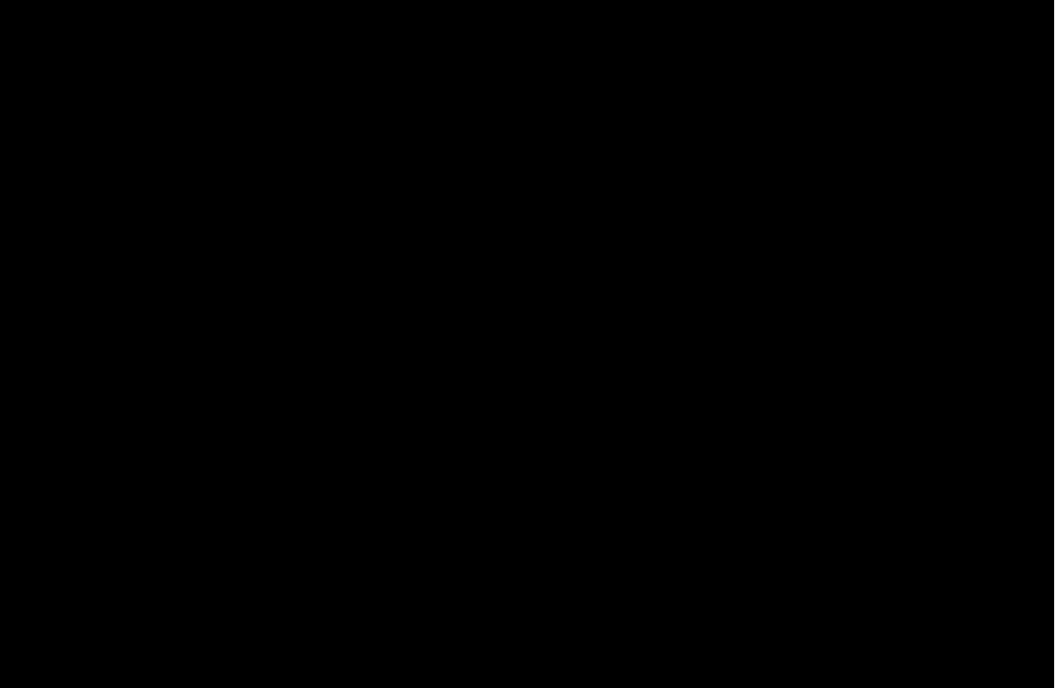}&
                \includegraphics[width=0.8in]{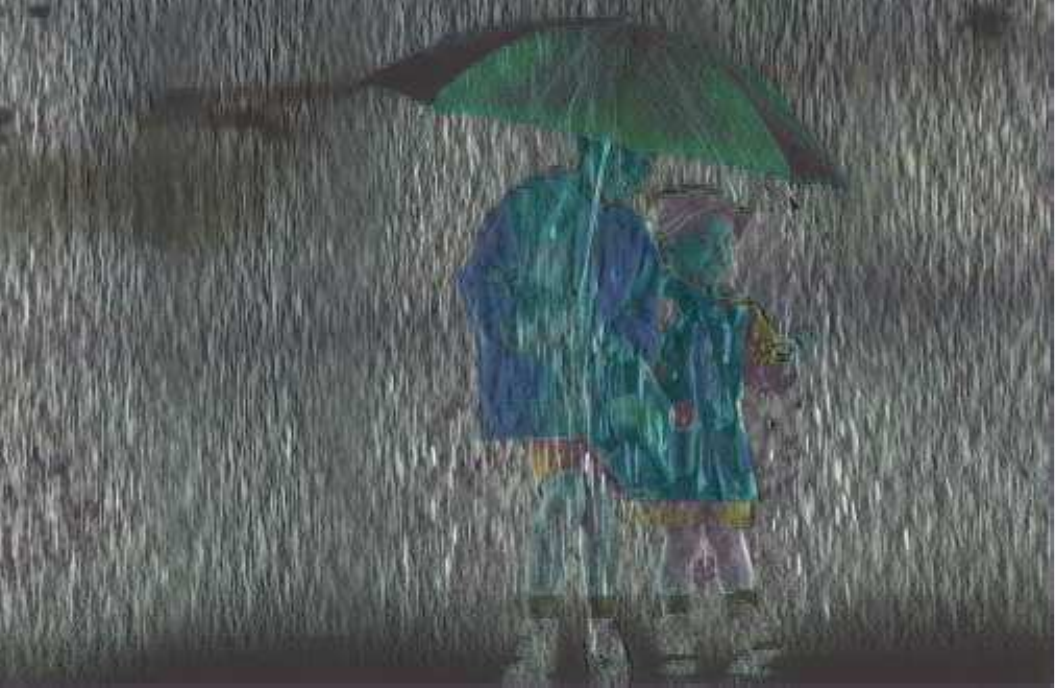}&
                \includegraphics[width=0.8in]{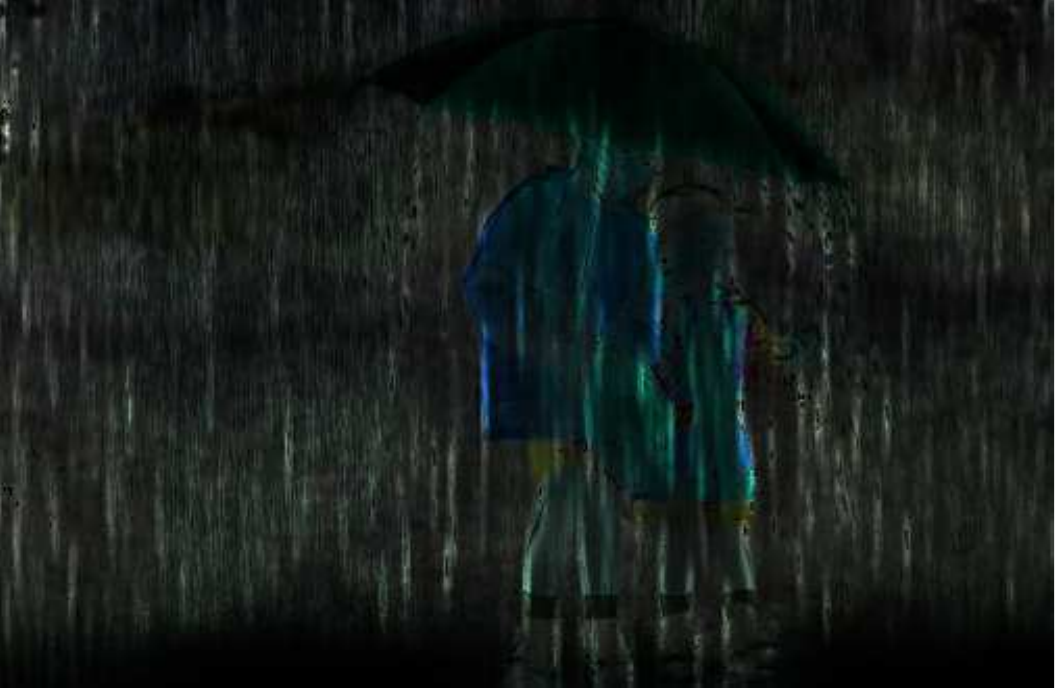}&
                \includegraphics[width=0.8in]{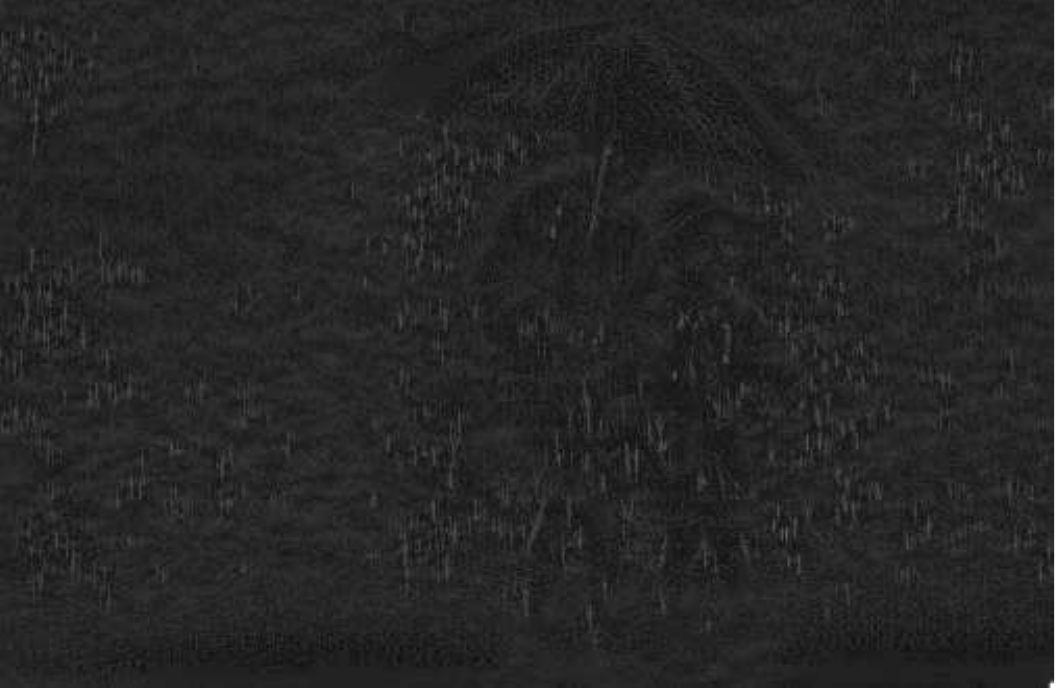}&
                \includegraphics[width=0.8in]{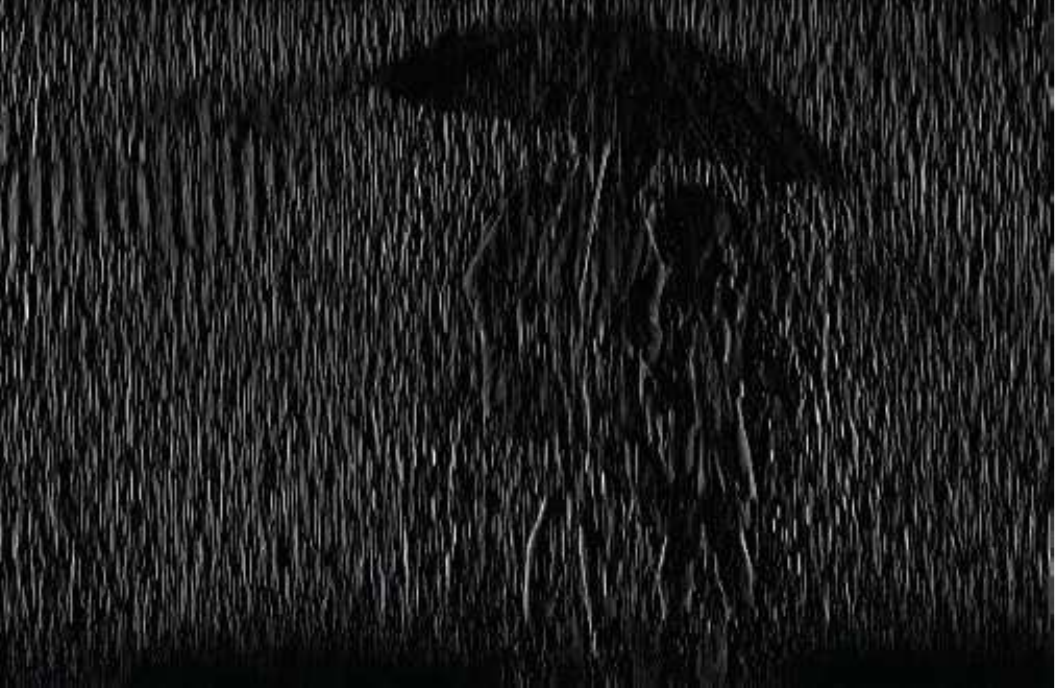}&
                \includegraphics[width=0.8in]{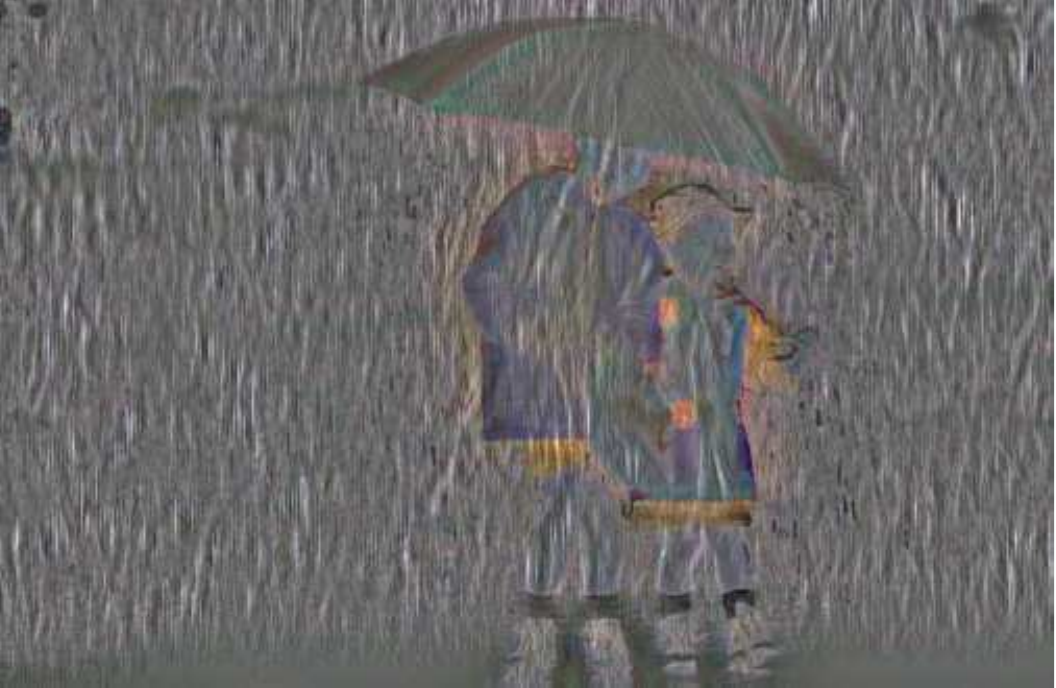}&
                \includegraphics[width=0.8in]{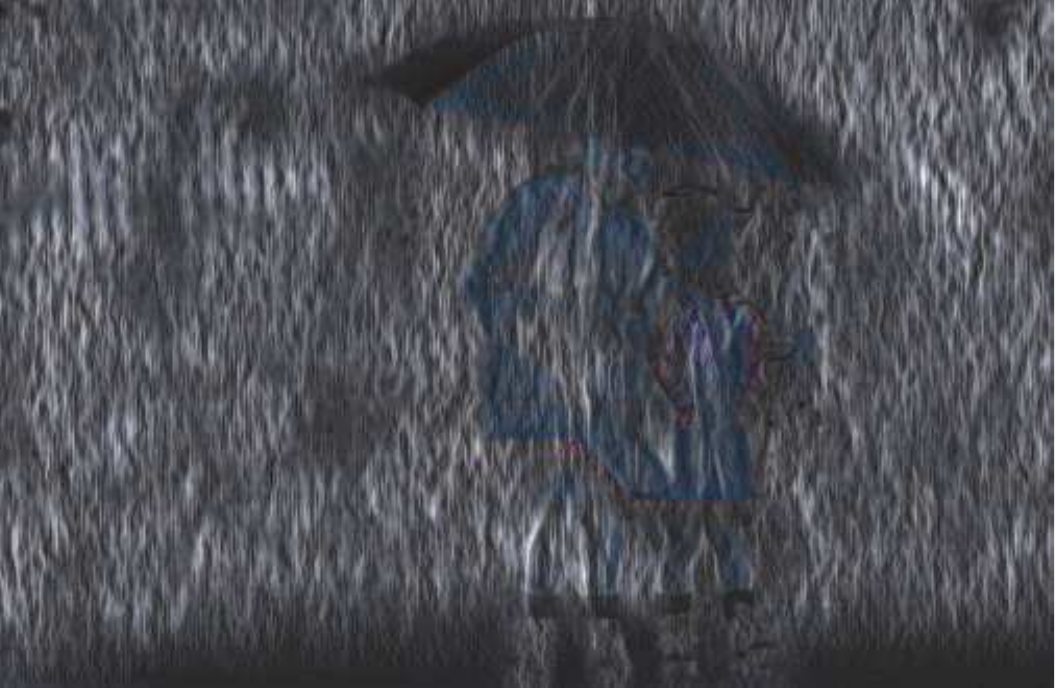}&
                \includegraphics[width=0.8in]{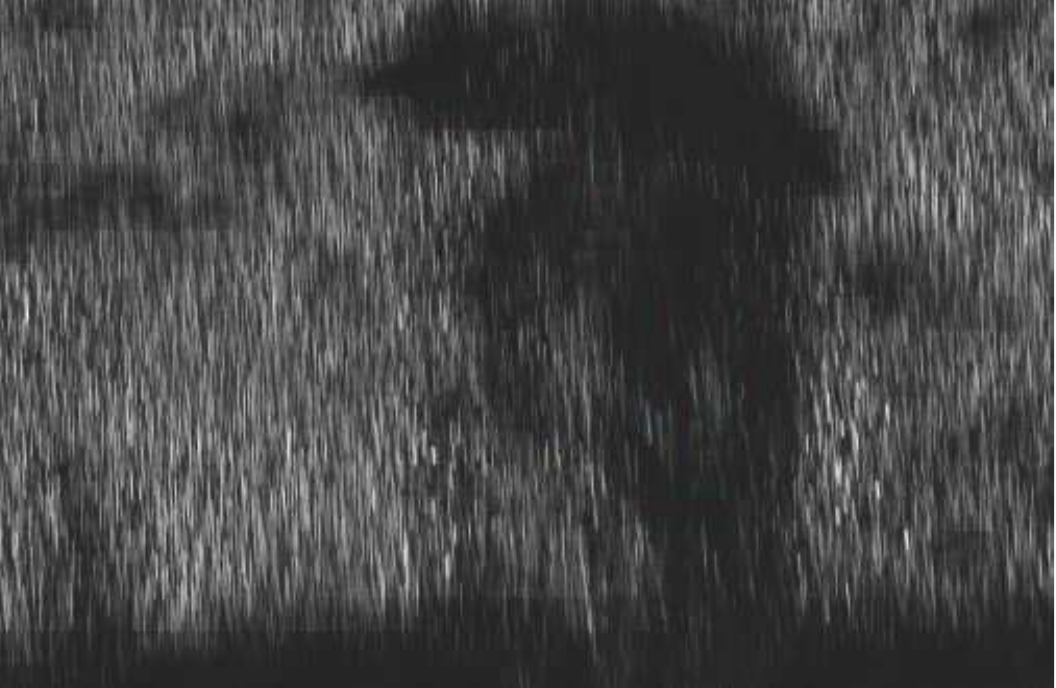}\\
                \vspace{0.5mm}

                \includegraphics[width=0.8in]{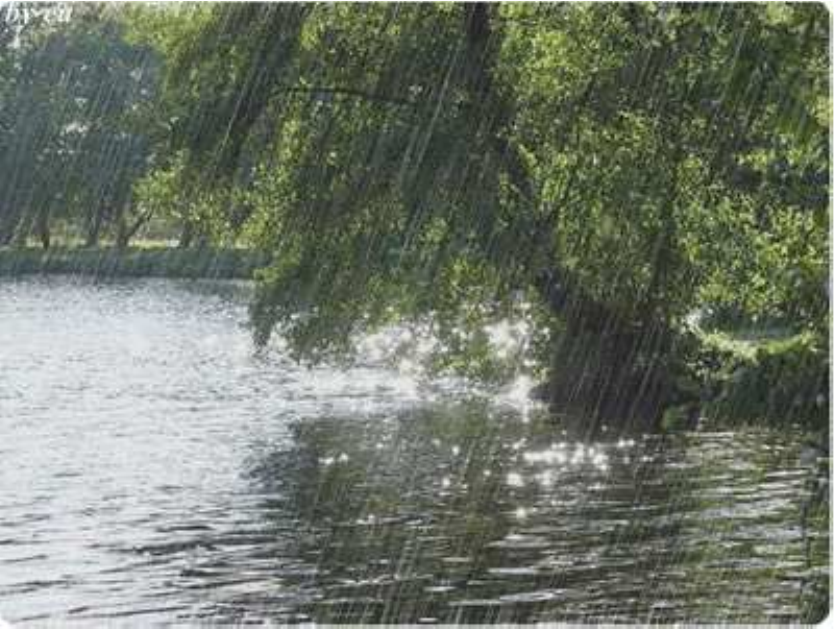}&
                \includegraphics[width=0.8in]{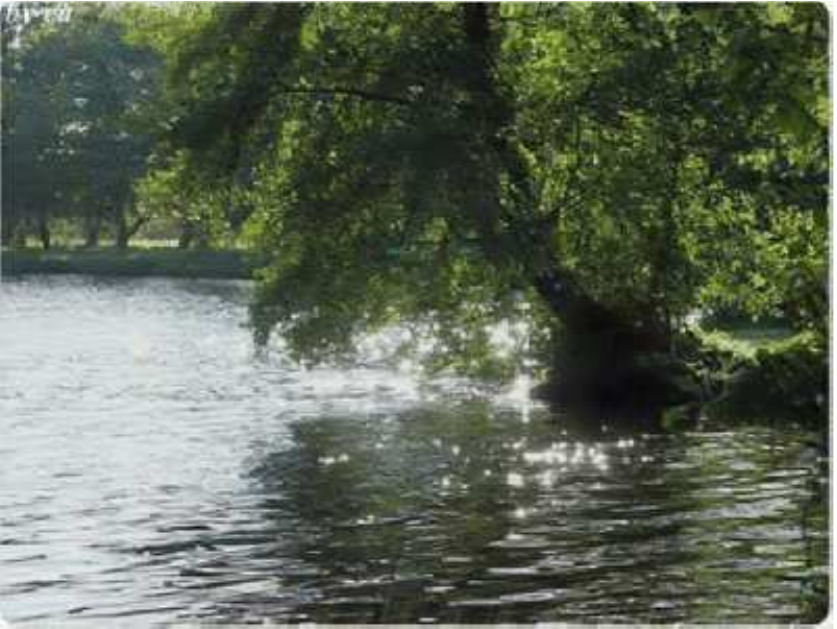}&
                \includegraphics[width=0.8in]{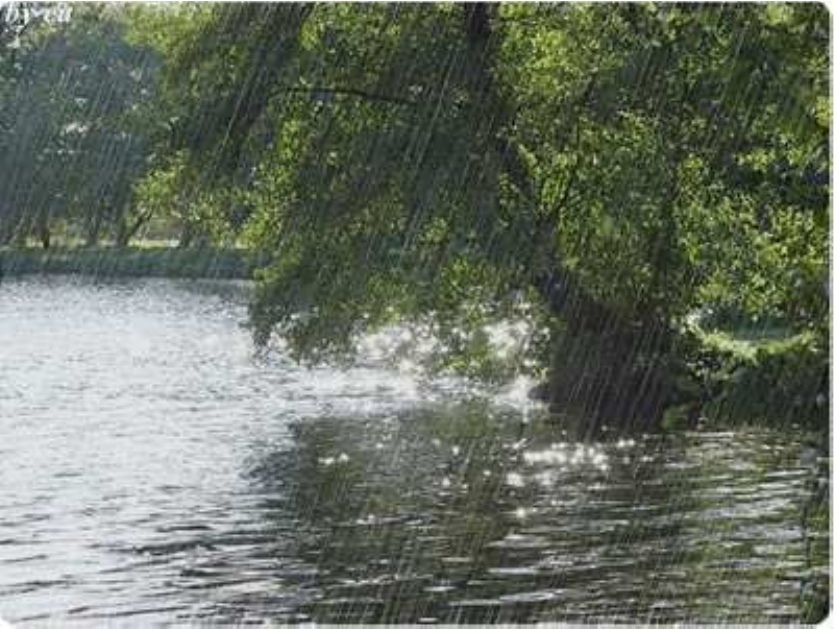}&
                \includegraphics[width=0.8in]{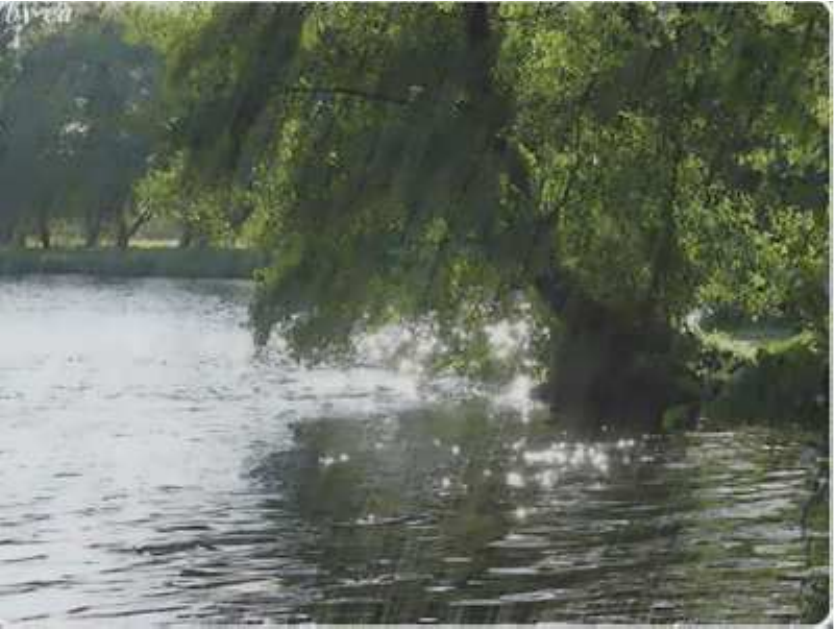}&
                \includegraphics[width=0.8in]{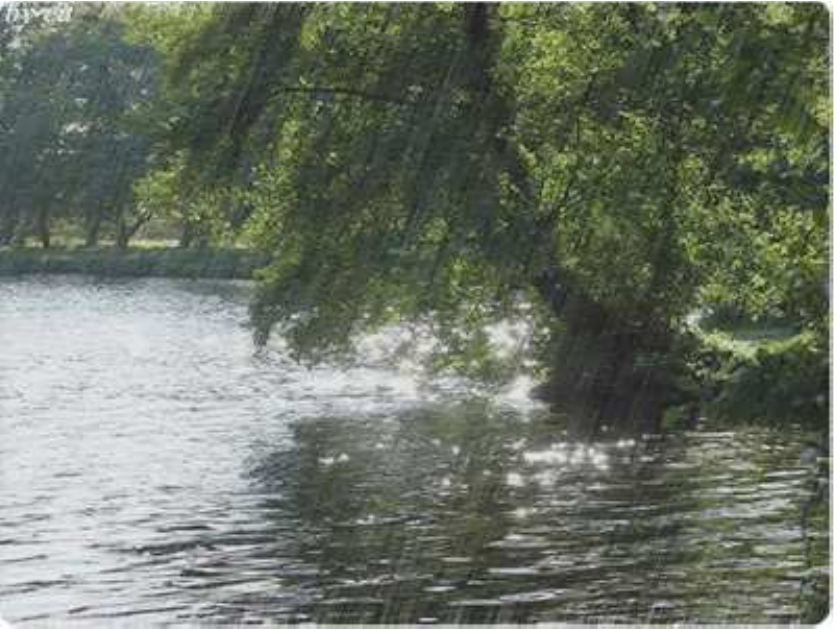}&
                \includegraphics[width=0.8in]{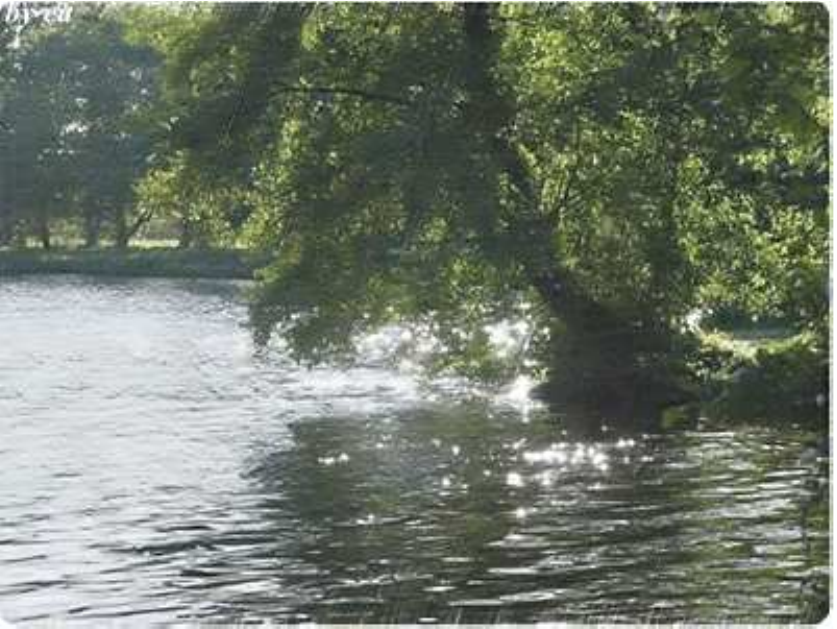}&
                \includegraphics[width=0.8in]{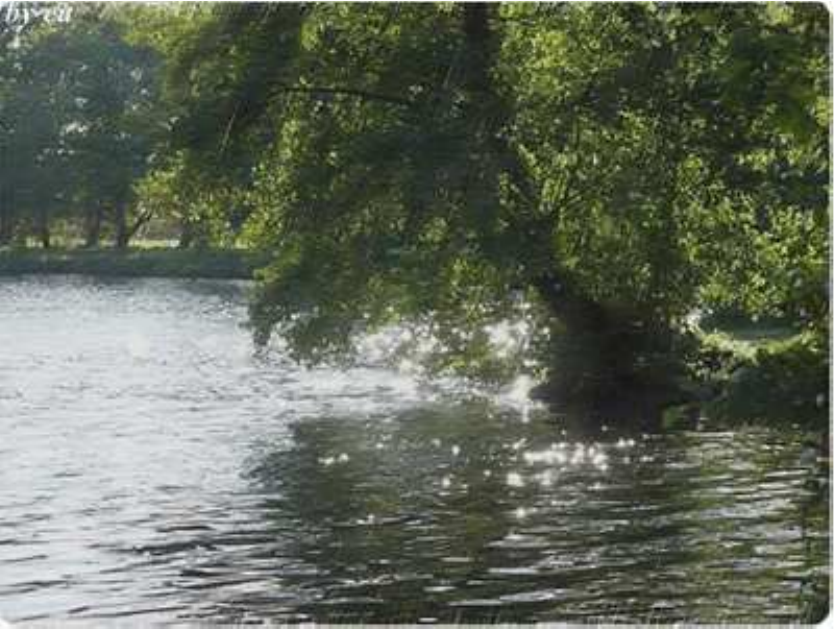}&
                \includegraphics[width=0.8in]{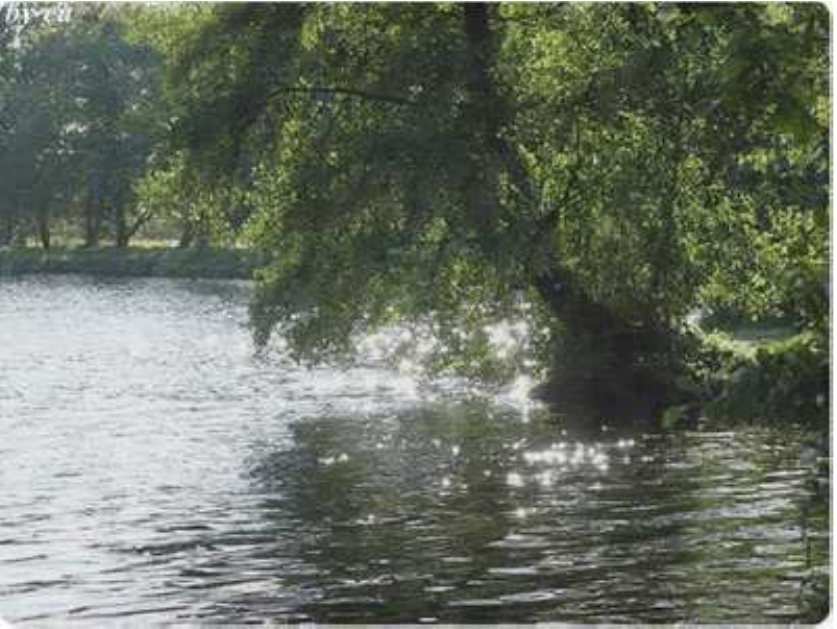}\\
                \vspace{0.5mm}

                \includegraphics[width=0.8in]{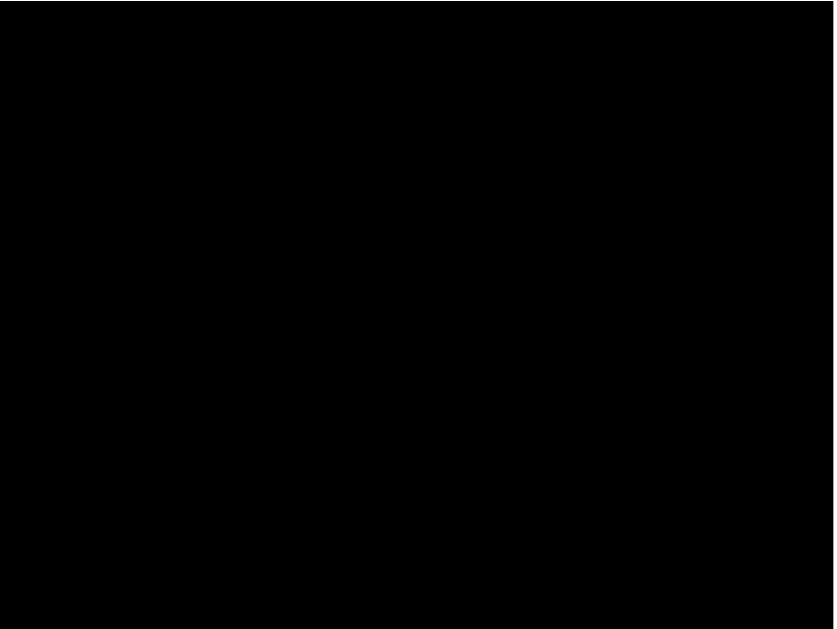}&
                \includegraphics[width=0.8in]{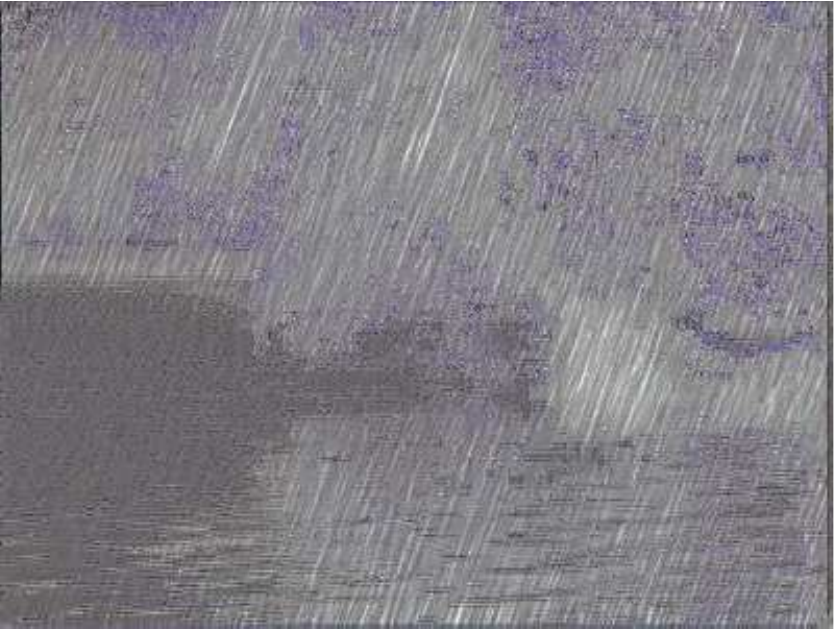}&
                \includegraphics[width=0.8in]{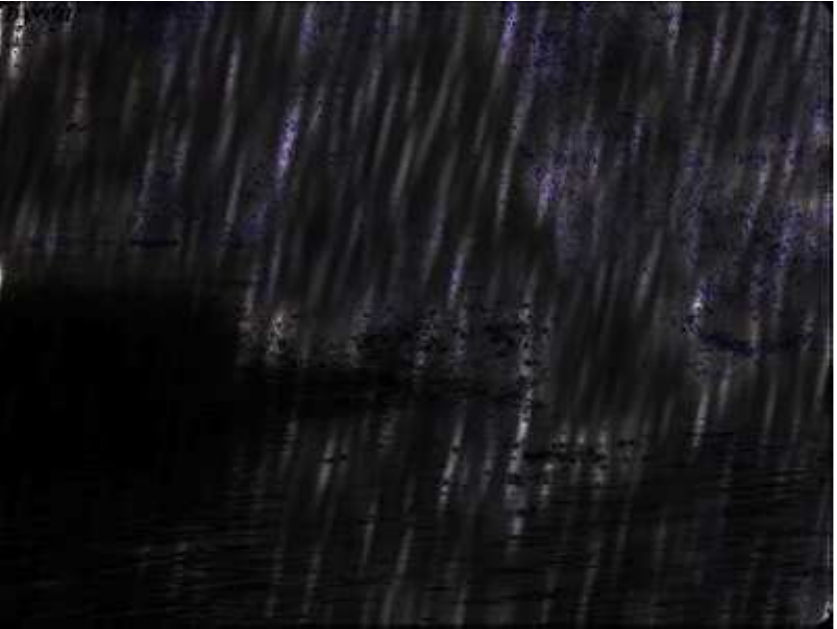}&
                \includegraphics[width=0.8in]{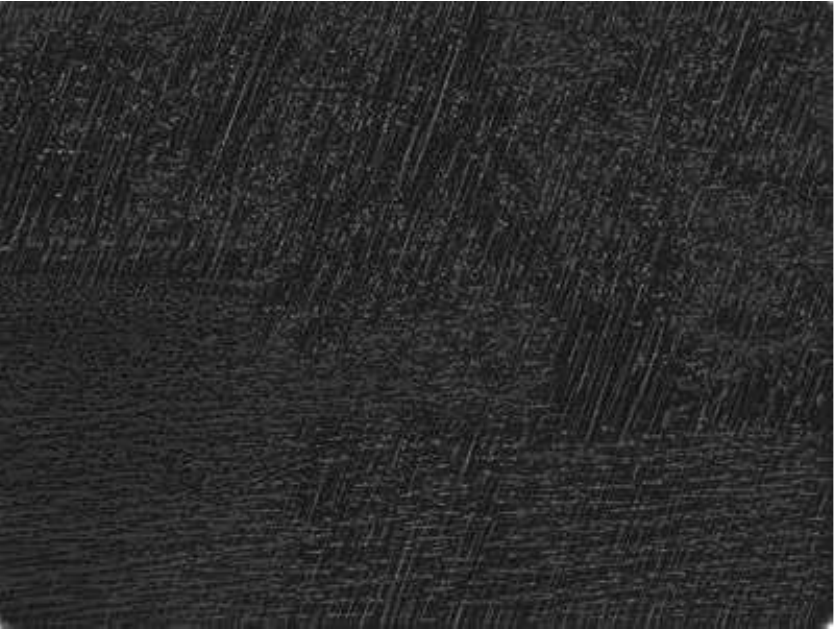}&
                \includegraphics[width=0.8in]{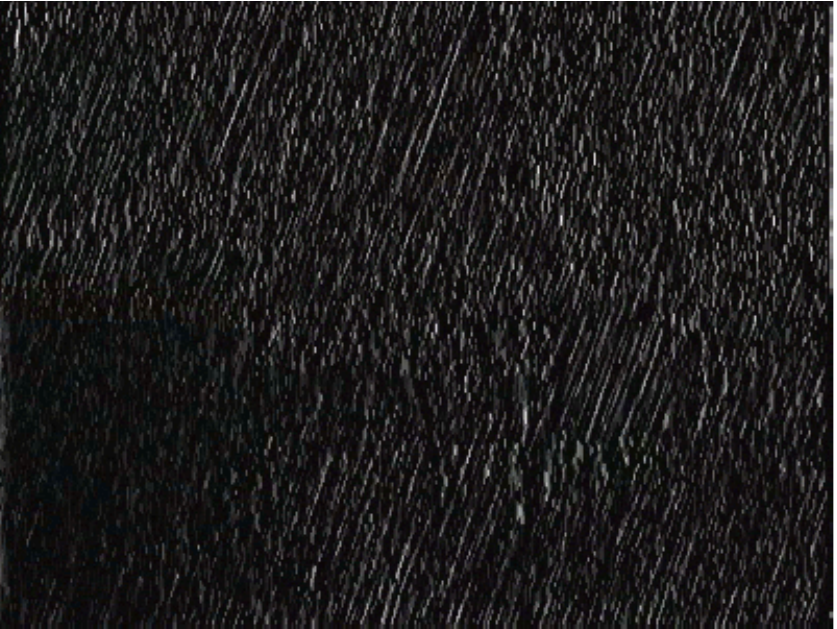}&
                \includegraphics[width=0.8in]{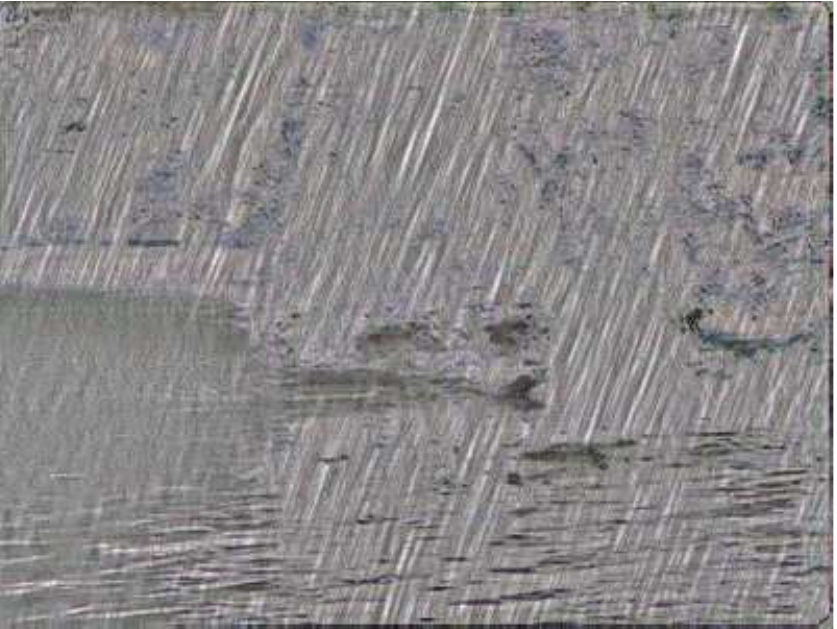}&
                \includegraphics[width=0.8in]{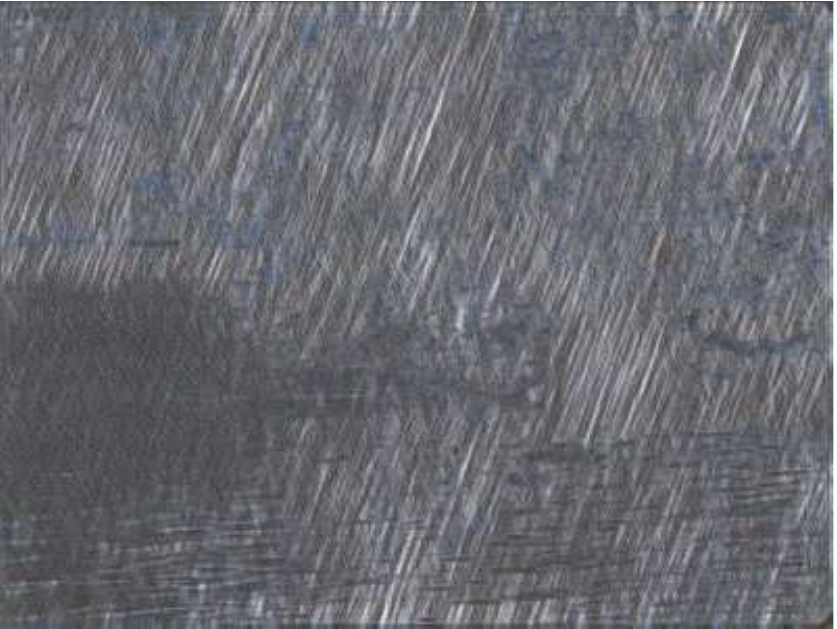}&
                \includegraphics[width=0.8in]{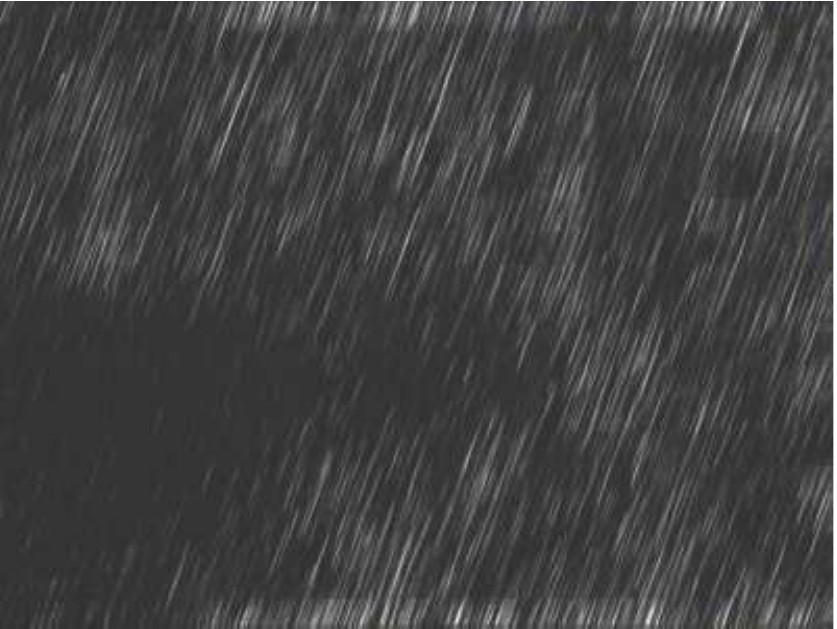}\\
                \vspace{0.5mm}

                \includegraphics[width=0.8in]{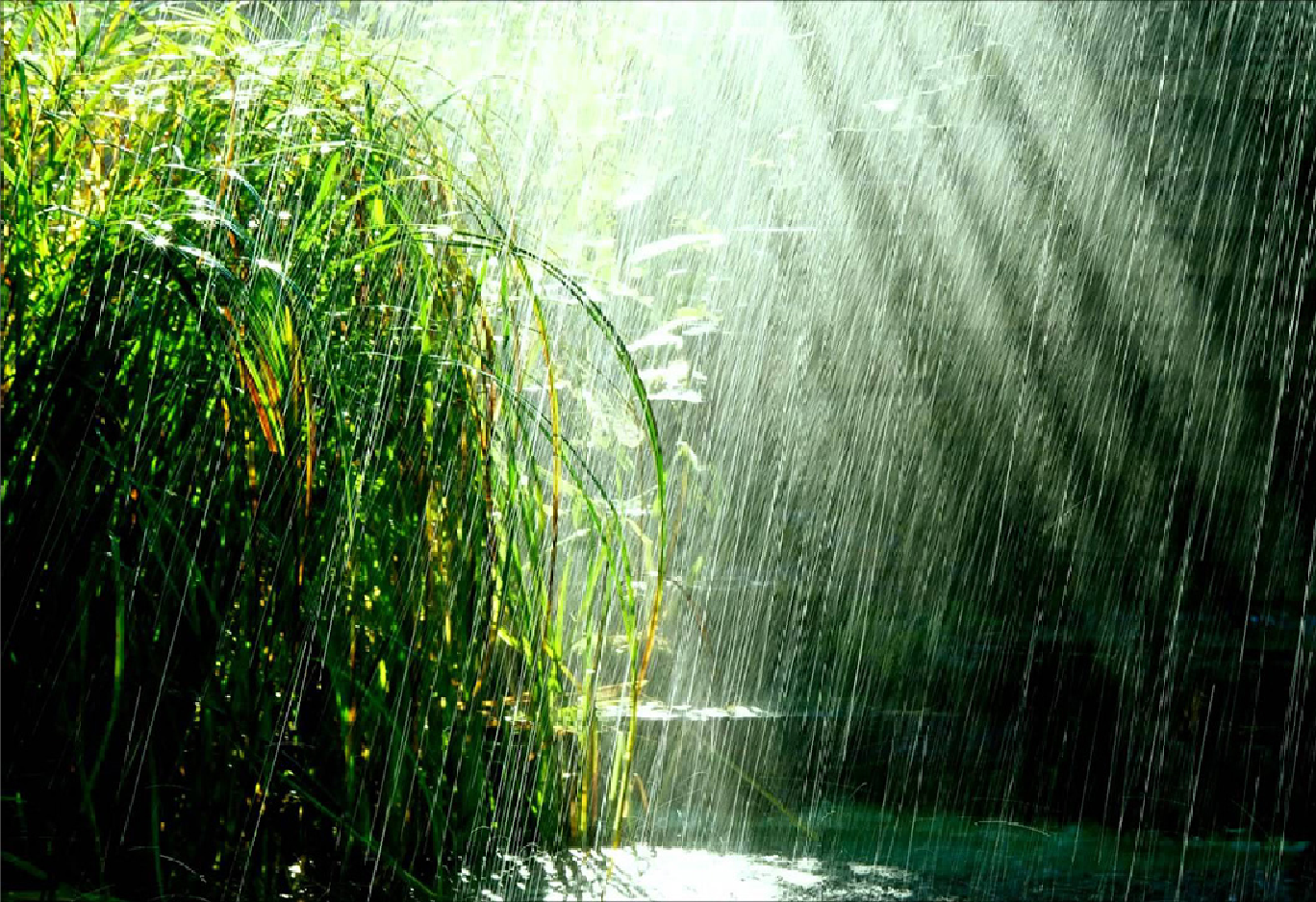}&
                \includegraphics[width=0.8in]{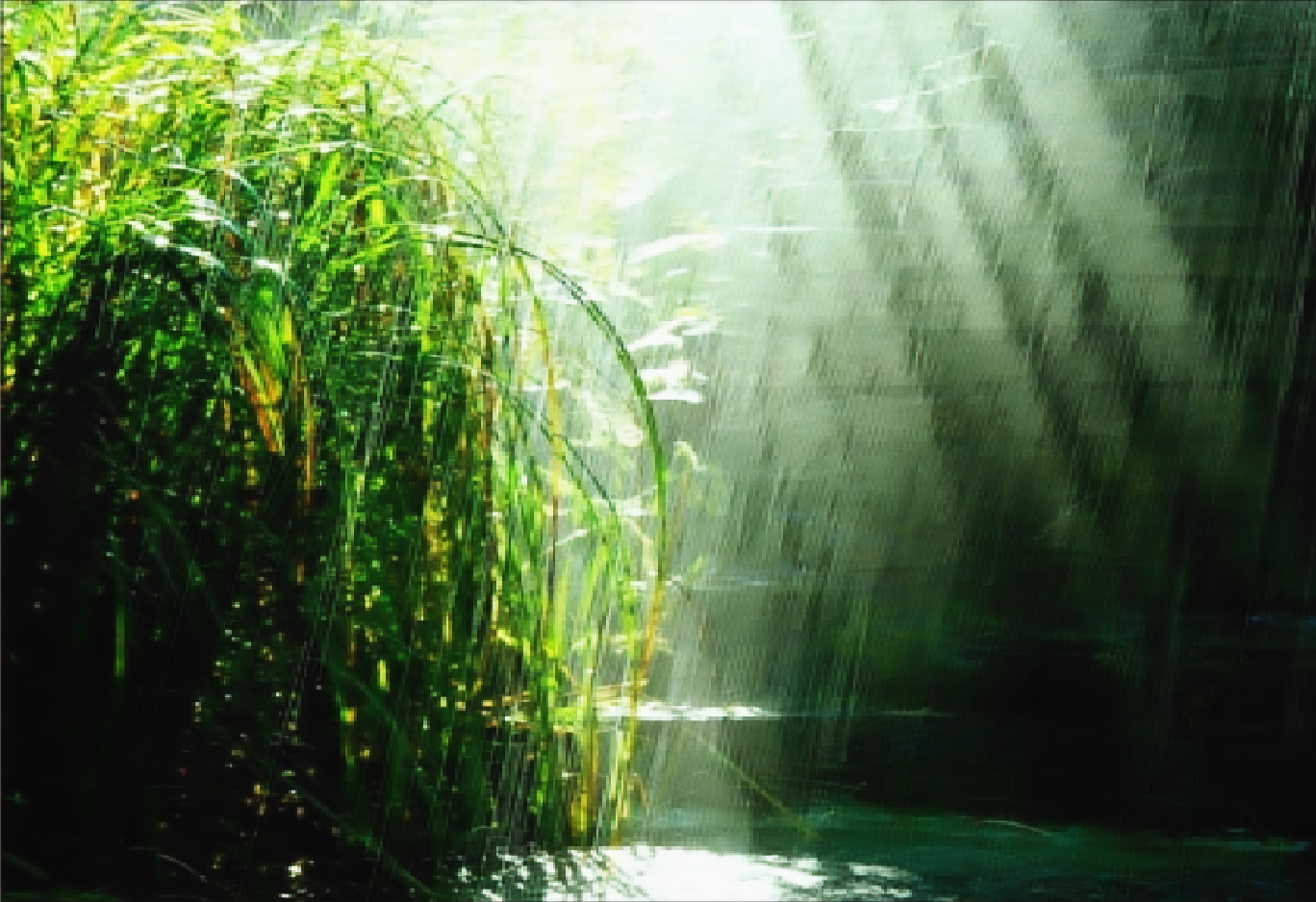}&
                \includegraphics[width=0.8in]{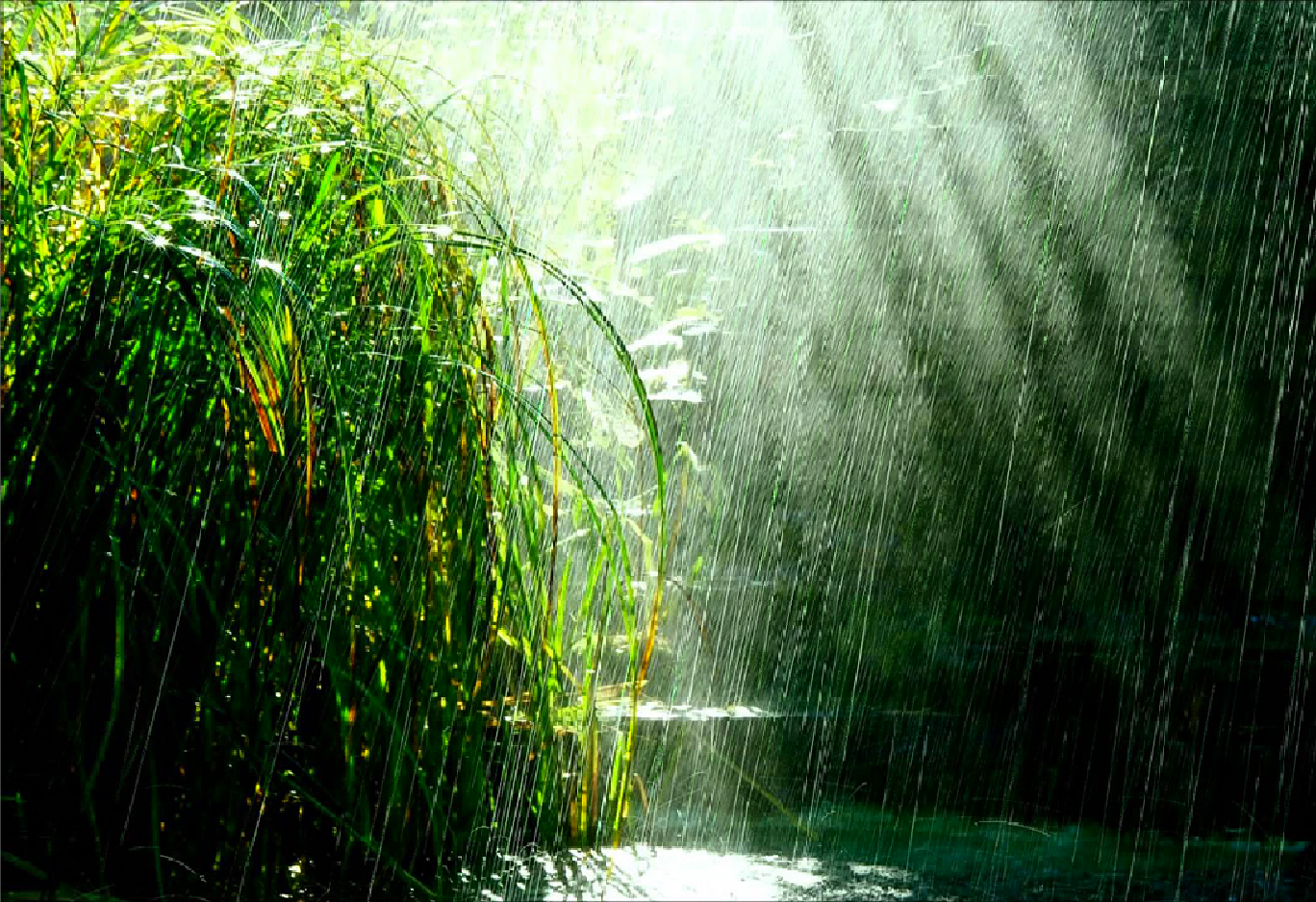}&
                \includegraphics[width=0.8in]{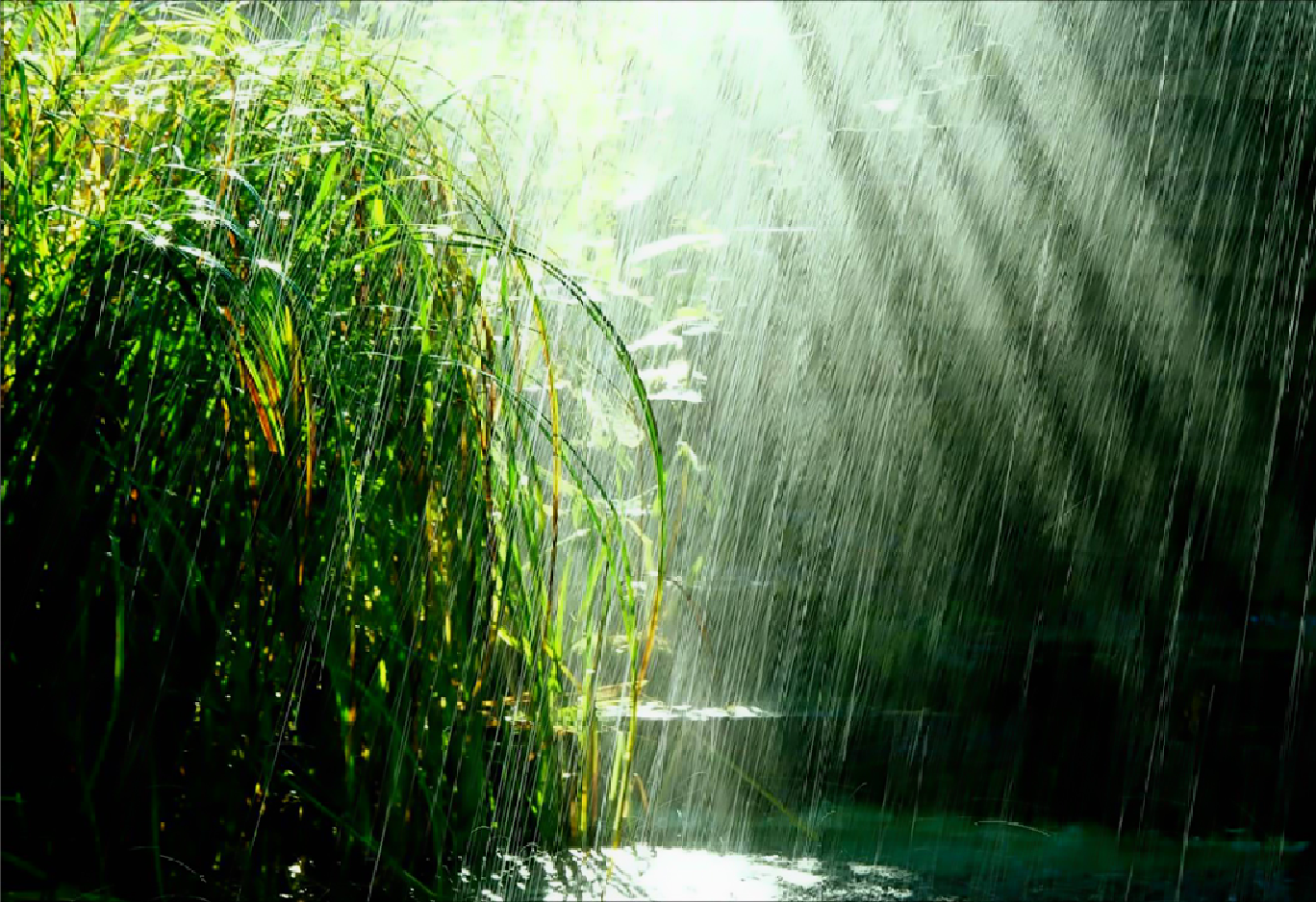}&
                \includegraphics[width=0.8in]{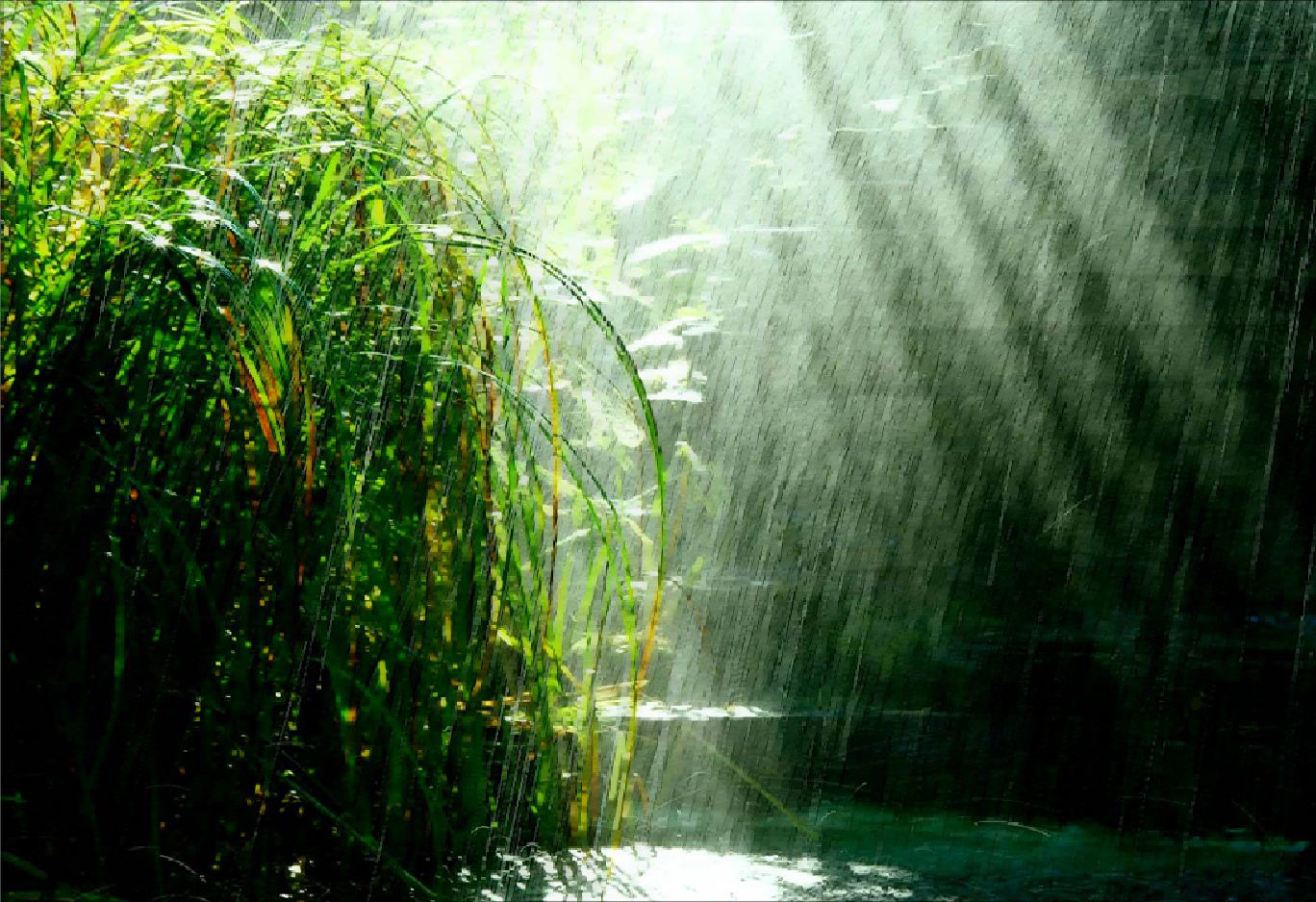}&
                \includegraphics[width=0.8in]{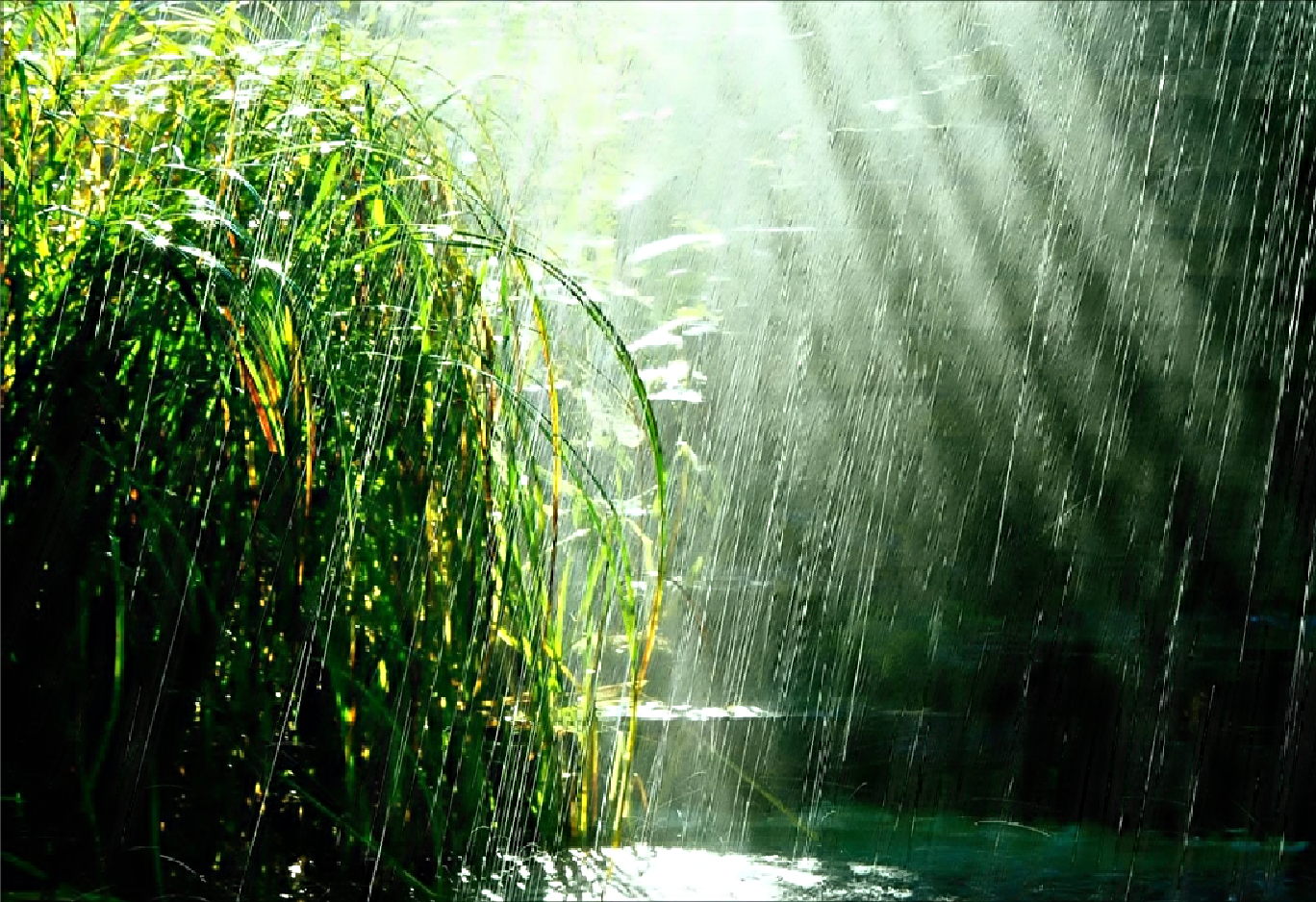}&
                \includegraphics[width=0.8in]{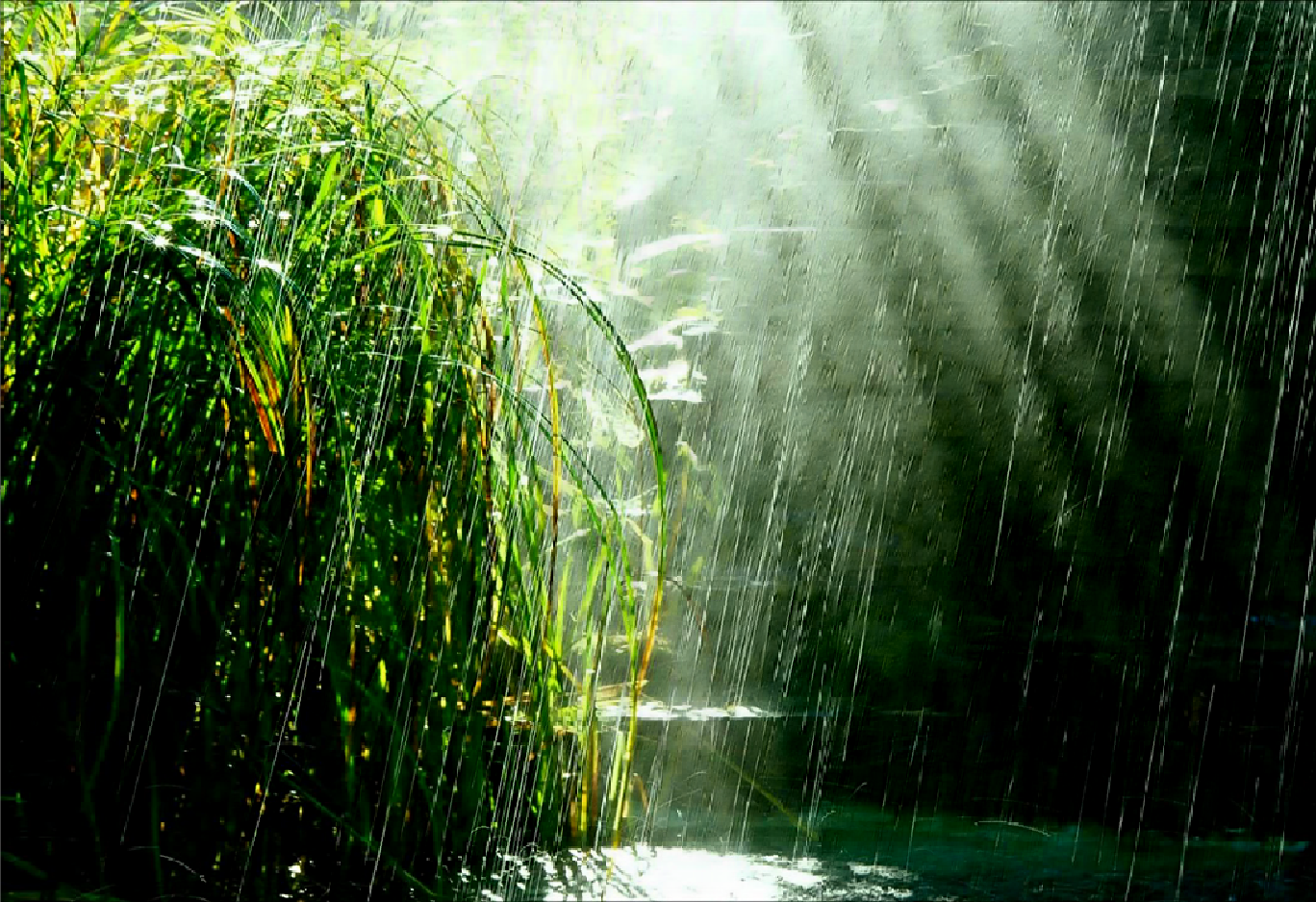}&
                \includegraphics[width=0.8in]{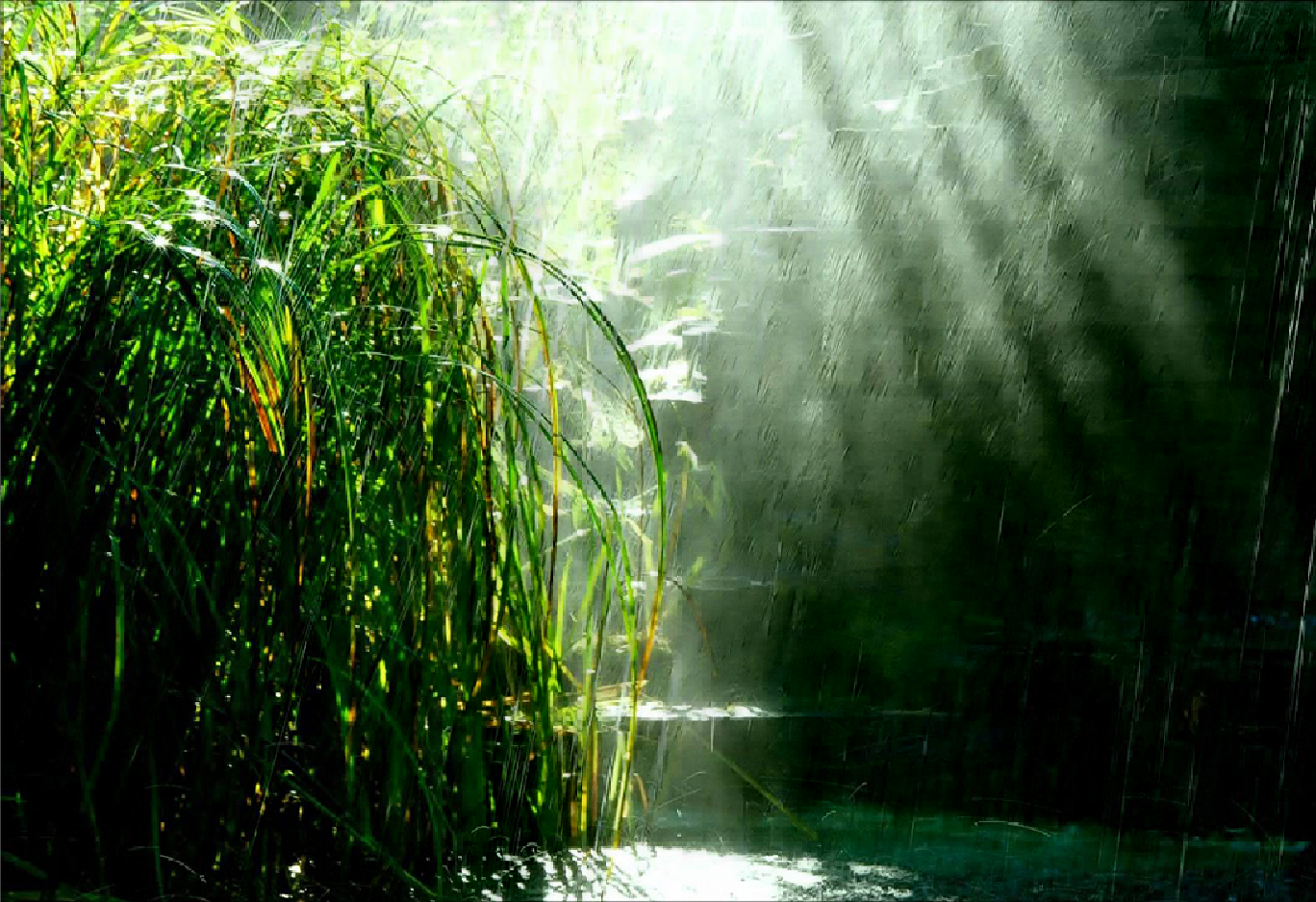}\\
                \vspace{0.5mm}

                \includegraphics[width=0.8in]{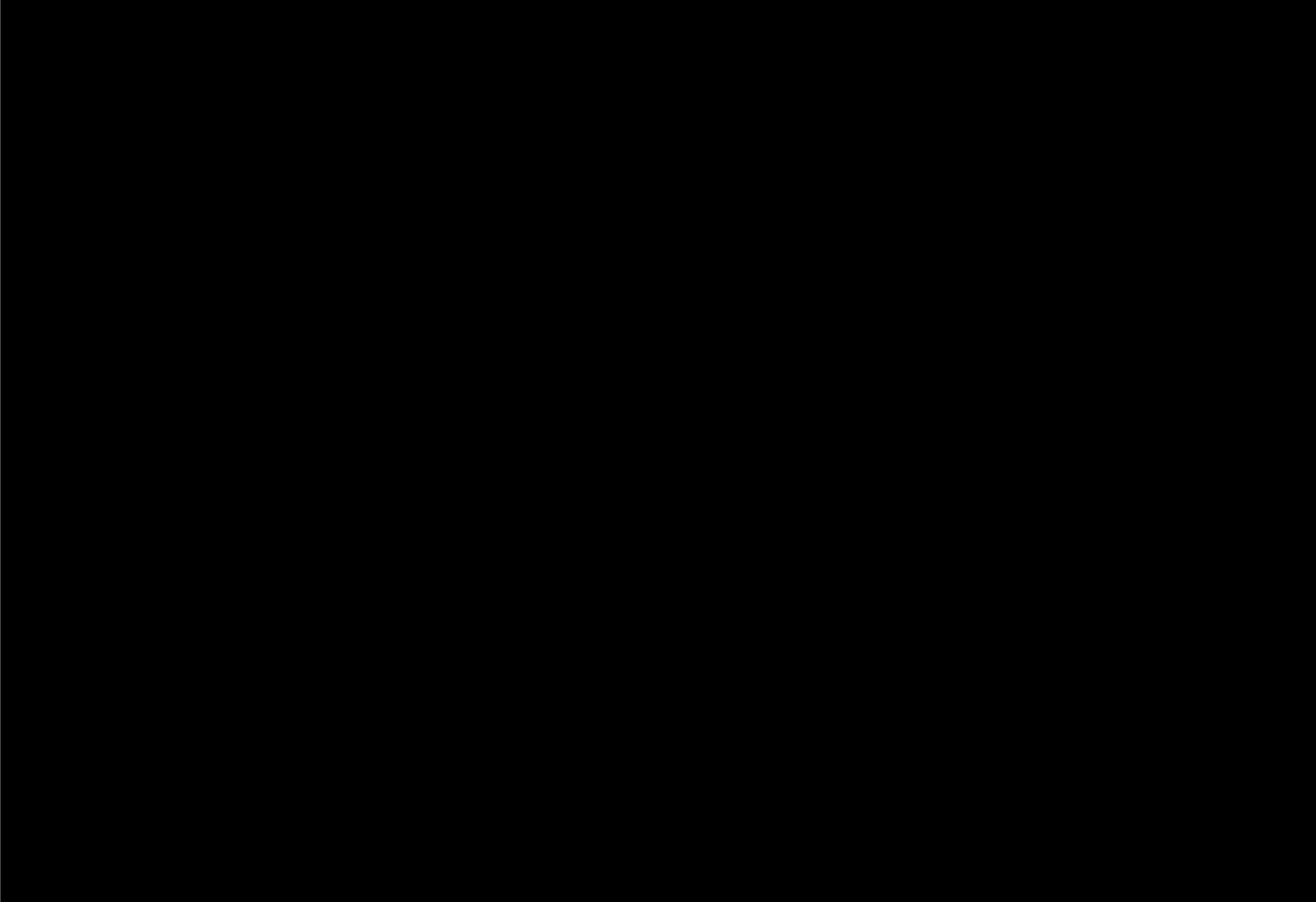}&
                \includegraphics[width=0.8in]{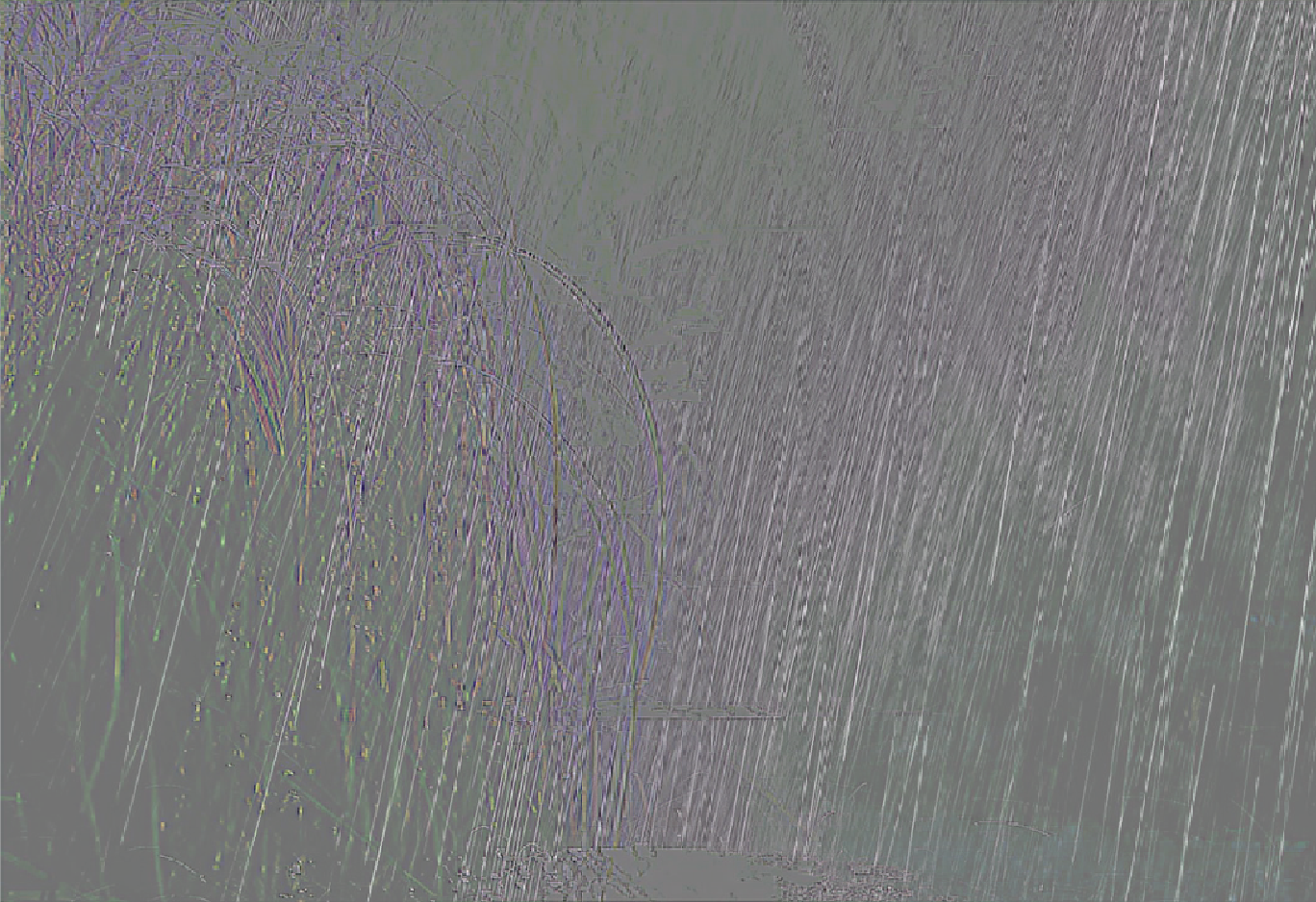}&
                \includegraphics[width=0.8in]{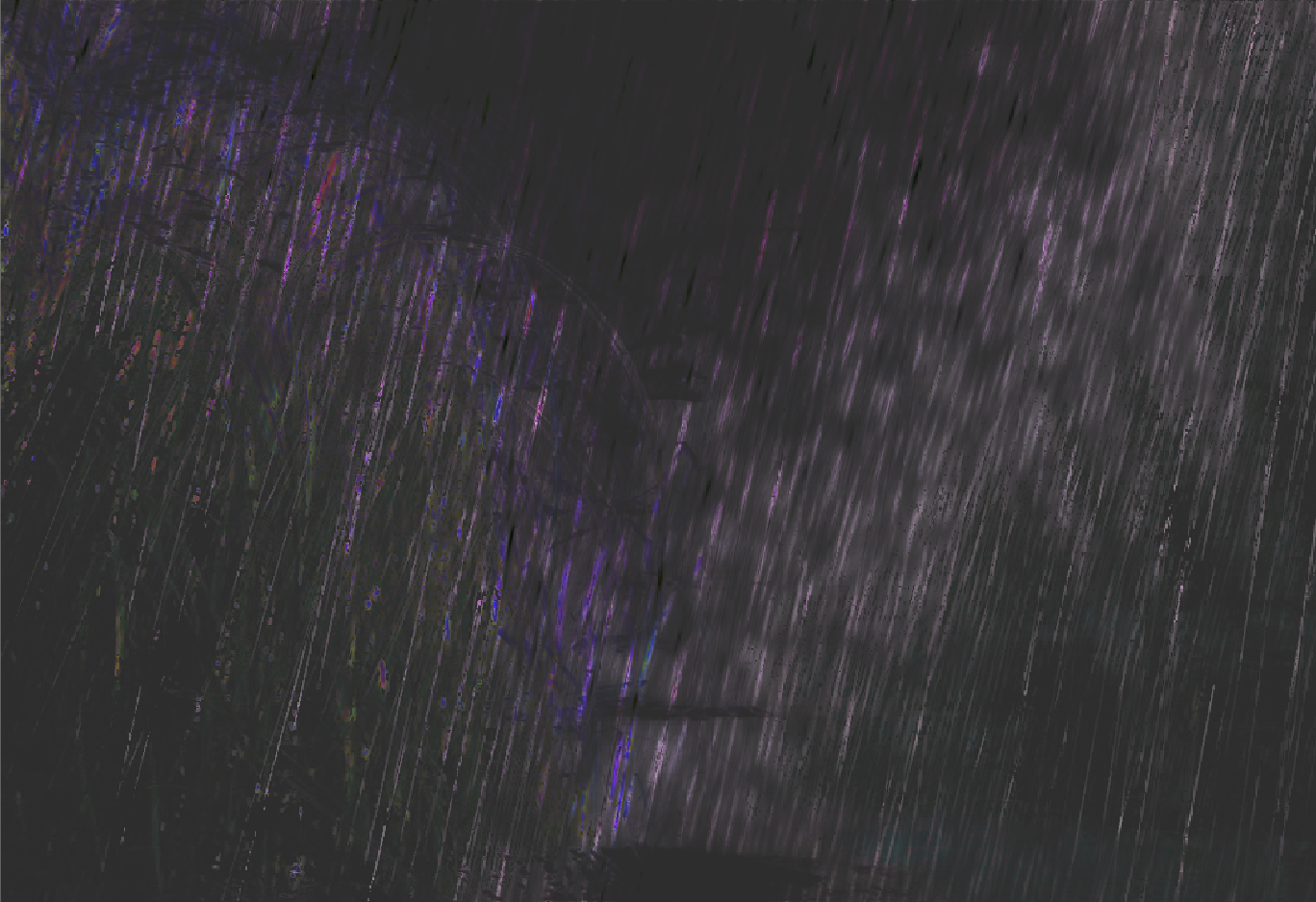}&
                \includegraphics[width=0.8in]{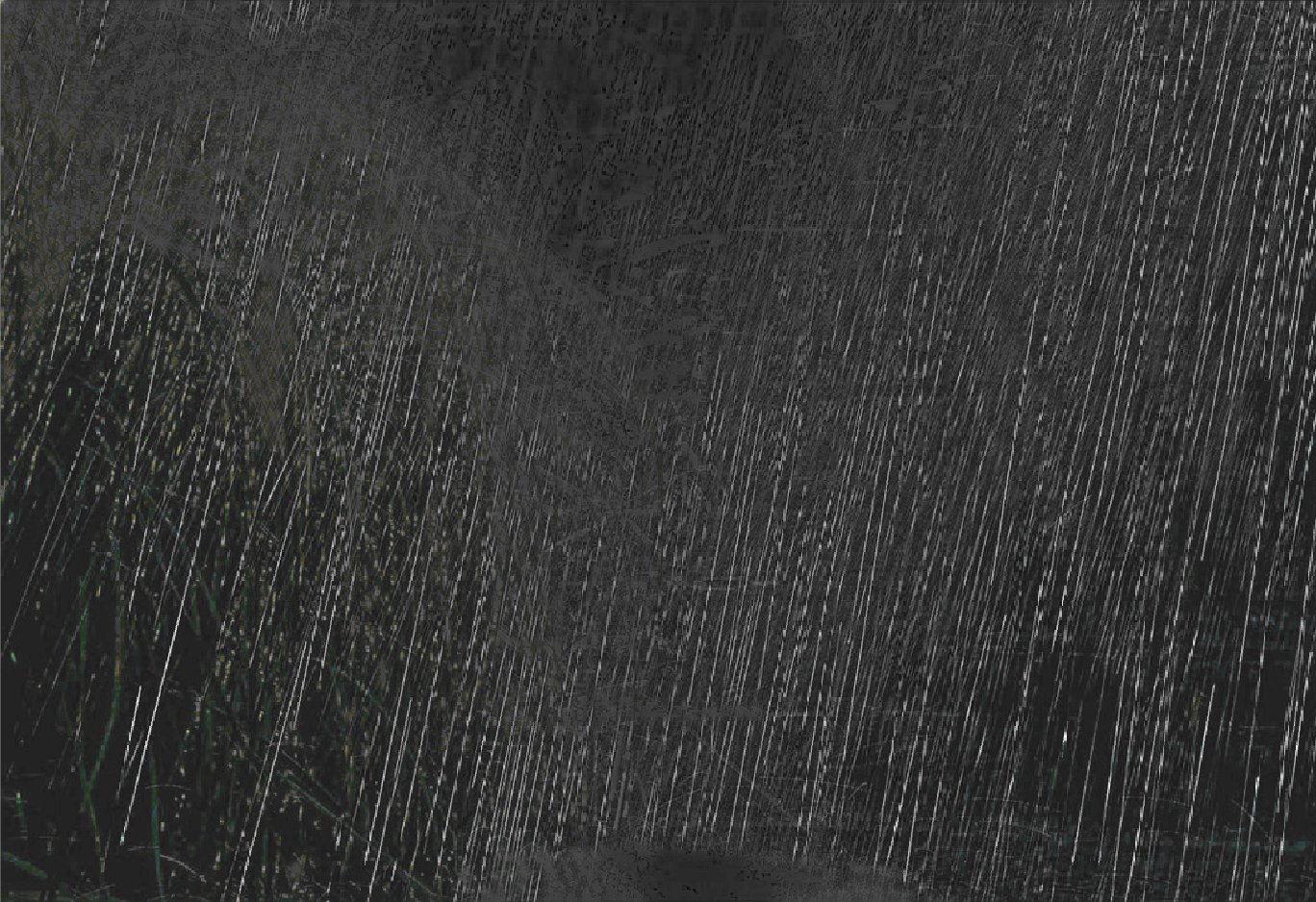}&
                \includegraphics[width=0.8in]{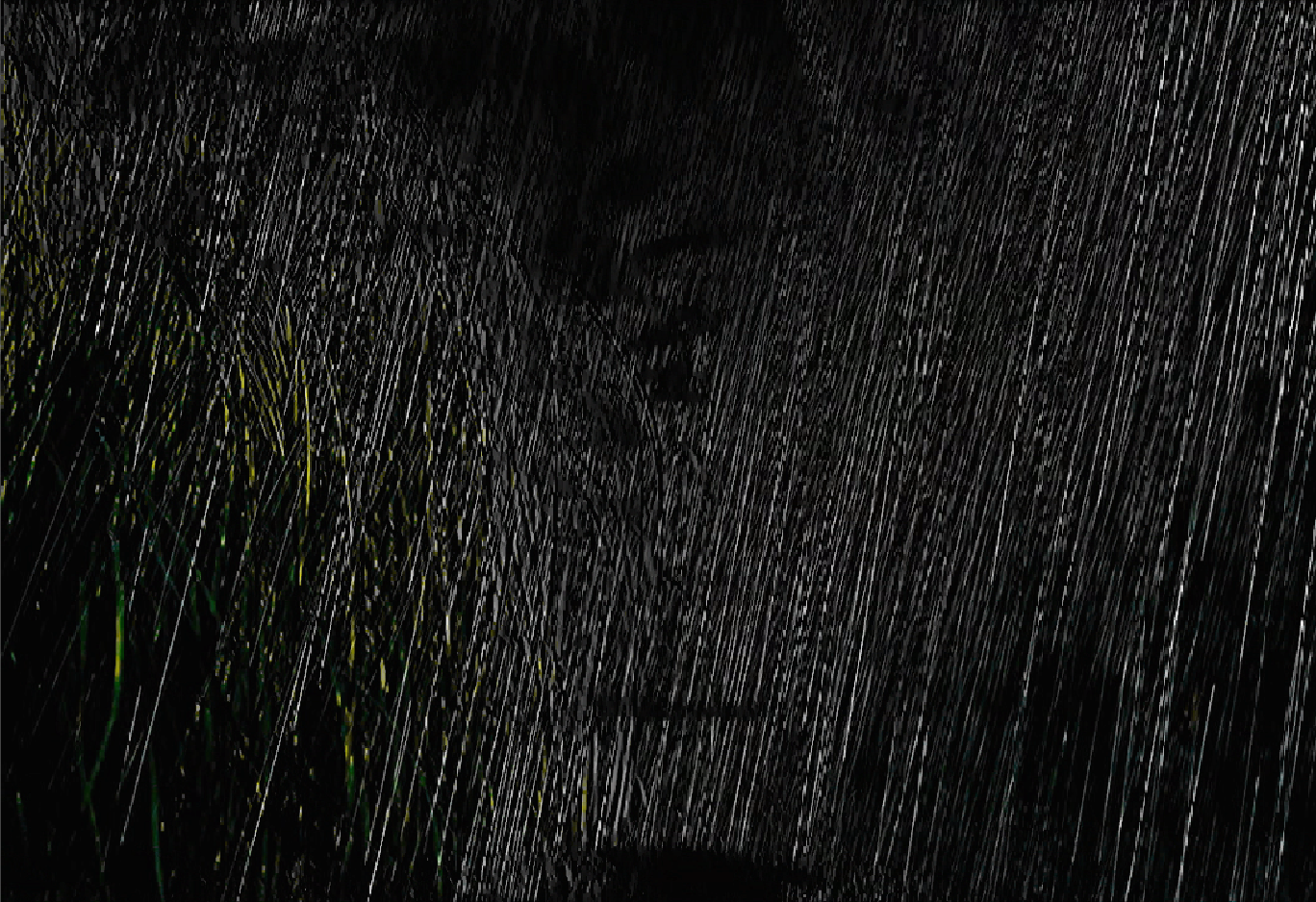}&
                \includegraphics[width=0.8in]{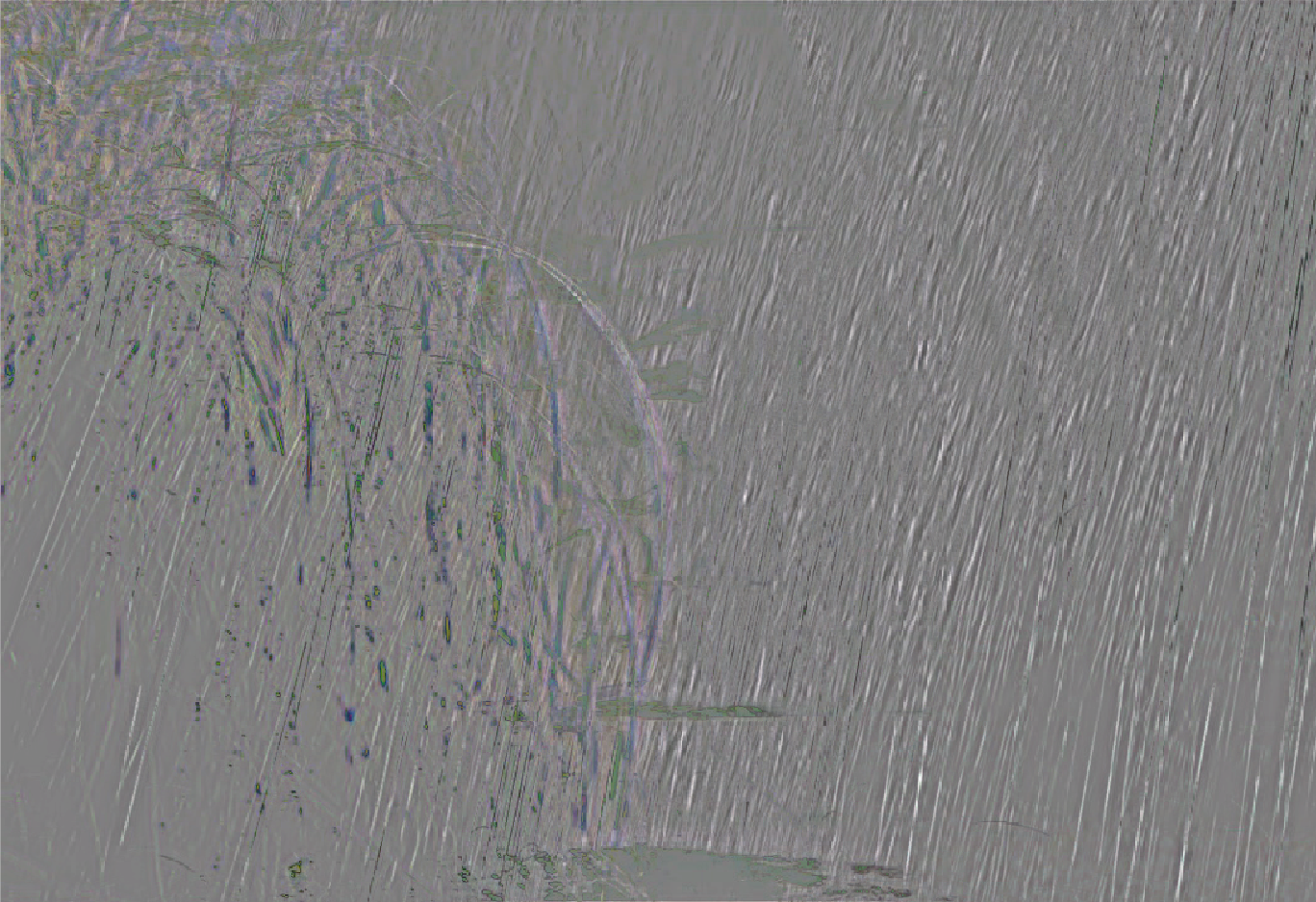}&
                \includegraphics[width=0.8in]{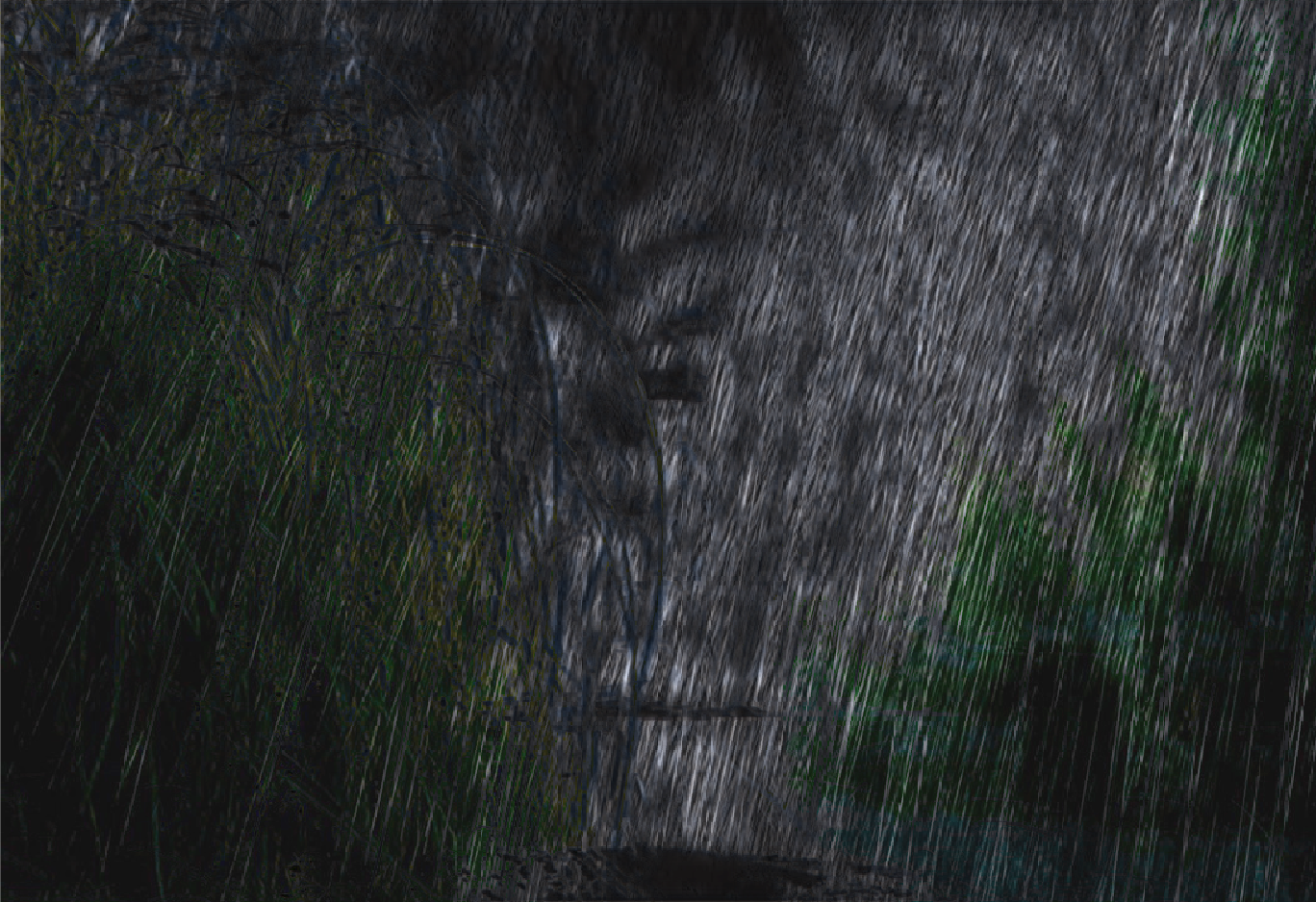}&
                \includegraphics[width=0.8in]{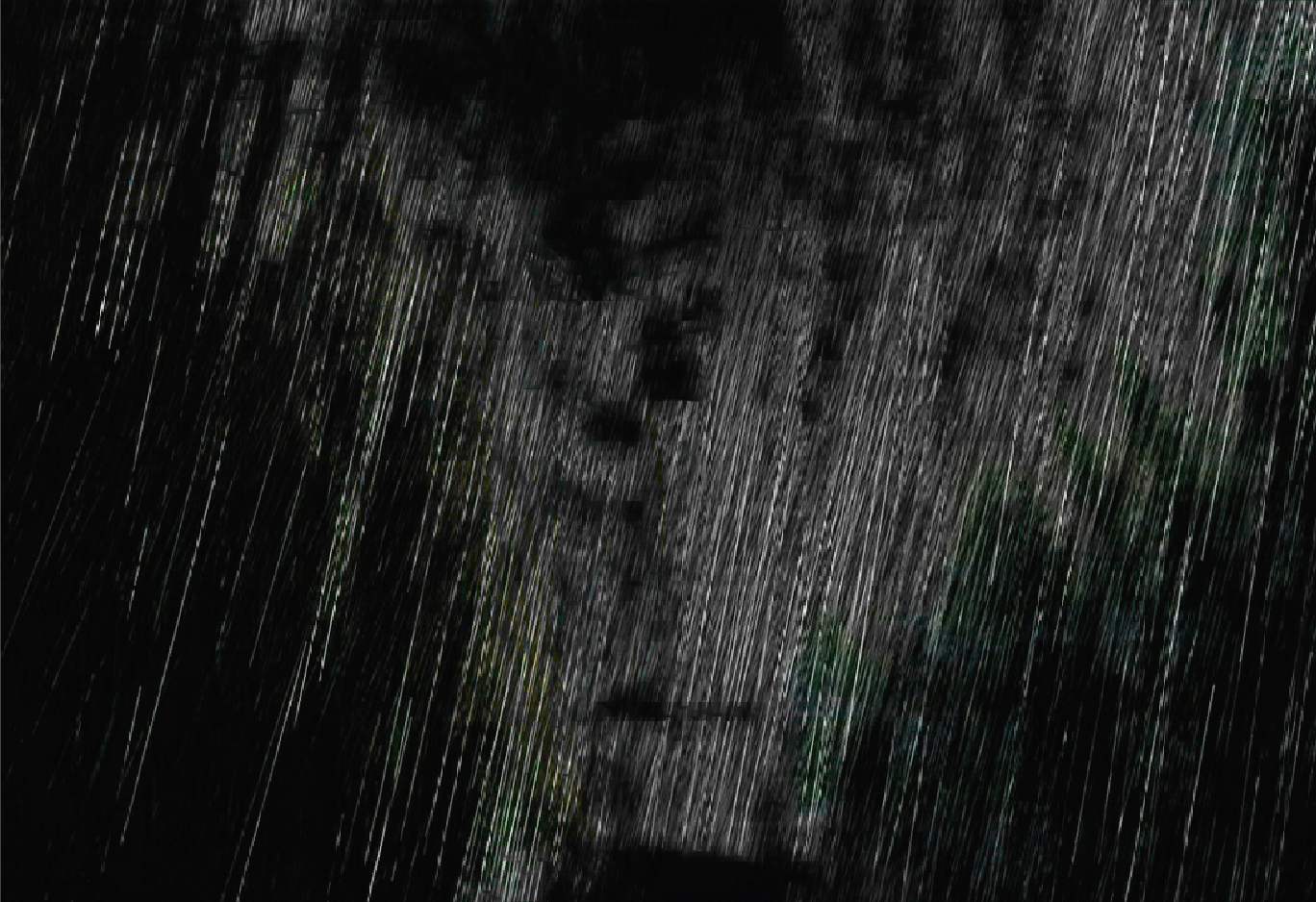}\\
                \vspace{0.5mm}

                \includegraphics[width=0.8in]{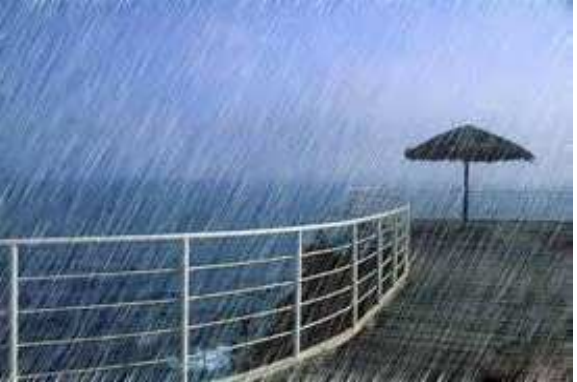}&
                \includegraphics[width=0.8in]{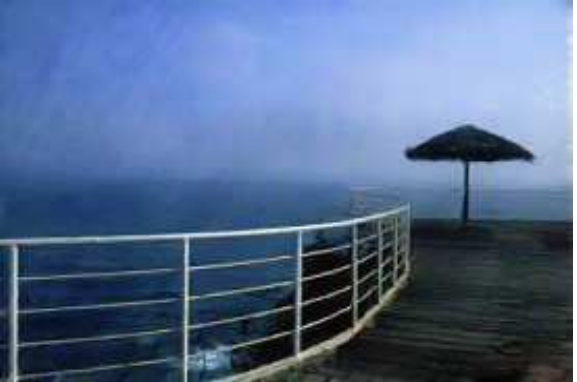}&
                \includegraphics[width=0.8in]{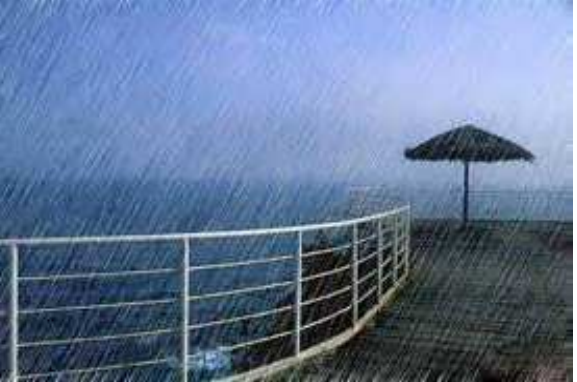}&
                \includegraphics[width=0.8in]{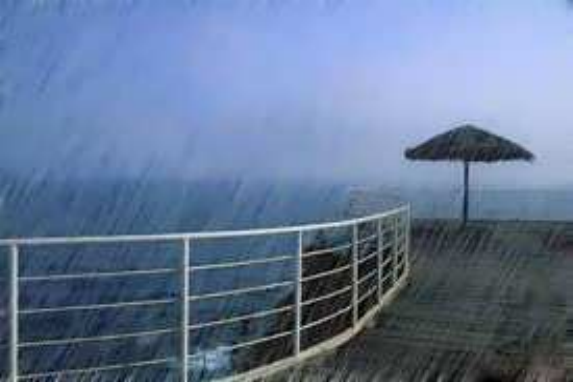}&
                \includegraphics[width=0.8in]{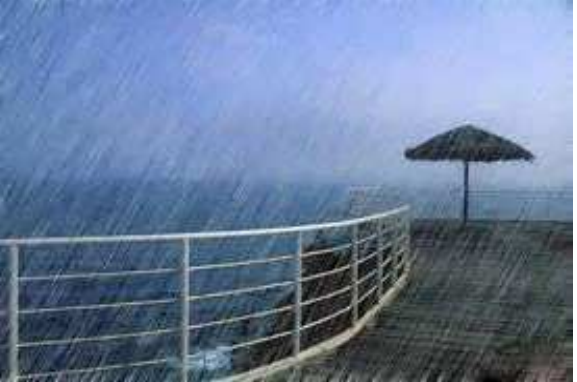}&
                \includegraphics[width=0.8in]{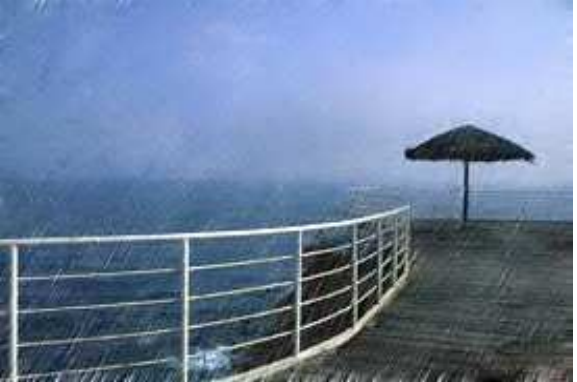}&
                \includegraphics[width=0.8in]{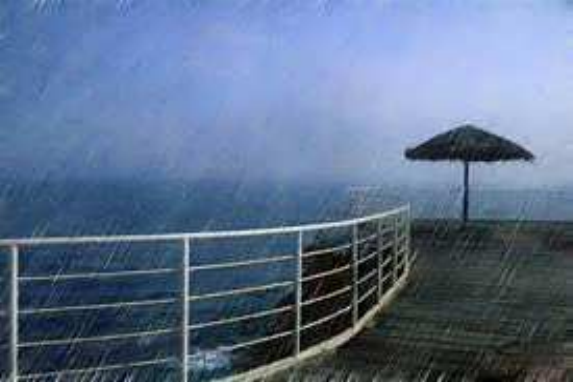}&
                \includegraphics[width=0.8in]{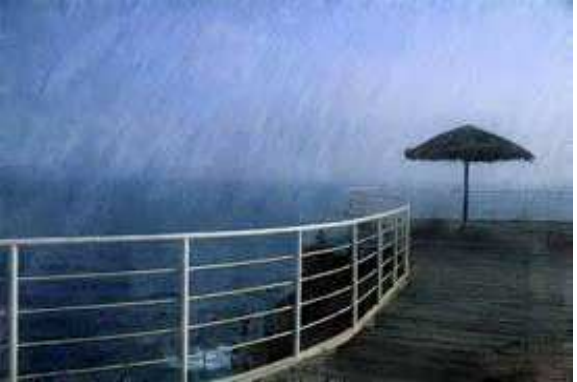}\\
                \vspace{0.5mm}

                \includegraphics[width=0.8in]{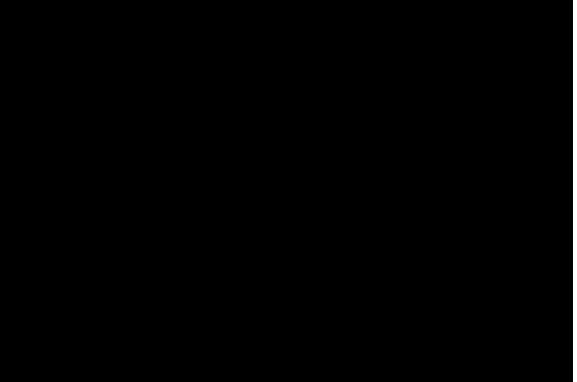}&
                \includegraphics[width=0.8in]{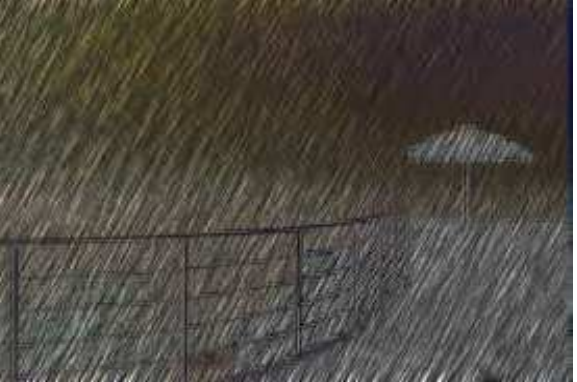}&
                \includegraphics[width=0.8in]{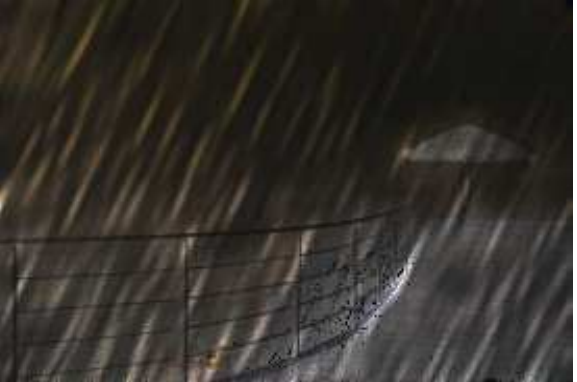}&
                \includegraphics[width=0.8in]{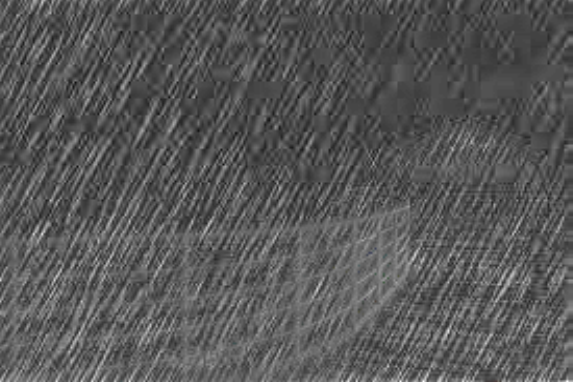}&
                \includegraphics[width=0.8in]{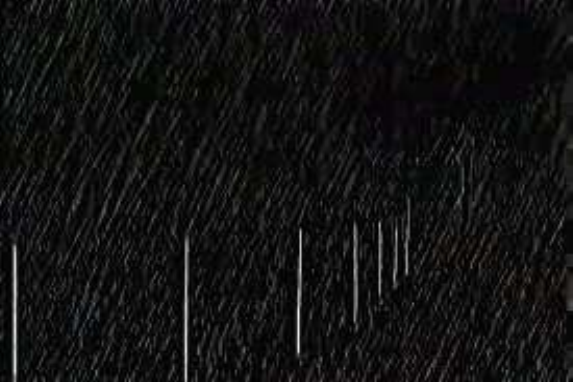}&
                \includegraphics[width=0.8in]{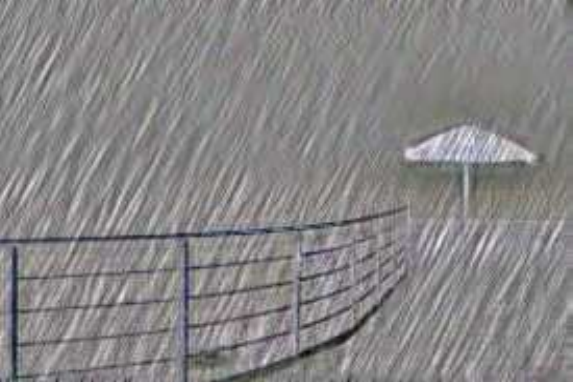}&
                \includegraphics[width=0.8in]{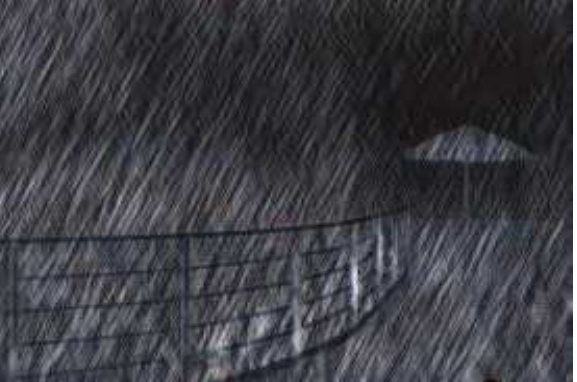}&
                \includegraphics[width=0.8in]{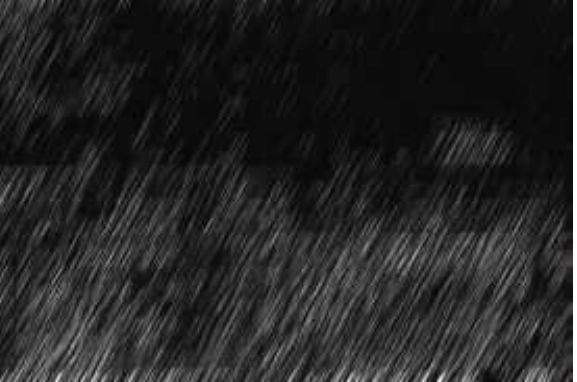}\\
                (a)&
                (b)&
                (c)&
                (d)&
                (e)&
                (f)&
                (g)&
                (h)\\

\end{tabular}
\caption{Rain streak removal results and rain streak images by different methods on real rainy images. From left to right: (a) the rainy images, the results by (b) DID \cite{zhang2018density}, (c) DSC \cite{luo2015removing}, (d) LP \cite{Li2014Single}, (e) UGSM \cite{Deng2018A}, (f) CNN \cite{fu2017clearing}, (g) DDN \cite{fu2017removing}, and (h) KGCNN.}
\label{real-visual3}
\end{center}
\end{figure*}

In Fig. \ref{real-visual2}, the KGCNN method removes the rain streaks completely while other approaches still exist obvious rain streaks.
From the rain streak images, our method could separate rain streaks excellently, while other method leaves some background texture to the separated rain streaks.
Especially the visual result by DID method shows a little of blur effect due to the loss of the texture information.
Particularly, readers can find more results in Fig. \ref{real-visual3} which also verifies the superiority of the proposed KGCNN method.

\subsection{Influence of kernel in the KGCNN method}\label{Sec:Exp:Kernel}
\begin{figure}[!htb]
\renewcommand\arraystretch{0.8}\setlength{\tabcolsep}{1.8pt}
\begin{center}
\begin{tabular}{cccccc}

                ~\includegraphics[width=0.64in]{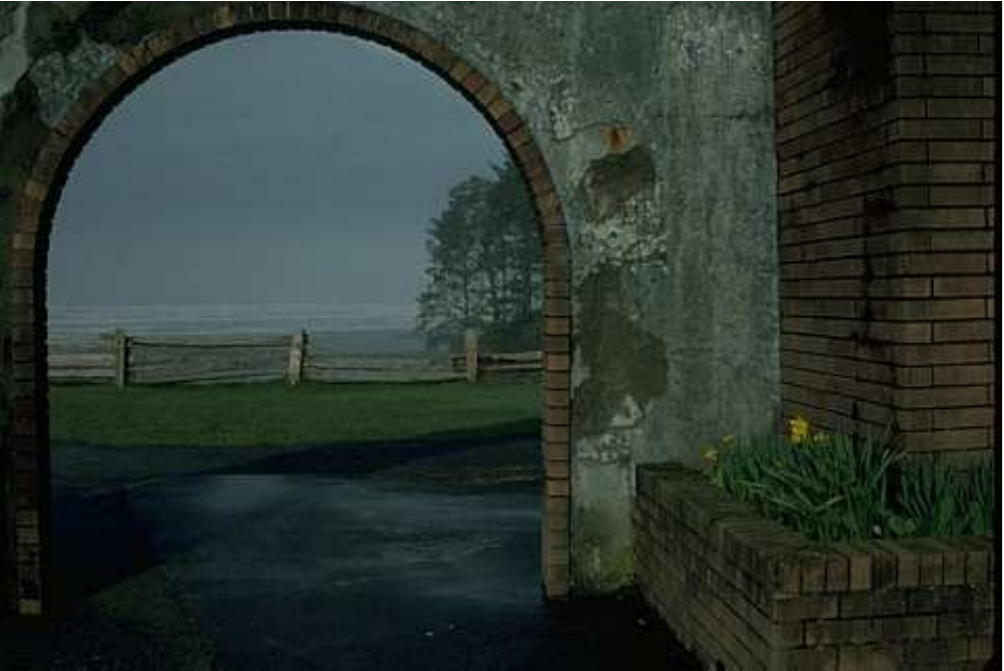}&
                \includegraphics[width=0.64in]{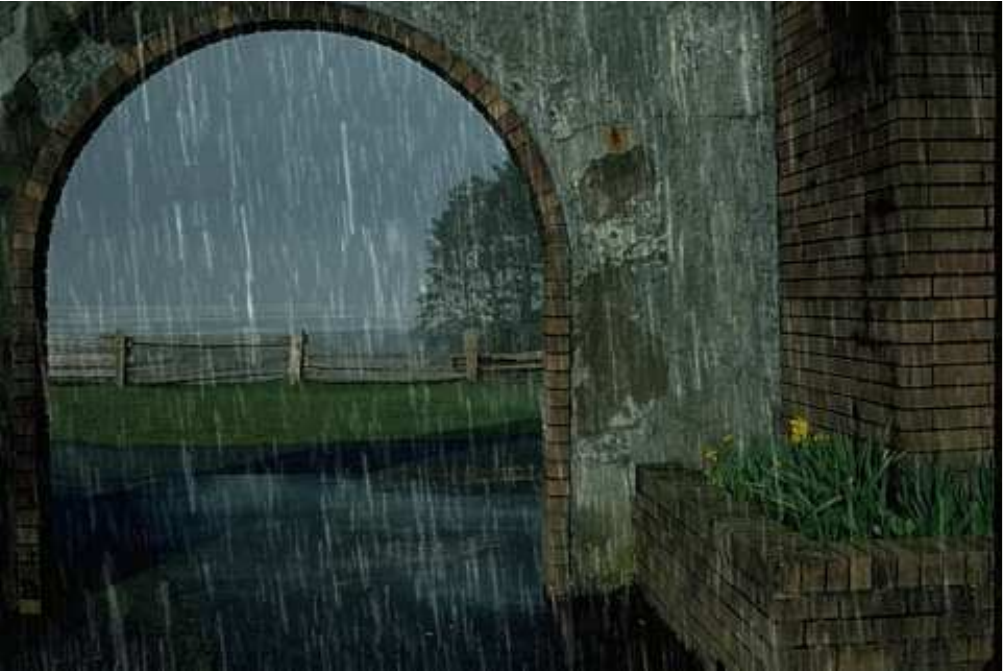}&
                \includegraphics[width=0.64in]{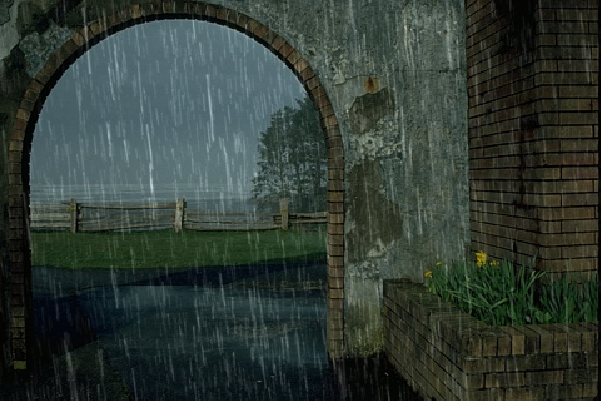}&
                \includegraphics[width=0.64in]{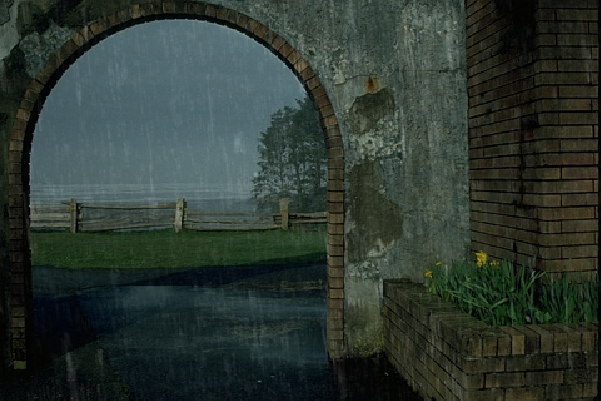}&
                \includegraphics[width=0.64in]{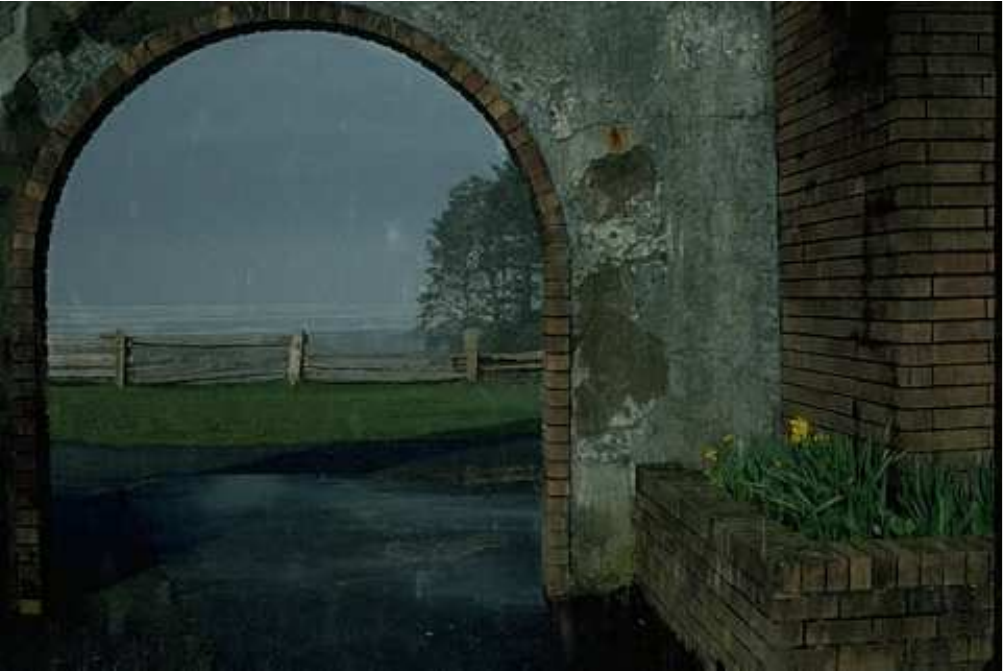}\\
                 \vspace{0.5mm}

                \includegraphics[width=0.64in]{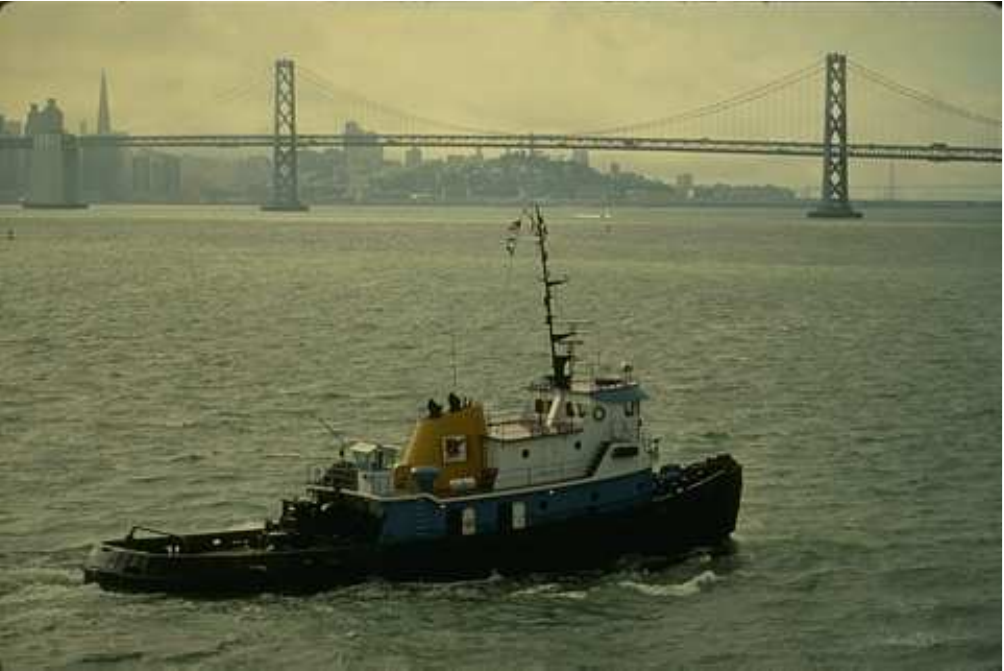}&
                \includegraphics[width=0.64in]{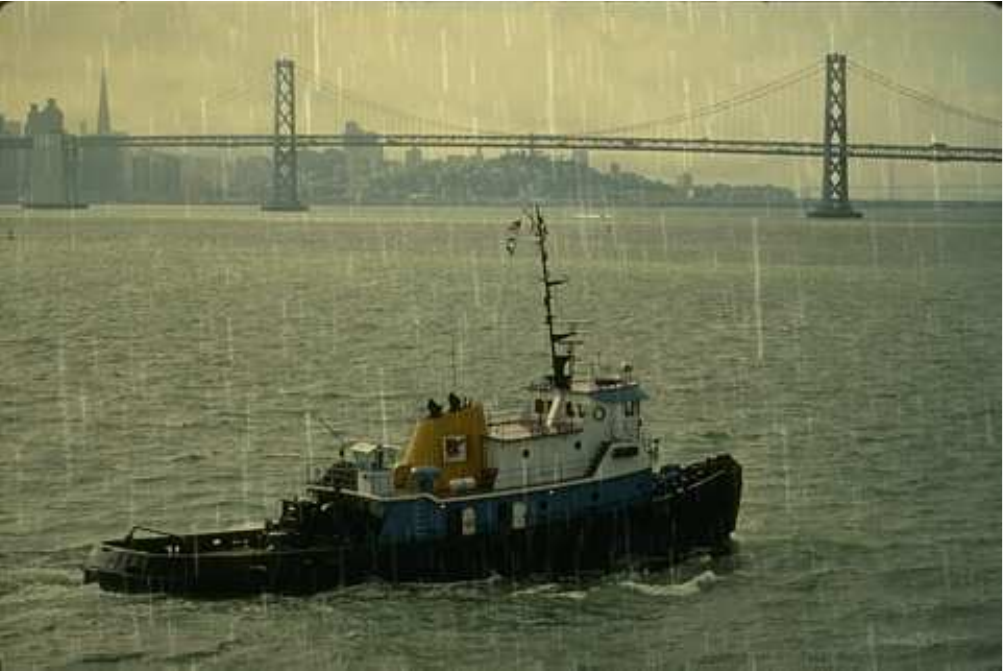}&
                \includegraphics[width=0.64in]{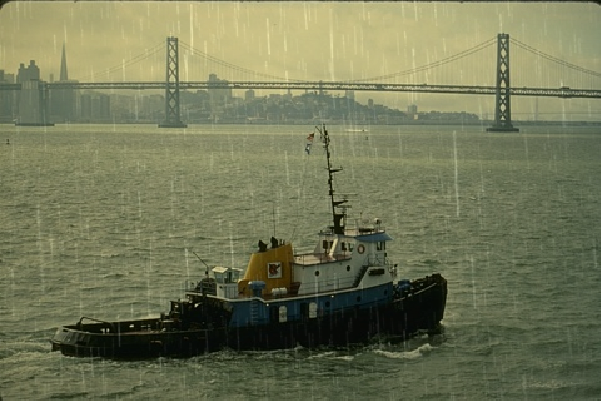}&
                \includegraphics[width=0.64in]{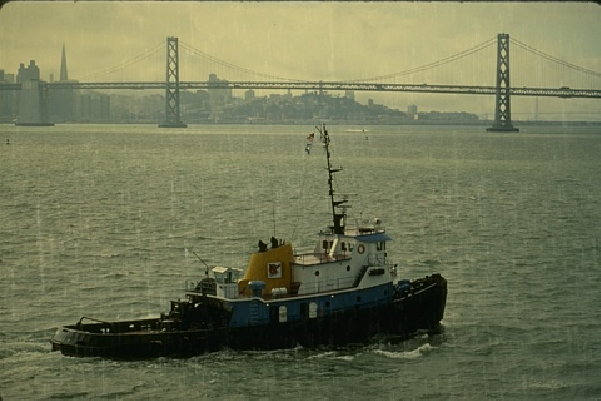}&
                \includegraphics[width=0.64in]{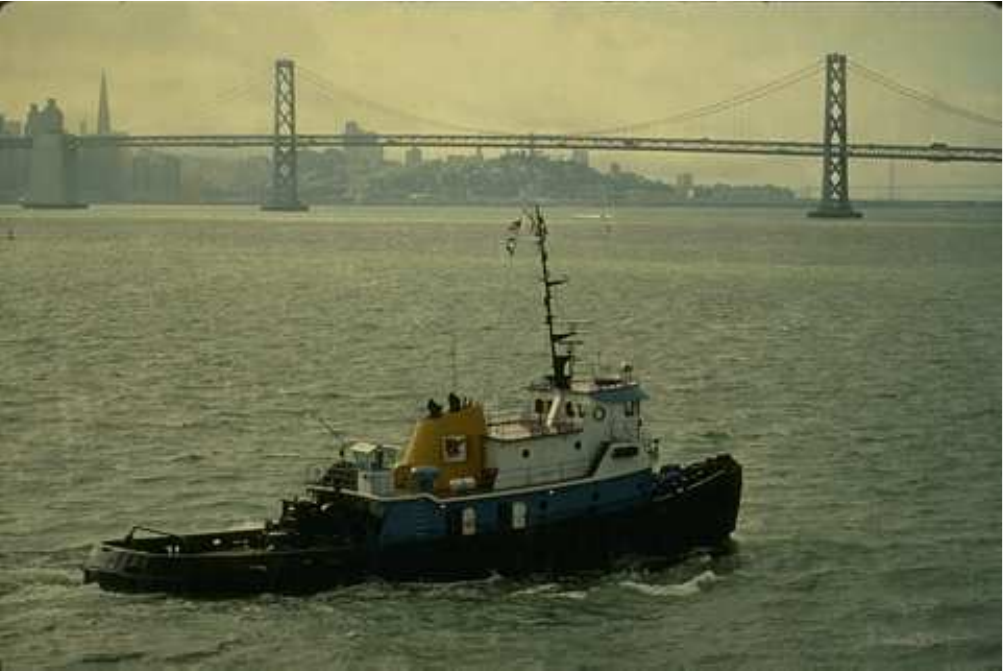}\\
                 \vspace{0.5mm}
                 \includegraphics[width=0.64in]{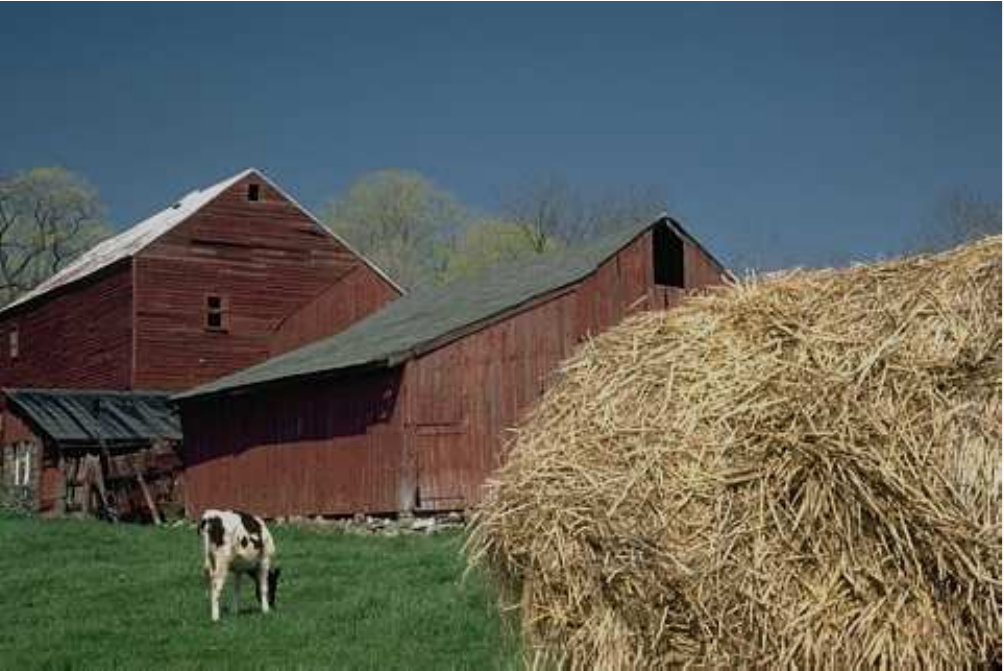}&
                \includegraphics[width=0.64in]{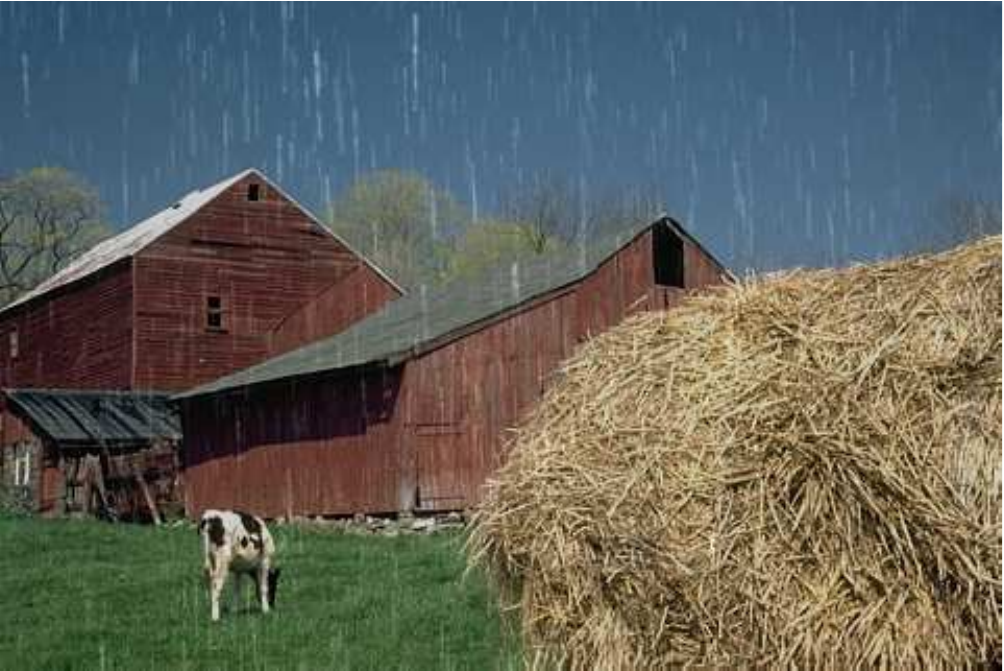}&
                \includegraphics[width=0.64in]{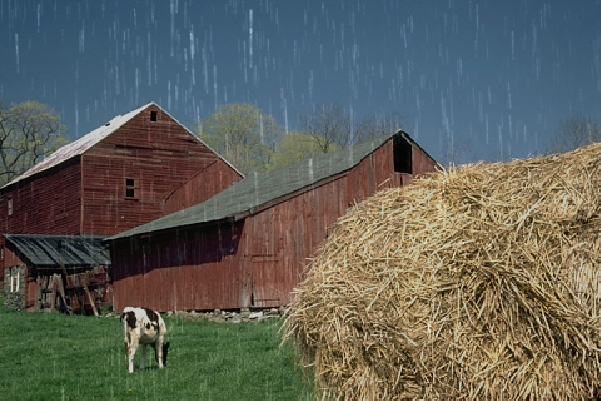}&
                \includegraphics[width=0.64in]{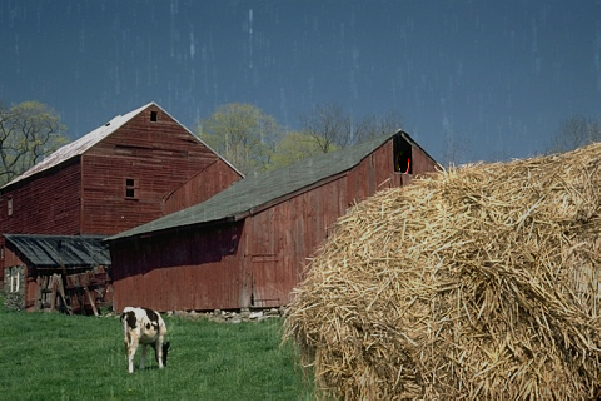}&
                \includegraphics[width=0.64in]{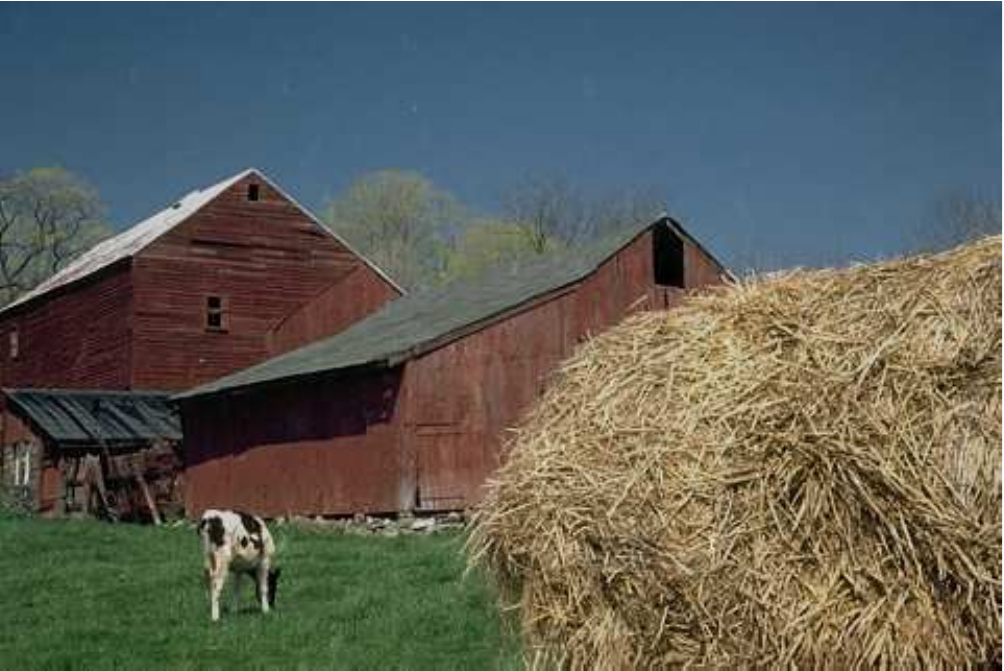}\\
                (a)&
                (b)&
                (c)&
                (d)&
                (e)\\

\end{tabular}
\caption{Rain streak removal results by different methods on 3 selected images from Rain12 dataset. From left to right: (a) the background, (b) the rainy images, the derain results by (c) KGCNN$^a$, (d) KGCNN$^b$, and (d) KGCNN.}
\label{noour-visual}
\end{center}
\end{figure}

In this paper, we propose a kernel guided CNN method for the image rain streak removal application.
The kernel plays a very important role to the KGCNN method.
There are still two problems.
(1) Dose the derain net output the rain streaks only using the information of rainy image and ignoring the kernel information?
(2) Dose the derain net work better if we retain it without the input of kernel?
To show the influence of the kernel in our method, we discard the kernel guided assumption to see the results what will happen with Rain12.
We use KGCNN to represent the proposed method, KGCNN$^a$ to represent the proposed method with kernel information being zero, and KGCNN$^b$ to represent the detrain net trained individually with our training data.
Fig. \ref{noour-visual} shows the visual results there methods.
It is easy to know that the kernel plays an import role in KGCNN.
The derain net dose use the kernel to output the rain streaks (see the result of KGCNN$a$) and even we train the derain net individually, the result is still good enough (see the result of KGCNN$^b$).
The quantitative results in Table \ref{no-quant} also demonstrate the similar conclusion.
In summary, the kernel guided assumption is quite important to the framework of the KGCNN method.

\begin{table}[htb]
\renewcommand\arraystretch{1.1}\setlength{\tabcolsep}{1.5pt}
\caption{Quantitative comparisons of rain streak removal results by KGCNN$^a$, KGCNN$^b$, and KGCNN on Rain12 (average value).}
\label{no-quant}
\begin{center}
\begin{tabular}{c|ccccc}
\Xhline{1.2pt}
      Method &PSNR &SSIM &FSIM &UIQI &GMSD \\
\Xhline{0.8pt}
      rainy &28.822 &0.910 &0.910 &0.968 &0.134      \\
        \hline
      KGCNN$^a$ &29.855 &0.924 &0.919 &0.967 &0.127 \\
      KGCNN$^b$     &33.194 &0.960 &0.950 &\bf{0.986} &0.065 \\
        \hline
      KGCNN &\bf{34.731} &\bf{0.971} &\bf{0.965} &0.983 &\bf{0.055} \\

\hline
\Xhline{1.2pt}
\end{tabular}
\end{center}
\end{table}

\subsection{Discussions on the depth and breadth}
Increasing the depth of network or increasing the filter number of network can improve a network's capacity.
We also investigate the optimal network design to achieve the best derain results.
In this section, we test the impact of network depth and width of KGCNN on Rain12.
Especially, we test for depth $\in \{18,26,34\}$ and filter number $\in \{24,36,48\} $.
Table \ref{net} shows the average values of the quantitative results.
As is clear, adding more hidden layers achieves better results over increasing the number of filters per layer while increasing computational time.
But we could see that there is over-fitting when depth is 34 and filter numbers is 48.
To balance the performance between avoiding the over-fitting and reducing the computation, we choose 26 as depth and 36 as filter number for our experiments above.
\begin{table}[htb]
\renewcommand\arraystretch{0.9}\setlength{\tabcolsep}{1.5pt}
\caption{Quantitative comparisons of rain streak removal results by different depth and filter number on Rain12(average value).}
\label{net}
\begin{center}
\begin{tabular}{c|c|ccccc}
\Xhline{1.2pt}
      depth &filter number &PSNR &SSIM &FSIM &UIQI &GMSD \\
\Xhline{0.8pt}
\multirow{3}[0]{*}{18}
                      &24 &34.049 &0.969 &0.963 &\bf{0.991} &0.057      \\
                      &36 &34.718 &\bf{0.973} &0.966 &0.985 &0.053 \\
                      &48 &34.738 &0.972 &\bf{0.967} &0.979 &0.052 \\
\hline
\multirow{3}[0]{*}{26}
                      &24 &34.614 &0.972 &0.965 &0.987 &0.053      \\
                      &36 &34.731 &0.971 &0.965 &0.983 &0.055 \\
                      &48 &\bf{34.941} &0.972 &0.966 &0.991 &0.053 \\
\hline
\multirow{3}[0]{*}{34}
                      &24 &34.794 &0.972 &0.966 &0.988 &\bf{0.051}      \\
                      &36 &34.853 &0.972 &0.966 &0.986 &0.051 \\
                      &48 &34.414 &0.969 &0.963 &0.983 &0.059 \\
\hline
\Xhline{1.2pt}
\end{tabular}
\end{center}
\end{table}

\section{Conclusion}\label{conclusion}

We have presented a deep learning architecture called KGCNN for removing rain streaks for single images.
Using guided kernel on the texture component, our approach learns the mapping function between rain image on detail component and rain streaks.
We show that convolutional neural networks, a technology widely used for high-level vision task, with guided kernel can also be exploited to successfully deal with natural images under bad weather conditions.
We also show that KGCNN noticeably outperforms other state-of-the-art methods with respect to image quality.
In addition, by using guided kernel, we are able to show that we do not need a very complex network to perform rain streak removal.


%

%

\section*{Acknowledgment}

The research is supported by NSFC (61876203, 61772003, 61702083) and the Fundamental Research Funds for the Central Universities (ZYGX2016J132, ZYGX2016KYQD142, ZYGX2016J129).
Thank the authors of DID \cite{zhang2018density}, DSC \cite{luo2015removing}, LP \cite{Li2014Single}, UGSM \cite{Deng2018A}, CNN \cite{fu2017clearing}, DDN \cite{fu2017removing} for providing the code.

\ifCLASSOPTIONcaptionsoff
  \newpage
\fi

{\footnotesize
\bibliographystyle{unsrt}
\bibliography{refference}
}

\end{document}